\documentclass[twoside]{report}
\usepackage{fancyhdr}
\usepackage{graphicx}
\usepackage{makeidx}

\catcode`\@=11                                                                  
                                                                                
\newcount\cnta\newcount\cntb\newcount\cntc

\def\showlasttree{%
     \g\cnta\count\l@stdiminfo
     \g\cntb\cnta
     \g\advance\cntb 5 
     \g\advance\cntb \count\l@sttreeheight
     \g\advance\cntb \count\l@sttreeheight
     \ifnum\count\l@sttreeheight=-1\relax
           \g\advance\cntb by 2
           \immediate\write16{Tree contour for dummy node:}
      \else\immediate\write16{Tree contour:}% 
        \fi
     \for\cntc:=\cnta\to\cntb\do\immediate\write16{\the\dimen\cntc}\od}

\newif\ifLaTeX                            % If \@@startpbox is undefined,
\ifx\@ndefined\@@startpbox\LaTeXfalse     % we assume that we are dealing        
    \else\LaTeXtrue\fi                    % with LaTeX.

\immediate\write16{This is TreeTeX, Version 2.3, for use with \ifLaTeX LaTeX%     
                   \else plain TeX\fi.}                                         
                                                                                
\ifLaTeX \let\lineseg\line              % latex_picture is part of latex.tex,   
   \else \let\@line\line                % so you don't need it if you use       
         \input l_pic                   % TreeTeX together with LaTeX. LaTeX    
         \let\lineseg\line              % has the command \line for geometric   
         \let\line\@line                % lines, and plain TeX has the same     
   \fi                                  % command for lines of text. Because    
\catcode`\@=11                                                                  
\let\g\global                                                                   
\def\gxdef{\global\xdef}                                                        
                                                                                
\def\newcount{\alloc@0\count\countdef\insc@unt}                                 
                                                                                
\def\for#1:=#2\to#3\do#4\od{%
   \def\f@rcount{#1}\def\upp@rlimit{#3}\def\b@dy{#4}\f@rcount=#2\relax\dof@r}

\def\dof@r{\ifnum\f@rcount>\upp@rlimit\relax\let\n@xt\relax
                 \else\b@dy\advance\f@rcount\@ne\let\n@xt\dof@r\fi
           \n@xt}
                                                                                
\newcount\@xcount
\newcount\t@mes                                                                    
                                                                                
\def\ex#1\times#2\xe{%                                                          
   \@xcount1 \t@mes#1\def\b@dy{#2}\do@x}

\def\do@x{\ifnum\@xcount>\t@mes\let\n@xt\relax
                \else\b@dy\advance\@xcount\@ne\let\n@xt\do@x\fi
          \n@xt}
                                                                              
\newskip\thickn@ss
\newskip\@nner
\newskip\@uter

\def\rect@ngle#1#2#3{\hbox to 0pt{%
     \thickn@ss#3%
     \g\@nner#2\g\advance\@nner-\thickn@ss
     \g\divide\@nner\tw@
     \g\@uter#2\g\advance\@uter\thickn@ss
     \g\divide\@uter\tw@
     \hskip 0pt minus .5fil%
     \vrule height\@uter depth\@nner width\thickn@ss
     \vrule height\@uter depth-\@nner width#1%
     \hskip 0pt minus 1fil%
     \vrule height-\@nner depth\@uter width#1%
     \vrule height\@nner depth\@uter width\thickn@ss
     \hskip 0pt minus .5fil%
     }% \hbox
     }% \def

\def\s@ries#1#2{%                                                               
     \g\t@mpcnta1                                                               
     \gdef\t@mp{#1}%                                                            
     \@ssign#2/\l@st  % \l@st is a sentinal element                                                        
     \expandafter\gdef\csname#1\endcsname##1{%                                  
                      \csname#1\romannumeral##1\endcsname}%                     
     }                                                                          
                                                                                
\def\@ssign#1/#2{%
      \expandafter\gdef\csname\t@mp\romannumeral\t@mpcnta\endcsname{#1}%    
      \g\advance\t@mpcnta\@ne                                            
      \ifx#2\l@st                                                               
          \g\let\n@xt\relax                                                     
     \else\g\let\n@xt\@ssign                                                    
       \fi                                                                      
      \n@xt}
                                                                                 
\newdimen\leftdist                                                              
\newdimen\rightdist                                                             
\newbox\TeXTree                                                                 
                                                                                
\newcount\sl@pe                                                                 
\newcount\l@vels                                                                
\newcount\s@ze                                                                  
                                                                                
\newbox\circleb@x                                                               
\newbox\squareb@x                                                               
\newbox\dotb@x                                                                  
\newbox\triangleb@x
\newbox\textb@x                                                             
\newbox\frameb@x
                                                                                
\newdimen\circlew@dth                                                           
\newdimen\squarew@dth                                                           
\newdimen\dotw@dth                                                              
\newdimen\trianglew@dth                                                         
\newdimen\textw@dth                                                             
\newdimen\framew@dth
                                                         
\newdimen\vd@st                                                                 
\newdimen\hd@st                                                                 
\newdimen\based@st                                                              
\newdimen\dummyhalfcenterdim@n                                                  
                                                                                
\newcount\t@mpcnta                                                              
\newcount\t@mpcntb                                                              
\newcount\t@mpcntc                                                              
\newcount\t@mpcntd                                                              
\newdimen\t@mpdima                                                              
\newdimen\t@mpdimb                                                              
\newdimen\t@mpdimc                                                              
\newbox\t@mpboxa                                                                
\newbox\t@mpboxb                                                                
                                                                                
\newbox\leftb@x                                                                 
\newbox\rightb@x                                                                
\newbox\centerb@x                                                               
\newbox\beneathb@x                                                              
\newtoks\typ@                                                                   
\newbox\centerb@@x                                                              
\newdimen\centerdim@n                                                           
\newdimen\halfcenterdim@n                                                       
                                                                                
\newdimen\mins@p                                                                
\newdimen\halfmins@p                                                            
\newdimen\tots@p                                                                
\newdimen\halftots@p                                                            
\newdimen\currs@p                                                               
\newdimen\adds@p                                                                
\newcount\l@ftht                                                                
\newcount\r@ghtht                                                               
\newcount\l@ftinfo                                                              
\newcount\r@ghtinfo                                                             
\newbox\l@ftbox                                                                 
\newbox\r@ghtbox                                                                
                                                                                
\newif\ifr@ghthigher  % true iff the right subtree is higher than the left one  
\newif\ifadds@p                                                                 
                                                                                
\newcount\@larg                                                                 
\newcount\@rarg                                                                 
                                                                                
\newif\ifl@fttop                                                                
\newif\ifl@ftonly                                                               
\newif\ifr@ghtonly                                                              
\newif\if@xt                                                                    
\newif\ifl@ftedge                                                               
\newif\ifr@ghtedge                                                              
\newif\ifext@nded                                                               
                                                                                
\newdimen\lm@ff                                                                 
\newdimen\rm@ff                                                                 
\newdimen\lb@ff                                                                 
\newdimen\rb@ff                                                                 
\newdimen\lt@p                                                                  
\newdimen\rt@p                                                                  

\newcount\l@sttreebox     % These four counter allocations have been copied
\newcount\l@sttreeheight  % to this position from the \Tree command
\newcount\l@stdiminfo     % (vs. 2.1). Previously each tree allocated its own
\newcount\l@sttreetype    % counters, using up counters for nothing.                                                 
                                                                                
\s@ries{xv@l}{0//1//1//1//1//2//1//3//2//3//4//5//1//6//%                       
              5//4//3//5//2//5//3//4//5//6}                                     
\s@ries{yv@l}{1//6//5//4//3//5//2//5//3//4//5//6//1//5//%                       
              4//3//2//3//1//2//1//1//1//1}

\def\hv@ldef{%                                                                  
     \for\t@mpcnta:=1\to24%                                                     
      \do\g\t@mpdima\vd@st\g\multiply\t@mpdima by\xv@l{\t@mpcnta}%              
         \g\divide\t@mpdima by\yv@l{\t@mpcnta}\g\multiply\t@mpdima by 2         
         \expandafter\gxdef\csname hv@l\romannumeral\t@mpcnta\endcsname{%       
                          \the\t@mpdima}%                                       
      \od}                                                                      
                                                                                
\def\hv@l#1{\csname hv@l\romannumeral#1\endcsname}

\def\p@s#1#2#3{%                                                                
     \g#1\csname#3info\endcsname                                                
     \gxdef\t@mp{\csname#3ht\endcsname}%                                        
     \ifnum\t@mp<0 \gxdef\t@mp{0}\fi                                            
     #2{#1}%                                                                    
     }                                                                          
                                                                                
\chardef\@lmoff0 \chardef\@rmoff1 \chardef\@ltop4 \chardef\@rtop5               
\chardef\@lboff2 \chardef\@rboff3 \chardef\@loff4 \chardef\@roff5               
                                                                                
\def\lmoff#1{\g\advance#1 by\@lmoff}                                            
\def\rmoff#1{\g\advance#1 by\@rmoff}                                            
\def\lboff#1{\g\advance#1 by\@lboff}                                            
\def\rboff#1{\g\advance#1 by\@rboff}                                            
\def\ltop#1{\g\advance#1 by\@ltop}                                              
\def\rtop#1{\g\advance#1 by\@rtop}                                              
\def\loff#1{\g\advance#1 by\@loff\g\advance#1 by\t@mp                           
     \g\advance#1 by\t@mp\relax}                                                
\def\roff#1{\g\advance#1 by\@roff\g\advance#1 by\t@mp                           
     \g\advance#1 by\t@mp\relax}                                                
                                                                                
\def\n@meinfo#1{%                                                               
     \n@me@nfo{#1}{lmoff}\n@me@nfo{#1}{rmoff}%                                  
     \n@me@nfo{#1}{lboff}\n@me@nfo{#1}{rboff}%                                  
     \n@me@nfo{#1}{ltop}\n@me@nfo{#1}{rtop}%                                    
     \n@me@nfo{#1}{loff}\n@me@nfo{#1}{roff}%                                    
     }                                                                          
                                                                                
\def\n@me@nfo#1#2{%                                                             
     \p@s\t@mpcnta{\csname#2\endcsname}{#1}%                                    
     \expandafter\gxdef\csname#1#2\endcsname{\dimen\the\t@mpcnta}}              
                                                                                
\def\n@metree#1#2#3#4#5{%                                                       
     \expandafter\gxdef\csname#5ht\endcsname{\count\the#1}%                     
     \expandafter\gxdef\csname#5info\endcsname{\count\the#2}%                   
     \expandafter\gxdef\csname#5box\endcsname{\the#3}%                          
     \expandafter\gxdef\csname#5type\endcsname{\toks\the#4}%                    
     \n@meinfo{#5}%                                                             
     }                                                                          
                                                                                
\chardef\@cntoff3 \chardef\@boxoff1 \chardef\@dimoff2 \chardef\@typeoff1        
                                                                                
\def\pr@vioustree{%                                                             
     \g\advance\l@sttreeheight by-\@cntoff                                      
     \g\advance\l@stdiminfo by-\@cntoff                                         
     \g\advance\l@sttreetype by-\@cntoff                                        
     \g\advance\l@sttreebox by-\@boxoff                                         
     \n@mel@st                                                                  
     }                                                                          
                                                                                
\def\@ddname#1#2{%                                                              
     \expandafter\gxdef\csname#2ht\endcsname{\csname#1ht\endcsname}%            
     \expandafter\gxdef\csname#2info\endcsname{\csname#1info\endcsname}%        
     \expandafter\gxdef\csname#2type\endcsname{\csname#1type\endcsname}%        
     \expandafter\gxdef\csname#2box\endcsname{\csname#1box\endcsname}%          
     \n@meinfo{#2}%                                                             
     }                                                                          
                                                                                
\def\n@xttree{%                                                                 
     \p@s\t@mpcnta\loff{l@st}\g\advance\t@mpcnta by\@dimoff                     
     \g\advance\l@sttreeheight by\@cntoff                                       
     \g\advance\l@stdiminfo by\@cntoff                                          
     \g\advance\l@sttreetype by\@cntoff                                         
     \g\advance\l@sttreebox by\@boxoff                                          
     \g\count\l@stdiminfo\t@mpcnta                                              
     }                                                                          
                                                                                
\def\@ppenddummy{% pushs a dummy onto the stack and names it `l@st'             
     \n@xttree \g\count\l@sttreeheight-\@ne\n@mel@st                            
     \l@sttype{circle}%                                                         
     \g\setbox\l@stbox\copy\voidb@x                                             
     \g\l@stlmoff=0pt\g\l@strmoff=0pt\g\l@stlboff=0pt\g\l@strboff=0pt%          
     \g\l@stltop=0pt\g\l@strtop=0pt\g\l@stloff=0pt\g\l@stroff=0pt%              
     }                                                                          
                                                                                
\def\g@tchildren{% enables us to talk about the left and the right child
     \ifl@fttop\@ddname{l@st}{l@ft}%                                            
               \pr@vioustree                                                    
               \@ddname{l@st}{r@ght}%                                           
          \else\@ddname{l@st}{r@ght}%                                           
               \pr@vioustree                                                    
               \@ddname{l@st}{l@ft}%                                            
            \fi                                                                 
     \ifnum\r@ghtht>\l@ftht\relax                                               
               \r@ghthighertrue                                                 
               \@ddname{r@ght}{m@x}%                                            
               \@ddname{l@ft}{m@n}%                                             
          \else\r@ghthigherfalse                                                
               \@ddname{l@ft}{m@x}%                                             
               \@ddname{r@ght}{m@n}%                                            
            \fi                                                                 
               }                                                                
                                                                                
\def\n@mel@st{%                                                                 
     \n@metree\l@sttreeheight\l@stdiminfo\l@sttreebox\l@sttreetype{l@st}}       
                                                                                
\def\beginTree{%                                                                
     \begingroup                                                                
     \unitlength 1pt%                                                           
     \divide\unitlength by 65536                                                
     \l@sttreebox\count14                                                       
     \l@sttreeheight\count10                                                    
     \advance\l@sttreeheight by \@ne                                            
     \count\l@sttreeheight=-1                                                   
     \l@stdiminfo\l@sttreeheight                                                
     \advance\l@stdiminfo by \@ne                                               
     \count\l@stdiminfo\count11                                                 
     \advance\count\l@stdiminfo by -5                                           
     \l@sttreetype\l@stdiminfo                                                  
     \advance\l@sttreetype by\@ne                                               
     \count\l@sttreetype\count15                                                
     \n@mel@st\ignorespaces                                                     
     \treef@nts  % vs. 2.2 introduces \treef@nts to keep font
     }

\def\endTree{%                                                                  
     \g\leftdist-\l@stlmoff\g\advance\leftdist by \l@stltop                     
     \g\rightdist\l@strmoff\g\advance\rightdist by\l@strtop                     
     \g\setbox\TeXTree\box\l@stbox\endgroup\ignorespaces}

\def\th@ck{\let\@linefnt\tenlnw                                                 
     \@wholewidth\fontdimen8\tenlnw\@halfwidth.5\@wholewidth}                   
                                                                                
\def\leftthick{\g\let\l@ftthick\th@ck}                                          
\def\rightthick{\g\let\r@ghtthick\th@ck}                                        
\def\lft#1{\g\setbox\leftb@x\hbox{#1\ }}                                        
\def\rght#1{\g\setbox\rightb@x\hbox{\ #1}}                                      
\def\cntr#1{\g\setbox\centerb@x\hbox{#1\strut}}                   
\def\bnth#1{\g\setbox\beneathb@x\hbox to0pt{\hss\strut#1\hss}}                        
\def\type#1{%                                                                   
     \g\setbox\centerb@@x\copy\csname#1b@x\endcsname                            
     \g\centerdim@n\csname#1w@dth\endcsname                                     
     \typ@{#1}%                                                                 
     \g\halfcenterdim@n=.5\centerdim@n}                                         
                                                                                
\def\ext@nded{\g\ext@ndedfalse} % This definition must precede                  

\def\node#1{%                                                                   
     \g\setbox\leftb@x\copy\voidb@x                                             
     \g\setbox\rightb@x\copy\voidb@x                                            
     \g\setbox\centerb@x\copy\voidb@x                                           
     \g\setbox\beneathb@x\copy\voidb@x                                          
     \type{circle}%                                                             
     \g\l@fttopfalse\g\l@ftonlyfalse\g\l@ftedgetrue                             
     \g\r@ghtonlyfalse\g\r@ghtedgetrue\g\@xtfalse\ext@nded\n@dummy              
     \g\let\l@ftthick\relax\g\let\r@ghtthick\relax                              
     #1% 
     \@pdcenter                                                                       
     \d@mmy
     \n@de
     \ignorespaces                                                              
     }

\def\@pdcenter{\csname\the\typ@ @cntr\endcsname}

\let\circle@cntr\relax
\let\square@cntr\relax
\let\triangle@cntr\relax
\let\dot@cntr\relax

\def\text@cntr{%
     \g\centerdim@n\wd\centerb@x
     \g\halfcenterdim@n.5\centerdim@n}                                                                          
     
\def\frame@cntr{% 
     \g\setbox\centerb@x\hbox{\ \unhcopy\centerb@x\ }
     \g\centerdim@n\wd\centerb@x
     \g\halfcenterdim@n.5\centerdim@n
     \g\setbox\centerb@@x\rect@ngle{\centerdim@n}{\squarew@dth}{.4pt}}
                                                                           
\def\leftonly{\g\l@ftonlytrue\g\r@ghtedgefalse\g\let\d@mmy\l@ftdummy}           
\def\rightonly{\g\r@ghtonlytrue\g\l@ftedgefalse\g\let\d@mmy\r@ghtdummy}         
\def\unary{\g\r@ghtedgefalse\g\let\d@mmy\@ndummy}                               
\def\external{\g\@xttrue\g\l@ftedgefalse\g\r@ghtedgefalse\g\let\d@mmy\@xtdummy} 
                                                                                
\def\lefttop{\g\l@fttoptrue}                                                    
                                                                                
\def\@xtdummy{%                                                                 
     \@ppenddummy                                                     
     \g\l@strtop-\halfmins@p                                         
     \@ppenddummy
     \g\l@stltop-\halfmins@p                                         
     }                                                                          
                                                                                
\def\n@dummy{\g\let\d@mmy\relax}                                                
                                                                                
\def\l@ftdummy{% cf. \g@tposition                                                               
     \@ppenddummy                                                      
     \g\l@stltop=\dummyhalfcenterdim@n                                          
     \g\l@strtop=\dummyhalfcenterdim@n                                          
     }                                                                          
                                                                                
\def\r@ghtdummy{% cf. \g@tposition                                                              
     \lefttop                                                                   
     \@ppenddummy                                                      
     \g\l@stltop=\dummyhalfcenterdim@n                                          
     \g\l@strtop=\dummyhalfcenterdim@n                                          
     }                                                                          
                                                                                
\def\@ndummy{%                                                                  
     \g\t@mpdima\l@strtop\relax                                                 
     \@ppenddummy                                                     
     \g\l@stltop-\mins@p\g\advance\l@stltop by-\t@mpdima                        
     \g\l@strtop=\t@mpdima                                                      
     }                                                                          
                                                                                
\def\n@de{%                                                                     
     \g@tposition       % naming children and calculating \sl@pe and \tots@p    
     \g@tlt@p\g@trt@p   % calculating \lt@p and \rt@p                           
     \g@tlm@ff\g@trm@ff % calculating \lm@ff and \rm@ff                         
     \g@tlb@ff\g@trb@ff % calculating \lb@ff and \rb@ff                         
     \@pdlroff          % updating loff and roff for all levels but the top one 
     \@pdloffl\@pdroffl % updating loff(1) and roff(1) of the parent tree             
     \@pddim            % updating ltop, rtop, lmoff, rmoff, lboff, and rboff   
     \@pdinfo\@pdht     % updating diminfo and treeheight                       
     \@pdbox            % updating treebox                                      
     \@pdtype           % updating type                                         
     \n@mel@st          % giving the name `l@st' to the new tree 
     \ignorespaces                                                              
     }                                                                          
                                                                                
\def\g@tposition{% naming children and calculating \sl@pe, \tots@p, and node offsets
     \g@tchildren\c@lcsep\c@lcslope\c@lcoffsets
     \ifext@nded\relax                                                          
           \else\ifl@ftonly\g\r@ghtrtop=-\tots@p                                
                           \g\advance\r@ghtrtop by\l@ftrtop                     
                        \fi                                                     
                \ifr@ghtonly\g\l@ftltop=-\tots@p                                
                            \g\advance\l@ftltop by\r@ghtltop                    
                         \fi                                                    
             \fi % cf. \l@ftdummy and \r@ghtdummy                                                               
     }                                                                          
                                                                                
\def\@pdinfo{% updating diminfo                                                 
     \g\l@stinfo=\m@xinfo\relax                                                 
     }                                                                          
                                                                                
\def\@pdht{% updating treeheight                                                
     \g\l@stht=\m@xht                                                           
     \g\advance\l@stht by\@ne                                                   
     }                                                                          
                                                                                
\def\@pdtype{% updating type                                                    
     \g\l@sttype\typ@                                                           
     }                                                                          
                                                                                
\def\g@tlt@p{% calculating \lt@p                                                
     \g\lt@p\wd\leftb@x\g\advance\lt@p by\halfcenterdim@n                       
     }                                                                          
                                                                                
\def\g@trt@p{% calculating \rt@p                                                
     \g\rt@p\wd\rightb@x\g\advance\rt@p by\halfcenterdim@n                      
     }                                                                          
                                                                                
\def\g@tlm@ff{% calculating \lm@ff                                              
     \g\lm@ff\l@ftlmoff                                                         
     \g\advance\lm@ff by-\l@ftltop                                              
     \g\advance\lm@ff by-\halftots@p                                            
     \g\advance\lm@ff by\lt@p\relax                                             
     \ifnum\l@ftht<\r@ghtht\relax                                               
           \g\t@mpdima\r@ghtlmoff                                               
           \g\advance\t@mpdima by-\r@ghtltop                                    
           \g\advance\t@mpdima by\halftots@p                                    
           \g\advance\t@mpdima by\lt@p\relax                                    
           \ifdim\t@mpdima<\lm@ff\relax                                         
                 \g\lm@ff\t@mpdima                                              
              \fi                                                               
        \fi                                                                     
     \ifdim0pt<\lm@ff\relax                                                     
           \g\lm@ff=0pt%                                                        
        \fi                                                                     
     }                                                                          
                                                                                
\def\g@trm@ff{% calculating \rm@ff                                              
     \g\rm@ff\r@ghtrmoff                                                        
     \g\advance\rm@ff by\r@ghtrtop                                              
     \g\advance\rm@ff by\halftots@p                                             
     \g\advance\rm@ff by-\rt@p\relax                                            
     \ifnum\r@ghtht<\l@ftht\relax                                               
           \g\t@mpdima\l@ftrmoff                                                
           \g\advance\t@mpdima by\l@ftrtop                                      
           \g\advance\t@mpdima by-\halftots@p                                   
           \g\advance\t@mpdima by-\rt@p\relax                                   
           \ifdim\t@mpdima>\rm@ff\relax                                         
                 \g\rm@ff\t@mpdima                                              
              \fi                                                               
        \fi                                                                     
     \ifdim0pt>\rm@ff\relax                                                     
           \g\rm@ff=0pt                                                         
        \fi                                                                     
     }                                                                          
                                                                                
\def\g@tlb@ff{% calculating \lb@ff                                              
     \if@xt\g\lb@ff0pt%
      \else\ifnum\l@ftht<\r@ghtht\relax                                               
                 \g\lb@ff\r@ghtlboff                                                  
                 \g\advance\lb@ff by-\r@ghtltop                                       
                 \g\advance\lb@ff by\halftots@p                                       
                 \g\advance\lb@ff by\lt@p\relax                                       
            \else\g\lb@ff\l@ftlboff                                                   
                 \g\advance\lb@ff by-\l@ftltop                                        
                 \g\advance\lb@ff by-\halftots@p                                      
                 \g\advance\lb@ff by\lt@p\relax                                       
              \fi
        \fi                                                                     
     }                                                                          
                                                                                
\def\g@trb@ff{% calculating \rb@ff                                              
     \if@xt\g\rb@ff0pt%
      \else\ifnum\r@ghtht<\l@ftht\relax                                               
                 \g\rb@ff\l@ftrboff                                                   
                 \g\advance\rb@ff by\l@ftrtop                                         
                 \advance\rb@ff by-\halftots@p                                        
                 \g\advance\rb@ff by-\rt@p\relax                                      
            \else\g\rb@ff\r@ghtrboff                                                  
                 \g\advance\rb@ff by\r@ghtrtop                                        
                 \g\advance\rb@ff by\halftots@p                                       
                 \g\advance\rb@ff by-\rt@p\relax                                      
              \fi
        \fi                                                                     
     }                                                                          
                                                                                
\def\@pdlroff{% updating loff and roff for all levels but the top one           
     \ifr@ghthigher\g\t@mpdima-\r@ghtltop\relax                                 
                   \g\t@mpdimb\l@ftlboff                                        
                   \g\advance\t@mpdimb by-\l@ftltop\relax                       
              \else\g\t@mpdima\l@ftrtop\relax                                   
                   \g\t@mpdimb\r@ghtlboff                                       
                   \g\advance\t@mpdimb by\r@ghtrtop\relax                       
                \fi                                                             
     \ifr@ghthigher\p@s\t@mpcnta\loff{m@n}% pointer to loff(1) of smaller tree  
                   \p@s\t@mpcntb\loff{m@x}% pointer to loff(1) of larger tree  
              \else\p@s\t@mpcnta\roff{m@n}% pointer to roff(1) of smaller tree  
                   \p@s\t@mpcntb\roff{m@x}% pointer to roff(1) of larger tree  
                \fi  % if the right tree is the higher one you have to shift    
     \ex\m@nht\times                                                            
        \g\advance\t@mpdima by\dimen\t@mpcntb                                   
        \g\dimen\t@mpcntb\dimen\t@mpcnta                                        
        \g\advance\t@mpcnta by-\@dimoff                                         
        \g\advance\t@mpcntb by-\@dimoff\relax                                   
     \xe                                                                        
     \ifnum\m@xht=\m@nht\relax                                                  
      \else\g\advance\dimen\t@mpcntb by\t@mpdima                                
           \ifnum\l@ftht<\r@ghtht\relax                                         
                 \g\advance\dimen\t@mpcntb by\tots@p                            
            \else\g\advance\dimen\t@mpcntb by-\tots@p                           
              \fi                                                               
           \g\advance\dimen\t@mpcntb by-\t@mpdimb                               
        \fi                                                                     
     }                                                                          
                                                                                
\def\@pdloffl{% updating loff(1) of parent tree                                 
     \p@s\t@mpcnta\loff{m@x}%                                                   
     \g\advance\t@mpcnta by \@dimoff\relax % pointer to loff(0) of parent tree  
     \g\dimen\t@mpcnta\lt@p                                                     
     \g\advance\dimen\t@mpcnta by-\halftots@p                                   
     \g\advance\dimen\t@mpcnta by-\l@ftltop\relax                               
     }                                                                          
                                                                                
\def\@pdroffl{% updating roff(1) of parent tree                                 
     \p@s\t@mpcnta\roff{m@x}%                                                   
     \g\advance\t@mpcnta by \@dimoff\relax % pointer to roff(0) of parent tree  
     \g\dimen\t@mpcnta-\rt@p                                                    
     \g\advance\dimen\t@mpcnta by\halftots@p                                    
     \g\advance\dimen\t@mpcnta by\r@ghtrtop\relax                               
     }                                                                          
                                                                                
\def\@pddim{% updating ltop, rtop, lmoff, rmoff, lboff, and rboff               
     \g\m@xlmoff=\lm@ff\g\m@xrmoff=\rm@ff                                       
     \g\m@xlboff=\lb@ff\g\m@xrboff=\rb@ff                                       
     \g\m@xltop=\lt@p\g\m@xrtop=\rt@p                                           
     }                                                                          
                                                                                
\def\@pdbox{% pushing the nodebox on the stack: updating treebox                
     \g\@xarg\xv@l\sl@pe\g\@yarg\yv@l\sl@pe                                     
     \ifnum\sl@pe=1 % vertical edge                                             
           \g\t@mpdima\vd@st                                                    
           \g\advance\t@mpdima by-\y@ff\typ@                                    
           \g\advance\t@mpdima by-\y@ff\l@fttype                                
           \g\@larg\t@mpdima % \@larg is a number register!                     
           \g\t@mpdima\vd@st                                                    
           \g\advance\t@mpdima by-\y@ff\typ@                                    
           \g\advance\t@mpdima by-\y@ff\r@ghttype                               
           \g\@rarg\t@mpdima % \@rarg is a number register!                     
      \else\g\t@mpdima\halftots@p                                               
           \g\advance\t@mpdima by-\x@ff\typ@                                    
           \g\advance\t@mpdima by-\x@ff\l@fttype                                
           \g\@larg\t@mpdima % \@larg is a number register!                     
           \g\t@mpdima\halftots@p                                               
           \g\advance\t@mpdima by-\x@ff\typ@                                    
           \g\advance\t@mpdima by-\x@ff\r@ghttype                               
           \g\@rarg\t@mpdima % \@rarg is a number register!                     
        \fi                                                                     
     \g\setbox\l@sttreebox\hbox{%
           \ifvoid\leftb@x\relax
             \else\hskip-\halfcenterdim@n\hskip-\wd\leftb@x
                  \unhcopy\leftb@x\hskip\halfcenterdim@n
               \fi
           \ifvoid\centerb@x\relax
             \else\g\t@mpdima-.5\wd\centerb@x\hskip\t@mpdima
                  \unhbox\centerb@x\hskip\t@mpdima
               \fi
           \ifvoid\rightb@x\relax
             \else\g\t@mpdima-\wd\rightb@x\hskip\halfcenterdim@n
                  \unhbox\rightb@x\hskip\t@mpdima\hskip-\halfcenterdim@n
               \fi
           \raise\based@st\copy\centerb@@x
           \if@xt\relax
                 \lower\s@ze pt\hbox to0pt{\hss\unhbox\beneathb@x\hss}%
            \else\hskip-\halftots@p
                 \lower\vd@st\box\l@ftbox
                 \ifl@ftedge\drawl@ftedge\else\hskip\halftots@p\fi
                 \ifr@ghtedge\drawr@ghtedge\else\hskip\halftots@p\fi
                 \lower\vd@st\box\r@ghtbox
                 \hskip-\halftots@p
              \fi
           }% of hbox
     }                                                                          
                                                                                
\def\drawl@ftedge{%                                                             
           \hskip\x@ff\l@fttype                                                 
           \g\t@mpdimc\y@ff\l@fttype\g\advance\t@mpdimc by\based@st
           \g\advance\t@mpdimc-\vd@st
         \ifx\l@ftthick\relax\relax
         \else
           \raise\t@mpdimc                                                      
           \hbox to 0pt{\hskip.3pt\l@ftthick\lineseg(\@xarg,\@yarg){\@larg}\hss}%
           \raise\t@mpdimc                                                      
           \hbox to 0pt{\hskip.6pt\l@ftthick\lineseg(\@xarg,\@yarg){\@larg}\hss}%
           \raise\t@mpdimc                                                      
           \hbox to 0pt{\hskip-.3pt\l@ftthick\lineseg(\@xarg,\@yarg){\@larg}\hss}%
         \fi
           \raise\t@mpdimc                                                      
           \hbox{\l@ftthick\lineseg(\@xarg,\@yarg){\@larg}}%
           \hskip\x@ff\typ@                                                     
     }                                                                          
                                                                                
\def\drawr@ghtedge{%                                                            
           \hskip\x@ff\typ@                                                     
           \g\t@mpdimc\vd@st                                                      
           \g\advance\t@mpdimc by \based@st
           \g\advance\t@mpdimc by -\y@ff\typ@\relax
           \g\advance\t@mpdimc by- \vd@st
         \ifx\r@ghtthick\relax\relax
         \else
           \raise\t@mpdimc                                                      
           \hbox to 0pt{\hskip.3pt\r@ghtthick\lineseg(\@xarg,-\@yarg){\@rarg}\hss}% 
           \raise\t@mpdimc                                                      
           \hbox to 0pt{\hskip-.3pt\r@ghtthick\lineseg(\@xarg,-\@yarg){\@rarg}\hss}% 
           \raise\t@mpdimc                                                      
           \hbox to 0pt{\hskip-.6pt\r@ghtthick\lineseg(\@xarg,-\@yarg){\@rarg}\hss}% 
         \fi
           \raise\t@mpdimc                                                      
           \hbox{\r@ghtthick\lineseg(\@xarg,-\@yarg){\@rarg}}% 
           \hskip\x@ff\r@ghttype                                                
     }                                                                          
                                                                                
\def\x@ff#1{%                                                                   
     \csname\the#1x@ff\endcsname\sl@pe                                          
     }                                                                          
                                                                                
\def\y@ff#1{%                                                                   
     \csname\the#1y@ff\endcsname\sl@pe                                          
     }                                                                          
                                                                                
\def\c@lcslope{%                                                                
     \g\sl@pe1                                                                  
       \loop                                                                    
      \ifdim\hv@l\sl@pe < \tots@p                                               
            \g\advance\sl@pe by1                                                
     \repeat
     \g\tots@p\hv@l\sl@pe                                                       
     \g\halftots@p.5\tots@p}

\def\c@lcsep{%                                                                  
     \g\tots@p\mins@p                                                           
     \g\advance\tots@p by\l@ftrtop                                              
     \g\advance\tots@p by\r@ghtltop\relax                                       
     \g\currs@p\mins@p                                                          
     \p@s\t@mpcnta\roff{l@ft}%                                                  
     \p@s\t@mpcntb\loff{r@ght}%                                                 
     \g\adds@pfalse                                                             
     \g\t@mpcntc\m@nht                               
     \ex\t@mpcntc\times                                                         
        \g\advance\currs@p by-\dimen\t@mpcnta                                   
        \g\advance\currs@p by \dimen\t@mpcntb                                   
        \ifdim\mins@p<\currs@p                                                  
         \else\g\adds@ptrue                                                     
           \fi                                                                  
        \ifdim\currs@p<\mins@p                                                  
              \g\advance\tots@p by\mins@p                                       
              \g\advance\tots@p by -\currs@p                                    
              \g\currs@p\mins@p                                                 
           \fi                                                                  
        \g\advance\t@mpcnta by -\@dimoff                                        
        \g\advance\t@mpcntb by -\@dimoff                                        
     \xe                                                                        
     \ifadds@p\g\advance\tots@p by\adds@p\fi}                                   
                                                                                
\def\tri@ngle{%                                                                 
     \vtop{\g\@xarg\xv@l\sl@pe \g\@yarg\yv@l\sl@pe                              
           \g\t@mpdimc\l@vels\vd@st 
           \g\advance\t@mpdimc by .5\squarew@dth
           \g\multiply\t@mpdimc\xv@l\sl@pe
           \g\divide\t@mpdimc\yv@l\sl@pe
           \g\@larg\t@mpdimc                                                    
           \offinterlineskip                                                    
           \vskip0pt% Force the reference point to the top                                                         
           \hbox to0pt{\hss\lineseg(\@xarg,\@yarg){\@larg}%                     
                       \hskip\t@mpdimc\rlap{\lineseg(-\@xarg,\@yarg){\@larg}}%  
                       \hss}%                                                   
           \setbox\t@mpboxa                                                          
           \hbox to0pt{\hss\vrule height.2pt depth.2pt width2\t@mpdimc\hss}%
           \t@mpdimc-.5\squarew@dth\advance\t@mpdimc\based@st
           \ht\t@mpboxa0pt\dp\t@mpboxa\t@mpdimc\copy\t@mpboxa    
          }%                                                                    
     }                                                                          
                                                                                
\def\lvls#1{\g\l@vels#1}                                                        
\def\slnt#1{\g\sl@pe#1}                                                         
                                                                                
\def\treesymbol#1{%                                                             
     \g\setbox\leftb@x\copy\voidb@x                                             
     \g\setbox\rightb@x\copy\voidb@x                                            
     \g\setbox\centerb@x\copy\voidb@x                                           
     \g\setbox\beneathb@x\copy\voidb@x                                          
     \lvls{1}\slnt{3}%                                                          
     #1%                                                                        
     \g\centerdim@n\trianglew@dth                                               
     \g\halfcenterdim@n.5\trianglew@dth                                         
     \n@xttree                                                                  
     \g\count\l@sttreeheight\l@vels% \g\advance\count\l@sttreeheight by\tw@  
     \g\toks\l@sttreetype{triangle}%                                            
     \n@mel@st                                                                  
     \g\hd@st\hv@l\sl@pe \g\divide\hd@st by\tw@                                 
     \g\l@stltop=\halfcenterdim@n\g\advance\l@stltop by\wd\leftb@x              
     \g\l@strtop=\halfcenterdim@n\g\advance\l@strtop by\wd\rightb@x             
     \g\l@stlboff=-\hd@st \g\multiply\l@stlboff by\l@vels   
     \g\advance\l@stlboff by\wd\leftb@x
     \g\l@strboff=\hd@st \g\multiply\l@strboff by\l@vels 
     \g\advance\l@strboff by-\wd\rightb@x
     \g\l@stlmoff=\l@stlboff\relax
     \ifdim\l@stlmoff>0pt\relax\g\l@stlmoff=0pt\fi   
     \g\l@strmoff=\l@strboff
     \ifdim\l@strmoff<0pt\relax\g\l@strmoff=0pt\fi 
     \g\t@mpcnta\l@stinfo\g\advance\t@mpcnta by6% preliminary                   
     \ex\l@vels\times                                                           
        \g\dimen\t@mpcnta-\hd@st\g\advance\t@mpcnta by\@ne                      
        \g\dimen\t@mpcnta\hd@st\g\advance\t@mpcnta by\@ne                       
     \xe                                                                        
     \g\advance\t@mpcnta by-\tw@                                                
     \g\advance\dimen\t@mpcnta by\wd\leftb@x                                    
     \g\advance\t@mpcnta by\@ne                                                 
     \g\advance\dimen\t@mpcnta by-\wd\rightb@x                                  
     \g\setbox\l@stbox\vtop % to\l@vels\vd@st
            {\offinterlineskip                   
             \g\setbox\t@mpboxa                                                 
             \hbox{\hskip-\halfcenterdim@n\hskip-\wd\leftb@x\unhbox\leftb@x
                   \hskip\halfcenterdim@n
                   \raise\based@st\tri@ngle
                   \hskip\halfcenterdim@n\t@mpdima-\wd\rightb@x
                   \unhbox\rightb@x\hskip\t@mpdima\hskip-\halfcenterdim@n}                                     
             \g\ht\t@mpboxa=0pt\box\t@mpboxa                                    
             \setbox\centerb@x\hbox to0pt{\hss\unhbox\centerb@x\hss}%
             \ht\centerb@x0pt\dp\centerb@x0pt\box\centerb@x
             \vskip\s@ze pt
             \ht\beneathb@x0pt\box\beneathb@x                                                 
             \vskip-\dp\beneathb@x\vskip-\ht\beneathb@x}%
     \ignorespaces                                                              
     }

\def\norm@ff{% everything set up for 10pt node size                                                                 
\s@ries{circley@ff}{0.50000pt//0.49320pt//0.49029pt//0.48507pt//%               
                    0.47434pt//0.46424pt//0.44721pt//0.42875pt//%               
                    0.41603pt//0.40000pt//0.39043pt//0.38411pt//%               
                    0.35355pt//0.32009pt//0.31235pt//0.30000pt//%               
                    0.27735pt//0.25725pt//0.22361pt//0.18570pt//%               
                    0.15811pt//0.12127pt//0.09806pt//0.08220pt}%                
     }                                                                          
                                                                                
\def\dotx@ff#1{0pt}
\def\doty@ff#1{0pt} 

\def\trianglex@ff#1{0pt}
\def\triangley@ff#1{0pt} 

\def\c@lcoffsets{%
     \ifnum\sl@pe=\@ne\relax
           \xdef\circlex@ff##1{0pt}%
      \else\g\t@mpcnta26 % number of slopes + 2
           \g\advance\t@mpcnta-\sl@pe
           \xdef\circlex@ff##1{\circley@ff\t@mpcnta}%
        \fi
     \ifnum\sl@pe<13\relax % incoming edge meets upper border of a square node
           \g\t@mpdima.5\squarew@dth
           \xdef\squarey@ff##1{\the\t@mpdima}%
           \g\multiply\t@mpdima\xv@l\sl@pe
           \g\divide\t@mpdima\yv@l\sl@pe
           \xdef\squarex@ff##1{\the\t@mpdima}%
      \else\g\t@mpdima.5\squarew@dth
           \xdef\squarex@ff##1{\the\t@mpdima}%
           \g\multiply\t@mpdima\yv@l\sl@pe
           \g\divide\t@mpdima\xv@l\sl@pe
           \xdef\squarey@ff##1{\the\t@mpdima}%
        \fi
     \g\t@mpdima.5\squarew@dth
     \xdef\texty@ff##1{\the\t@mpdima}%
     \g\multiply\t@mpdima\xv@l\sl@pe
     \g\divide\t@mpdima\yv@l\sl@pe
     \xdef\textx@ff##1{\the\t@mpdima}% 
     \let\framex@ff\textx@ff
     \let\framey@ff\texty@ff
    }
                                                                             
\def\upds@ze#1{%                                                                
     \for\t@mpcntc:=1\to24                                                      
      \do\g\t@mpdimc=\csname#1\romannumeral\t@mpcntc\endcsname\relax            
         \g\multiply\t@mpdimc by\s@ze                                           
         \expandafter\gxdef\csname#1\romannumeral\t@mpcntc\endcsname            
                            {\the\t@mpdimc}%                                    
      \od}                                                                      
                                                                                
\def\nodes@ze{%                                                                 
     \begingroup                                                                
     \unitlength 1pt%                                                           
     \divide\unitlength by 65536                                                
     \g\based@st\s@ze pt\g\divide\based@st by 10 % \based@st is 10 % of         
     \g\dummyhalfcenterdim@n=\s@ze pt\g\divide\dummyhalfcenterdim@n by\tw@      
     \g\circlew@dth=\s@ze pt%                                                   
     \g\t@mpcntc\s@ze\g\multiply\t@mpcntc by 65536                              
     \g\setbox\circleb@x\hbox to0pt{\circle{\t@mpcntc}\hss}%                     
     \upds@ze{circley@ff}%                                  
     \g\squarew@dth.9pt\g\multiply\squarew@dth by\s@ze                          
     \g\setbox\squareb@x\rect@ngle{\squarew@dth}{\squarew@dth}{.4pt}%      
     \g\dotw@dth=\s@ze pt\g\divide\dotw@dth by 5                                
     \ifdim\dotw@dth < 1pt\relax                                                
           \g\dotw@dth1pt\relax                                                 
        \fi                                                                     
     \g\t@mpcntc\dotw@dth                                               
     \g\setbox\dotb@x\hbox to 0pt{\circle*{\t@mpcntc}\hss}%
     \g\trianglew@dth=\s@ze pt\g\multiply\trianglew@dth by \tw@                 
     \g\divide\trianglew@dth by 3 
     \g\textw@dth=0pt%
     \g\setbox\textb@x\copy\voidb@x
     \g\framew@dth0pt%
     \g\setbox\frameb@x\copy\voidb@x
     \hv@ldef                                                                   
     \endgroup                                                                  
     }                                                                          
                                                                                
\def\treefonts#1{\g\def\treef@nts{#1}} % vs. 2.2 introduces \treef@nts
\def\vdist#1{\g\vd@st=#1\relax}                                                 
\def\minsep#1{\g\mins@p=#1\relax\g\halfmins@p=.5\mins@p}                        
\def\addsep#1{\g\adds@p=#1\relax}                                               
\def\extended{\def\ext@nded{\g\ext@ndedtrue}}                                   
\def\noextended{\def\ext@nded{\g\ext@ndedfalse}}                                
\def\nodesize#1{\g\t@mpdima=#1\relax\g\s@ze=\t@mpdima                           
     \g\divide\s@ze by 65536\relax} % conversion from dimension to number       
\def\Treestyle#1{\norm@ff#1\nodes@ze\ignorespaces}                                           
                                                                                
\input classes                                                             
                                                                                
\Treestyle{%                                                                    
     \ifLaTeX\treefonts{\normalsize\rm}%                                        
        \else\treefonts{\tenrm}%                                                
          \fi                                                                   
     \vdist{60pt}%                                                              
     \minsep{20pt}%                                                             
     \addsep{0pt}%                                                              
     \nodesize{20pt}%                                                           
     }

\Treestyle{%
  \minsep{10pt}%
  \vdist{30pt}%
  \nodesize{14pt}%
 }    % smaller than default 

\newcommand{\ABSSS}{{MT-SSS*}}
\newcommand{\ABDUAL}{{MT-DUAL*}}
\newcommand{\Keyano}{{Keyano}}
\newcommand{\Chinook}{{Chinook}}
\newcommand{\Phoenix}{{Phoenix}}
\newcommand{\Test}{{Test}}
\newcommand{\MT}{{MT}}
\newcommand{\NS}{{Nega\-Scout}}
\newcommand{\PVS}{{PVS}}

\newcommand{\MTD}{{MTD}}
\newcommand{\SSS}{{SSS*}}
\newcommand{\DUAL}{{DUAL*}}
\newcommand{\AB}{{Alpha-Beta}}
\newcommand{\MTDf}{{MTD$(f)$}}
\newcommand{\AspNS}{{Aspiration Nega\-Scout}}

\newcommand{\MTDp}{{MTD$(+\infty)$}}
\newcommand{\MTDm}{{MTD$(-\infty)$}}
\newcommand{\MTDs}{{MTD(step)}}
\newcommand{\MTDb}{{MTD(bi)}}
\newcommand{\MTDd}{{MTD(best)}}

\makeindex

\title{\sf\Huge\bf\vbox{
{\mbox{RESEARCH}\\ \mbox{RE:\,SEARCH \&}\\ \mbox{RE-SEARCH}}
}}
\author{Aske Plaat}

\begin{document}
\thispagestyle{empty}

\ \vspace{3cm}

\begin{center}
{\sf\bf\Huge \vbox{\mbox{RESEARCH}\\
\mbox{RE:\,SEARCH \&}\\
\mbox{RE-SEARCH}}}

\vspace{2cm}
{\sf\large Aske Plaat}
\end{center}
\newpage

\setcounter{page}{1}

%\evensidemargin 74pt
%\maketitle
%\evensidemargin 36pt
\pagestyle{empty}
\thispagestyle{empty}

\ \vspace{3cm}

\begin{center}
{\sf\bf\Huge \vbox{\mbox{RESEARCH}\\
\mbox{RE:\,SEARCH \&}\\
\mbox{RE-SEARCH}}}
\end{center}
\newpage

\ 

\vspace{15cm}
\noindent
This book is number 117 of the Tinbergen Institute Research Series.
This series is established through cooperation between Thesis
Publishers and the Tinbergen Institute. A list of books which already
appeared in the series can be found in the back.

 \newpage

\begin{center}
{\sf\bf\Huge \vbox{\mbox{RESEARCH}\\
\mbox{RE:\,SEARCH \&}\\
\mbox{RE-SEARCH}}}

\vspace{2cm}
{\sc
Proefschrift\\
\ \\
ter verkrijging van de graad van doctor aan de\\
Erasmus Universiteit Rotterdam op gezag van de\\
rector magnificus\\
\ \\
prof.dr. P.W.C. Akkermans M.A.\\
\ \\
en volgens besluit van het college voor promoties.\\
\ \\ 
\ \\
\ \\ 
De openbare verdediging  zal plaatsvinden op\\
donderdag 20 juni 1996 om 13.30 uur\\
\ \\
\ \\
\ \\
door\\
\ \\
\ \\
\ \\
Aske Plaat\\
geboren te Odoorn.
}

\end{center}
\newpage

\noindent{\large Promotiecommissie}
\ \\

\begin{tabbing}
mmmmmmmmmm\=  \kill 
Promotores: \> prof.dr.~A.~de~Bruin\\
            \> prof.dr.~J.~Schaeffer\\
\\
Overige leden: \> dr.~J.C.~Bioch\\
               \> prof.dr.ir.~R.~Dekker\\
               \> prof.dr.~H.J. van den Herik\\
               \> dr.~W.H.L.M.~Pijls (tevens co-promotor)
\end{tabbing}

\mbox{ \normalsize } \clearpage
% \pagestyle{empty}\setcounter{page}{0} 
% \mbox{ \normalsize } \clearpage

%\cleardoublepage
\begin{center}
{\sf\bf\Large \vbox{\mbox{RESEARCH \ RE:}\\
\mbox{SEARCH \& RE-SEARCH}}}
\mbox{}\\
Aske Plaat\\
\mbox{}\\
{\bf\uppercase{Abstract}}
\end{center}
%\section*{Abstract}
%\input abstract2
%\input ks-abstract3
Search algorithms are often categorized by their node expansion
strategy.  One option is the depth-first strategy, a simple
backtracking strategy that traverses the search space in the order
in which successor nodes are generated.  An alternative is the best-first
strategy, which was designed to make it possible to use
domain-specific heuristic information.  By exploring promising parts
of the search space first, best-first algorithms are usually more
efficient than depth-first algorithms.

In programs that play minimax games such as chess and checkers, the
efficiency of the search is of crucial importance.
Given the success of best-first algorithms in other domains, one would
expect them to be used for minimax games too.
However, all high-performance game-playing
programs are based on a depth-first algorithm.

This study takes a closer look at a depth-first
algorithm, {\AB}, and a best-first algorithm, {\SSS}.  
The prevailing opinion on these algorithms is that {\SSS} offers the
potential for a more efficient search, but that its complicated
formulation and exponential memory requirements render it impractical.
The theoretical part of this work shows that there is a
surprisingly straightforward link between the two algorithms---for all
practical purposes, {\SSS} is a special case of {\AB}.  
Subsequent empirical evidence proves the prevailing opinion on {\SSS}
to be wrong: it is not a complicated algorithm, it does not need too
much memory, and it is also {\em not\/} more efficient than
depth-first search. 

Over the years, research on {\AB} has yielded many enhancements, such
as transposition tables and minimal windows with re-searches, that are
responsible for the success of depth-first minimax search.  
The enhancements have made it possible to use a depth-first procedure
to expand nodes in a best-first sequence.
Based on these insights, a new algorithm is presented, {\MTDf}, which
out-performs both {\SSS} and {\NS}, the {\AB} variant of choice by
practitioners. 

In addition to best-first search, other ways for improvement
of minimax search algorithms are investigated. 
The tree searched in {\AB}'s best case is usually considered to be
equal to the minimal tree that has to be searched by any algorithm in
order to find and prove the minimax value.
We show that in practice this assumption is not valid. 
For non-uniform trees, the real minimal tree---or rather, graph---that
proves the minimax value is shown to  be
significantly smaller than {\AB}'s best case.  Thus, there is more
room for improvement of full-width minimax search than is generally
assumed.

\pagestyle{empty}
\cleardoublepage
 \chapter*{Preface}
 \markboth{Preface}{Preface}
\thispagestyle{empty}
 I would like to thank my advisors
% The research described in this thesis has been performed in close
% co-operation with three people.
% This work is the result of
% the close co-operation of a number of people. 
Arie de Bruin, Wim Pijls, and Jonathan Schaeffer, for being just as
eager as I to understand minimax trees. 
We have spent countless hours discussing trees and algorithms, trying
to find the hidden essence. 
% I have often been amazed at how naturally new ideas were born in these
% sessions. 
Their experience with minimax search, as well as their 
different backgrounds, were of great value.
They know how much this research owes
to their willingness to spend so much time 
discussing ideas, combining theory and practice,
trying to understand trees. 
It has been
a great time, for which I thank you three. 

This work started in May 1993 in Rotterdam, the Netherlands,
with Arie and Wim, at the department of Computer Science of the
Erasmus University. 
I thank my colleagues then at the department of Computer Science 
for creating an interesting environment to work in.
I thank Jan van den Berg for helping me experience how enjoyable
research can be, which made me decide to pursue a PhD.
It has been good talking to Roel van der Goot on subjects ranging from
parsing to politics. 
Maarten van Steen answered many
questions on parallel computing. Harry Trienekens and Gerard Kindervater
know much about parallel branch and bound as well as operating systems.
Reino de Boer knows \LaTeX\ inside out. 

From October 1994 to October 1995 I spent most of my time at the
University of Alberta in Edmonton,
Canada, with Jonathan Schaeffer. Working together with someone with so
much experience and energy, and so many ideas, has been an
extraordinary experience.  Jonathan, I thank you for giving me the
opportunity to do research with you and to learn from you; but most of
all, I thank you for good times.  %I hope that one day I can repay you
%for all you did for me. 
In Edmonton many people did their best to make my stay as pleasant as
possible. Jonathan and Stephanie Schaeffer made sure that
I felt at home, helping me wherever and whenever they could. 
I thank the people of the 
department of Computing Science of the University of Alberta for
creating a good working environment. 
I have had many discussions on minimax search with  Mark Brockington, 
author of Keyano, one of the top Othello programs. Yngvi Bj\"ornsson and
Andreas Junghanns, who started building a new chess program, were
always keen to discuss new ideas too. Rob Lake answered my
troff questions. 
I thank all of the members of the Games research group for making both
the scheduled and the informal meetings  interesting and enjoyable.

A number of people took the time to comment on or discuss various
parts of this research. I thank Victor Allis, Don Beal, Hans
Berliner, Murray Campbell, Rainer Feldmann, 
Jaap van den Herik, Hermann Kaindl, Laveen Kanal, Rich Korf,
Patrick van der Laag, Tony Marsland, Chris McConnell, Alexander
Reinefeld, and especially George Stockman, the inventor of {\SSS}, for 
their helpful comments. 

I thank Victor Allis for teaching a class at the Free University in
Amsterdam, in September 1993, in which he explained the concept of the
minimal tree in a way that (to me) was strangely reminiscent to
solution trees, planting a seed of the {\SSS}--{\AB} relation. 

In addition, research meetings with Victor Allis, Henri Bal, Dennis
Breuker, Arie de Bruin, Dick Grune, Jaap van den Herik, Gerard
Kindervater, Wim Pijls, John Romein, Harry Trienekens, and Jos
Uiterwijk on search, combinatorial optimization and parallelism
provided a valuable broader perspective for this work.   

I thank 
%Cor Bioch,
Mark Brockington, Arie de Bruin, 
Jaap van den Herik, Andreas Junghanns,  Wim Pijls,
Jonathan Schaeffer, and Manuela Sch\"one for their careful and
detailed reading of drafts of this thesis. Their many comments and
suggestions have greatly enhanced its quality.  
Furthermore, Peter van Beek and Rommert Dekker made valuable
suggestions for improvement of parts of this work. Thanks to Gerard
van Ewijk for reporting an especially nasty bug.

Through all of this, Saskia Kaaks has been a great friend. I thank you
for everything, with love.\\

\noindent %Aske Plaat\\
{\em Rotterdam,\\
March 31, 1996.}

\pagestyle{empty}
\cleardoublepage
\markboth{}{}
\thispagestyle{empty}
\tableofcontents
\thispagestyle{empty}
\listoffigures
\thispagestyle{empty}

% uncomment this for double side printing
% comment it four quadruple sided printing

\cleardoublepage
\pagestyle{fancyplain}
\renewcommand{\chaptermark}[1]{\markboth{#1}{}}
\renewcommand{\sectionmark}[1]{\markright{\thesection\ #1}}
\lhead[\fancyplain{}{\sf\thepage}]{\fancyplain{}{\small\sf\sl\rightmark}}
\rhead[\fancyplain{}{\small\sf\sl\leftmark}]{\fancyplain{}{\sf\thepage}}
%\cfoot{\fancyplain{\sf\thepage}{}}
\cfoot{\fancyplain{}{}}

\bibliographystyle{plain}

\cleardoublepage
\setcounter{page}{1}
\chapter{Introduction}
% \markboth{Chapter 1}{Chapter 1}
%\markboth{Chapter 1}{Introduction}
% 
% 
% 
% 1 intro: sell the stuff
% 
% 
% 
% 
% make the reader think that this is exciting stuff, and that it's a
% shame he doesn't have the time to read it all.
% 
% 
% *** WHY BOTHER? *** (motivation)
% 
% game playing is really interesting and central to all of CS
% searching deeper is good for you
% 
\section{Games}
Game playing \index{game playing}%
is one of
the classic problems of artificial intelligence.
\index{artificial intelligence}%
The idea of creating 
a machine that can beat humans at chess 
\index{chess}%
has a continued fascination
for many people.  
Ever since computers were built, people have tried
to create game-playing programs. 

Over the years researchers have put
much  effort into improving the quality of play. 
They have been quite successful. Already, non-trivial games like
nine-men's-morris, connect-four 
and qubic are solved; 
\index{nine-men's-morris}\index{connect-four}\index{qubic}%
they have been searched to the end and their 
game-theoretic value has been determined 
\cite{Alli94b,Gass94}. In backgammon \index{backgammon}%
the BKG
\index{BKG}%
program received some
fame in 1979 for defeating the World Champion \index{world champion}%
in a short match
\cite{Berl80a,Berl80b}. Today, computer
programs are among the strongest backgammon players \cite{Tesa95}.  In
Othello  \index{Othello}%
computer programs are  widely considered to be superior  to the
strongest human players \cite{Lee90}, the  best program of the moment
being Logistello \index{Logistello}%
\cite{Buro95a}. In checkers
\index{checkers}%
the current World 
Champion is the program Chinook \index{Chinook}%
\cite{Scha92,Scha95}. In chess,
computer programs play better than all but the strongest human players
and have occasionally beaten players such as the World Champion in
isolated  games \cite{Frie94b,Frie94,Uite96}. Indeed, many believe that it is
only a matter of time until a computer defeats the human world
champion in a match \cite{Heri82,Hsu90b,Uite96}.
%Commenting on the playing style of IBM's chess machine
%Deep Blue\index{Deep Blue} in a match in February 1996, PCA World
%Champion Garry 
%Kasparov\index{Kasparov} made the following remark: ``Just a few more
%positions per second, reaching somewhat more deeply, might make the
%machine beat me.'' \cite{Uite96}. 

For other games,
such as bridge, go, shogi, and chinese chess,
\index{bridge}\index{go}\index{shogi}\index{chinese chess}%
humans
outplay existing computer programs by a wide margin. Current research
has not been able to show that the successes can be repeated for these
games. An overview of some of this research can be found in a study by
Allis \index{Allis}%
\cite{Alli94b}. 
% Allis has studied which games are solved,
% which could be solved in the near future, in which games computers may
% beat the human world champion, and
% which will survive \cite{Alli94b}. 

Computer game playing \index{game playing!computer game playing}%
has a natural
appeal to both amateurs and 
professionals, both of an intellectual and a competitive
nature. Seen from a scientific perspective, games provide a
conveniently closed 
domain, that reflects  many 
real-world properties, making it well-suited for experiments with
new ideas in problem solving. According to Fraenkel \index{Fraenkel}%
\cite{Frae95},
applications of the 
study of two-player games include areas within 
mathematics, computer science and 
economics \index{mathematics}\index{computer science}\index{economics}%
such as game
theory, combinatorial games, complexity theory, 
combinatorial optimization, logic 
programming, theorem proving,
constraint satisfaction,  parsing, pattern recognition, connectionist 
networks and parallel computing.\index{game theory}\index{combinatorial games}\index{complexity theory}\index{combinatorial optimization} \index{logic programming}\index{theorem proving}\index{constraint satisfaction}\index{parsing}\index{pattern recognition}\index{connectionist networks}\index{parallel computing}
See for example the
bibliographies \index{bibliograhy}%
by Stewart, Liaw and White
\index{Stewart}%
\cite{Stew92}, 
by Fraenkel \index{Fraenkel}%
\cite{Frae95}  and Levy's \index{Levy}%
computer chess compendium 
\cite{Levy88}.  With this long list of applications
\index{game playing!applications of}%
one would expect
results in game playing to have a profound influence on artificial
intelligence and operations research. For a number of 
reasons, not all scientific, this is not the case \cite{Dons89}. Ideas
generated in game playing find their way into main stream research at
a slow pace.

\section{Minimax Search} \index{search!minimax search}\index{minimax search}%
Of central importance in most game-playing programs is the search
algorithm. In trying to find the move to make, a human player would
typically try
to look ahead a few moves, predicting the replies of the opponent to
each move (and the responses to these replies, and so on) and select the
move that looks most promising. In other 
words, the space of possible moves is searched trying to find the best
line of 
play. Game-playing programs 
mimic this behavior. They search each line of play to a certain depth
and evaluate the position. 
Assuming that both players choose
the move with the highest probability of winning for them, in each position the
value of the best  move is returned to the parent position. Player A
tries to maximize the chance of winning the game; player B tries to
maximize B's chance, which is equivalent to 
minimizing A's chances. Therefore, the  process of backing up the value
of the best 
move for alternating sides is called minimaxing; two-player search
algorithms\index{two-player search}\index{two-agent search}\index{game playing!minimax}
are said
to perform a {\em minimax search}. 

Searching deeper generally improves the quality of the decision
\index{decision quality}%
\cite{Hsu90,Scha93,Thom82}.
Quite a number of researchers have studied minimax search
algorithms to improve their efficiency, effectively allowing them to
search more deeply within a given real-time constraint. 
In game playing a move typically has to be made every few
minutes. For many games this constraint is too tight to 
allow optimizing 
strategies---a  look-ahead search to the end of the game is infeasible.
Minimax search is a typical example of satisficing, \index{satisficing}%
heuristic search.  
Game-playing programs are real-time systems where the
utility of actions is strongly time-dependent. The recent interest in
anytime algorithms acknowledges \index{anytime algorithm}%
the practical importance of this class of decision making
agents \cite{Dean88,Zilb95}.    
% There have been quite a number of researchers  
% Trying to find ways to increase the depth of the search, .

Real-time search \index{search!real-time}%
is central to many
areas of computer science, so 
it is fortunate that a number of the ideas created in
minimax search have proved useful in other search domains. For 
example,
% single-agent search as well. 
% Examples are 
iterative deepening (IDA*) \index{iterative deepening}\index{IDA*}%
\cite{Korf85}, transposition tables \index{transposition table}%
\cite{Kain95,Mars94}, pattern databases \cite{Culb93}, real time search
(RTA*) \index{RTA*}%
\cite{Korf90} and 
bi-directional  search \index{bi-directional search}%
(BIDA*) \index{BIDA*}%
\cite{Kwa89,Manzini95} have been
applied in both two-agent and single-agent search.
These are all examples of a successful technology transfer from game playing
to other domains, supporting the view of games as a fertile  environment
for new ideas. However, the length of Fraenkel's \cite{Frae95}\index{Fraenkel}
list of applications
of game playing suggests that
there ought to be more examples.

% WHAT'S IT ALL ABOUT? (problem statement, general setting)
% 
% finding better minimax algs 
% 

\subsection{A Tale of Two Questions}
The main theme of the research behind this thesis has been to find
ways to improve the performance of algorithms that search for the
minimax value of a (depth-limited) tree in real time, through a better
understanding 
of the search trees that these algorithms generate. 
This goal is pursued along two lines: by looking into the
best-first/depth-first issue, and by examining the concept of the
minimal tree. 

\index{Alpha-Beta}%
\index{depth-first|see{search, depth-first}}%
\index{best-first|see{search, best-first}}%
\index{search!depth-first}%
\index{search!best-first}%
\subsubsection{Best-First and Depth-First}
Most successful game-playing programs are based on the {\AB}
algorithm, a simple recursive depth-first minimax
procedure  invented in the late 1950's. 
In its basic form a
depth-first algorithm traverses nodes using a rigid left-to-right
expansion sequence \cite{Pear84}.
% , decades of research have been
% unable to replace {\AB} as the basis for search algorithms of most
% succesful 
% game-playing programs. 
There is an exponential gap in the size of trees built by
best-case and worst-case {\AB}. 
This led to numerous enhancements to the basic algorithm, including
\index{Alpha-Beta!enhancements}\index{enhancements}%
iterative  deepening, transposition tables, the history heuristic,
and narrow search windows (see for example \cite{Scha89b} for an
assessment). \index{iterative deepening}\index{transposition table}%
\index{history heuristic}\index{narrow search window}%
These enhancements have improved the performance of depth-first
minimax search considerably.

An alternative to depth-first search is best-first
search. This expansion strategy makes use of extra heuristic
information, which is used to select nodes that appear more
promising first, 
increasing the likelihood of reaching the goal sooner.
Although best-first approaches have been successful in other search
domains, 
minimax search in practice has been almost exclusively based on
depth-first strategies. (One could argue that the enhancements to
{\AB} have transformed it into best-first
search. However, the designation ``best-first'' is normally reserved for
algorithms that deviate from {\AB}'s left-to-right strategy.)

In 1979 {\SSS} was introduced, a best-first algorithm
for searching AND/OR trees \cite{Stoc79}. 
\index{SSS*}\index{AND/OR tree}%
For minimax trees, {\SSS} was
proved never to 
build larger trees than  
{\AB}, and simulations \index{simulation}%
showed that it had the potential to build
significantly smaller trees. 
% Thus it looked like in minimax too, as
% had happened
% in single-agent search,
% simple depth-first search would be replaced by more advanced schemes.
% ,
% although a few years later {\AB} was proven to be asymptotical
% optimal \cite{Pear80,Pear82}.  
% 
% This at first sight paradoxical situation, has
% 
% {\SSS}  has been analyzed by many
% researchers since its introduction.
Since its introduction, many  
researchers have analyzed {\SSS}.  Understanding this complicated
algorithm turned out to be challenging. The premier
artificial intelligence  journal, {\em Artificial Intelligence}, 
\index{artificial intelligence journal}%
has published six articles in which 
{\SSS} plays a major role
\cite{Camp83,Ibar86,Kuma83,Mars87,Rein94b,Roiz83}, in addition to 
Stockman's original publication \cite{Stoc79}, and our own forthcoming
paper \cite{Plaa95b}.   
Most authors conclude that {\SSS} does indeed evaluate less nodes, but
that its complex formulation and exponential memory requirements are
serious problems for practical use.
% The complex formulation
% of {\SSS} is another drawback for practical use, since it complicates
% the use of search enhancements.  
% Best-first approaches were more complex and reportedly required more memory,
% both being serious impediments to their general acceptance.
% % {\SSS}, a best-first algorithm, will provably never build larger trees
% than {\AB} and generally builds significantly smaller trees
% \cite{Camp81,Kain91,Mars82a,Mars87,Rein89,Stoc79}.
Despite the potential, the algorithm remains largely ignored in
practice. However, the fact that it searches smaller trees casts doubts on the
effectiveness of {\AB}-based approaches, stimulating the search for
alternatives. 

This brings us to the first theme of our research: the relation between
{\SSS} and {\AB}, between best-first and depth-first. Simply put, in
this work we
try to find out which is best.

\subsubsection{The Minimal Tree} \index{minimal tree}%
The second issue concerns the notion of the minimal tree. 
In a seminal paper in 1975 Knuth and Moore \index{Knuth and Moore}%
proved that to find the 
minimax value, any algorithm has to search at least the 
{\em minimal tree\/} \cite{Knut75}. The concept of the minimal tree has
had a profound 
impact on our understanding of minimax algorithms and the structure of
their search trees. 

The minimal tree can be used as a standard for algorithm performance.
\index{performance standard}%
Since
any algorithm has to 
expand at least the minimal tree, it is a limit on the
performance of all 
minimax algorithms. Many authors of game-playing programs use it as a
standard benchmark \index{performance benchmark}%
for the quality of their algorithms.
The size of the minimal tree is defined for uniform
trees. For use with real minimax trees, with
irregular branching factor and depth, the minimal tree is usually
defined as {\AB}'s best case.  However, this is not necessarily the
smallest possible tree
that defines the minimax value. Due to transpositions
\index{transposition}\index{branching factor}%
and the
irregularity of the branching factor it might be significantly
smaller. 

The second theme of this research is to find out by how
much. 
A minimal tree that is significantly smaller would have two
consequences. First, it would mean that the relative performance of
game-playing programs has dropped. Second, and more important, it would show
where possible improvements of minimax algorithms could be found.

% \\
% 
% The first question is concerned with how algorithms try to construct
% the same minimal tree as efficiently as possible.
% The second question is concerned with showing that their minimal
% tree is not the smallest one possible. By showing that
% smaller ones  exist, the standard is raised.

\section{Contributions}
\subsection{Best-First and Depth-First}
The first main result of our research is that {\SSS} can be
reformulated as a special case of {\AB}. Given that best-first {\SSS}
is normally portrayed as an entirely different approach to minimax
search than depth-first {\AB}, this is a surprising result.
% This paper presents the surprising result that best-first {\SSS}
% can be reformulated as a special case of depth-first {\AB}. 
{\SSS} is now easily implemented in existing {\AB}-based
game-playing programs. The reformulation solves all of the perceived
drawbacks of the 
{\SSS} algorithm.  Experiments conducted with three tournament-quality
game-playing 
programs show that in practice {\SSS} requires as much memory as {\AB}.
When given identical memory resources,
{\SSS} does not evaluate significantly less nodes than {\AB}.
It is typically out-performed by {\NS} \cite{Fish81,Pear84,Rein85},
the current depth-first {\AB} variant of choice.  
In effect, the reasons for ignoring {\SSS} have been eliminated,
but the reasons for using it are gone too.

\index{Alpha-Beta}\index{SSS*}\index{NegaScout}\index{C*}%
\index{search!best-first}\index{search!depth-first}%
\index{reformulation}%

The ideas at the basis of the {\SSS} reformulation are generalized to
create a framework for best-first fixed-depth minimax search that is based on
depth-first null-window {\AB} calls.  The framework is called {\MT},
for Memory-enhanced Test (see page~\pageref{Pearl}).  
A number of existing algorithms,
including {\SSS}, {\DUAL} \cite{Kuma84,Mars87,Rein89} and C*
\cite{Copl82}, are special cases of this  
framework.  In addition to reformulations, we introduce new instances
of this framework. One of the instances, called {\MTDf}, out-performs
all other  
minimax search algorithms that we tested, both on tree size and on
execution time.

In the new formulation, 
{\SSS} is equivalent to a special case of {\AB}; tests show that it
is out-performed by other {\AB} variants.
%  (both best-first and depth-first).
In light of this, we believe that {\SSS} should from now on be
regarded as a footnote in the history of game-tree search.
\index{SSS*!footnote}%

The  results  contradict the prevailing view in the
literature on 
{\AB} and {\SSS}.  
How can it be that the conclusions in the literature are so 
different from what we see in practice? Probably due to the complex
original 
formulation of {\SSS}, previous work mainly used theoretical analyses
and simulations 
to predict the performance of {\SSS}. However, there are
many differences between simulated algorithms and trees, and those
found in practice. Algorithmic enhancements have improved
performance considerably. 
Artificial trees lack essential properties
of trees as they are searched by actual
game-playing programs. Given the fact
that there are numerous high-quality game-playing
programs available,  
there is no valid reason to use  simulations for performance
assessments of minimax algorithms. Their 
unreliability can lead to conclusions that are the opposite of
what is seen in practice.

\subsection{The Minimal Tree}
Concerning the second issue, we have found that 
in practice the minimal tree can indeed be improved upon.
The irregular branching factor and transpositions mean that the {\em
real\/} minimal tree (which 
should really be called minimal {\em graph\/}) is significantly
smaller. Our approximations  show the difference to be at least a factor of
1.25 for 
chess to  2 for checkers. This means that there
is more room for improvement of minimax algorithms than is generally
assumed \cite{Ebel87,Feldmann93,Schaeffer86}. 
\index{minimal tree}\index{minimal graph}\index{ETC}%

We present one such improvement, called Enhanced Transposition Cutoff
(ETC), a simple way to make 
better use of transpositions that are found during a search.

\subsection{List of Contributions}
The contributions of this research can be summarized as follows:
\begin{itemize}
\item {\em {\SSS} = $\alpha$-$\beta$ + transposition tables}\\
  The obstacles to efficient {\SSS} implementations have been solved,
  making the algorithm a practical  {\AB} variant. 
  By reformulating the algorithm,
  {\SSS} can be expressed simply and intuitively as a series of
  null-window calls
  to {\AB} with a transposition table (TT), yielding a new formulation called
  {\ABSSS}. 
  {\ABSSS} does not need an expensive OPEN list; a familiar
  \index{OPEN list}%
  transposition table performs as well. In effect:
 % In effect, {\SSS} can be reformulated to use well-known technology,
 % as a special case of the {\AB} procedure enhanced with transposition
 % tables: 
 {\SSS} = $\alpha$-$\beta$ + TT.

\item {\em A framework for best-first minimax search based on depth-first
search}\\
Inspired by the {\ABSSS} reformulation, a new framework for
minimax search is
  introduced. It is based on the procedure {\MT}, which is a memory-enhanced
  version of Pearl's 
  {\Test} procedure \cite{Pear84}, also known as null-window {\AB} search. 
  We present a simple framework of {\MT} drivers ({\MTD}) that make
  repeated calls to {\MT} to home in on the minimax value. Search
  results from previous passes are stored in memory and re-used.
  {\MTD} can be used to construct a variety of best-first
  search algorithms  using depth-first search. Since {\MT} can be
  implemented using {\AB} with transposition tables, the instances of
  this framework are readily incorporated into existing game-playing
  programs. 

\item {\em Depth-first search can out-perform best-first}\\
Using our new framework, we were able to compare the performance
  of a number of best-first algorithms to some well-known depth-first
  algorithms. Three high performance game-playing programs were used to
ensure the generality and reliability of the outcome. The
  results of these 
  experiments were quite surprising, since they contradict the large
  body of published results based on simulations. Best-first searches
  and depth-first searches are roughly
 comparable in performance, with \mbox{\NS}, a
depth-first algorithm, out-performing 
  {\SSS}, a best-first algorithm.

  In previously published experimental results, depth-first and
best-first minimax search algorithms 
  were allowed different memory requirements. To our knowledge,
  we present the first
  experiments that compare them using {\em identical storage\/}
  requirements. 

\item {\em Real versus artificial trees}\\
  In analyzing why our results differ from simulations, we identify a
  number of differences between real and artificially-gen\-er\-at\-ed game
  trees. Two important factors are transpositions and value
  interdependence between parent and child nodes. In game-playing
  programs these factors are
  commonly exploited by transposition tables and iterative deepening
  to yield  \index{simulation}%
  large performance gains---making it possible that depth-first
  al\-go\-rithms  out-perform best-first. Given that most simulations
  neglect to include important properties of trees built in practice,
  of what value are the previously published simulation results?
%We believe that
%  it will be very hard to construct a reliable simulation model of the
%  multitude of interconnected phenomena occurring in real game trees.  

\item {\em Memory size}\\
In the {\MT} framework the essential part of the search tree is
formed by a max and/or a min solution tree of size $O(w^{d/2})$ for
trees with branching factor $w$ and depth $d$. Our experiments show
that for game-playing programs under tournament conditions, these
trees fit in memory without problems.  \index{memory}%
The reason that the exponential space complexity is not a problem
under tournament conditions is that the time complexity of a search
is at least  $O(w^{d/2})$; time runs out before the memory is
exhausted. 

\item {\em Domination and dynamic move re-ordering}\\
  With dynamic move reordering schemes like iterative
  deepening,
  {\SSS} and its dual {\DUAL} \cite{Kuma83,Rein89} are no longer
guaranteed to expand fewer  
    leaf nodes  than {\AB}. 
\index{domination}\index{surpassing}\index{Stockman}%
  The conditions for Stockman's proof \cite{Stoc79} are not met in
  practice.

\item {\em {\MTDf}}\\ 
  We formulate a new algorithm, {\MTDf}. It out-performs our best
  {\AB} variant, {\NS},  on leaf nodes, \index{MTD$(f)$}%
  total nodes, and execution time for our test-programs. The
improvement was bigger 
than the improvement of {\NS} over {\AB}. Since
  {\MTDf} is an instance of the {\MT} framework, \index{MT framework}%
  it is easily implemented in existing programs: just add one loop to
  an {\AB}-based program.

\item {\em The real minimal graph}\\ 
The minimal tree is a limit on the performance of minimax algorithms.
For search spaces with transpositions and irregular $w$ and $d$, most
researchers redefine the minimal ``tree'' as the minimal graph
searched by {\AB}. However, this is not the smallest graph that proves
the minimax \index{real minimal graph}\index{minimal graph}%
value. Because of transpositions and variances in $w$ and $d$,
cutoffs may exist that are cheaper to
compute. The size of the real minimal graph is shown to be
significantly smaller. This implies that algorithms are not as close
to the optimum, leaving more room for improvement than is generally
assumed.

\item {\em ETC}\\
We introduce a technique to take better advantage of available
transpositions. The technique is called Enhanced Transposition Cutoff
(ETC). It \index{ETC}%
reduces the search tree size for checkers and chess between  20\%--30\%,
while incuring only a small computational overhead.
% at
% a small cost.

\item {\em Solution trees}\\
Theoretical
abstractions of actual search trees have been useful in building models
to reason about the behavior of various minimax algorithms, explaining
experimental evidence and facilitating the construction of a new
framework. \index{solution tree}%
The key abstractions have been the dual notions of the max and min
solution tree, defining an upper and a lower bound on the minimax value
of a tree.

\end{itemize}
% For the minimax theoretician, 
% there is positive news, but also reason for concern. 
% On the other hand, this research showed that the accuracy of
% theoretical models of minimax trees is 
% limited. For performance assessments of simulated algorithms  the
% accuracy is 
% insufficient.
% This has lead to incorrect conclusions in the literature.
% Real applications should be used for performance assessments.
One of the most striking results, besides the simulation versus
practice issue, is perhaps 
% For other researchers the most interesting parts, 
% in addition to the simulation versus
% practice issue, is perhaps 
that
best-first algorithms can be expressed in a depth-first
framework. (In contrast to the IDA*/A* case \cite{Korf85},
\index{IDA*}\index{A*}%
the algorithms in this work are
reformulations of the best-first originals. They evaluate exactly the same
nodes and
the space complexity does not change.) \index{memory}%
Furthermore, it is surprising  that despite the exponential storage
requirements of  best-first algorithms  their memory consumption
does not render them impractical.
% \section{Summary}
% 
% 
% In doing so we have found that {\SSS}, a best-first full-width minimax
% algorithm, can be expressed using depth-first searches. 
% To put it differently, {\SSS} is another {\AB} enhancement, like {\NS}
% (where the 
% latter is more effective in practice). Tests show that
% in practice {\SSS} may build trees that are bigger than {\AB},
% contradicting the literature.
% 
% Using ideas behind the {\AB} {\SSS}
% reformulation, we have created {\MTDf}. In practical tests this
% algorithm out-performs all other tested algorithms.
% Given that the new formulation for best-first full-width  minimax
% search is more general, clearer, and performs better, we believe that
% {\SSS} should become 
% a footnote in the history of minimax search.
% % 
% % we think that
% % the old {\SSS} is history.
% 
% The tests showed that simulating the performance of minimax algorithms
% is a hazardous business, that can easily lead to wrong conclusions. 
% 
% In practice the minimal tree, defined as best-case {\AB},  is not the
% smallest proof of  the minimax value. We show that the real minimal
% graph must be significantly smaller. There is more room for
% improvement of minimax algorithms than is generally assumed.

\section{Overview} 
This section gives a short overview of the rest of the thesis.
% showing where 
% the preceding topics can be found in the thesis.
% are treated in the following chapters.
Chapter~\ref{chap:litt} provides some background on minimax
algorithms. {\AB} and {\SSS} are discussed.
The chapter  includes an
explanation 
of bounds, solution trees, the minimal tree, and narrow-window calls,
notions that form the basis of the next
chapters.

Chapter~\ref{chap:mt} introduces {\MT}, a framework for
best-first full-width minimax search, based on null-window {\AB}
search enhanced with memory. 
The main benefit of the 
framework is its simplicity, which makes it possible to use a number of
real applications to test the instances. 

Chapter~\ref{chap:exper} reports on results of  tests with
{\AB}, \mbox{\NS}, and {\MT} instances. We have tested
both performance and 
storage requirements of the algorithms, using three tournament
game-playing programs.
As stated, the test results paint an entirely different picture
of {\AB}, {\NS}, and {\SSS} than the literature does.
Section~\ref{sec:simsucks} discusses the reason: differences between real and
artificial search trees.  

% Chapter~\ref{chap:anal} goes deeper into explaining the test
% results. 
% For example, {\MTDf}  out-performed all
% other tested algorithms. Section~\ref{sec:bestmtdf} explains some
% reasons for {\MTDf}'s performance. 
Chapter~\ref{chap:bestall} 
discusses the minimal tree in the light of the test results. This
chapter discusses
how the size of a more realistic minimal tree can be computed.
% Section~\ref{sec:deeper} uses theoretical notions to provide an
% explanation for a number of phenomena that occurred in the
% experiments. 

Chapter~\ref{chap:concl} presents the conclusions of this work.
% Chapter~\ref{chap:summ} contains a summary.

In appendix~\ref{app:ex} we give examples of how {\AB}, {\SSS},
and {\ABSSS} traverse a
tree. 
% Appendix~\ref{app:sssex} shows how {\SSS} traverses
% it. Appendix~\ref{app:absssex} shows how our reformulation of {\SSS}
% traverses the tree.

Appendix~\ref{app:equiv} is concerned with the equivalence of {\SSS}
and {\ABSSS}. It contains a detailed technical treatment
indicating why our reformulation is equivalent to the original.

% Appendix~\ref{app:oddeven} contains a detailed analysis of an odd/even
% oscillation of the move ordering as seen in figure~\ref{fig:mvord}.

Appendix~\ref{app:testpos} lists test positions that were used for the
experiments as well as  some numerical results of these tests.

The ideas of chapter~\ref{chap:mt} and \ref{chap:exper} have
appeared in \cite{Plaa95a,Plaa95b}, and also in  
\cite{Plaa94a,Plaa94b,Plaa95d,Plaa95e,Scha96b}. 
The ideas on solution trees in chapter~\ref{chap:litt} have appeared
in \cite{DeBruin94,Icca94}. The work on the real minimal
graph in chapter~\ref{chap:bestall} has appeared in \cite{Plaa96b},
and also in \cite{Plaa94c,Scha96b}.

\cleardoublepage
\chapter{Background---Minimax Search}% : Fixed Depth Minimax Algorithms}
\label{chap:litt}
% \markboth{Chapter 2}{Chapter 2}
%\markboth{Chapter 2}{Background}
% 
% 
% literature on algorithms (background):
%   - mm, ab, ns, sss, fd/var depth, mv ord, TT, asp win, search tree/tt
% 
% 2 litt background
%   - litt overview: mm, ab, ns, sss, fd/var depth, mv ord, TT, asp
%                    win, search tree/tt 
%   - the definitive AB code???
\section{Minimax Trees and {\AB}}\label{sec:ab}
% Objectives: 
% Clear and simple introduction to AB
% Explain searching game trees, cutoffs
% Give simplest AB pseudocode + example
% Might as well introduce solution trees
% Later on talk about enhancements, and alternatives to AB like SSS*,
%   and var depth like B*
%
% Outline:
% two person zero sum perfect information game like tic-tac-toe
% minimax function
% minimax tree + figure
% minimax code
% cutoffs
% search tree is subtree of minimax tree
% alpha-beta code
% alpha-beta tree, example run
% bounds
% solution trees
% crit tree, best case alpha-beta
% solution trees NWS
%
%
%
%
This chapter provides some background on minimax algorithms.
We briefly introduce the minimax function, and the concept of a
cutoff. To find the value of the minimax function, one does not have
to search the entire problem space. Some parts can be pruned; they are
said to be {\em cut off}.
The basic algorithm of minimax search, {\AB}, is discussed.
Next we discuss common enhancements to {\AB}, as well
as alternatives, notably {\SSS}. 

Other general introductions into this matter
can be found in, for example, \cite{Frey77,Nils71,Pear84}. 

\subsection{Minimax Games}
This research is concerned with (satisficing) search algorithms for
minimax games. \index{minimax games}\index{satisficing}%
In a zero-sum game \index{zero-sum game}%
the
loss of one player is the gain of the other. Seen from the viewpoint
of a single  player, one player tries to make moves that maximize the
likelihood of an outcome that is positive for this person, while the
other tries to 
minimize this likelihood. The pattern of alternating turns by the
maximizer and minimizer 
has caused these games to be called minimax games. Likewise, the 
function describing the outcome of a minimax game is called the
minimax function, denoted by $f$.

\begin{figure}
\begin{center}
\includegraphics[width=9.3cm]{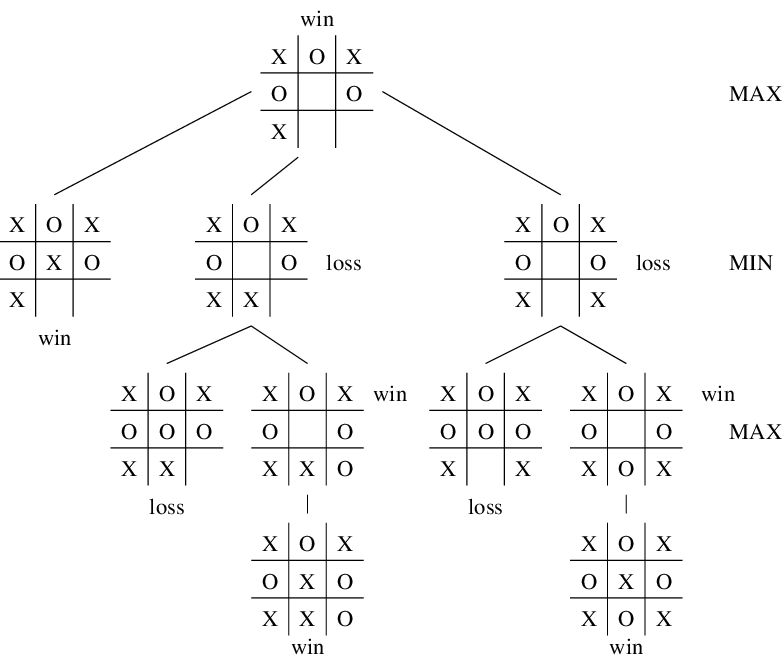}
\end{center}
\caption{Possible Lines of Play for Tic-tac-toe Position}\label{fig:ttt1}
\end{figure}

\begin{figure}
  \begin{center}
\includegraphics[width=5cm]{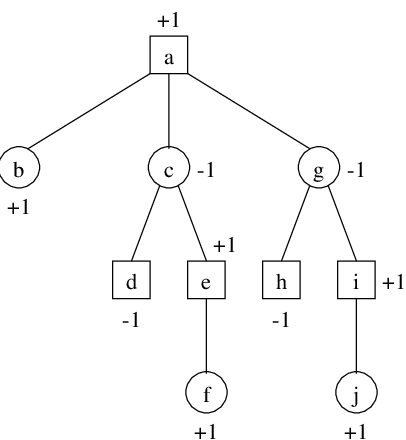}
  \end{center}
  \caption{Minimax Tree of the Possible Lines of Play}\label{fig:ttt-tree}
\end{figure}

As an example,  let us look at tic-tac-toe. \index{tic-tac-toe}%
\index{minimax function}%
Figure~\ref{fig:ttt1} gives the possible moves from a certain
position.  A  graph representation is shown in
figure~\ref{fig:ttt-tree}. This graph is called the minimax
tree. For reasons that will become clear soon, player X is called
``Max'' and player O is called ``Min.'' Board positions are
represented by square or circle nodes. 
Nodes where the Max player moves are shown
as squares, nodes where the Min player moves are shown as
circles. The possible moves from a position are represented by unlabeled
links in the graph. The node at the top represents the
actual start  
position. In the start position, Max has
three possible moves, leading to nodes $b, c$ and $g$. By considering
the options of Min in each of these nodes, and the responses by Max,
the tree in 
figure~\ref{fig:ttt-tree} is constructed.

\index{root node}%
The start position (node $a$) is called the {\em root\/} of the
tree. The other nodes 
represent the possible lines of play. Nodes $a, c, e, g$ and $i$ are 
called interior nodes. Nodes $b, d, f, h$ and $j$ are the leaf nodes of this
tree. Considering the possible lines of play to find the best one is
usually referred to as {\em searching the tree}. A move by one player
is often called a half-move or a {\em ply}. \index{ply}\index{lines of play}% 
\index{leaf node}\index{interior node}%

At the leaf nodes in the tree the game ends, and the
value of the minimax function $f$ can be determined.
A win for max is denoted by $+1$, \index{win}\index{loss}%
a  $-1$ denotes a loss. The value 0 would represent  a draw. 
% Player X (Max) will want to choose branches
% that maximize the minimax value $f$. Player O (Min) will want to
% minimize it. 
The root 
of the tree is at level 0. \index{even level}\index{odd level}%
At even levels player Max will make a move to maximize the
outcome $f$, while at odd levels player Min will move to minimize
$f$. Therefore, at even levels $f$ equals the maximum of
\index{maximum}\index{minimum}%
the value of its children and at odd levels $f$ equals the
minimum. In this way $f$ is recursively defined for all positions.
In the  \index{minimax example}%
example position its value is +1, Max wins.
  
%                       -1 
% 
%                     x o x
%                     o o                                max-to-move x
%                     x               
%    
%   x o x             x o x            x o x
%   o o x -1          o o   -1         o o   -1          min-to-move o
%   x                 x   x            x x       
%      
% o x   x o x     x o x   x o x    x o x   x o x
% o x   o o x     o o     o o o    o o o   o o           max-to-move x
% o     x   o     x o x   x   x    x x     x x o
% -1               -1      -1       -1       
%       x o x                              x o x
%       o o x                              o o x         
%       x x o                              x x o
%         0                                  0
%                       +1 
% 
%                     x o x
%                     o   o                              max-to-move x
%                     x               
%    
%   x o x             x o x            x o x
%   o x o             o   o -1         o   o -1          min-to-move o
%   x                 x   x            x x       
%     +1
%                 x o x   x o x    x o x   x o x
%                 o o o   o   o    o o o   o   o         max-to-move x
%                 x   x   x o x    x x     x x o
%                  -1                -1     
%                         x o x            x o x
%                         o x o            o x o         
%                         x o x            x x o
%                          +1                +1
% 
% 
% 
% 
\begin{figure}
{\small
\begin{tabbing}
mmmmmmmmmmmm\=mm\=mm\=mm\=mm\kill
\> {\bf function} minimax$(n) \rightarrow f$;\\
\> \> {\bf if} $n$ = leaf {\bf then return} eval$(n)$;\\
\> \> {\bf else if} $n$ = max  {\bf then}\\
\> \> \> $g := -\infty$;\\
\> \> \> $c := $ firstchild$(n)$;\\
\> \> \> {\bf while} $c \neq \bot$ {\bf do}\\
\> \> \> \> $g := $ max$\big(g,$ minimax$(c)\big)$;\\
\> \> \> \> $c := $ nextbrother$(c)$;\\
\> \> {\bf else} /* $n$ is a min node */ \\ % {\bf then}\\
\> \> \> $g := +\infty$;\\
\> \> \> $c := $ firstchild$(n)$;\\
\> \> \> {\bf while} $c \neq \bot$ {\bf do}\\
\> \> \> \> $g := $ min$\big(g,$ minimax$(c)\big)$;\\
\> \> \> \> $c := $ nextbrother$(c)$;\\
\> \> {\bf return} $g$;
\end{tabbing}
}
\caption{The Minimax Function}\label{fig:mmcode}
\end{figure}
Figure~\ref{fig:mmcode} gives the recursive minimax function in a Pascal-like
pseudo code. The code takes a node $n$ as input parameter and
returns $f_n$, the minimax value for node $n$. The code contains a number of
abstractions. First, every node is either a leaf, a min, or a max
node. Second, an evaluation \index{evaluation function}%
function, {\em eval}, exists that returns the minimax value (win,
loss, or draw) 
for each board position 
at a leaf node. 
%It is called
%{\em eval\/} in the figure. 
Third, functions \index{firstchild}\index{nextbrother}%
{\em firstchild\/} and {\em nextbrother\/} exist, returning the
child node and brothers. (They return the value $\bot$ if no child 
or brother \index{$\bot$}%
exist.)
Note that the given minimax function traverses the tree in a
depth-first \index{search!deth-first}% 
order. The min and max
operations implement the backing-up \index{back-up}%
of the scores of nodes from a deeper level. The value of $g$
represents an intermediate value of a node. When all children have
been searched $g$ becomes $f$, the final minimax value.

Strictly speaking, a game-playing program does not need to know the minimax
value of the root. All it needs is the best move. \index{best move}%
By
searching for the 
minimax value, the best move is found too. Later on (in
section~\ref{sec:mt}) % and \ref{sec:bestmove}
we will discuss
algorithms that stop searching as soon as the best move is known,
giving a more efficient search.

In tic-tac-toe, the tree is small enough to traverse all possible
paths to the end of the game within a reasonable amount of time. The
evaluation function is simple,  all it has to do is determine
whether an end position is a
win, loss, or draw. In many games  it is not
feasible to search 
all paths to the end of the game, because the complete minimax tree
would be too
big. For these games (such as chess, checkers and Othello) a different
approach is taken.  The evaluation 
function is changed to return a heuristic assessment. Now it can be
called on any position, not just where the game has ended. 
To facilitate a better discrimination between moves, the range of
return values \index{evaluation function range}%
is usually substantially wider than just
win, loss, or draw. From the perspective of the search algorithm, the
change is minor. The minimax
function does not change at all, only the evaluation function is
changed (and the 
definition of a leaf node).
% Now it has
% the task of co a part of the lookahead 
% eshould now Usually the merit of 
% child nodes is changed so that it can  related to that of the
% parent. Therefore, it is .
% possible to use an assessment of a position before the end of the game
% as an approximation of the outcome of a line of play. The algorithm
% can stop searching 
% a branch before the end of the game has
% been reached, and have the
% evaluation function  return a heuristic assessment of the
% node. 
In its simplest form, the search of the minimax tree is stopped at
a fixed depth from the root and the position is evaluated. 
% Now the evaluation
% function has to know more about the game than just a few  rules of when a
% player wins. 
% % A starting point for chess would be counting the material
% % of both sides, for Othello the number of moves.
% If adding knowledge makes an evaluation function slower, then
% there is a trade-off between search and knowledge. It has been
% observed many times that {\em removing\/} a certain piece of knowledge
% made a program play better, because the faster evaluation function
% made it possible to search deeper, which more than compensated for the
% lack of knowledge.
% The creation of a well balanced evaluation function for games
% like Chess, Checkers and Othello is an error-prone and
% time-consuming process. Some pointers to evaluation function construction
% are \cite{Slag69,Slat77,Ebel87,Schaeffer86}.
% (If no strong correlation between parent and child values exists,
% searching deeper using 
% the minimax back-up rule does not work
% very well. More on this phenomenon, called pathology, can be found in
% \cite{Nau83,Pear83}.)

Evaluation functions are necessarily application dependent. In chess
they typically include features such as material, mobility, center
\index{material}\index{mobility}\index{center control}%
\index{king safety}\index{pawn structure}\index{positional}%
control, king safety and pawn structure. In Othello the predominant
feature is mobility. Positional features are often assessed using
patterns. A heuristic evaluation function has \index{heuristic evaluation function}%
to collapse a part of the look-ahead search into a
single numeric value---a task that is much harder to do well than
recognizing a win, loss, or draw \index{win}\index{loss}\index{draw}%
at the end of the game (see  
section~\ref{sec:vardeep} for a discussion of alternatives to the
single numeric back-up value). The reason 
that the scheme works so well in 
practice, is that the value of the evaluation of a child node is
usually related to that of its parent. If no strong correlation
\index{correlation, parent-child value}\index{pathology}%
between parent and child values exists, 
searching deeper using 
the minimax back-up rule does not work
very well. More on this phenomenon, called pathology, can be found in
\cite{Nau83,Pear83}.

\subsection{The Minimal Tree}\label{sec:bounds}
This section discusses bounds, solution trees and the minimal tree, to
show why algorithms like {\AB} can cut off certain parts of the minimax tree.
\index{bound}\index{solution tree}\index{minimal tree}%

In increasing the search depth of a minimax tree, the number
of nodes grows exponentially. For a minimax tree of uniform search
\index{uniform depth}\index{uniform width}%
depth $d$ and uniform branching factor $w$ (the number of children of each
interior node), the number of leaf nodes is $w^d$. 
% For all but the
% smallest branching factors, the number of leaves dominates the number
% of interior nodes. In analyses interior nodes are usually ignored. 
The
exponential rate of growth would 
limit the search depths in game-playing programs
to small numbers. For chess, given a rate of a million positional
evaluations per second (which is relatively fast by today's
standards),  a branching factor of about $35$, 
\index{branching factor}%
no transpositions (see 
chapter~\ref{chap:anal}),  and the requirement to
compute 40 moves in two hours (the standard tournament rate),
\index{tournament conditions}%
it would mean that under tournament conditions the search
would be limited to depths of 5 or 6. (Such a machine
could evaluate on average
180 million positions for each of the 40 moves, which lies between
$35^5 \approx 53$ million and $35^6
\approx 1838$ million.) The search depth \index{search depth}%
would have been 
limited to 3 or at most 4 in the 1960's and 
1970's. A number of studies have shown a strong positive correlation between
deeper search and better play \cite{Hsu90,Scha93,Thom82}. The studies show
that a program that is \index{look ahead}%
to play chess at grand-master level has to look significantly further ahead
than 5 ply.
If it were not for the early introduction of a powerful
pruning technique that made this possible, 
game-playing programs would not have been so successful. 
\index{pruning}%

The minimax back-up rule alternates taking the maximum and the
minimum of a set of numbers. Suppose we are at the root of the tree in
figure~\ref{fig:ttt1}. The first child returns $+1$. The root is a max
node, so its $f$ can only increase by further expanding nodes, $f_{root}
\geq +1$. 
We know that the range of possible values for $f_{root}$ is limited to
$\{ -1, 0, +1\}$. Thus $+1$ is an upper bound on $f_{root}$. Now we
have $f_{root} \geq +1$ and $f_{root} \leq +1$, or $f_{root} =
+1$. The interpretation is that further search can never
change the minimax  
value of the root. The first win returned is
as good as 
any other win. In figure~\ref{fig:ttt1} only one child of the root has to be
expanded to determine the minimax value. The rest can be eliminated,
or carrying the tree analogy further, pruned. 

In the more common situation where the range of values for $f$ is much
\index{evaluation function range}%
wider, say $[-1000, +1000]$, the probability of one of the moves returning
exactly a win  of $+1000$ is small. However, the fact that the
output value of the 
maximum function never decreases can still be
exploited. In the pseudo code of the minimax function in
figure~\ref{fig:mmcode}, the  
variable $g$ is equal to the value of the highest child seen so far
(for a max node).
As return values of child nodes come in, $g$ is never decreased.
At any point 
in time $g$ represents a lower bound on the return value of the max
node, $f_{max} \geq g_{max}$.
If no child has yet returned, it is $-\infty$, a trivial lower bound.
Likewise, at min nodes $g$ is an upper bound, $f_{min} \leq
g_{min}$.

\Treestyle{%
%  \addsep{1pt}%
  \minsep{4pt}%
  \vdist{30pt}%
  \nodesize{13pt}%
 }    % smaller than default 

\begin{figure}
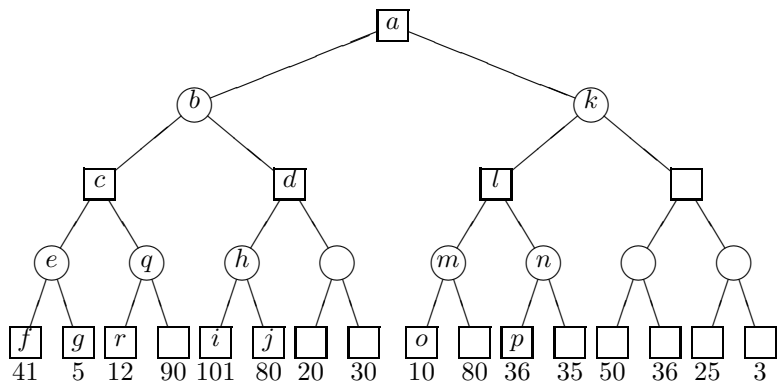

{\small
 \begin{Tree}
   \node{\external\type{square}\cntr{$f$}\bnth{$41$}}
   \node{\external\type{square}\cntr{$g$}\bnth{$5$}}
   \node{\cntr{$e$}}
   \node{\external\cntr{$r$}\type{square}\bnth{$12$}}
   \node{\external\type{square}\bnth{$90$}}
   \node{\cntr{$q$}}
   \node{\type{square}\cntr{$c$}}
   \node{\external\type{square}\cntr{$i$}\bnth{$101$}}
   \node{\external\type{square}\cntr{$j$}\bnth{$80$}}
   \node{\cntr{$h$}}
   \node{\external\type{square}\bnth{$20$}}
   \node{\external\type{square}\bnth{$30$}}
   \node{}
   \node{\type{square}\cntr{$d$}}
   \node{\cntr{$b$}}

   \node{\external\type{square}\cntr{$o$}\bnth{$10$}}
   \node{\external\type{square}\bnth{$80$}}
   \node{\cntr{$m$}}
   \node{\external\type{square}\cntr{$p$}\bnth{$36$}}
   \node{\external\type{square}\bnth{$35$}}
   \node{\cntr{$n$}}
   \node{\type{square}\cntr{$l$}}
   \node{\external\type{square}\bnth{$50$}}
   \node{\external\type{square}\bnth{$36$}}
   \node{}
   \node{\external\type{square}\bnth{$25$}}
   \node{\external\type{square}\bnth{$3$}}
   \node{}
   \node{\type{square}}
   \node{\cntr{$k$}}
   \node{\type{square}\cntr{$a$}}
\end{Tree}
\hspace{5.5cm}\usebox{\TeXTree}
}
\caption{Example Tree for Bounds}\label{fig:mmtree}
\end{figure}

\Treestyle{%
  \addsep{2pt}%
  \minsep{10pt}%
  \vdist{30pt}%
  \nodesize{14pt}%
 }    % smaller than default 

\index{solution tree example}%
\index{bound tree example}%
Up to now we have looked at single nodes. Things get more interesting
if we try to extend the concept of bounds to the rest of the minimax
tree.  For this we will use a bigger example
tree, as shown in figure~\ref{fig:mmtree}. 
%Max nodes are
%depicted by squares and min nodes by circles. 
To show a
sufficiently interesting depth, the branching factor (or width) of the
nodes has been 
restricted to 2. (The tree is based on one in \cite{Pear84}. It is
also used in the {\AB} and {\SSS} examples in the appendices. To
illustrate the concept 
of a deep cutoff in {\AB}, the value of node $o$ has been changed to
10.)\index{lower bound}\index{upper bound}

\begin{figure}
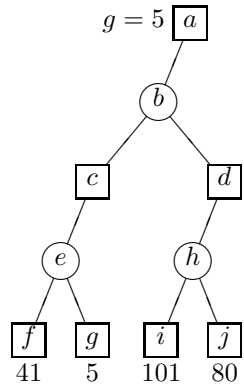

{\small
 \begin{Tree}
   \node{\external\type{square}\cntr{$f$}\bnth{$41$}}
   \node{\external\type{square}\cntr{$g$}\bnth{$5$}}
   \node{\cntr{$e$}}
   \node{\type{square}\cntr{$c$}\leftonly}
   \node{\external\type{square}\cntr{$i$}\bnth{$101$}}
   \node{\external\type{square}\cntr{$j$}\bnth{$80$}}
   \node{\cntr{$h$}}
   \node{\type{square}\cntr{$d$}\leftonly}
   \node{\cntr{$b$}}

   \node{\type{square}\cntr{$a$}\leftonly\lft{$g = 5$}}
\end{Tree}
\hspace{6.4cm}\usebox{\TeXTree}
}
\caption{A Tree for a Lower Bound, a Min Solution Tree}\label{fig:minsoltree}
\end{figure}

As we saw in the tic-tac-toe case, the search of a node $n$ can be stopped
as soon as an upper and lower bound of equal value for $n$
exist (where node $n$ is any node in a minimax tree, not just the one
with the same name in figure~\ref{fig:mmtree}).
%(which can be before all nodes in the minimax tree have been
%expanded). 
Therefore we will examine what part of the minimax tree has to be
searched by the minimax function to find such bounds. Suppose we wish
to find a better lower bound (other 
than $-\infty$) for
the root, node $a$. This amounts to finding non-trivial lower bounds
($f \geq g > -\infty$)
for all nodes 
in a sub-tree below $a$. At node $a$, a max node, all we need
is to have {\em one\/} child  return a lower bound $> -\infty$, to have the
variable $g_a$ change its value to $> -\infty$. Subsequent children
cannot decrease the value of $g$, so we will then have that $f_a \geq g_a >
-\infty$. 
We turn to finding a lower bound $> -\infty$ for 
this first child of node
$a$. 
Node $b$ is a min node. 
% Now the problem is not getting the value
% of $g_b$ away from $-\infty$, but to make sure it is a true lower bound
% on $f_b$. 
% At this node the minimum function causes any
% expansion of a child to lower the value of $g_b$. 
We are not sure that $g_b$
is a lower bound unless all children have been expanded. So, the values
of the children $c$ and $d$ have to be determined. These are max nodes,
and one child suffices to have their return value conform to 
$f \geq g > -\infty$. Thus, the value of
nodes $e$ and $h$ has to be determined. These are min nodes, so the
value of all
children $f, g$ and $i, j$ is needed. These are leaf nodes, for which
the relation $f \geq g > -\infty$ always holds. 
We have come to the end of our recursive search for a lower
bound. The sub-tree that we had to expand below node $a$ is shown in
figure~\ref{fig:minsoltree}. 

If we apply the minimax back-up rule to
figure~\ref{fig:minsoltree}, we find the value of the lower bound: 5. 
This is the lowest of the leaves of this sub
tree. The tree has the special property that only one child of each max node
is part of it, which renders the max-part of the minimax rule
redundant. For this reason these trees are called {\em min
solution trees}, denoted by $T^-$, their minimax value, denotred by
$f^-$, is the {\em minimum\/} of their  
leaves. (Min solution trees originated from the study of AND/OR
\index{AND/OR tree}%
trees, where they were called simply solution trees. The term {\em
solution\/} refers to the fact that they 
represent the solution to a problem
\cite{Kuma83,Nils71,Pear87,Stoc79}.)\index{solution tree, max}\index{solution tree, min}%

An intuitive interpretation of min solution trees in the area of 
game\index{strategy}
playing is that of a {\em strategy}. A min solution tree represents all
responses from the Min player to one of the moves by the Max
player. It is like reasoning: ``If I do this move, then Min has all
those options, to which my response will be $\ldots$'' A confusing
artefact of this terminology is that a {\em min\/}
solution tree is a strategy
for the {\em Max\/} player. 

\begin{figure}
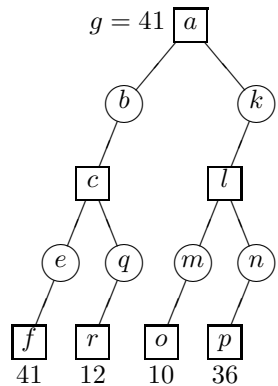

{\small
 \begin{Tree}
   \node{\external\type{square}\cntr{$f$}\bnth{$41$}}
   \node{\leftonly\cntr{$e$}}
   \node{\external\type{square}\cntr{$r$}\bnth{$12$}}
   \node{\leftonly\cntr{$q$}}
   \node{\type{square}\cntr{$c$}}
   \node{\leftonly\cntr{$b$}}

   \node{\external\type{square}\cntr{$o$}\bnth{$10$}}
   \node{\leftonly\cntr{$m$}}
   \node{\external\type{square}\cntr{$p$}\bnth{$36$}}
   \node{\leftonly\cntr{$n$}}
   \node{\type{square}\cntr{$l$}} 
   \node{\leftonly\cntr{$k$}}

   \node{\type{square}\cntr{$a$}\lft{$g = 41$}}
\end{Tree}
\hspace{5.5cm}\usebox{\TeXTree}
}
\caption{A Tree for an Upper Bound, a Max Solution Tree}\label{fig:maxsoltree}
\end{figure}

A min solution tree at node $n$  determines the value of a lower bound
on $f_n$. The 
sub-tree that determines an {\em upper\/} bound ($f^+$) is called, not
surprisingly,  a max solution
tree ($T^+$). It consists of all children at max nodes, and one at min
nodes. An example is shown in figure~\ref{fig:maxsoltree}. Its value
is the highest of the leaves of the max solution tree: 41. The
interpretation of a max solution tree is that of a strategy for the
Min player. 

Now we have the tools to construct upper and lower bounds at the
root, and thus to find (prove) the minimax value. 
This can be done cheaper than
by traversing the entire minimax tree, since solution trees are much
smaller.
%  To find an upper bound, we have
% to construct a max solution tree. 
In a solution tree,  at half of the nodes 
 only one child is traversed. Therefore, the exponent determining the
number of leaf nodes is halved. To find an upper bound we have to
examine only $w^{\lceil d/2 \rceil}$ leaf nodes. Since the root is a
max node, for odd search depths
lower bounds are even cheaper,  requiring $w^{\lfloor d/2 \rfloor}$
leaves to be searched. 
% Much less than $w^d$.
\index{best case}\index{odd/even search depth}\index{minimal tree}%
With this, we can determine the best case of any algorithm that wants to
prove the minimax value, for minimax trees of uniform $w$ and uniform
$d$. If the minimax tree is ordered so that the first child of a max
node is its highest and the first child of a min node is its lowest,
then the first max solution tree and the first min solution tree that
are traversed prove the minimax value of the root. We note that these
two solution 
trees overlap in one leaf, leaf $f$ in figure~\ref{fig:minsoltree} and
\ref{fig:maxsoltree} (in the figures the upper bound is not equal to
the lower bound, so although these are the left-most solution trees,
they do not prove the minimax value). Thus, the number of leaves of
the minimal tree 
that proves the value of $f$ is $w^{\lfloor d/2 \rfloor} + w^{\lceil
d/2 \rceil} - 
1$. This is a big improvement over the number of leaves
of the minimax 
tree $w^d$. It means that with pruning, programs can search up to twice
the search depth of full minimax. 

The tree that is actually traversed by an algorithm is called the
{\em search tree}. With pruning, the search tree has become a sub-set
of the minimax tree.

This concept of a minimal tree (or critical tree, or proof tree) was
\index{critical tree|see{minimal tree}}%
\index{proof tree|see{minimal tree}}\index{Knuth and Moore}%
introduced by Knuth and Moore \cite{Knut75}. They introduced it as
the best case of the {\AB} algorithm, using a categorization of three
different types of nodes, not in terms of solution trees (see
figure~\ref{fig:crittree} for a minimal tree with node types).  
\index{node type (Knuth and Moore)}%
The treatment of the minimal tree in terms of bounds and solution trees,
is based on \cite{DeBruin94,Icca94,Pijl95}, analogous to the use
of strategies by Pearl \cite[p.~222--226]{Pear84} and by Nilsson 
\cite[p.~110]{Nils71}.  \index{Pearl}\index{Nilsson}%
(Regrettably, they use $T^+$ to denote
a max strategy, which we  refer to as a min solution tree,
denoted by $T^-$.) \index{strategy}\index{solution tree}%
Other works on solution trees are
\cite{Ibar86,Kuma84,Pear87,Pijls90,Rein89,Stoc79}. 
% 
% This concept of a minimal tree (or critical tree, or proof tree) was
% introduced by Knuth and Moore \cite{Knut75}. They introduced it as
% the best case of the {\AB} algorithm, using a categorization of three
% different types of nodes, not in terms of solution trees (see
% figure~\ref{fig:crittree} for a minimal tree with node types).  
% The treatment of the minimal tree in terms of bounds and solution trees,
% is based on \cite{DeBruin94,Icca94,Pijl95}. That work
% draws on other work on {\SSS} and solution trees
% \cite{Ibar86,Kuma84,Pear87,Pijls90,Rein89,Stoc79}. 

\subsection{{\AB}} \index{Alpha-Beta}\index{pruning}\index{cut off}%
% Now that we have seen that 
Pruning can yield sizable improvements, potentially reducing the
complexity of finding the minimax value to the square root.
The {\AB} algorithm enhances 
the minimax 
function with pruning.   {\AB} has been in use by the
computer-game-playing community since the end of the
1950's. It seems to have been
conceived of independently by several people. The first
publication describing a form of pruning is 
%(although without deep cutoffs---see the
%example in appendix~\ref{sec:abex}) 
\index{Newell, Shaw and Simon}%
\index{McCarthy}\index{Knuth and Moore}%
by Newell, Shaw and Simon \cite{Newe58}. John McCarthy is
said to have had the original idea,
and also to have coined the term {\AB} 
\cite{Knut75}.  However, according to Knuth and Moore \cite{Knut75},
Samuel has stated that
the idea was already present in his checker-playing programs of the
late 1950's \cite{Samu59}, but he did not mention this because he
considered other aspects of his program to be more significant.
The first accounts of the full algorithm in Western literature appear
at the end of the 1960's, by Slagle and Dixon \cite{Slag69}, and by
Samuel \cite{Samu67}\index{Samuel}. However, Brudno 
already described an algorithm identical to {\AB} in Russian in 1963
\cite{Brud63}\index{Brudno}. 
A comprehensive analysis of the
algorithm, introducing the concept of the minimal tree,
has been published by Knuth and Moore in 1975
\cite{Knut75}. This classical work also contains a brief historic
account, which has been summarized here. Other works analyzing {\AB} are
\cite{Baud78b,Icca94,Pear82,Pijls94,Pijl95}. In the following we pursue a
description in intuitive terms.

% - code

\begin{figure}
{\small
\begin{tabbing}
mmmmmmmmm\=mm\=mm\=mm\=mm\kill
\> {\bf function} alphabeta$(n, \alpha, \beta) \rightarrow g$;\\
\> \> {\bf if} $n$ = leaf {\bf then return} eval$(n)$;\\
\> \> {\bf else if} $n$ = max  {\bf then}\\
\> \> \> $g := -\infty$;\\
\> \> \> $c := $ firstchild$(n)$;\\
\> \> \> {\bf while} $g < \beta$ {\bf and}  $c \neq \bot$ {\bf do}\\
\> \> \> \> $g := $ max$\big(g,$ alphabeta$(c, \alpha, \beta)\big)$;\\
\> \> \> \> $\alpha := \max(\alpha, g)$;\\
\> \> \> \> $c := $ nextbrother$(c)$;\\
\> \> {\bf else} /* $n$ is a min node */ \\ % {\bf then}\\
\> \> \> $g := +\infty$;\\
\> \> \> $c := $ firstchild$(n)$;\\
\> \> \> {\bf while} $g > \alpha$ {\bf and} $c \neq \bot$ {\bf do}\\
\> \> \> \> $g := $ min$\big(g,$ alphabeta$(c, \alpha, \beta)\big)$;\\
\> \> \> \> $\beta := \min(\beta, g)$;\\
\> \> \> \> $c := $ nextbrother$(c)$;\\
\> \> {\bf return} $g$;
\end{tabbing}
}
\caption{The {\AB} Function}\label{fig:abcode}
\end{figure}

Figure~\ref{fig:abcode} gives the pseudo code for {\AB}. It consists of the
minimax function, plus
two extra input parameters and cutoff tests. The $\alpha$ and
$\beta$ parameters together are called the {\em search window}. At max
nodes, $g$ is a lower bound on the return value. This lower bound is passed
to children as the $\alpha$ parameter. Whenever any of these children
finds it can no longer return a value above that lower bound,
further searching  is useless and is stopped. This can happen in
children of type 
min, since there $g$ is an upper bound on the return value. Therefore,
figure~\ref{fig:abcode} contains for min nodes the line ``{\bf while} $g >
\alpha$.'' 
\begin{figure}
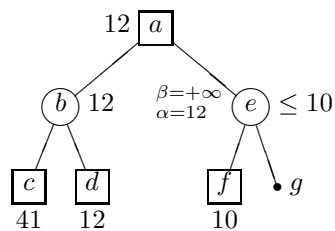

{\small
 \begin{Tree}
   \node{\external\type{square}\cntr{$c$}\bnth{$41$}}
   \node{\external\type{square}\cntr{$d$}\bnth{$12$}}
   \node{\cntr{$b$}\rght{$12$}}

   \node{\external\type{square}\cntr{$f$}\bnth{$10$}}
   \node{\external\type{dot}\rght{$g$}}
   \node{\cntr{$e$}\rght{$\leq 10$}\lft{$\ ^{\beta = +\infty}_{\alpha = 12}$}}

   \node{\type{square}\cntr{$a$}\lft{$12$}}
\end{Tree}
\hspace{5.5cm}\usebox{\TeXTree}
}
\caption{Node $g$ is Cut Off}\label{fig:cutoff}
\end{figure}
At min nodes $g$ is an upper bound. Parameter $\beta$ passes
the bound on so that any max children with a lower bound $\geq \beta$ can
stop searching when necessary. 
Together, $\alpha$ and $\beta$ form a search window which can be
regarded as a task for a node to return a value that lies inside the
window. If a node finds that its return value is proven to lie not
within  the
search window, the search is stopped. This is illustrated in
figure~\ref{fig:cutoff}. Assume that {\AB} is called with an initial
window of $\langle -\infty, +\infty\rangle$. Node $e$ is searched with
a window of 
$\langle 12, +\infty\rangle$. After node $f$ returns, the return value
of  node $e$ is $\leq 10$, which lies outside the search window, so
the search is stopped before node $g$ is traversed.

As children of a node are expanded, the $g$-value in that node is a
bound. The bound is established because {\AB} has traversed a solution
tree that defines its value. As more nodes are 
expanded, the bounds become tighter, until finally a min and a max
solution tree of equal value prove the minimax value of the root. In
the following postcondition of 
{\AB} these solution trees are explicitly present, following
\cite{Icca94,Pijl95}. 
\index{Alpha-Beta!precondition}\index{Alpha-Beta!postcondition}%
The precondition of a call {\AB}$(n, \alpha,
\beta)$ is $\alpha < \beta$.  As before, $g$ denotes
the return value of the call, $f_n$ denotes
the minimax value of node $n$, $f^+_n$ denotes the minimax value of a
max solution tree $T^+_n$, which is an upper bound on $f_n$,  and
$f^-_n$ denotes the minimax value of a min solution tree $T^-_n$,
which is a lower bound on $f_n$. The postcondition
has three cases \cite{Icca94,Knut75,Pijl95}:
\begin{enumerate}
\item $\alpha < g_n < \beta$ (success). $g_n$ is equal to
$f_n$. {\AB} has traversed at least a $T^+_n$ and a $T^-_n$ with 
$f(T^+_n) = f(T^-_n) = f_n$. 
\item $g_n \leq \alpha$ (failing low). $g_n$ is an upper bound
$f^+_n$, or $f_n \leq g_n$. {\AB} has traversed at least a $T^+_n$ with
$f(T^+_n) = f^+_n$. 
\item $g \geq \beta$ (failing high). $g_n$ is a lower bound $f^-_n$,
or $f_n \geq g_n$. {\AB} has traversed at least a $T^-_n$ with $f(T^-_n) = f^-_n$. 
\end{enumerate}\label{sec:pcab}\label{sec:abpc}
%  the last two cases can only occur with
% children of root $n$, as we shall see shortly in the example. 
If the return value of a node 
lies in the search window, then its minimax value
has been found. Otherwise the return value represents a
bound on it. From these cases we can infer that in order to be sure to
find the game value 
{\AB} must be called as {\AB}$(n, -\infty, +\infty)$.
(Older versions
of {\AB} returned  $\alpha$ or 
$\beta$ at a fail low or fail high. The version returning a bound is called {\em
fail-soft\/} \index{fail-soft}\index{fail high}\index{fail low}%
{\AB} in some publications \cite{Camp83,Fish80,Rein89}, because a fail high
or fail low still returns useful information. 
We use the term {\AB} to denote the fail-soft
version. Furthermore, implementations of {\AB} generally use
the negamax \index{negamax view}%
formulation since it is more compact \cite{Knut75}. 
For reasons of clarity, we use
the minimax view.) \index{minimax view}%

Appendix~\ref{app:abex} contains a detailed example of how a tree is
searched by {\AB}.

\subsection{Plotting Algorithm Performance}
The purpose of this thesis is to find better minimax search
algorithms. In comparing algorithms, we will 
use a grid that shows two key performance
parameters, allowing us to position an algorithm at a glance.
% This thesis discusses quite a number of algorithms. To reduce any
% confusion,
% we will use a
% picture that shows the place of an algorithm relative to two 
% key criteria. 

For game-playing programs the quality of an
algorithm is determined (a) by the speed with which it finds the best move
for a given position and (b) by the amount of storage it needs in
doing that. The speed is largely influenced by how many cutoffs it finds.
Often there is a trade-off between speed and storage: more memory gives
a faster search. In performance comparisons this trade-off introduces the
danger of unfair biases. Luckily, for the algorithms that are studied in
this work, this relation stabilizes at some point to the extent that
adding more memory does not improve performance measurably. To avoid this
bias, all algorithms were tested with the same amount of storage, which
was known to be large enough for all to achieve stability.
% The fact that we were able to do this with conventional hardware
% implies that for these algorithms,
% for search depths typical for current game-playing programs,
% speed is the critical factor, not memory.

Execution time is generally considered to be a difficult performance
metric, since it can vary widely for different programs and hardware.
Instead, the size of the  tree that is searched by an algorithm is
often used in comparisons.
% since it makes comparisons with other publications easier is the
% size of the search tree. 
Though better than execution time, this
metric is still not without problems, since a program usually spends a
different amount of time in processing leaf nodes and interior nodes. 
Many publications only report the number of leaf nodes searched by an
algorithm. 
This introduces a bias towards algorithms that revisit interior nodes
frequently.  
In many programs leaf evaluations are slowest,
followed by making and generating moves, which occur at interior nodes. 
However, in some programs the reverse holds. Compared to the time spent
on these actions, transposition nodes are fast (see section~\ref{sec:tt}).

The algorithms in this thesis are judged on their performance. In the
experiments we have counted leaf nodes, interior 
nodes, and transpositions. The most
important parameter is the size of the search tree generated by the
algorithm. This parameter will be used as an indicator of time
efficiency. By finding more 
cutoffs, an algorithm performs better on this 
parameter. A second parameter is the amount of memory an algorithm
needs. Though not a limiting factor in their practical applicability,
this remains a factor of importance. \index{performance picture}%
\begin{figure}
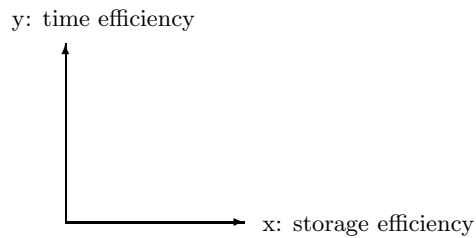

{\small
\begin{center}
\input twodim.latex
\end{center}
}
\caption{Two Performance Dimensions}\label{fig:3dim}
\end{figure}
Thus we
have two parameters to judge algorithms on:
\begin{enumerate}
\item {\em Cutoffs}\\
      The degree to which an algorithm finds useful cutoffs to reduce
      its node count is influenced by the quality of move ordering. A
higher quality of move ordering causes more cutoffs and thus a smaller
search tree. \index{move ordering, quality of}%
\item {\em Storage efficiency}\\
      The amount of storage an algorithm needs to achieve high
      performance. \index{memory}%
\end{enumerate}
Figure~\ref{fig:3dim} shows a grid of these two dimensions. It will be
used to summarize the behavior of conventional implementations of a
number of algorithms (and not more than that; the graphs do not
explain behavior, they only picture it).  
Algorithms should go as far as possible on the $x$ and $y$ axis,
to achieve 
high performance; slow algorithms that use a lot of memory are close
to the origin.

% The first parameter yields the greatest possible gain: a
% reduction of the search tree from $O(w^d)$ to $O(w^{d/2})$ for trees of
% unifrom width $w$ and depth $d$.  Algorithms
% try to increase the number of cutoffs by improving the degree of move
% ordering, using smaller search windows and using best-first instead of
% depth-first search, as we will see further on in this chapter.
% 
% The second parameter is storage.
% An algorithm that needs too much storage to achieve high performance is
% useless. As was indicated before,  the algorithms that are discussed in
% depth in this work are all practical.

% These two parameters depend on each other. For example, 
% more memory can improve the number of 
% cutoffs in a best-first algorithm. 
% This picture is not  an all-encompassing taxonomy of minimax
% algorithms. It uses two parameters that are thought to be of key
% importance to judge performance, to make it easier to keep track of
% the multitude of different 
% algorithms and enhancements that the following chapters will
% bring. Figure~\ref{fig:bfdftable} and \ref{fig:factors} show factors
% that can be used to categorize the algorithms qualitatively.

\begin{figure}
{\small
\begin{center}
\setlength{\unitlength}{0.009375in}%
\begingroup\makeatletter\ifx\SetFigFont\undefined
% extract first six characters in \fmtname
\def\x#1#2#3#4#5#6#7\relax{\def\x{#1#2#3#4#5#6}}%
\expandafter\x\fmtname xxxxxx\relax \def\y{splain}%
\ifx\x\y   % LaTeX or SliTeX?
\gdef\SetFigFont#1#2#3{%
  \ifnum #1<17\tiny\else \ifnum #1<20\small\else
  \ifnum #1<24\normalsize\else \ifnum #1<29\large\else
  \ifnum #1<34\Large\else \ifnum #1<41\LARGE\else
     \huge\fi\fi\fi\fi\fi\fi
  \csname #3\endcsname}%
\else
\gdef\SetFigFont#1#2#3{\begingroup
  \count@#1\relax \ifnum 25<\count@\count@25\fi
  \def\x{\endgroup\@setsize\SetFigFont{#2pt}}%
  \expandafter\x
    \csname \romannumeral\the\count@ pt\expandafter\endcsname
    \csname @\romannumeral\the\count@ pt\endcsname
  \csname #3\endcsname}%
\fi
\fi\endgroup
\begin{picture}(220,180)(5,625)
\thinlines
\put(182,700){\circle*{6}}
\put( 75,640){\vector( 0, 1){140}}
\put( 75,640){\vector( 1, 0){140}}
\multiput( 75,650)(7.74194,0.00000){16}{\line( 1, 0){  3.871}}
\multiput( 75,770)(7.74194,0.00000){16}{\line( 1, 0){  3.871}}
\put( 45,790){\makebox(0,0)[lb]{\smash{\SetFigFont{9}{10.8}{rm}y: time efficiency}}}
\put(225,635){\makebox(0,0)[lb]{\smash{\SetFigFont{9}{10.8}{rm}x: storage efficiency}}}
\put( 72,650){\makebox(0,0)[rb]{\smash{\SetFigFont{9}{10.8}{rm}minimax tree}}}
\put( 72,770){\makebox(0,0)[rb]{\smash{\SetFigFont{9}{10.8}{rm}minimal tree}}}
\put( 75,625){\makebox(0,0)[lb]{\smash{\SetFigFont{9}{10.8}{rm}$O(w^{d/2})$}}}
\put(175,625){\makebox(0,0)[lb]{\smash{\SetFigFont{9}{10.8}{rm}$O(d)$}}}
\put(190,695){\makebox(0,0)[lb]{\smash{\SetFigFont{9}{10.8}{rm}Alpha-Beta}}}
\end{picture}
\end{center}
}
\caption{Alpha-Beta's Performance Picture}\label{fig:abpic}
\end{figure}
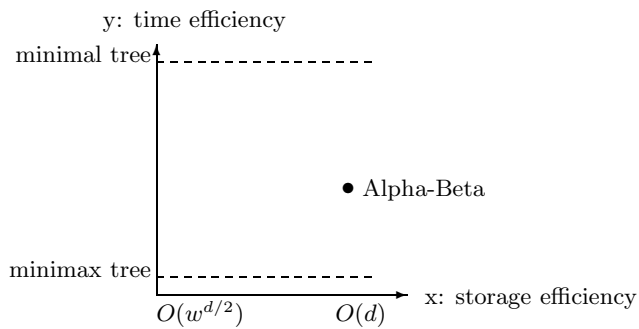

\section{{\AB} Enhancements} \index{Alpha-Beta!enhancements}%
The example in appendix~\ref{app:abex} illustrates that {\AB} can miss
some cutoffs. 
% Although pruning occurs in the example, {\AB} builds a tree that is
% larger than the minimal tree, as can be seen
% by comparing figures~\ref{fig:abtree} and \ref{fig:critabtree}. {\AB}
% expands 11 leaves, less than the minimax tree of 16, but more than the
% minimal tree of 7.
Figure~\ref{fig:abpic} illustrates that {\AB}'s performance picture lies
between that of minimax and the minimal tree. Its storage needs
are excellent, only $O(d)$, the recursion depth. (The $y$ axis is not
drawn to scale.) 

The performance of {\AB} depends
on the ordering of child nodes in the tree.\index{move ordering, quality of}
This brings us to the question of whether there are ways to enhance
{\AB}'s performance to bring it closer to the theoretical best case,
or whether there are alternative ways to implement 
pruning. The remainder of this section  discusses the former, {\AB}
enhancements. Section~\ref{sec:alt} will discuss the latter,
alternatives to {\AB} 
pruning.

\subsection{Smaller Search Windows} \index{narrow search window}%
{\AB} cuts off a sub-tree when the value of a node falls outside the
search window. One idea to increase tree pruning is to search with a
smaller search window. 
Assuming $c \leq a$
and $b \leq d$, a search with (wider) window $\langle c,d \rangle$
will visit at least every single node that the search with (smaller)
window $\langle 
a,b \rangle$ will visit (if both search the same minimax
tree). Normally the wider search will visit extra nodes 
\cite{Camp83,Pear84}.
However, {\AB} already uses all return values of the depth-first
search to reduce 
the window as much as possible. Additional search window reductions
run the risk that {\AB} may not be able to find the minimax
value. According to the postcondition in section~\ref{sec:pcab}, one
can only be sure that the minimax value is  
found if $\alpha < g < \beta$. If the return value lies outside the
window, then all we are told is that a bound on the minimax value is
found. To find the 
true minimax value in that case, a
re-search with the right window is necessary. In practice the savings
of the tighter window 
out-weigh the overhead of additional re-searches, as has been
reported in many studies (see for example \cite{Camp83,Pear84,Rein89}).  In
addition, the use of storage can reduce the re-search overhead
(see section~\ref{sec:tt} and chapter~\ref{chap:mt}). 

Here we will describe two widely used  techniques to benefit from the
extra cutoffs that artificially-narrowed search windows yield.

\begin{figure}
{\small
\begin{tabbing}
mmmmmmmmmmmm\=mm\=mm\=mm\=mm\kill
\> {\bf function} aspwin$(n,$ estimate, delta$) \rightarrow f$;\\
\> \> $\alpha := $ estimate $ - $ delta;\\
\> \> $\beta := $ estimate $ + $ delta;\\
\> \> $g := $ alphabeta$(n, \alpha, \beta);$\\
\> \> {\bf if} $g \leq \alpha$ {\bf then}\\
\> \> \> $g := $ alphabeta$(n, -\infty, g);$\\
\> \> {\bf else if} $g \geq \beta$ {\bf then}\\
\> \> \> $g := $ alphabeta$(n, g, +\infty);$\\
\> \> {\bf return} $g$;
\end{tabbing}
}
\caption{Aspiration Window Searching}\label{fig:aspwincode}
\end{figure}
\subsubsection{Aspiration Window}
In many games the values of parent and child nodes are correlated.
\index{correlation, parent-child value}\index{aspiration window}%
Therefore we can obtain cheap estimates of the result that a search to a
certain depth will return.  (We can do a relatively cheap
search to a shallow depth to obtain this estimate.)  This estimate can
be used to create a small search window, in chess typically $\pm$~the
value of a pawn. This window is known as an
{\em aspiration window}, since we aspire that the result will be
within the bounds of the window.
With this window an {\AB} search is performed. If it ``succeeds'' (case 1
of the postcondition on page~\pageref{sec:abpc}), then we
have found the minimax value cheaply. If it ``fails,'' then a re-search
must be performed. Since the failed search has returned a bound, this
re-search can also benefit from a window smaller than $\langle -\infty,
+\infty\rangle$.

Aspiration window searching is commonly used at the root of the
tree. 
Figure~\ref{fig:aspwincode} gives the pseudo code for this standard
technique.  One option for the estimate is to evaluate the current
position. Assuming that a pawn is given the value of 100, the call
``aspwin$(n, $ eval$(n), 100)$'' would find us the minimax value and 
usually do so more efficiently than the call ``{\AB}$(n, -\infty,
+\infty)$.'' Some 
references to this technique are \cite{Baud78b,Camp83,Frey77}.

% 
% The rest of the tree can be searched with the traditional
% $\langle\alpha, \beta\rangle$ window. However, also for these nodes an
% alternative to straight {\AB} searching exists. This will be
% described next. 

\begin{figure}
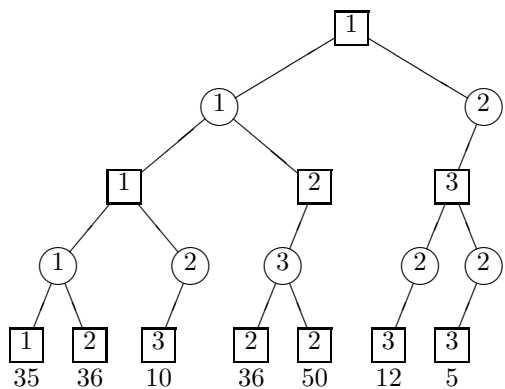

{\small
 \begin{Tree}
   \node{\external\type{square}\cntr{$1$}\bnth{$35$}}
   \node{\external\type{square}\cntr{$2$}\bnth{$36$}}
   \node{\cntr{$1$}}
   \node{\external\cntr{$3$}\type{square}\bnth{$10$}}
   \node{\cntr{$2$}\leftonly}
   \node{\type{square}\cntr{$1$}}
   \node{\external\type{square}\cntr{$2$}\bnth{$36$}}
   \node{\external\type{square}\cntr{$2$}\bnth{$50$}}
   \node{\cntr{$3$}}
   \node{\type{square}\cntr{$2$}\leftonly}
   \node{\cntr{$1$}}

   \node{\external\type{square}\cntr{$3$}\bnth{$12$}}
   \node{\cntr{$2$}\leftonly}
   \node{\external\type{square}\cntr{$3$}\bnth{$5$}}
   \node{\cntr{$2$}\leftonly}
   \node{\type{square}\cntr{$3$}}
   \node{\cntr{$2$}\leftonly}
   \node{\type{square}\cntr{$1$}}
\end{Tree}
\hspace{6.4cm}\usebox{\TeXTree}
}
\caption{Minimal Tree (with node types)}\label{fig:crittree}
\end{figure}
\subsubsection{Null-Window Search and Scout}
Pushing the idea of a small-search-window-plus-re-search to the limit
is the use of a null-window. The values
for $\alpha$ and $\beta$ are now chosen so that {\AB} will always
return a bound. Case 1 of the postcondition in section~\ref{sec:abpc}
can be eliminated by choosing $\alpha = \beta - 1$, assuming
integer-valued leaves, or $\alpha = \beta - \varepsilon$ in general,
where 
$\varepsilon$ is a number less than the smallest difference between
any two leaf values. (The precondition of {\AB} demands $\alpha <
\beta$, so $\alpha = \beta$ won't work.) 

An {\AB} search window of $\alpha = \beta - 1$ ensures the highest
number of cutoffs. 
% In
% effect, it builds just a solution tree, as the postcondition shows. 
On the downside, it also guarantees that a re-search is necessary to
find the 
minimax value. Since we usually want a minimax value at the root, it
makes more sense to use a wider aspiration window there, with a low
probability of the need for a re-search. However, if we look at the
structure of the minimal tree, we see that for most interior nodes a
bound is all that is needed.
Figure~\ref{fig:crittree} shows an ordered version of  the minimal
tree of the {\AB} 
example, with the nodes labeled with the three node types as defined
by Knuth \index{node type (Knuth and Moore)}%
and Moore in \cite{Knut75}. Nodes that have only one child in the minimal
tree are type~2, nodes with all children included are 
type~3, and nodes that are part of both the max and the min solution
tree are type~1. The type~1 nodes form the path from the root to
the best leaf. This intersection of the two solution trees is also
known as the critical path or the \index{principal variation}%
\index{critical path|see{principal variation}}%
principal variation (PV). Its interpretation is that of 
the line of play that the \index{line of play}%
search predicts.
For the type~1 nodes the minimax value is computed. In a depth $d$
tree, there are only $d+1$ type~1 nodes. The only task of
the type~2 and type~3 nodes
% ---the vast majority of the nodes, try
% picturing a tree with $w = 35$ instead of 2---
is to
prove that it does not make sense to 
search them any further, because they are worse than their type~1
relative. 

\index{null-window|see{narrow search window}}%
For the nodes off the PV it makes sense to use a
null-window search, because a 
bound is all that is needed. 
% There is no danger of the need for a
% re-search, in the ideal case. 
In real trees that are not perfectly \index{real tree}%
ordered, the PV node is not known {\em a priori\/} so there is a
danger of having to re-search. Once again, in 
practice the small-window savings out-weigh the re-search overhead
\index{re-search overhead}%
(see section~\ref{sec:tiny} and \cite{Camp83,Mars82a,Pear84}).
 
% 
% \begin{figure}
% {\small
% \begin{tabbing}
% mmmmmmmmmmmm\=mm\=mm\=mm\=mm\=mm\kill
% \> {\bf function} P-alphabeta$(n) \rightarrow f$;\\
% \> \> {\bf if} $n$ = leaf {\bf then return} eval$(n)$;\\
% \> \> $c := $ firstchild$(n)$;\\
% \> \> $g := $ P-alphabeta$(c)$;\\
% \> \> $c := $ nextbrother$(c)$;\\
% \> \> {\bf if} $n$ = max  {\bf then}\\
% \> \> \> {\bf while} $c \neq \bot$ {\bf do}\\
% \> \> \> \> $t := $ alphabeta$(c, g, g + 1)$;\\
% \> \> \> \> {\bf if} $t > g$ {\bf then} $g := $ alphabeta$(c, t, +\infty)$;\\
% %\> \> \> \> \> $g := $ alphabeta$(c, t, +\infty)$;\\
% \> \> \> \> $c := $ nextbrother$(c)$;\\
% \> \> {\bf else} /* $n$ is a min node */ \\ % {\bf then}\\
% \> \> \> {\bf while} $c \neq \bot$ {\bf do}\\
% \> \> \> \> $t := $ alphabeta$(c, g - 1, g)$;\\
% \> \> \> \> {\bf if} $t < g$ {\bf then}  $g := $ alphabeta$(c, -\infty, t)$;\\
% %\> \> \> \> \> $g := $ alphabeta$(c, -\infty, t)$;\\
% \> \> \> \> $c := $ nextbrother$(c)$;\\
% \> \> {\bf return} $g$;
% \end{tabbing}
% }
% \caption{P-alphabeta}\label{fig:pab}
% \end{figure}
Thus we come to an algorithm that uses a wide search window for the
first child, hoping it will turn out to be part of the PV, and a
null-window for the other children.
At a max node the first node should be the
highest. If one of the null-window searches returns a bound that is
higher, then this child becomes the new PV candidate and should be
re-searched with a wide window to determine its value. For a min node
the first node should 
remain the lowest. If the null-window searches show one of the
brothers to be lower, then that one replaces the PV candidate.

% \begin{figure}
% {\small
% \begin{tabbing}
% mmmmmmmmmmmm\=mm\=mm\=mm\=mm\=mm\kill
% \> {\bf function} Scout$(n) \rightarrow f$;\\
% \> \> {\bf if} $n$ = leaf {\bf then return} eval$(n)$;\\
% \> \> $c := $ firstchild$(n)$;\\
% \> \> $g := $ Scout$(c)$;\\
% \> \> $c := $ nextbrother$(c)$;\\
% \> \> {\bf if} $n$ = max  {\bf then}\\
% \> \> \> {\bf while} $c \neq \bot$ {\bf do}\\
% \> \> \> \> {\bf if} TEST$(c, g, >)$ {\bf then} $g := $ Scout$(c)$;\\
% %\> \> \> \> \> $g := $ Scout$(c)$;\\
% \> \> \> \> $c := $ nextbrother$(c)$;\\
% \> \> {\bf else} /* $n$ is a min node */ \\ % {\bf then}\\
% \> \> \> {\bf while} $c \neq \bot$ {\bf do}\\
% \> \> \> \> {\bf if} TEST$(c, g, <)$ {\bf then} $g := $ Scout$(c)$;\\
% %\> \> \> \> \> $g := $ Scout$(c)$;\\
% \> \> \> \> $c := $ nextbrother$(c)$;\\
% \> \> {\bf return} $g$;
% \end{tabbing}
% }
% \caption{Scout}\label{fig:scout}
% \end{figure}

\begin{figure}
{\small
\begin{tabbing}
mmmmmm\=mm\=mm\=mm\=mm\=mm\kill
\> {\bf function} {NegaScout}$(n, \alpha, \beta) \rightarrow g$;\\
\> \> {\bf if} $n$ = leaf {\bf then return} eval$(n)$;\\
\> \> $c := $ firstchild$(n)$;\\
\> \> $g := $ \mbox{NegaScout}$(c, \alpha, \beta)$;\\
\> \> $c := $ nextbrother$(c)$;\\
\> \> {\bf if} $n$ = max  {\bf then}\\
\> \> \> $g := $ max$(g, \alpha);$\\
\> \> \> {\bf while} $g < \beta$ {\bf and} $c \neq \bot$ {\bf do}\\
\> \> \> \> $t := $ \mbox{NegaScout}$(c, g, g + 1)$;\\
\> \> \> \> /* the last two ply of the tree return an accurate value */\\
\> \> \> \> {\bf if} $c = $ leaf {\bf or} firstchild$(c)$ = leaf {\bf then} $g := t$;\\
\> \> \> \> {\bf if} $t > g$ {\bf and} $t < \beta$ {\bf then} $g := $ {NegaScout}$(c, t, \beta)$;\\
%\> \> \> \> \> $g := $ \mbox{NegaScout}$(c, t, \beta)$;\\
\> \> \> \> $g := $ max$(g,t)$;  $c := $ nextbrother$(c)$;\\
\> \> {\bf else} /* $n$ is a min node */ \\ % {\bf then}\\
\> \> \> $g := $ min$(g, \beta);$\\
\> \> \> {\bf while} $g > \alpha$ {\bf and} $c \neq \bot$ {\bf do}\\
\> \> \> \> $t := $ \mbox{NegaScout}$(c, g - 1, g)$;\\
\> \> \> \> {\bf if} $c = $ leaf {\bf or} firstchild$(c)$ = leaf {\bf then} $g := t$;\\
\> \> \> \> {\bf if} $t < g$ {\bf and} $t > \alpha$ {\bf then} $g := $ {NegaScout}$(c, \alpha, t)$;\\
%\> \> \> \> \> $g := $ \mbox{NegaScout}$(c, \alpha, t)$;\\
\> \> \> \> $g := $ min$(g,t)$;  $c := $ nextbrother$(c)$;\\
\> \> {\bf return} $g$;
\end{tabbing}
}
\caption{{\NS}}\label{fig:ns}
\end{figure}

\begin{figure}
{\small
\begin{center}
\setlength{\unitlength}{0.009375in}%
\begingroup\makeatletter\ifx\SetFigFont\undefined
% extract first six characters in \fmtname
\def\x#1#2#3#4#5#6#7\relax{\def\x{#1#2#3#4#5#6}}%
\expandafter\x\fmtname xxxxxx\relax \def\y{splain}%
\ifx\x\y   % LaTeX or SliTeX?
\gdef\SetFigFont#1#2#3{%
  \ifnum #1<17\tiny\else \ifnum #1<20\small\else
  \ifnum #1<24\normalsize\else \ifnum #1<29\large\else
  \ifnum #1<34\Large\else \ifnum #1<41\LARGE\else
     \huge\fi\fi\fi\fi\fi\fi
  \csname #3\endcsname}%
\else
\gdef\SetFigFont#1#2#3{\begingroup
  \count@#1\relax \ifnum 25<\count@\count@25\fi
  \def\x{\endgroup\@setsize\SetFigFont{#2pt}}%
  \expandafter\x
    \csname \romannumeral\the\count@ pt\expandafter\endcsname
    \csname @\romannumeral\the\count@ pt\endcsname
  \csname #3\endcsname}%
\fi
\fi\endgroup
\begin{picture}(220,180)(5,625)
\thinlines
\put(182,680){\circle*{6}}
\put(190,675){\makebox(0,0)[lb]{\smash{\SetFigFont{9}{10.8}{rm}Alpha-Beta}}}
\put(182,695){\circle*{6}}
\put( 75,640){\vector( 0, 1){140}}
\put( 75,640){\vector( 1, 0){140}}
\multiput( 75,650)(7.74194,0.00000){16}{\line( 1, 0){  3.871}}
\multiput( 75,770)(7.74194,0.00000){16}{\line( 1, 0){  3.871}}
\put( 45,790){\makebox(0,0)[lb]{\smash{\SetFigFont{9}{10.8}{rm}y: time efficiency}}}
\put(225,635){\makebox(0,0)[lb]{\smash{\SetFigFont{9}{10.8}{rm}x: storage efficiency}}}
\put( 72,650){\makebox(0,0)[rb]{\smash{\SetFigFont{9}{10.8}{rm}minimax tree}}}
\put( 72,770){\makebox(0,0)[rb]{\smash{\SetFigFont{9}{10.8}{rm}minimal tree}}}
\put( 75,625){\makebox(0,0)[lb]{\smash{\SetFigFont{9}{10.8}{rm}$O(w^{d/2})$}}}
\put(175,625){\makebox(0,0)[lb]{\smash{\SetFigFont{9}{10.8}{rm}$O(d)$}}}
\put(190,690){\makebox(0,0)[lb]{\smash{\SetFigFont{9}{10.8}{rm}NegaScout}}}
\end{picture}
\end{center}
}
\caption{{\NS}'s Performance Picture}\label{fig:abnspic}
\end{figure}

% There are small variations in the way a number of algorithms implement
% these ideas.
% An algorithm called P-alphabeta (figure~\ref{fig:pab} shows
% a minimax version)
% \cite{Camp83,Fish80} uses {\AB} for the re-search. The re-search
% window uses the bound returned by the null-window call. Another algorithm
% is Scout 
% \cite{Pear80,Pear84}. Figure~\ref{fig:scout} shows the pseudocode of a
% minimax version, the
% original publication shows the algorithm as a flow-chart. It uses a
% recursive 
% function called TEST, which returns a Boolean value indicating whether
% the  assertion $f > g$ or $f < g$ is true for the sub-tree at a node.
% The function TEST is less
% useful
% than a null-window fail-soft {\AB} call, because it does not return a
% bound. Therefore the re-search cannot 
% benefit from the 
% null-window search like  P-alphabeta. This re-search is not
% performed by {\AB}, but by a  recursive call to Scout. Since Scout
% uses null-windows, and {\AB} does not, a recursive call  is more
% efficient.

Early algorithms that use the idea of null-windows are P-alpha\-beta
\cite{Camp83,Fish80}, Scout \cite{Pear80,Pear84} and PVS
\index{P-alpha-beta}\index{PVS}\index{NegaScout}\index{Pearl}\index{Marsland}\index{Campbell}\index{Fishburn}\index{Finkel}\index{Reinefeld}\cite{Camp83,Mars87,Mars93}. Reinefeld has studied these algorithms
as well as a number of other improvements \cite{Mars87,Rein89,Rein85}. 
This has resulted in {\NS}, the algorithm that is used by most 
high-performance game-playing programs. 
Figure~\ref{fig:ns} shows a minimax version of {\NS}. The
choice of the name
{\NS} is a bit unfortunate, since it suggests that this is just a negamax
formulation of the Scout algorithm, instead of a new, cleaner, and more
efficient algorithm. An extra
enhancement in {\NS} is that it 
is observed that a fail-soft search of the last two ply of a tree
always returns an exact value, even in cases 2 and 
3 of the postcondition of section~\ref{sec:pcab}, so no re-search is
necessary (figure~\ref{fig:ns} assumes a fixed-depth tree). 
% (This is the only difference between {\NS} and PVS as it
% appeared in \cite{Mars93}. An older version of PVS used {\AB} for the
% re-search \cite{Camp83}.) Note 
% that P-alphabeta and Scout are not 
% called with $\alpha$ and $\beta$ parameters, whereas {\NS} does need
% them. {\NS} has the same postcondition as {\AB}
% \cite{Rein89}. P-alphabeta and Scout do not; they always return the
% minimax value. Note that {\NS} does not call {\AB} or TEST anymore.
% It is self-contained.

Empirical evidence has shown that, even without the use of extra memory,
null-window algorithm {\NS} finds more cutoffs than wide-window 
algorithm {\AB}$(n, -\infty, +\infty)$. Figure~\ref{fig:abnspic}
illustrates this point.  (The $y$ axis is not drawn to
scale.) 

Most successful game-playing programs use a combination of {\NS} with
aspiration window searching at the root. We will call
this combination {\AspNS}.

Another idea is to do away with wide-windowed re-searches
altogether, and use null-windows only. This idea is further analyzed
in chapter~\ref{chap:mt}.

\subsection{Move Ordering} \index{move ordering, quality of}%
A different way to improve the effectiveness of {\AB} pruning is to
improve the order in which child positions are examined. On a perfectly
ordered uniform tree {\AB} will cut off the maximum number
of nodes. 
A first approach is to use application-dependent knowledge to order
the moves. For example, in chess it is often wise to try moves
that will capture an opponent's piece first, and in Othello certain moves
near the corners are often better. However, there exist also a
number of techniques that work independently of the application at
hand. 

\subsubsection{The History Heuristic}\index{history heuristic}%
We start with a technique that is in spirit still close  to the use of
application-dependent knowledge. In most games there are moves that
are good in many positions, for example, the two types of moves just
mentioned for chess and Othello. Schaeffer introduced the history
heuristic, a technique to identify moves that were
repeatedly good automatically
\cite{Schaeffer86,Scha89b}. It maintains a 
table in which for each move a score is increased whenever that move
turns out to be the best move or to cause a cutoff. At a node, moves are
searched in order of history heuristic scores. In this way the
program learns which moves are good in a certain position in a certain
game and which are not, in an
application-independent fashion. There exist a number of older
techniques, such as killer moves and refutation tables, which the
history heuristic 
generalizes and improves upon. Schaeffer has published an 
extensive study on the relative performance of these search
enhancements \cite{Scha89b}.

\subsubsection{Transposition Tables and Iterative Deepening}
\label{sec:idtt}\label{sec:tt}\index{transposition table}%
\index{iterative deepening}%
In many application domains of minimax search algorithms,
the search \index{graph}%
space is a graph, whereas minimax-based algorithms are suited for {\em tree\/}
search. Transposition tables (TT) are used to enhance the efficiency of
tree-search algorithms
by preventing the re-expansion of children with multiple parents
(transpositions)
\cite{Mars82a,Scha89b}. A transposition table is usually implemented
as a hash
\index{collision}%
table in which searched nodes are stored (barring collisions, the search tree).
The tree-search algorithm is modified to look in this table before
it searches a node and, if it finds the node,
uses the value instead of searching. 
In application domains where there are many paths leading to a 
node, this scheme leads to a substantial reduction of the search
space. (Although technically incorrect, we will stick to the usual
terminology and keep using terms like minimax {\em tree\/} search.)
Transposition tables are becoming a popular way to enhance the
performance of single  agent search programs as well \cite{Korf85,Mars94}. 

Most game-playing programs use
iterative deepening \cite{Mars82a,Scha89b,Slat77}. It is based on the
assumption that a shallow search is  a good
approximation of a deeper search. It starts off by doing a depth one
search, which terminates almost immediately. It then increases the
search depth step by step, each time restarting the search over and
over again. Due to the exponential growth of the tree the former
iterations usually take a negligible amount of time compared to the
last iteration. Among the benefits of iterative deepening (ID) in
game-playing programs are better move ordering 
(if used  with transposition tables), and advantages for
tournament time control information. (In 
the area of one-player games it is used as a way of reducing
the \index{one-player games}%
space complexity of best-first searches \cite{Korf85}.)

Transposition tables are often used in conjunction with iterative
deepening to achieve a partial move ordering.
The search value and the branch leading to the highest
score (the best move) are saved for each node.
When iterative deepening searches one level deeper and revisits nodes,
this information is used to search 
the previously best move first. Since we assumed that a shallow
search is a good approximation of a deeper search, this best move for
depth $d$ will often turn out to be the best move for depth $d+1$ too.
% Good move ordering increases the pruning power of algorithms like
% {\AB} and {\SSS}.

\begin{figure}[t]
{\small
\begin{tabbing}
mmmmmmmmm\=mm\=mm\=mm\=mm\kill
\> {\bf function} alphabeta$(n, \alpha, \beta) \rightarrow g$;\\
\> \> {\bf if} retrieve$(n)$ = ok {\bf then}\\
\> \> \> {\bf if} $n.f^- \geq \beta$ {\bf then return} $n.f^-$;\\
\> \> \> {\bf if} $n.f^+ \leq \alpha$ {\bf then return} $n.f^+$;\\
\> \> \> $\alpha := $ max$(\alpha, n.f^-)$;\\ 
\> \> \> $\beta := $ min$(\beta, n.f^+)$;\\ 
\> \> {\bf if} $n$ = leaf {\bf then} $g := $ eval$(n)$;\\
\> \> {\bf else if} $n$ = max  {\bf then}\\
\> \> \> $g := -\infty$;\\
\> \> \> $c := $ firstchild$(n)$;\\
\> \> \> {\bf while} $g < \beta$ {\bf and}  $c \neq \bot$ {\bf do}\\
\> \> \> \> $g := $ max$\big(g,$ alphabeta$(c, \alpha, \beta)\big)$;\\
\> \> \> \> $\alpha := \max(\alpha, g)$;\\
\> \> \> \> $c := $ nextbrother$(c)$;\\
\> \> {\bf else} /* $n$ is a min node */ \\ % {\bf then}\\
\> \> \> $g := +\infty$;\\
\> \> \> $c := $ firstchild$(n)$;\\
\> \> \> {\bf while} $g > \alpha$ {\bf and} $c \neq \bot$ {\bf do}\\
\> \> \> \> $g := $ min$\big(g,$ alphabeta$(c, \alpha, \beta)\big)$;\\
\> \> \> \> $\beta := \min(\beta, g)$;\\
\> \> \> \> $c := $ nextbrother$(c)$;\\
\> \> {\bf if} $g < \beta$ {\bf then} $n.f^+ := g$;\\
\> \> {\bf if} $g > \alpha$ {\bf then} $n.f^- := g$;\\
\> \> store $n.f^-, n.f^+$;\\ 
\> \> {\bf return} $g$;
\end{tabbing}
}
\caption{The {\AB} Function for Use with Transposition
Tables}\label{fig:ttabcode}\index{transposition table, code}%
\end{figure}

Thus, transposition tables in conjunction with ID are typically used
to enhance the performance of algorithms in two ways:
\label{sec:ttidx}%
\index{transposition table, uses of}%
\begin{enumerate}
\item
improve the quality of the move ordering, and
\item
detect when different paths through the search space transpose into
the same state,  to prevent the re-expansion of that node.
\end{enumerate}
In the case of an algorithm in which each ID iteration may perform
re-searches, like {\NS} and aspiration window searching,
there is an additional use for the TT:
\begin{enumerate}
\item[3.]
prevent the re-search of a node that has been searched in a previous
pass, in the {\em current\/} ID iteration.
\end{enumerate}

In game-playing programs {\NS} is almost always used in combination
with a transposition table. Figure~\ref{fig:ttabcode} shows a
version of {\AB} for use with transposition tables
\cite{Mars86b,Mars82a,Schaeffer86}.  Not shown is the
code for retrieving and storing the best move, nor the fact that in
most implementations usually one bound is stored, instead of both
$n.f^+$ and $n.f^-$  (see also the remark on
page~\pageref{sec:boundsremark}).  

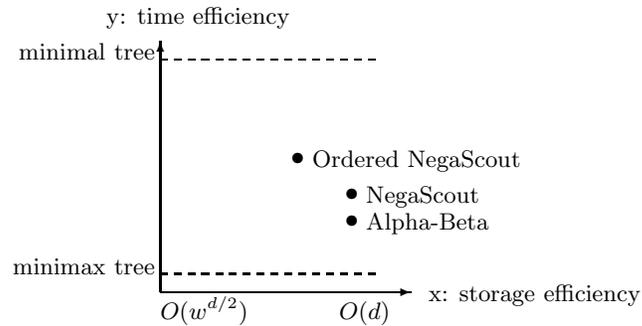
\begin{figure}
{\small
\begin{center}
\setlength{\unitlength}{0.009375in}%
\begingroup\makeatletter\ifx\SetFigFont\undefined
% extract first six characters in \fmtname
\def\x#1#2#3#4#5#6#7\relax{\def\x{#1#2#3#4#5#6}}%
\expandafter\x\fmtname xxxxxx\relax \def\y{splain}%
\ifx\x\y   % LaTeX or SliTeX?
\gdef\SetFigFont#1#2#3{%
  \ifnum #1<17\tiny\else \ifnum #1<20\small\else
  \ifnum #1<24\normalsize\else \ifnum #1<29\large\else
  \ifnum #1<34\Large\else \ifnum #1<41\LARGE\else
     \huge\fi\fi\fi\fi\fi\fi
  \csname #3\endcsname}%
\else
\gdef\SetFigFont#1#2#3{\begingroup
  \count@#1\relax \ifnum 25<\count@\count@25\fi
  \def\x{\endgroup\@setsize\SetFigFont{#2pt}}%
  \expandafter\x
    \csname \romannumeral\the\count@ pt\expandafter\endcsname
    \csname @\romannumeral\the\count@ pt\endcsname
  \csname #3\endcsname}%
\fi
\fi\endgroup
\begin{picture}(220,180)(5,625)
\thinlines
\put(182,680){\circle*{6}}
\put(190,675){\makebox(0,0)[lb]{\smash{\SetFigFont{9}{10.8}{rm}Alpha-Beta}}}
\put(182,695){\circle*{6}}
\put(190,690){\makebox(0,0)[lb]{\smash{\SetFigFont{9}{10.8}{rm}NegaScout}}}
\put(152,715){\circle*{6}}
\put( 75,640){\vector( 0, 1){140}}
\put( 75,640){\vector( 1, 0){140}}
\multiput( 75,650)(7.74194,0.00000){16}{\line( 1, 0){  3.871}}
\multiput( 75,770)(7.74194,0.00000){16}{\line( 1, 0){  3.871}}
\put( 45,790){\makebox(0,0)[lb]{\smash{\SetFigFont{9}{10.8}{rm}y: time efficiency}}}
\put(225,635){\makebox(0,0)[lb]{\smash{\SetFigFont{9}{10.8}{rm}x: storage efficiency}}}
\put( 72,650){\makebox(0,0)[rb]{\smash{\SetFigFont{9}{10.8}{rm}minimax tree}}}
\put( 72,770){\makebox(0,0)[rb]{\smash{\SetFigFont{9}{10.8}{rm}minimal tree}}}
\put( 75,625){\makebox(0,0)[lb]{\smash{\SetFigFont{9}{10.8}{rm}$O(w^{d/2})$}}}
\put(175,625){\makebox(0,0)[lb]{\smash{\SetFigFont{9}{10.8}{rm}$O(d)$}}}
\put(160,710){\makebox(0,0)[lb]{\smash{\SetFigFont{9}{10.8}{rm}Ordered NegaScout}}}
\end{picture}
\end{center}
}
\caption{Performance Picture with Better Ordering}\label{fig:abnsordpic}
\end{figure}
Figure~\ref{fig:abnsordpic} shows that a better move ordering enables an
algorithm to find more cutoffs. Usually a better move ordering comes at
the cost of storing some part of the search tree, which is shown in the
picture. Chapter~\ref{chap:exper} determines the storage needs of algorithms
in game-playing programs that typically use quite a number of
enhancements.

Originally transposition tables were introduced to  prevent
the search of transpositions in the search space---hence their name.
However, as 
algorithms grew more sophisticated, the role of the transposition
table evolved to that of storing the search tree.
It has become a cache \index{cache of nodes}%
of nodes that may or may
not be of use in future 
iterations or re-searches.
Calling it a transposition table does not do justice to
the central role that it plays in a modern algorithm like ID
{\AspNS}---{\NS} enhanced with an initial aspiration window, and called
in an iterative deepening framework. However, to prevent confusion we
will keep using the term ``transposition table.'' 

Noting that current game-playing programs store (part of) the search
tree in memory is a point that has to be taken into account when
reasoning about the behavior of search algorithms.
Many analyses and performance simulations of {\AB} and its variants 
do not use a transposition table
(for example, \cite{Baud78b,Camp83,Kain91,Mars87,Pear84,Rein85}). 
Since a 
transposition table makes the overhead of doing re-searches negligible,
this is a serious omission. The consequences of this
point will be discussed in a broader scope in section~\ref{sec:simsucks}.
The fact that the search tree is effectively stored in memory also
means that many theoretical schemes in which an explicit  search tree
plays a central role are closer to
practice than one would generally assume \cite{Ibar86,Pear80,Pijls90,Pijls93}.

Of central importance for the validity of the {\em TT = Search Tree\/}
assumption is the question of how big the search tree is
that is generated by an algorithm, whether it fits in the available
memory. This question is answered in section~\ref{sec:space}.

\subsection{Selective Search}\label{sec:abext}%
\index{search!selective}\index{search!fixed-depth}\index{search!full-width}%
Up to now we have been dealing exclusively with fixed-depth,
full-width algorithms. The pruning method used is sometimes called
{\em backward\/} pruning, since the backed-up return values of {\AB}
are used for the cutoff decisions. Studying  this class
of algorithms has the advantage that performance improvements are
easily measurable; one has only to look at the size of the tree. 
Although our research is concerned with fixed-depth full-width search,
a short overview of selective search is useful to put things in
perspective. (In practice a fixed-depth search may not always search all
moves equally deep, for example by finding that a move leads to an
early end of the game.)

In this section we will discuss algorithms that do not necessarily
search full-width. It has long been known that a flaw of the {\AB}
algorithm is that it searches all nodes to the same depth. No matter
how bad a move is, it gets searched as deep as the most promising
move \cite{Shan50}. Of course, backward pruning will probably make
sure that most of 
the nodes in the subtree of the bad move get pruned, but a more
selective search strategy could perhaps make sure that really bad
moves are not considered at all. These strategies are said to use selective
searching (in contrast to full-width), variable depth (in contrast to
fixed-depth), or forward pruning (in contrast to backward
pruning). An algorithm that uses selective deepening can search a line
of play more deeply in one iteration than in another. Forward pruning is
a bit more aggressive. An algorithm that
uses forward pruning will not select a pruned node again
in later iterations, unless the score drops drastically. 
The problem with these ideas is to find a method to decide reliably 
which move is bad. Some moves may look bad at first sight, but turn
out to be a winning move after a deeper search. 

\index{forward pruning}%
Comparing the quality of selective deepening and forward pruning
enhancements is 
hard. Tree size is 
no longer the only parameter. With these schemes the quality of the decision
becomes important as well, since one could say that search extensions try
to mend deficiencies in the evaluation function by searching
deeper. Selective deepening adds a second dimension to performance
comparisons. In fixed-depth searching, improvements mean more cutoffs
in the search; algorithm performance measurements are only a matter of
counting nodes. 

One way to 
compare selective deepening algorithms is by having different versions
of the program play 
matches. A problem with this approach is that deficiencies in both
versions can distort the outcome. Another problem is that these
matches take up a large amount of time, especially if they are to be
played with regular tournament time controls.

\subsubsection{{\AB} Search Extensions}\index{search!extensions}%
Most game-playing programs are still based on the {\AB}
algorithm. In this setting search extensions are often introduced to
make up for features that a static evaluation cannot uncover. A well-known
problem of fixed-depth searching is the horizon effect \cite{Berl74}. Certain
positions are more dynamic than others, which makes them hard to
assess correctly by a static evaluation. For example, in chess
re-captures and check evasions can change the static assessment of a
board position drastically. The inability of an evaluation function
to assess tactical features in a position is called the horizon
effect. To reduce it, most
\index{quiescence search}%
\index{search!quiescence}%
programs use {\em quiescence search\/} \cite{Beal89,Frey77}. A quiescence
search extends the search at a leaf position until a quiet position is
reached. In chess ``quiet'' is usually defined as no captures present
and not in check.

% Other methods go a step further towards the ideal of selective
% searching.  Over the years a number of techniques have been developed
% to selectively deepen the search along promising lines. Contrasting
% them to the more conservative backward pruning, they are sometimes
% called forward pruning, meaning the disregard of a node before it has
% been searched. Strictly speaking any method
% that searches one move a ply deeper in favor of another uses forward
% pruning, including quiescence search.
% However, the term forward pruning is usually reserved for the
% newer, more aggressive, methods that will be discussed next. 

Going a step further are
\index{singular extensions}%
\index{null-move heuristic}%
techniques such as singular extensions  and null-moves.  Singular extensions
try to find, through \index{search!shallow}%
shallow searches, a single move
in a position that is more promising than all others. This node is
then searched more deeply (details can be found in
\cite{Anan90b,Anan90}). A null-move is a move in which the side to play
passes (the other side gets two moves in a row). In most positions
passing is worse than any other move. The idea behind null-moves is to
get a lower 
bound  on a position in a cheap way, because all
other moves are 
presumably better. (This is not true for all
games. For example, in checkers and 
Othello null-moves do not work. Also, when in zugzwang in chess (a
series of positions with no good move),
passing can be beneficial.) 
If this lower bound causes a cutoff, then it has been found in a cheap
way. If it does not, then little effort is wasted.
Many people have
experimented with different uses of the null-move. More can be found in
\cite{Adel75,Beal89,Beal90,Goet88,Pala85,Scha87}.

Another source of information on which nodes to prune is the result of the
shallow search itself. A successful example of this idea is the
ProbCut \index{ProbCut}%
algorithm, formulated by Buro \cite{Buro95}. 
ProbCut performs a shallow null-window {\AB} 
search to find nodes whose value will fall with a certain probability
outside the search window.  These nodes can be left out of the
full-depth search. Although many researchers
have used similar ideas---sometimes for move ordering, sometimes for
selective search, sometimes for forward pruning---Buro was the first
to use statistical
analysis to determine the most effective values for the search windows.

An open problem in the use of selective search methods and
\index{search!inconsistencies}%
\label{searchinconsistencies}%
transposition tables are {\em search inconsistencies}, mentioned by
Schaeffer \cite{Jonathan}. Searching a position to different depths
can yield different values. Transpositions can cause nodes to return
results for a deeper search.
% 
% For example, due to a transposition, a
% node returns a result that is based on a deeper search than that
% of its brothers. 
This can cause cutoff and score problems.
% errors in move ordering, cutoff
% decisions and the history heuristic. 
% Finding a deeper result thus causes a wrong decision.
Other than not using the deeper search result (which is wasteful) we
do not know of a method to solve this problem.

A related problem of selective deepening is that one can use too much of
it \cite{Jonathan}. Selectively extending certain lines causes the
\label{sec:searchincon} search to become uneven. As an algorithm is
made more selective, 
comparing scores for different depths gives more and more
inconsistencies. Furthermore, moves that  look less promising at first
sight are not given enough resources, causing the program to miss good
moves that a full-width search (or at least less selectivity) with the
same resources would have found. Thus, selective search should be used
with care.

% Suppose a position would cause a cutoff if
% it returned a value of 12 or higher. If a deep search result, due to
% a transposition, would yield a value of 15, then the brothers of this
% position will be cutoff. However, it can very well be that the search
% result for the regular search depth would have returned 8, not causing
% a cutoff. It is also possible that one of the brothers will have a
% value of 13 for the regular search depth. In this scenario the
% transposition has caused a a wrong decision. 

There has been some debate over the pros and cons of selective
searching, ever since Shannon's original article \cite{Shan50}. An
interesting historical account is given by Beal \cite{Beal90}, which
we summarize in the following. In the 
early days of computer chess the lack of power of computers almost
forced researchers to use forward pruning
to achieve reasonable
results. For example, Bernstein's program of around 1957 \cite{Bern58}
\index{Bernstein}%
looked at the \index{best 7 moves}%
``best 7'' moves at every node. Ten years later Greenblatt's \index{Greenblatt}%
program \cite{Gree67}, using carefully chosen quiescence rules, 
became the best of its time.
By 1973 computers had grown powerful enough to make fixed-depth
full-width searching the winning technique in the program Chess 4.0
\index{Chess 4.0 (program)}%
\cite{Slat77}, although a quiescent search 
at the leaves of the fixed-depth tree remained an important component. 

The 1970's and 1980's showed interesting research into selective
search algorithms  \cite{Adel75,Berl79,McAl88,Pala85}, as well as the drive
towards using more powerful computers, such as parallel computers and
special hardware
\cite{Cond82,Ebel87,Hsu89,Mars85,Scha89a}.  The strongest
programs emphasized full-width searching based on powerful hardware.
These two trends have continued into the 1990's. Programs running on
massively parallel computers, \index{Deep Thought}%
such as Deep Thought \cite{Hsu90},
*Socrates \cite{Joer94,Kusz94}, \index{*Socrates}%
\index{Zugzwang}%
Zugzwang \cite{Feldmann93,Feld90a} and Frenchess
\index{Frenchess}%
\cite{Weil95} compete 
against 
commercially available 
programs running on personal computers, such as Fritz, \index{Fritz}%
Chess Genius, \index{Chess Genius}%
and WChess. \index{WChess}%
These
programs use highly  tuned  
aggressive selective deepening and forward pruning techniques, often
based on null-moves 
and shallow searches, combined with lightweight \index{evaluation function}%
evaluation functions. After the successes of full-width search \index{search!full-width}%
in the 1970's and 1980's, the fact that personal computer programs
are not crushed by the parallel power is an indication of the
viability of selective search techniques, although we should not
forget that personal computers have become quite fast as well, over the
years. Also, there is some evidence that the utility of searching an
extra ply deeper is less for deeper searches. For deep searches this
increases the relative importance of a good, highly 
tuned, evaluation function \cite{Scha93}.

\section{Alternative Algorithms}\label{sec:alt}\index{Alpha-Beta!alternatives to}%
The {\AB} tree-searching algorithm 
has been in use since the end of the 1950's.
No other minimax search algorithm has achieved
the wide-spread use in practical applications that {\AB} has. 
Thirty years of research has found ways of improving the
algorithm's efficiency,
and variants such as {\NS} and
{\PVS} are quite popular.
Interesting alternatives to depth-first searching,
such as breadth-first and best-first strategies,
have been largely ignored.

In this section we will look at these alternative algorithms.

\subsection{Best-First Fixed-Depth: {\SSS}}\label{sec:sssxx}
\index{search!best-first}\index{SSS*}\index{Stockman}%
In 1979 Stockman introduced {\SSS}, a fixed-depth full-width algorithm
which looked like a radically different approach from {\AB} for searching
minimax trees \cite{Stoc79}.
It builds a tree in a best-first fashion by
visiting the most promising nodes first.
{\AB}, in contrast, uses a depth-first, left-to-right traversal of the tree.
Intuitively, it would seem that a best-first strategy should
prevail over a rigidly ordered depth-first one.
Stockman proved this intuition true: {\SSS} dominates {\AB};
it never evaluates more leaf nodes than {\AB}. Consequently, {\SSS} has
drawn considerable attention in the literature (for example,
\cite{Bhatta93,Camp81,Camp83,Kuma83,Kuma84,Kuma88,Leif85,Mars82a,Mars87,Pijl91,Pijls90,Rein89,Rein94b,Roiz83}).
%When both algorithms are given a perfectly ordered game tree to search,
%they visit the same leaves,
On average {\SSS} evaluates considerably fewer leaf nodes.
This has been repeatedly demonstrated in the literature by numerous
simulations \index{simulation}%
(for example,
\cite{Kain91,Mars87,Musz85,Rein89,Rein94b,Roiz83}). 

% There exists a dual formulation of {\SSS}, which is created by
% exchanging max by min, descending by ascending, and $+\infty$ by
% $-\infty$. In the literature this version, called {\DUAL}, is further
% analyzed  
% \cite{Kuma84,Rein89}. 

\begin{figure}
\vspace{-0.4cm}
{\small
\begin{tabbing}
mmm\= \kill 
{\bf Stockman's {\SSS}} (including Campbell's correction \cite{Camp81}) \\ 
(1) \> Place the start state $\langle n = root, s = \mbox{LIVE}, \hat{h} =
       +\infty\rangle$ on a list called OPEN.\\
(2) \> Remove from OPEN state $p = \langle n, s, \hat{h}\rangle$ with largest merit
$\hat{h}$. OPEN is a\\ \> list kept in non-decreasing order of merit, so
$p$ will be the first in the\\ \> list.\\
(3) \> If $n = root$ and $s = \mbox{SOLVED}$ then $p$ is the goal state
so terminate with\\ \> $\hat{h} = f(root)$ as the minimax evaluation of the
game tree.\\ \> Otherwise continue.\\
(4) \> Expand state $p$ by applying state space operator $\Gamma$ and
queuing all\\ \> output states $\Gamma(p)$ on the list OPEN in merit
order. Purge redundant\\ \> states from OPEN if possible. The specific
actions of $\Gamma$ are given in \\ \> the table below.\\
(5) \> Go to (2)
\end{tabbing}
\begin{tabular}{lll}
\multicolumn{3}{c}{State space operations on state$\langle
n,s,\hat{h}\rangle$ (just removed from top of OPEN)} \\ \hline \hline
$\Gamma$ & Conditions satisfied & Actions of $\Gamma$ in creating \\
  & by input state $\langle n, s, \hat{h}\rangle$ & new output
states\\ \hline \hline
$-$ & $s = $ SOLVED & Final state reached, exit algorithm \\
 & $n = $ ROOT & with $g(n) = \hat{h}$.\\ \hline
1 & $s = $ SOLVED & Stack $\langle m = \mbox{parent}(n), s,
\hat{h}\rangle$ on OPEN 
list. \\
& $n \neq $ ROOT & Then purge OPEN of all states $(k,s,\hat{h})$ \\
& type$(n)$ = MIN & where $m$ is an ancestor of $k$ in the game
tree.\\ \hline
2 & $s = $ SOLVED & Stack $\langle$next$(n)$, LIVE, $\hat{h}\rangle$ \\
  & $n \neq$ ROOT & on OPEN list \\
  & type$(n) = $ MAX & \\
  & next$(n) \neq $ NIL \\ \hline
3 & $s = $ SOLVED & Stack $\langle$parent$(n), s, \hat{h}\rangle$ \\
  & $n \neq$ ROOT & on OPEN list \\
  & type$(n) = $ MAX & \\
  & next$(n) = $ NIL \\ \hline
4 & $s = $ LIVE & Place $\langle n, $ SOLVED, min$(\hat{h}, f(n))\rangle$ on \\
  & first$(n) = $ NIL & OPEN list (interior) in front of all states of \\
  &             & lesser merit. Ties are resolved left-first. \\ \hline
5 & $s = $ LIVE & Stack $\langle$first$(n), s, \hat{h}\rangle$ \\
  & first$(n) \neq $ NIL & on (top of) OPEN list.\\
  & type(first$(n)) = $ MAX \\ \hline  
6 & $s = $ LIVE & Reset $n$ to first$(n)$.\\
  & first$(n) \neq $ NIL & While $n \neq$ NIL do\\
  & type(first$(n)) = $ MIN & \ \ \ queue $\langle n, s,
\hat{h}\rangle$ on top of OPEN 
list \\ 
  &   & \ \ \ reset $n$ to next$(n)$ \\ \hline \hline

\end{tabular}
\vspace{-0.3cm}
\caption{Stockman's {\SSS}
\protect\cite{Pear84,Stoc79}}\label{fig:sss}\label{fig:ssscodeex}  
}
\end{figure}

The code of {\SSS} is shown in figure~\ref{fig:sss} (taken from
\index{SSS*!code}%
\cite{Camp81,Pear84,Stoc79}). 
% {\SSS} is a difficult algorithm to understand, as
% can be appreciated by looking at the
% code.
{\SSS} works by manipulating a list of 
nodes, the OPEN list, using six ingeniously inter-locking cases
of the so-called $\Gamma$ operator.
The nodes have a status associated with them, 
either live or solved, and a merit, denoted $\hat{h}$.
The OPEN list is sorted in descending order,  \index{OPEN list}%
so that the entry with highest merit (the ``best'' node) is at the front,
to be selected for expansion. {\SSS} is a variation of AO*, an
algorithm for searching AND/OR graphs, or \index{AO*}\index{A*}%
\index{AND/OR tree}\index{problem-reduction space}\index{state-space}%
problem-reduction-spaces. The more widely known A* algorithm searches
plain OR graphs, 
or state-spaces
\cite{Nils71,Nils80,Pear84}. 

% In the next chapter  we will present a clearer formulation that has the added
% advantage of solving a number of obstacles that have hindered {\SSS}'s
% use in practice in game-playing programs. Appendix~\ref{app:sssex}
% contains an example showing how {\SSS} traverses a tree.
% 

At first sight the code in figure~\ref{fig:sss} may look
overwhelmingly complex. However, once we see through the veil of the
code, we 
find that {\SSS}'s idea of best-first node selection is quite
straightforward. 
% {\SSS} finds the minimax value in a sequence of passes. In each pass,
% a max solution tree is constructed, that defines an upper bound on the
% minimax value of the root. Subsequent passes select the ``best'' open
% node for expansion. 
{\SSS} finds the minimax value through a successive lowering of an upper
bound on it. As 
the example of appendix~\ref{app:sssex} illustrates, a number of
passes can be distinguished in this process. In each pass the value of
a new (better) upper 
bound is computed. What makes {\SSS} such an interesting algorithm, is
that in each of these passes it  selects nodes in a best-first
order.

\begin{figure}
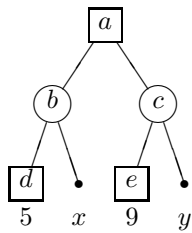

{\small
 \begin{Tree}
   \node{\external\type{square}\cntr{$d$}\bnth{$5$}}
   \node{\external\type{dot}\bnth{$x$}}
   \node{\cntr{$b$}}
   \node{\external\type{square}\cntr{$e$}\bnth{$9$}}
   \node{\external\type{dot}\bnth{$y$}}
   \node{\cntr{$c$}}
   \node{\type{square}\cntr{$a$}}
\end{Tree}
\hspace{5.3cm}\usebox{\TeXTree}
}
\caption{Which is ``Best:'' $x$ or $y$?}\label{fig:best}
\end{figure}
``Best'' in the {\SSS} sense is defined in a simple 
\index{best, ``SSS*-best''}\index{best-move example}%
manner. The upper bound of each pass is defined by a max solution
\index{upper bound}\index{solution tree, max}%
tree (possibly under construction), whose minimax value is the maximum
of its leaves. The leaf (or leaves) defining this value is 
the critical leaf at the end of the principal variation. See
figure~\ref{fig:best} for 
an example. In the figure the \index{principal variation}%
PV consists of nodes $a, c, e$. The value at the root is an upper
bound of 9. Expanding node $x$ will not change the minimax
value of the tree. Only expanding the brother of the critical leaf $e$
might give a better (sharper) upper bound, if its value happens to be less than
9. Thus, 
the {\SSS}-best node to expand in figure~\ref{fig:best} is  node
$y$. (If there is more than one critical leaf, {\SSS} selects the
left-most.) 

The ``best'' node in {\SSS} terms is the left-most and
deepest node whose expansion can possibly lower the upper bound.
% Now, the max solution tree that {\SSS} creates is part of the minimal
% tree that {\AB} would traverse. 
Intuitively, the fact that {\SSS} dominates {\AB} is based on the
property of the \index{domination}% 
best node that it cannot be cut off by some others: node $x$ can
possibly be cut off by  
node $y$ (for example, if $y$ would get value $8$) however, node $y$
cannot be cut off by expansion of node $x$.  
Any node that {\AB} could cutoff is
not ``best'' in the {\SSS} sense. 
% This is the idea behind Stockman's
% proof that {\SSS} never selects a node
% that {\AB} will cutoff. 
% Appendix~\ref{app:equiv} provides a more detailed analysis of the
% relation between {\AB} and {\SSS}.

Appendix~\ref{app:sssex} contains a detailed example of how {\SSS}
finds the minimax value of a tree in a best-first manner.
% Confused by the example? Don't worry, you're not alone.
% {\SSS}, as formulated by Stockman, has several problems.

\subsection{The Conventional View on {\SSS}}\label{sec:sssviewxx}
The literature lists a number of problems with {\SSS}.
\index{SSS*!conventional view}\index{SSS*!list of problems}%
First, it takes considerable effort to understand how the
algorithm works 
from a minimax point of view
and still more to see its relation to {\AB}, as a glance at
figure~\ref{fig:sss} and the examples may have suggested. 
Although it is possible to follow the individual steps in the example, one of
the problems with 
Stockman's  formulation is that it is 
hard to see what is ``really'' going on.  The high level explanation
provided with figure~\ref{fig:best} is not obtained easily.
% To put it differently,
% we would like to 
% understand it in terms of concepts above the level of which $\Gamma$
% case happens when and does what. 
An understanding at a higher level of abstraction is essential for
those who wish to work with this algorithm. For example, parallelizing
the original Stockman formulation is hard, and getting good results is
even harder, as is shown by a number of \index{SSS*!parallel}%
(simulation) attempts \cite{Amig91,Bhat,Dide92,Kraas,Kuma84,Vorn87}.
%We believe to have found this
%understanding with the sequence of low failing null-window {\AB}
%calls, that create a sequence of upper bounds on the minimax value, by
%refining in each pass a single max solution tree stored in memory.
%{\ABSSS} is a much simpler formulation than Stockman's
%original {\SSS} code. 

Another problem caused by the complexity of the algorithm is that to
our knowledge
nobody has reported measurements on how good {\SSS} performs
in actual high performance
game-playing programs;
all published performance assessments are based on
simulations. 
(One could argue that work published in 1987 by Vornberger and
Monien is an exception \cite{Vorn87}, since chess positions
were used in these measurements. However, one of the authors of the
chess program reports that the work was performed with a
less-than-state-of-the-art chess program, without any enhancements
such as minimal 
windows, transposition tables, the history heuristic, or iterative
deepening \cite{Mysl96}.  As can be expected, the results showed 
that {\SSS} searched substantially less leaf nodes than {\AB}, in
line with the simulation literature. 
Since 1987 the quality of their 
program has increased considerably
\cite{Feldmann93}. Regrettably, they
%\cite{Feldmann93,Feld90res,Scha89comm,Scha90rej}. Regrettably, they
have not reported tests with {\SSS} since then.)
 
Second, a drawback of {\SSS} is its memory usage. \index{memory}%
SSS* maintains an OPEN list,
similar to that found in single-agent best-first search algorithms like A*
\cite{Pear84}.
The size of this list grows exponentially with the depth of the search tree.
This has led many authors to conclude that {\SSS} is effectively disqualified
from being useful for real applications like game-playing programs
\cite{Kain91,Musz85,Roiz83,Stoc79}.

The OPEN list must be kept in sorted order. \index{OPEN list}%
Insert and (in particular) delete/pur\-ge operations on the OPEN list
are slow. They can dominate the execution time of any program using SSS*.
Despite the promise of expanding fewer nodes,
the disadvantages of {\SSS} have proven a significant deterrent in practice.
The general view of {\SSS} then is that:\label{sec:viewsss}\label{sec:sssview}
\begin{enumerate}
\item it is a complex algorithm that is difficult to understand,
\item it has large memory requirements that make the algorithm
impractical for real applications,
\item it is ``slow'' because of the overhead of
maintaining the sorted OPEN list,
\item it has been proven to dominate {\AB} in terms of the number of
leaf nodes evaluated, and 
\item it evaluates significantly fewer leaf nodes than {\AB} in simulations. 
\end{enumerate}
\begin{figure}
{\small
\begin{center}
\setlength{\unitlength}{0.009375in}%
\begingroup\makeatletter\ifx\SetFigFont\undefined
% extract first six characters in \fmtname
\def\x#1#2#3#4#5#6#7\relax{\def\x{#1#2#3#4#5#6}}%
\expandafter\x\fmtname xxxxxx\relax \def\y{splain}%
\ifx\x\y   % LaTeX or SliTeX?
\gdef\SetFigFont#1#2#3{%
  \ifnum #1<17\tiny\else \ifnum #1<20\small\else
  \ifnum #1<24\normalsize\else \ifnum #1<29\large\else
  \ifnum #1<34\Large\else \ifnum #1<41\LARGE\else
     \huge\fi\fi\fi\fi\fi\fi
  \csname #3\endcsname}%
\else
\gdef\SetFigFont#1#2#3{\begingroup
  \count@#1\relax \ifnum 25<\count@\count@25\fi
  \def\x{\endgroup\@setsize\SetFigFont{#2pt}}%
  \expandafter\x
    \csname \romannumeral\the\count@ pt\expandafter\endcsname
    \csname @\romannumeral\the\count@ pt\endcsname
  \csname #3\endcsname}%
\fi
\fi\endgroup
\begin{picture}(220,180)(5,625)
\thinlines
\put(182,680){\circle*{6}}
\put(190,675){\makebox(0,0)[lb]{\smash{\SetFigFont{9}{10.8}{rm}Alpha-Beta}}}
\put(182,695){\circle*{6}}
\put(190,690){\makebox(0,0)[lb]{\smash{\SetFigFont{9}{10.8}{rm}NegaScout}}}
\put( 85,710){\circle*{6}}
\put( 75,640){\vector( 0, 1){140}}
\put( 75,640){\vector( 1, 0){140}}
\multiput( 75,650)(7.74194,0.00000){16}{\line( 1, 0){  3.871}}
\multiput( 75,770)(7.74194,0.00000){16}{\line( 1, 0){  3.871}}
\put( 45,790){\makebox(0,0)[lb]{\smash{\SetFigFont{9}{10.8}{rm}y: time efficiency}}}
\put(225,635){\makebox(0,0)[lb]{\smash{\SetFigFont{9}{10.8}{rm}x: storage efficiency}}}
\put( 72,650){\makebox(0,0)[rb]{\smash{\SetFigFont{9}{10.8}{rm}minimax tree}}}
\put( 72,770){\makebox(0,0)[rb]{\smash{\SetFigFont{9}{10.8}{rm}minimal tree}}}
\put( 75,625){\makebox(0,0)[lb]{\smash{\SetFigFont{9}{10.8}{rm}$O(w^{d/2})$}}}
\put(175,625){\makebox(0,0)[lb]{\smash{\SetFigFont{9}{10.8}{rm}$O(d)$}}}
\put( 80,720){\makebox(0,0)[lb]{\smash{\SetFigFont{9}{10.8}{rm}SSS*}}}
\end{picture}
\end{center}
}
\caption{Performance Picture of Best-First
Search Without Enhancements}\label{fig:mmabnsssspic}
\end{figure}
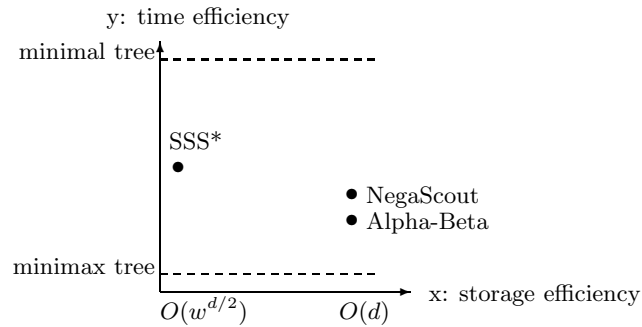
Figure~\ref{fig:mmabnsssspic} illustrates the point that at the cost
of storing a 
solution tree of size $O(w^{d/2})$ in memory, the number of cutoffs can
be improved through a best-first expansion sequence. This picture
shows the conventional view in the literature of {\SSS}, where it is
compared against 
un-enhanced algorithms (note that here we call {\NS} an un-enhanced
algorithm, while elsewhere we consider it to be an {\AB} variant
enhanced with null-windows). In chapter~\ref{chap:exper} an enhanced
version of {\SSS} will be 
compared against enhanced versions of {\AB} and {\NS}.

In the next chapter we present results of our attempts to
understand how and why {\SSS} works and see whether its drawbacks can be
solved. 

Not everybody agrees with our describing {\SSS} as difficult to
understand. \index{SSS*!Stockman's view}%
Stockman's original application had little to do with minimax search or
game-playing. His application was an analyzer for medical data related
to EKG's (electrocardiograms). {\SSS} was used in parsing syntactic
descriptions (AND/OR graphs) of
waveforms to recognize certain types of waves \cite{Stoc77}.
Commenting on our (game-tree) view on {\SSS}, Stockman remarks: ``So, 
you have yet another way and think that my
reasoning was opaque---it just seemed to lay out that way naturally!
Actually, I was competing multiple parse trees against each other in
a waveform parsing application. I wanted an algorithm where the 
currently best parse tree was the one extended and this got me into
SSS*. [$\ldots$] The version of SSS* that I actually implemented in the
thesis could develop trees bottom-up or even middle out and not just
top-down as game-tree searches do.'' \cite{Stockman}.

Stockman views {\SSS} as a problem-reduction search method such as
AO*, in which one searches for the best solution tree. This view is in
line with Nilsson's explanation of ``an ordered search algorithm for
AND/OR trees'' which was later called AO* \index{AO*}\index{Nilsson}%
\index{state-space}\index{problem-reduction space}%
\cite{Nils71,Nils80}. Nilsson stresses the 
difference between the state-space and  problem-reduction-space 
approach as follows: ``The appropriate ordering technique involves
asking `Which is the most promising potential solution tree to
extend?' rather than asking `Which is the most promising node to
expand next?'\,'' \cite[p.~129]{Nils71}. In chapter~\ref{chap:mt} we
will go against this advice and look at {\SSS} asking which {\em
node\/}  to expand next. Applying a state-space view leads in this
case to a reformulation that is much clearer.

\subsection{Variable Depth}\label{sec:vardeep}\index{search!variable depth}%
Sections~\ref{sec:sssxx} and \ref{sec:sssviewxx} presented an
alternative to {\AB} 
in the form of 
the best-first fixed-depth full-width {\SSS} algorithm.
Variable search depth algorithms provide  another alternative to {\AB}.
It has always been felt that one of the biggest drawbacks
of {\AB} is its rigid search depth. Section~\ref{sec:abext} discussed
{\AB} search extensions---enhancements to reduce this drawback while
staying within the {\AB} framework. In this section we briefly review
algorithms that do away with {\AB} completely. 

The
frustration with {\AB} has been well described
in an article by Hans Berliner \cite{Berl79}. \index{Berliner}\index{B*}%
He proposes a more human-like search method, B*, which manipulates two
heuristic bounds, an optimistic bound and a pessimistic 
bound.  Lines of play that are
easily refuted are not searched deeply, whereas more promising lines are
searched more deeply. Furthermore, B* stops searching as soon as the backed-up
value of the pessimistic bound of the best move is greater or equal
to the backed-up value of the optimistic bound. By stopping the
search before the minimax value is known, less nodes have to be
expanded. A problem with B* is to find reliable heuristic bounds. This
has turned out to be an obstacle. 

A second problem of the {\AB} framework is that the evaluation function
returns a single value, without an assessment of the quality of the
evaluation. In certain types of positions the evaluation function may be
quite good at assessing the merit, whereas in other positions the
return value is less trustworthy. 
B*'s optimistic and pessimistic bounds are better, in the sense that a
narrow gap  between the bounds can be interpreted as a high confidence, and a
wide gap signals a low confidence.
The work on B* was extended by Palay to include
probability distributions, leading to an algorithm called PB*
\index{Palay}\index{PB*}\index{McConnell}%
\cite{Pala85}. In a recent paper, Berliner and McConnell showed the
performance of a  new version of B* to approach that of a high-quality
{\AB}-based program \cite{Berl95}. They argue that there is still room for
improvement,  although it is also clear that the current
implementation of the B* idea has become quite complex, when compared
to the original idea in \cite{Berl79}.  

A potential disadvantage of  more elaborate back-up
schemes is that the 
extra information gained by having a backed-up probability
distribution can be offset by the longer time it takes to compute
it. 
However, the idea of backing up more
information than just a single value is quite appealing, and this is
an area of active research.
% Backing up a single integer value is usually faster.
See for example the work of \index{Russell}\index{Wefald}%
\index{Rivest}\index{Baum}\index{Junghanns}%
Russell and Wefald \cite{Russ91}, Ballard \cite{Ball83}, Rivest
\cite{Rive87}, and Baum  
\cite{Baum93}. Junghanns describes related experiments using fuzzy
numbers \cite{Jung95}. 
Some results show an algorithm
beating unenhanced text-book versions of {\AB}, for example
\cite{Baum93,Russ91} (and even \cite{Stoc79}). However, despite
all efforts, we 
do not know of a result that proves the superiority of backing up
probability distributions over programs based on
{\AB}. Beating {\AB}  is not an easy task,
because the yardstick that 
has to be beaten has been considerably enhanced, amongst others by
selective deepening techniques (section~\ref{sec:abext}). 

Another idea for variable depth search is Conspiracy Numbers
\cite{McAl88,Scha90}, and its \index{conspiracy numbers}\index{proof numbers}%
Boolean variant, Proof Numbers \cite{Allis94}.  These algorithms can be
seen as a ``least-work-first'' approach.  For every possible outcome
of a node $n$
these algorithms compute how many nodes in $n$'s sub-tree must change
their value to 
have $n$ take on that value. In Proof Number search the value is either
a 1 or a 0. 
% For highly irregular trees (varying in depth and width)
% these methods often find a solution before {\AB} does. 
These ideas have 
proven to be useful for certain problems. For minimax trees that
are highly irregular---such as chess mating problems, the game of
Awari, and for solving certain games---Proof Numbers can build smaller
trees than {\AB} \cite{Alli94b}. If there is no clear solution to be
found, then Proof Numbers does not perform well.
In \cite{Icca94} a relation between
proof trees and solution trees is discussed. 

Recently another algorithm for performing variable depth minimax search has
been proposed by Korf and Chickering \cite{Korf93,Korf94}. This algorithm,
\index{Korf}\index{Chickering}\index{search!best-first minimax}%
called Best-First Minimax Search (BFMMS), starts by expanding the root
position and evaluating the children. Then it selects the best child
and expands it, after which it evaluates its children, and so on and
so forth. Although both BFMMS and {\SSS} are ``best-first''
algorithms, they traverse the search space in quite different ways.
{\SSS} is in principal a fixed-depth algorithm, BFMMS is inherently
variable-depth. 
{\SSS} expands the {\em brother\/} of
the critical leaf (figure~\ref{fig:best}), BFMMS expands its {\em
children}. 
\index{A*}%
BFMMS appplies a single-agent, A*-like, expansion strategy to the
domain of two-agent search.

Korf and Chickering report first results with this \index{Othello}%
algorithm that show it to perform well on moderate depth simulated 
Othello-like
trees. For deeper searches the trees become too unbalanced. Here
a hybrid algorithm with {\AB}, using BFMMS as a kind of 
{\AB} search extension, seems more effective.

\section{Summary}
Despite the considerable interest in new algorithms for minimax
search, most successful game-playing programs are still based on the 
{\AB} algorithm with enhancements like iterative deepening,
transposition tables, narrow windows, the history heuristic, and
search extensions.  
Many other algorithms seemed or seem promising. One of them is {\SSS}. It is a
relatively  conservative algorithm in that it does not rely on
selective search, just like the original {\AB} algorithm. This makes
performance comparisons between {\AB} and {\SSS} relatively easy,
since one can compare tree sizes. Furthermore, there exist 
theoretical models---notably solution trees---to guide one's thoughts
in trying to understand their search trees. 

% The main problem
% of {\SSS} is that it is hard to understand. In the next chapter we
% will  show how this can be solved.

 % litt
\cleardoublepage
% \chapter{MT---A Deceptively Simple Framework for Best-First
\chapter{The MT Framework}\label{chap:mt}  \index{MT framework}
% \markboth{Chapter 3}{Chapter 3}
%\markboth{Chapter 3}{The MT Framework}
% \chapter{{\SSS} = $\alpha$-$\beta$ + TT}\label{chap:absss}
% 
% 
% SSS* problems, RecSSS*, SSS-2 as solutions, now ABSSS solves it
% - example absss, expand on example, give more background of 
% how people see SSS*, and what AB is doing in it, how it does it,
% giving my kind of proof
% talk about TT, best-first seq
% 
% 
% 
% 3 MT -- the framework that makes it all much clearer
%   - problems with SSS*, solution with AB+TT, and how it works (bf)
%   - how ABSSS solves SSS*'s problems, incl example
%   - MTD(f), ideas, intuition
%   - storage
%   - performance
% 
% 

The previous chapter showed how the null-window idea is used in {\NS}.
This chapter takes the idea further.
In  section~\ref{sec:absss} we present a reformulation of {\SSS} that uses
null-window {\AB} calls.
It has the 
advantage of solving a number of obstacles that have hindered {\SSS}'s
use in game-playing programs.
The reformulation is based on the {\AB}
procedure. It examines the same leaf nodes in the same order as
{\SSS}. It is called {\ABSSS} and the code is shown  in
figure~\ref{fig:sss2}.   

% We will discuss the relationship between {\SSS} and
% {\AB}. Appendix~\ref{app:absssex} contains a detailed example.
% , we will stress 
% higher-level concepts like bounds and solution trees. Since each
% null-window {\AB} search traverses a solution tree, this is easier to
% see in {\ABSSS} than in Stockman's formulation with six different
% operators. 

In section~\ref{sec:mt} we will generalize the ideas behind {\ABSSS}
into a new framework that elegantly ties together a number of
algorithms that are perceived to be dissimilar. 

Results from this chapter have been published in \cite{Plaa95a,Plaa95b}.

\section{\ABSSS}\label{sec:absss} \index{MT-SSS*} \index{SSS*}
\index{Alpha-Beta} 
{\SSS} finds the minimax value through a successive lowering of an upper
bound.
In each of a number of passes it  selects nodes in a best-first
order. 
The upper bound of each pass is
defined by a max solution\index{upper bound}\index{solution tree, max}\index{search!best-first}\index{principal variation} 
tree, whose minimax value is the maximum of its leaves. The ``best''
node to expand next is a brother of the left-most leaf 
defining this value---the critical leaf at the end of the principal
variation.  (See 
also figure~\ref{fig:best}.) 

Here we reformulate {\SSS} as a sequence of upper bounds, generated
by null-window {\AB} calls.
The basis of our reformulation of {\SSS} is the realization that
a conventional version of {\AB} that uses transposition tables,
such as the one in figure~\ref{fig:ttabcode} or in
\cite{Mars86b,Mars82a}, can be used to traverse the principal
variation. The node that
it will expand next (if called on the same max solution tree with the
right search window) is
precisely the one that is ``{\SSS}-best,'' the one that {\SSS}
\index{best, ``SSS*-best''} would
select.

\begin{figure}
{\small
\begin{tabbing}
mmmmmmmm\=mm\=mm\=mmmmmmmmmmmmmmmmmmmmm\=mm\=mm\=  \kill
\>{\bf function} MT-SSS*$(n) \rightarrow f$; \\
\>   \>$g := +\infty$;\\
\>   \>{\bf repeat}                      \\
\>   \>   \>$\gamma := g$;        \\ 
\>   \>   \>$g := $ MT($n,  \gamma$);\\
\>   \>   \>/* the call $g := $ Alpha-Beta($n, \gamma -1, \gamma$); is
equivalent */\\
\>   \>{\bf until} $g = \gamma$;          \\
\>   \>{\bf return} $g$;                 
\end{tabbing}}
  \caption{{\SSS} as a Sequence of Memory-Enhanced {\AB} Searches} \label{fig:sss2}
\end{figure}

{\AB}'s postcondition on page~\pageref{sec:pcab} shows that for a fail low,
a max solution tree is constructed and an upper bound is
returned. Assuming that the solution trees that {\AB} and {\SSS}
generate are the same, then all the ingredients for a reformulation of
{\SSS} are available. The code in figure~\ref{fig:sss2} implements the idea of
calling {\AB} to generate a sequence of upper bounds. By using a
window of $\langle \beta-1, \beta\rangle$, where $\beta$ is the
previous upper bound, a return value of $g \leq \beta - 1$ is a fail low where
$g$ equals the new upper bound, and a return value of $g \geq \beta$
is a fail high where $g$ equals a lower bound. (Since $\beta$ is always
an upper bound, the $g > \beta$ part of $g \geq \beta$ will never
occur, assuming there are no search inconsistencies, see
section~\ref{sec:searchincon}.).\index{fail low}\index{fail high}\index{narrow search window} The example in 
appendix~\ref{app:absssex} shows how {\AB} 
constructs max solution trees by selecting nodes best-first.
A full formal proof of the equivalence of {\ABSSS} and {\SSS} can be
found in \cite{Pijl95}, an extended outline in \cite{Plaa95b,Plaa95e}.
An ``informal proof'' describing in detail the link between the six {\SSS}
$\Gamma$ cases and the {\AB} code can be found in
appendix~\ref{app:equiv}. 

% The idea
% behind the {\ABSSS} reformulation is that these upper bounds can also
% be found using a
% null-window {\AB} call. The null-window call creates  the
% solution tree. This solution tree is stored in memory, so that it can
% be refined in later passes. 

% {\AB} is used to construct solution trees. 
% The postcondition of the {\AB} procedure in section~\ref{sec:pcab} on
% page~\pageref{sec:pcab} shows that 
% using outcome 2, we can have it return an upper
% bound if we make it fail low. To create a fail low, {\AB} must be called
% with a search window greater than any
% possible leaf node value. Since both {\AB} and {\SSS}
% expand the children of a node in a left-to-right order, {\AB}, when called
% with such a window, will find the same upper
% bound, and expand the same max solution tree, as {\SSS}. 
% 
\begin{figure}
{\small
    \begin{tabbing}
mmmmm\=mm\=mm\=mm\=mm\=mmmmm\=mmmmmmmmmmmmmmmmmm\=  \kill
/* MT: storage enhanced null-window {\AB}$(n, \gamma-1, \gamma)$.\>\>\>\>\>\>\>*/\\
/* $n$ is the node to be searched, $\gamma-1$ is  $\alpha$ and
$\gamma$ is  $\beta$ in the call.\>\>\>\>\>\>\>*/\\
/* 'Store' saves search bound information in memory;\>\>\>\>\>\>\>*/\\
/* 'Retrieve' accesses these bounds.\>\>\>\>\>\>\>*/\\
{\bf function} MT$(n, \gamma) \rightarrow g$; \\

\> {\bf if} $n$ = leaf {\bf then } \\
\>  \> retrieve $n.f^-,n.f^+$; /* non-existing bounds are $\pm\infty$ */\\
\>  \> {\bf if} $n.f^- = -\infty$ {\bf and} $n.f^+ = +\infty$ {\bf then} \\
\>   \> \> $g :=$ eval$(n)$;\\
\>   \> {\bf else if} $n.f^+ = +\infty$ {\bf then} $g := n.f^-$; {\bf else}
                                                 $g := n.f^+$;\\
 
\> {\bf else if} $n$ = max {\bf then}\\
\>   \> $g := -\infty$;\\
\>   \> $c := $ firstchild$(n)$;\\
\>   \> /* $g \geq \gamma$ causes a beta cutoff ($\beta = \gamma$) */ \\
\>   \> {\bf while} $ g < \gamma$ {\bf and} $c \neq \bot$ {\bf do}\\
 
\>   \> \> retrieve $c.f^+$;\\
\>   \> \> {\bf if} $c.f^+ \geq \gamma $ {\bf then} \\
\>   \> \> \> $g{\prime}:=$ MT$(c, \gamma)$;\\ 
\>   \> \> {\bf else}  $g{\prime}:= c.f^+$;\\ 
\>   \> \> $g :=$ max$(g,g{\prime})$;\\
\>   \> \> $c := $ nextbrother$(c)$;\\
 
\> {\bf else} /* $n$ is a  min node */ \\ %{\bf then}\\
\>   \> $g := +\infty$;\\
\>   \> $c := $ firstchild$(n)$;\\
\>   \> /* $g < \gamma $ causes an alpha cutoff ($\alpha = \gamma
            -1$) */ \\
\>   \> {\bf while} $ g \geq \gamma$ {\bf and} $c \neq \bot$ {\bf do}\\
\>   \> \> retrieve $ c.f^-$;\\
\>   \> \> {\bf if} $c.f^- < \gamma$ {\bf then} \\
\>   \> \> \> $g{\prime}:=$ MT$(c, \gamma)$; \\
\>   \> \> {\bf else} $g{\prime} := c.f^-$;\\
\>   \> \> $g :=\min(g,g{\prime})$; \\
\>   \> \> $c := $ nextbrother$(c)$;\\
\> /* Store one bound per node. Delete any old bound (see
                page~\pageref{sec:boundsremark}). */ \\ 
\> {\bf if} $g \geq \gamma$ \={\bf then} $n.f^- := g$; store $n.f^-$;\\
\>                          \>{\bf else} $n.f^+ := g$; store $n.f^+$;\\

\> {\bf return} $g$;
\end{tabbing}
 }
  \caption{{\MT}: Null-window Alpha-Beta With Storage for Search Results}
  \label{fig:mmab}\index{MT code}
\end{figure}

{\ABSSS} only uses null-windows, so there really is no need to carry around
the two $\alpha$ and $\beta$ bounds. 
Figure~\ref{fig:mmab} shows a null-window-only
version of {\AB}, enhanced with
memory.\index{Pearl}\label{Pearl}
Pearl  introduced the procedure {\Test}
\cite{Pear80,Pear82}, a routine that tests whether the value of a sub
tree lies above or below a certain threshold.  We have named our  
procedure {\MT}, for Memory-enhanced {\Test}. {\MT} returns
a bound, not just a Boolean value. It is a\index{Test (procedure)}\index{fail-soft}
{\em fail-soft\/} {\Test}.  
% (We will interchangeably use the terms
% {\MT} and {\AB} procedure to denote the same null-window call. In
% other places the term {\AB} is meant to denote the {\em algorithm\/}
% {\AB}$(n, -\infty, +\infty)$ that finds the minimax value.)

In accessing storage, most {\AB} implementations descend to a child node,
retrieve the bounds stored previously in memory, and check whether an
immediate cutoff 
occurs (see for example the code in figure~\ref{fig:ttabcode}). 
In our pseudo code for {\MT}, we have taken a different
approach. {\MT} checks 
whether a child bound will cause a cutoff before calling itself recursively.
In this way we save a recursive call, and it simplifies 
showing that {\SSS} and {\ABSSS} are equivalent. In
figure~\ref{fig:mtcode} a version of {\MT} is shown where the six
{\SSS} list operations are inserted, which is used to show the equivalence.
However, there is no conceptual difference. Other {\AB}
implementations expand the same nodes, and can be used
just as well, as long as values of stored nodes are treated as in
figure~\ref{fig:ttabcode}. 

\begin{figure}
{\small
\begin{tabbing}
mmmmmmmm\=mm\=mm\=mmmmmmmmmmmmmmmmmmmmm\=mm\=mm\=  \kill
\>{\bf function} MT-DUAL*$(n) \rightarrow f$; \\
\>   \>$g := -\infty$;\\
\>   \>{\bf repeat}                      \\
\>   \>   \>$\gamma := g$;        \\ 
\>   \>   \>$g := $ MT($n, \gamma + 1$);\\
\>   \>   \>/* the call $g := $ Alpha-Beta($n, \gamma, \gamma + 1$); is
equivalent */\\
\>   \>{\bf until} $g = \gamma$;          \\
\>   \>{\bf return} $g$;                 
\end{tabbing}}
  \caption{{\DUAL} as a Sequence of Memory-Enhanced {\AB} Searches} \label{fig:dual2}\index{MT-Dual*}
\end{figure}

One of the problems with Stockman's original {\SSS}
formulation is that we found it very 
hard to understand what is ``really'' going on. 
Part of the reason of the problem is the iterative nature of the
algorithm.  
This has been the motivation behind the development of other
algorithms, notably RecSSS*\index{RecSSS*} \cite{Bhatta93} and SSS-2\index{SSS-2} \cite{Pijls90}, which
are recursive formulations of 
{\SSS}. Although clarity is a subjective issue, 
it seems simpler to express {\SSS} in terms of a well-understood algorithm
({\AB}), rather than inventing a new formulation.
Comparing the
figures~\ref{fig:sss} and \ref{fig:sss2} shows why we believe to have
solved this
point. Furthermore, figure~\ref{fig:dual2}, which gives the code for
our reformulation of {\DUAL}, shows the versatility of this
formulation. In section~\ref{sec:mt} we pursue this point further by
presenting a generalization of these codes.

% show the general functions (not very long, since most of the
% explaining of nws+tt has already been done in the background
% chapter)
% And there's more...
% MT (NWS) based searching, MT drivers
% \input ai-sec5 % from MT sec2,3
% don't forget that nws surpasses wws story

\section{Memory-enhanced Test: A Framework}\label{sec:mt}
The relatively simple and well-known concept of
null-window 
{\AB} search is 
powerful or versatile enough to create the best-first behavior of a complicated
algorithm. 
This section introduces a generalization of the
ideas behind {\ABSSS}, in the form of a new framework
for best-first minimax algorithms. To put it succinctly: this framework
uses {\em depth-first\/} procedures to implement {\em best-first\/} algorithms.
Memory is used to pass on previous
search results to later passes, allowing selection of the ``best''
nodes based on the available information from previous passes.

The previous section showed how null-window {\AB} searches can be used as
an efficient method to compute bounds, and thus form the core of our
{\SSS} reformulation. 
So far, we have discussed the following two mechanisms to be used in
building efficient algorithms: 
\begin{enumerate}
\item 
null-window searches cut off more nodes than wide search windows, and

\item 
we can \index{memory}%
use storage to glue multiple passes of null-window calls
together, so that they can be used to home in on the minimax value,
without re-expanding  nodes searched in previous passes, creating a
best-first expansion sequence. 
\end{enumerate}
A general driver routine to call {\MT} repeatedly can be constructed.
One idea, {\SSS}'s idea, for such a driver is  to start at an
upper bound 
for the minimax value, $f^+ = +\infty$.
Subsequent calls to {\MT} can lower this bound until the minimax value
is reached, as shown in
figure~\ref{fig:sss2}.

Having seen the two drivers for {\MT} in figure~\ref{fig:sss2} and
\ref{fig:dual2}, 
the ideas can be encompassed in a generalized driver routine.
The driver provides a series of calls
to {\MT} to refine bounds on the minimax value succesively.
The driver code can be parameterized so that one
piece of code can construct a variety of algorithms.
The three parameters needed are:
\begin{itemize}
\item {\em n}\\ The root of the search tree.
\item {\em first}\\ The {\em first} starting bound for {\MT}.
\item {\em next}\\ A search has been completed.
Use its result to determine the {\em next} bound for {\MT}.
\end{itemize}
Using these parameters, an algorithm using our {\MT} driver, {\MTD},
can be expressed as {\em MTD(n, first, next)}. 
The last parameter is not a value but a piece of code.
The  pseudo code of the driver can be found in figure~\ref{fig:drivers}.
%The constant $\varepsilon$ is introduced to create a null-window out
%of one input bound. Assuming integer valued leaves, $0 < \varepsilon < 1$.
A number of interesting algorithms can 
be constructed using {\MTD}, of which we present the following
examples (see also figure~\ref{fig:sssfriends}):
\begin{figure}
{\small
  \begin{center}
    \leavevmode
    \input{sssfrnd2.pic}    
  \end{center}
}
  \caption{{\MT}-based Algorithms}
  \label{fig:sssfriends}
\end{figure}
\begin{figure}
{\small
\begin{tabbing}
mmmmmmmmm\=mm\=mm\=mm\=mmmmmmmmmmmmmmmmmmmmm\=mm\=mm\=  \kill
\> {\bf function} MTD($n$, first, next) $ \rightarrow f$; \\
\> \>$f^+ := +\infty;$ $ f^- := -\infty$; \\
\> \>$bound :=$ first;\\
\> \>{\bf repeat}                      \\
\> \>   \>$g := $ MT$(n, bound)$;\\
\> \>   \>{\bf if} $g < bound$ {\bf then} $f^+ := g$ {\bf else} $f^- := g$;\\
\> \>   \> /* The {\em next\/} operation must set the variable {\em bound}\ */\\
\> \>   \>next;        \\ 
\> \>{\bf until} $f^- = f^+$;          \\
\> \>{\bf return} $g$;                 
\end{tabbing}}
  \caption{A Framework for {\MT} Drivers} \label{fig:drivers}
\end{figure}

\begin{figure}
{\small
\begin{tabbing}
mmmmmmmmm\=mm\=mm\=mmmmmmmmmmmmmmmmmmmmm\=mm\=mm\=  \kill
\> {\bf function} MTD($n, f) \rightarrow f$; \\
\> \>$g := f$; \\
\> \>$f^+ := +\infty;$ $ f^- := -\infty$; \\
\> \>{\bf repeat}                      \\
\> \>   \>{\bf if} $g = f^-$ {\bf then} $\gamma := g + 1$ {\bf else}
$\gamma := g$;\\
\> \>   \>$g := $ MT($n, \gamma$);\\
\> \>   \>/* $g := $ Alpha-Beta($n, \gamma -1, \gamma$) is equivalent */\\
\> \>   \>{\bf if} $g < \gamma$ {\bf then} $f^+ := g$ {\bf else} $f^- := g$;\\
\> \>{\bf until} $f^- = f^+$;          \\
\> \>{\bf return} $g$;                 
\end{tabbing}}
  \caption{{\MTDf}} \label{fig:mtdfcode}
\end{figure}

\index{MTD}%

\begin{itemize}
  \item  \index{MTD$(+\infty)$}%
        {\MTD}$(n, +\infty, bound := g)$\\
        This is just {\ABSSS}. For brevity
we call this driver {\MTDp}.

  \item \index{MTD$(-\infty)$}\index{MT-Dual*}\index{Dual*}%
        {\MTD}$(n, -\infty, bound := g + 1)$\\
        This is {\ABDUAL}, which we refer to as {\MTDm}.

  \item \index{MTD$(f)$}%
        {\MTD}$\big(n, $ approximation, {\bf if} $g < bound$ {\bf then}
        $bound := g$ {\bf else} $bound := g + 1\big)$\\ 
        Rather than arbitrarily using an extreme value as a starting point,
        any information on where the value is likely to lie can be used
        as a better approximation. (This assumes a relation between
        start value and search effort that is discussed in
        section~\ref{sec:start}.) 
        Given that iterative deepening is used in many application domains,
        the obvious approximation for the minimax value is the result of
        the previous iteration.
        This algorithm, which we call {\MTDf},
        can be viewed as starting close to $f$, and then doing
        either {\SSS} or {\DUAL}, skipping a large part of their
        search path.  Since the generic {\MTD} code may be 
        confusing at first sight, we give {\MTDf}'s pseudo code
        (slightly reorganized to make it look better) in
        figure~\ref{fig:mtdfcode}. This figure also facilitates a
        direct comparison 
        with figures~\ref{fig:sss2} and \ref{fig:dual2}.

  \item  \index{MTD(bi)}%
        {\MTD}$\big(n, \lceil$average$(+\infty, -\infty)\rceil, bound :=
        \lceil$average$(f^+, f^-)\rceil\big)$ \\
        Since {\MT} can be used to search from above ({\SSS}) as well as
        from below ({\DUAL}), an other try is to bisect the interval and
        start in the middle.  Since each pass produces an upper or
        lower bound, we can take some pivot value in between as the
        next center for our search. This algorithm, called {\MTDb} for short, 
        bisects the range of interest, reducing the number of {\MT} calls.
        To reduce big swings in the pivot value, some kind of
        aspiration searching may be beneficial in many application domains
        \cite{Scha89b}. Looking at {\AB}'s postcondition the bisection
        idea comes to mind easily. 
        It is further discussed in
        section~\ref{sec:oldstuff}. 

%         He does not state the link with best-first {\SSS}-like behavior,
%         but does prove that C* dominates {\AB} in the number of
%         leaf nodes evaluated, provided there is enough storage. (C*
%         has also been discussed in \cite{Weil92}.)

  \item \index{MTD(step)}%
        {\MTD}$\big(n, +\infty, bound := \max(f^-_{n} + 1, g -
        \mbox{\em stepsize})\big)$\\
        Instead of making tiny jumps from one bound to the next,
        as in {\MTDp}, {\MTDm} and {\MTDf},
        we could make bigger jumps. By adjusting the value of
        {\em stepsize\/} to some suitably large value, we can reduce the
        number of calls to {\MT}. This algorithm is called {\MTDs}.

  \item {\MTDd}\\
        If we are not interested in the game value itself, \index{MTD(best)}%
        but only in the best move, then a stop criterion suggested by
        Hans Berliner in the B* algorithm can be used \cite{Berl79}. 
        Whenever the lower bound of one move is not lower than the 
        upper bounds of all other moves, it is certain that this must
        be the best move. 
        To prove this, we have to do less work than when we try to
        determine $f$, since no upper bound on the value of the best
        move has to be 
        computed.  We can use either a
        disprove-rest strategy (establish a lower bound on one move and
        then try to create an upper bound on the others) or prove-best
        (establish an upper bound on all moves thought to be inferior
        and try to find a lower bound on the remaining move).
        A major difference between B* and {\MTDd} is that we
        use fixed-depth search, and the bounds are backed-up leaf values,
        instead of two separate heuristic bounds.
        The stop criterion in the ninth line of
        figure~\ref{fig:drivers} must be changed to $f^-_{best
        move} \geq f^+_{other moves}$.
        Note that this strategy has the potential to build search trees
        {\em smaller} than the minimal search tree, because it does not
        prove the minimax value.
       
        The indication which move should be regarded as best,
        and a suitable value
        for a first guess, can be obtained from a previous iteration
        in an iterative deepening scheme.
        This notion can change during the search, which makes for a
        more complicated implementation. 
Section~\ref{sec:bestmove} contains a short analysis of the best-move
cutoff. In
\cite{Plaa94a} we report on tests with this variant. 

\end{itemize}
%In
%figure~\ref{fig:sssfriends} the essence of a number of algorithms
%previously thought 
%to be totally unrelated is concisely captured in one figure. Using
%{\MT} we can abstract from search details, 
%
Note that while all the above algorithms use storage for bounds,
not all of them need to save both $f^+$ and $f^-$ values.
{\MTDp}, {\MTDm} and {\MTDf}
refine a single solution tree at a time. {\MTDb} and {\MTDs}
usually refine a union of two solution trees, where nodes on the
\index{principal variation}\index{both bounds}%
intersection (the principal variation) should store both an upper and
lower bound at the same time (see also a remark on
page~\pageref{sec:remark} and \cite{Pijls93}).  
We refer to section~\ref{sec:space} for data indicating that these
memory requirements are acceptable.

%Figure~\ref{fig:sssfriends} illustrates the different strategies used
%by the above algorithms for converging on the minimax value.
%

Some of the above instances are new, some are not, and some are small
adaptations of known ideas. The value of this framework does not lie so
much in the newness of some of the instances, but in the way how it enables
one to formulate the behavior of a number of algorithms. 
%The {\MTD} framework has a number of important advantages for
%reasoning about game-tree search algorithms.
Formulating a seemingly diverse collection of algorithms into one
unifying framework allows us to focus attention on the fundamental
differences in the algorithms
(see figure~\ref{fig:sssfriends}).
For example, the framework allows the reader to see just how similar
{\SSS} and {\DUAL} really are, that these are just special cases
of calling {\AB}.
The drivers concisely capture the algorithm differences.
{\MTD} offers  a high-level paradigm that facilitates the
reasoning about issues like algorithm efficiency and memory usage,
without the need for low-level details like search trees,
solution trees and which node to expand next.

\subsection{Related Work} \label{sec:gengame}

\subsubsection{Other Work on Null-Windows}
All the {\MTD} algorithms are based on {\MT}.
Since {\MT} is equivalent to a null-window {\AB} call (plus storage),
they search less nodes than the inferior
one-pass/wide-window {\AB}$(n, -\infty, +\infty)$ algorithm. 

There have
been other attempts with algorithms that
solely use 
null-window {\AB} searches \cite{Nagl89,Schaeffer86}.
Many people have noted that null-window searches have a great
potential, since tight windows usually generate more cutoffs than
wider windows
\cite{Allis94,Camp83,Copl82,Fish81,Pear80,Schaeffer86}. However, it
appears that the realization that the transposition table can be used
to create algorithms that
retain the efficiency of null-window searches by gluing them
together without {\em any\/} re-expansions---and create an
{\SSS}-like best-first expansion sequence---is new. 
%This is not too surprising, since 
This idea is supported by the fact that previous algorithms all
tried to minimize the number of small-window {\AB} calls in some
way, causing those algorithms to make sub-optimal decisions (see
section~\ref{sec:bestmove}). The 
notion that the value of a bound on the minimax 
value of the root of a tree is determined by a solution tree
\index{solution tree}%
was not widely known among
researchers, let alone that this sub-tree fits in memory. When seen in
this light, it 
is not  too surprising that the idea 
of using depth-first null-window {\AB} searches to model best-first
algorithms like {\SSS} is
new, despite their widespread use by the game-tree search community.

\subsubsection{Other Frameworks} \index{frameworks}%
The literature describes two frameworks that generalize
best-first and depth-first minimax search---or rather,
{\SSS} and {\AB}. Ibaraki\index{Ibaraki} created one such framework,
called Gsearch \cite{Ibar86}. He concentrated on
the\index{Ibaraki}\index{Gsearch} 
informed model where, inspired by the ideas of B*, heuristic upper
\index{heuritic bound}%
and lower bounds \index{B*}%
are available for every node.  In the 
uninformed version---disregarding the heuristic bounds---it is
possible to instantiate the framework to both {\AB} and an {\SSS}
variant by
choosing an appropriate function for the so-called select rule.
Ibaraki has used Gsearch for analyzing domination of
algorithms \cite{Ibar87}. Pijls\index{Pijls} and De Bruin\index{Bruin, de}
\cite{Pijls93} show that the {\SSS}-like
Gsearch instance differs slightly from
Stockman's {\SSS}. The Gsearch instance, which they call Maxsearch, is 
a bit more efficient because it stores both upper {\em and\/} lower
bounds for every node, wheras {\SSS} works with only one. This makes
that {\SSS} misses cutoffs in certain special cases \cite{Pijls93}.
 A recursive version of Gsearch,   
called Rsearch, allows restricted-memory
versions of {\SSS} (or rather, Maxsearch) to be formulated
\cite{Ibar91b}. Rsearch  \index{Rsearch}\index{Maxsearch}%
instances typically 
restrict the search trees to be built to a certain depth,
so that a best-first select rule can only work in a part of the
tree. Pijls and De Bruin \cite{Pijls94} show that when the size of the
resident tree is reduced to zero, the resulting Rsearch instance
selects the same nodes as {\AB}. Thus, in Rsearch one
can instantiate {\AB} in two ways: by using a left-first select rule, or
by reducing the size of the stored search tree.\index{GenGame}

Another framework for {\AB} and {\SSS} is proposed
by Bhattacharya and Bagchi \cite{Bhat94}.  This framework is called
GenGame. It is based on their work on RecSSS* \cite{Bhatta93}. GenGame
suffers from this background, in that it is a
relatively complex framework. Also, it uses an {\SSS}-like OPEN list. Just
like the Gsearch/Rsearch duo, instances of GenGame  are created by
changing certain 
rules, and instances vary in the amount of memory that is
needed. The  
authors 
\index{QuickGame}%
describe how the framework can be adapted to yield two variations
(QuickGame and MGame) that both
have  reduced memory requirements, at the cost of expanding more nodes
than {\SSS}. The  QuickGame variation is not dominated by {\AB}, it
misses some of 
{\AB}'s deep cutoffs, although Bhattacharya reports that on average
QuickGame tends to expand less nodes on artificial random trees
\cite{Bhat95}.  The 
other variation, MGame, appears to be using the same idea as Rsearch 
to achieve
a compromise between memory efficiency and search efficiency. On his
part, Ibaraki,
the inventor of Rsearch, makes no mention of the work on
\index{SSS*!Staged SSS*}\index{SSS*!recursive SSS*}\index{Campbell}%
Staged {\SSS} by Campbell \cite{Camp81,Camp83} which also describes
the idea of applying {\SSS} recursively. The idea of
using stages to control memory requirements of 
best-first algorithms goes back to standard texts on search such as Pearl
\cite[p.~68]{Pear84} and Nilsson 
\cite[p.~71]{Nils71}. 
\index{Pearl}\index{Nilsson}

The strength of both frameworks is that they isolate
the differences between best-first
and depth-first search in certain parts of the framework. Furthermore,
they provide a model that helps in reasoning 
about {\AB} and {\SSS}-like algorithms. This applies especially to
Gsearch, which is a 
rather abstract, but clear, framework. 

A drawback of Gsearch and GenGame is that they are not easily amenable
to an implementation.  They are both high-level frameworks that require a
non-trivial 
amount of detail to be filled in to yield an efficient design that can be
implemented in an {\AB}-based game-playing program.\index{MT framework}

Compared to Gsearch and GenGame, {\MT}  is a 
``focused'' framework.  It is only concerned with the manipulation of
bounds on the value of the root, and  only with
best-first algorithms. The details of how to traverse a tree are
performed by a transposition table-enhanced 
{\AB} procedure that acts  as a
black-box for the  calculation of  bounds. This makes {\MT} a  
simpler framework that nevertheless has the power to model a complex
algorithm such as
{\SSS}. 
% (It models {\SSS} as
% a special case of {\AB}.) 
Since {\MT} is based on {\AB},
it is a very practical
framework. Combined with its simplicity, {\MT} is well-suited
for experimental validation of ideas for new best-first
algorithms. {\MTDf} shows that this process can be successful (see  
section~\ref{sec:time}). 

Gsearch, and especially GenGame, are designed to capture {\SSS}'s
behavior. The  relatively complex best-first selection strategy of
{\SSS}    is their basis. Modeling the simple left-to-right {\AB}
behavior is achieved by  disabling some of the features of the
framework. In effect, they use advanced, complex
machinery to model 
simple behavior. They start at the complex side, and make modifications to
end up at the simple side. {\MT}, on the other hand, does not attempt
to model
{\AB}, but uses it as a building block to model best-first algorithms.
It is based on a few items of simple machinery that  are used to
model complex behavior. {\MT} starts at the simple side, adds memory
and a loop, and ends up explaining the complex side.

A disadvantage of {\MT} is that it is less general than Gsearch or
GenGame because it does 
not model the depth-first wide-window {\AB}
algorithm. The essence of {\MT} is that it performs  null-window {\AB}
\index{narrow search window}%
calls. Of course, it is 
straightforward 
to extend {\MT} to use wider windows, or even to become wide-window
{\AB} \cite{Pijl95}. However, this complicates reasoning about its
tree-traversing behavior.

\section{Null-Window {\AB} Search Algorithms}\label{sec:nws}
In this section two ideas for {\MT} instances will be analyzed more
deeply by looking at previous work: the bisection idea and searching for
the  best move.  
% based on previous work 
% We will look especially at the relation with other
% research. 

\subsection{Bisection}
\label{sec:oldstuff}
\index{C*}\index{Coplan}%
In addition to the widely used Scout family, 
there have been experiments with other null-window {\AB} based
algorithms.
% Where Scout uses a wide-window re-search after a fail,
Coplan's C* uses  bisection  to find the minimax
value with a relatively small number of re-searches \cite{Copl82,Weil92}. 
% Looking at the bounds {\AB}'s postcondition the idea for applying bisection
% to an interval comes to mind easily; it  is also mentioned by others
% \cite{Allis94}. 
Although C* was introduced in a rather specialized context, that
of a 
hardware implementation of a chess-endgame solver, and later for
Othello endgames, C* is conceptually equivalent
to {\MTDb}. 

Coplan proposes a pointer based structure to store visited
nodes, which incurs some storage management overhead. It seems more
attractive to use a hash table, as is normally used for storing
transpositions by many {\AB} implementations \cite{Mars82a}. 

\index{bisection}%
Bisecting a wide
interval of interest usually gives big swings in the value of $\gamma$
between successive {\AB} calls. 
Section~\ref{sec:swallow} will show that 
it is efficient to keep the null-window as close to the minimax value
as possible. The swings should therefore be reduced as much as
possible, for example by the use of an aspiration window. 

In a specialized endgame search the \index{evaluation function range}%
range of values is usually much smaller than in ordinary mid-game
search, so that the number of re-searches will be small.
From the perspective of a reduction of the number of {\AB} calls, C*'s
bisection method makes sense. When C* was created it was generally
assumed that search trees were too big to be stored in memory (in
section~\ref{sec:space} we will show that they do fit in memory).
% solution trees was not widely known in the minimax research community,
% let alone that {\AB} created them, or that they can be stored without
% problem in the available memory, so that the inefficiencies caused by
% re-searches are small. 
In retrospect, the reduction of 
the number of {\AB} calls should not have been such a high priority
and should  not be done at the cost of using inefficient
values for the {\AB} calls. 

\subsection{Searching for the Best Move}\label{sec:bestmove}
\index{Berliner}\index{best move}%
Berliner has suggested that the search can be stopped as soon as a
lower bound on the value of the best move is equal to an upper bound
on the value of the other moves $(f^-_{best} \geq f^+_{rest})$
\cite{Berl79}. We will analyze the possible savings in the case the
bounds are non-heuristic bounds, whose value is defined by 
a max or min solution tree, as is the case for {\MTDd}. First we will look
at the  best case, using the structure of the minimal
tree to see how much savings can be achieved.

\begin{figure}
\begin{center}
\includegraphics[width=8cm]{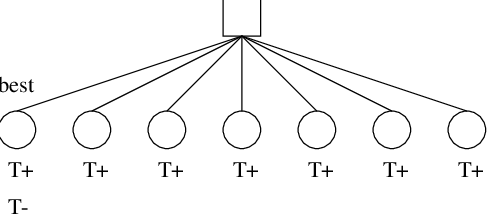}
\end{center}
\caption{Best Move}\label{fig:bestmove}
\end{figure}

With $f^-_{root} = f^+_{root}$ as the termination condition, a
max and a min solution tree have to be constructed at the 
root. Figure~\ref{fig:bestmove} shows the root position and its
moves. The picture shows where the max and min solution trees at these
moves are created: a
max solution tree $(T^+)$  at all of the moves, and a min
solution tree 
$(T^-)$ at the best move.

With the new termination condition $f^-_{best} \geq f^+_{rest}$ the
$T^+_{best}$ is no longer necessary. Thus, searching for the best move
saves in the best case the construction of a max solution tree below a
single move. For all the other moves the max solution trees still have
to be constructed. We conclude that, although there are definitely some 
savings, the order of the size of the search tree is unchanged, for all
but the smallest branching factors. 

Next, we look at  the  average case. Now  some effort
has to be expended in order to find the
solution trees that make up the minimal tree of the best
case. Although the search effort is bigger, the 
reasoning that the solution trees of the rest of the moves determine the
order of the size of the search tree still holds. This analysis suggests
that searching for the best move using backed-up leaf values as bounds
will lead to small improvements.  Tests with {\MTDd} show improvements
of a few percentage points \cite{Plaa94a}. It performs relatively better
in  ``quiet'' test positions, where the principal variation does not
change between  iterations, so  {\MTDd}  predicts the best move
correctly, and the value does not change much between iterations.

% \subsection{Alpha Bounding}
\label{sec:alphabounding}\index{Alpha Bounding}%
Another algorithm implementing this idea is Alpha Bounding,
introduced by \mbox{Schaeffer} \index{Schaeffer}\index{Nagl}%
% achieve Schaeffer has experimented with an algorithm called Alpha Bounding
\cite{Schaeffer86}. (Nagl has independently
re-invented this algorithm three years later \cite{Nagl89}.)
% It uses only null-window {\AB} calls, like the
% {\MTD} family. 
Alpha Bounding tries to be as efficient as possible by using two
ideas: (a) null windows and (b) best-move 
searching.
%  to avoid the minimax wall at the end of the search.

\begin{figure}
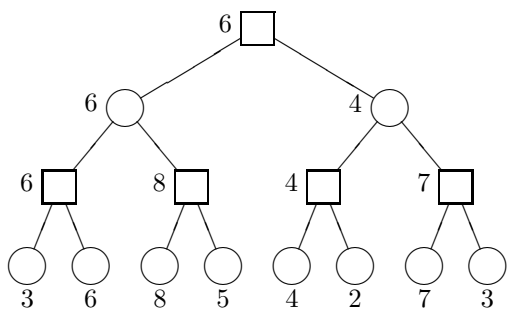

\begin{center}
    \begin{Tree}
   \node{\external\bnth{3}}
   \node{\external\bnth{6}}
   \node{\type{square}\lft{6}}
   \node{\external\bnth{8}}
   \node{\external\bnth{5}}
   \node{\type{square}\lft{8}}
   \node{\lft{6}}

   \node{\external\bnth{4}}
   \node{\external\bnth{2}}
   \node{\type{square}\lft{4}}
   \node{\external\bnth{7}}
   \node{\external\bnth{3}}
   \node{\type{square}\lft{7}}
   \node{\lft{4}}
   \node{\type{square}\lft{6}}
 \end{Tree}
 \vspace*{0.3cm}
\hspace{1cm}\usebox{\TeXTree}
\end{center}
\caption{Alpha Bounding Expands 7}\label{fig:alphabounding}
\end{figure}

The start value of the null-window searches 
should stay on the low side of the minimax value as much as
possible, since a lower bound on the best move has to be
created. 
Starting with the supposedly best move from a previous iteration,
Alpha Bounding searches all moves at the root with a value that will 
make a fail 
high likely below the best move, so that a lower bound will be found. If
this is successful, a min solution tree is constructed below
the best move. 
% each move. 
% 
% ******
% Alpha Bounding works with moves, because it wants to cutoff as soon as
% the best move is found. That's why it didn't start at the root.
% So it is too greedy going after the best  move cutoff, while
% neglecting to look carefully at the solution tree sizes.
% 
% If you're going after the best move, then it's only netural to
% concentrate on the singular moves. In contrast to what I first
% thought, Alpha Bounding is not so much afraid of the minimax wall,
% but it is really the BMC that they're after. And that's why they make
% the solution tree oversight. Since BMC is crucial, they couldn't have
% stumbled on {\MTDf}.
% 
% They were looking in the wrong direction: the BMC, instead of the
% minimax wall, which could have led them to the solution trees and
% possibly even {\MTDf}.
% ******

% Although this  scheme looks efficient enough at first sight---after
% all the same scheme was invented by two people indepently---there is a
% flaw. 
Unlike the {\MT} algorithms, Alpha Bounding may sometimes evaluate
nodes that are not visited by
{\AB}$(n, -\infty, +\infty)$. Figure~\ref{fig:alphabounding} gives
an example of this situation. Assuming that the start value of the
search is below 3, Alpha Bounding would expand the leaf with value 7
in the first search pass ($\alpha$ bound $= 3$). In general, with a
low start value, it could construct min solution trees at {\em each\/}
move at the root.  However, as we can see in figure~\ref{fig:bestmove}, 
to find the best move, an algorithm has to create {\em max\/} solution
trees below the inferior moves. All the min solution trees, except the
single one below the best move, are
overhead. Thus, a start value that is too low can result in overhead.
Likewise, if the start value was chosen too high, then all
moves will fail 
low and  will be re-searched with a lower start value.
Tests showed this algorithm to perform
not too bad, although {\NS} was 18\% better \cite{Schaeffer86}.

As was the case with C*, Alpha Bounding uses ideas that the {\MT}
framework also has. And as with C*, a better model of the structure of
search trees would probably have resulted in an improved algorithm. One such
model is the solution tree, the notion behind {\MT}. 
% in a way  it was ahead of its
% time, focussing on a single idea without the help of a framework to
% provide insight into the structure of the search trees.
% % C* concentrated on a reduction of the number of {\AB} calls, Alpha
% % Bounding on finding the best move. 
% By recognizing that {\AB}
% constructs solution trees one can see more easily how these
% algorithms work and how they can be enhanced.

% The problem with Alpha Bounding is that it focussed on finding the
% best move cutoff---avoiding to build the entire minimal tree---without
% realizing that it has to create max 
% solution trees below some nodes in any case. Although the ideas behind
% Alpha Bounding do use the bounds returned by fail-soft {\AB}, it does
% not take full advantage of the max and min solution
% trees that are created by the null-window {\AB} searches.
% Construction of the full minimal tree can be
% avoided by not building a full max solution tree at the root. Creating a
% min solution tree at all {\em children\/} of the root is something
% different.  Since solution trees
% cannot now and certainly not at the time when this algorithm was
% conceived (and re-conceived) be considered  mainstream knowledge, this
% should not come as 
% much of a surprise.

\subsubsection{Background}
Wim Pijls\index{Pijls} and Arie de Bruin\index{Bruin, de} have been investigating {\SSS} 
for some time \cite{Pijl91,Pijls90,Pijl92c}. In 1989 they found 
that
there is another view on {\SSS}---the max-solution-tree view. They
introduced a
two-procedure algorithm 
embodying these ideas called
SSS-2 \cite{Pijls90}. In May 1993 I joined this research, trying to
find my way in solution trees, {\AB}, and the critical tree.  After
some time our discussions also turned to {\NS} and
null-window search. In the spring of 1994 we found out about
the link between solution trees and null-window {\AB} searches, and
that there are other ways of using sequences 
of null-windows searches, which showed up favorably in simulations
\cite{DeBruin94}. 

The elegant idea of proving the equivalence of
{\SSS} and {\ABSSS} by 
inserting the list operations into the {\AB} code 
(appendix~\ref{app:equiv}) was suggested by
Wim Pijls, based on earlier work by him and Arie de Bruin. He created
an SSS-2 like two-procedure version and
helped by discussing and pointing out errors in earlier versions of
the subsequent one-procedure {\MT} code. The full {\ABSSS}/{\SSS} 
equivalence proof in 
\cite{Pijl95} was primarily worked out by him. Work on an alternative 
equivalence proof, based on static features of the search
trees of {\SSS} and {\ABSSS}, was also primarily done by him
\cite{Pijl95}. 
 
% \section{The Common View on {\SSS}} 
% 1
% 2
% 3
% 4
% 5
% 6 par

% Or, the need for clearity. 

% \section{MT}
% Intro on {\MT}, ides behind {\MT}
% - nws
% - mem
% - bf???
% - surpass ab

% In prev sect we saw that nws is good, but re-searches
% we also saw memory, in the form of TT has a ``Side-effect'' of making
% {\NS}'s re-searches very cheap (assuming the sol trees fit in memory)
% 
% Those two ingredients form the basic behind {\MT}. Use nws for maximum
% cutoffs, and memory to retain that efficiency. Further analysis of the
% behavior of the algorithms that can be created shows that they exhibit
% best-first behavior, they select only nodes that cannot be killed by
% the expansion of other nodes
% 
% Introducing Memory enhanced Test
% 
% 
% 
% A list of {\MTDp}, {\MTDm}, {\MTDf}, etc.
% bf sequences
% 
% 
% \subsection{{\ABSSS} or {\MTDp}}

% \subsection{Traversing Trees with Null-Windows --- Example and Intuition}
% Explaining bf seq
% 
% Proofy thing, on how we traverse the trees
% 
% \subsection{{\MTDf}}
% Intuition, ideas. Inquiry into validity of idea is deferred to chapter
% on analysis.

 % mt
\ \newpage

%\cleardoublepage

\thispagestyle{empty}

\ \newpage

%\cleardoublepage
%\thispagestyle{empty}
\chapter{Experiments}\label{chap:exper}
% \markboth{Chapter 4}{Chapter 4}
%\markboth{Chapter 4}{Experiments}
% \chapter{Storage}\label{chap:stor}
% 
% 
% 
% (graphs SSS/AB) TT/S incl MTD(f) and AspNS graphs
% cut & paste experiments & conclusions
% 
% 
% 
% 4 They Were Wrong, but why?
%   - SSS misconceptions
%   - SimSucks, differences between artificial  & real trees
% 
% 
% 

% \section{Storage}
% Graphs
% NS & MTD(f) in tiny memory
% finally prove that the stuff fits into memory
% Experiments with memory
% \input ai-sec3 % from SSS=AB+TT sec7 plus explanation that ID-AB
               % needs same order of memory, and that TT is more
               % flexible than list or array

The prevailing view in the literature is that {\SSS} is not practical,
because it uses  too much memory, is too slow, and too complicated. With {\ABSSS}, we
oppose this view on 
all points. There is nothing complex about implementing {\ABSSS}, it
is as simple (or hard) as implementing {\AB}. This enabled us to test
{\ABSSS} in real game-playing programs, and see whether the other points
were true. In this chapter we report on these experiments. First we
look at how much storage is needed for {\MT}-based algorithms in
typical  applications. Next we
look at their performance. 

Results from this chapter have been published in \cite{Plaa95a,Plaa95b}.

%\section{{\ABSSS} in Practice}\label{sec:space}
\section{All About Storage}\label{sec:space}
\index{memory}\index{storage|see{memory}}%
The literature portrays storage as the biggest  problem with {\SSS}.
The way it was dealt with in Stockman's original formulation
\cite{Stoc79} gave \index{Stockman}%
rise to two points of criticism:
\begin{enumerate}
\item {\em {\SSS} is slow.\/}
  Some operations on the sorted OPEN list
  have a time complexity that is non-polynomial in the search depth.
  As a results, measurements show 
  that the purge 
  operation of $\Gamma$ case 1 consumes about 90\% of {\SSS}'s runtime
  \cite{Mars87}.  
\item {\em {\SSS} has unreasonable storage demands.\/}
  Stockman states that his OPEN list needs to store at most
  \index{OPEN list}%
  $w^{\lceil d/2 \rceil}$ entries for a
  game tree of uniform branching factor $w$ and uniform depth $d$,
  the number of leaves
  of a max solution tree. The example in appendix~\ref{app:sssex} illustrates that a
  single max solution tree is manipulated. \index{solution tree, max}%
  (In contrast, {\DUAL},
  requires $w^{\lfloor d/2 \rfloor}$ entries, 
  the number of leaves of a min solution tree.)
  This is usually perceived as being unreasonably large storage
requirements. 

\end{enumerate}
Several alternatives to the {\SSS} OPEN list have been proposed.
One solution implements the storage as an unsorted array,
alleviating the need for the costly {\em purge\/} operation
by overwriting old entries
(RecSSS* \cite{Bhat90,Bhatta93,Rein94b}).
By organizing this data as an implicit tree,
there is no need to do any explicit sorting, since the principal variation
can be traversed to find the critical leaf.
Another alternative is to use a pointer-based tree,
the text-book implementation of a recursive data structure. 

Our solution is to extend {\AB} to include transposition tables (see
section~\ref{sec:idtt}).\index{transposition table}
As long as the transposition table is large enough to store at least
the min or  
max solution trees that are essential for
the efficient operation of \index{MT-SSS*}\index{MT-Dual*}%
{\ABSSS} and {\ABDUAL}, it provides for fast access and
efficient storage. (Including the direct children of nodes in 
the max solution tree. These can be skipped by optimizations 
in the {\AB}
code, in the spirit of what Reinefeld has done for Scout 
\cite{Rein83,Rein89,Rein85}.)
{\ABSSS} will in principle operate when the table is too
small, at the cost of extra re-expansions. The flexibility of the
transposition table allows experiments with different memory sizes.
In section~\ref{sec:results} we will see how big the transposition table
should be for {\ABSSS} to function efficiently.
{\SSS} stores the leaf nodes of a max solution tree. {\ABSSS} stores the
interior nodes too. In this way the PV can be traversed to the critical
leaf, without the need for time-consuming sorting.

%A potential drawback of most transposition-table implementations is
%that they do not handle hash-key collisions well.
%In \cite{Plaa95e} it is shown that this is not a problem in practice.
%Collisions occur since the hash
%function maps more than one node to one table entry (since
%there are many more possible nodes than entries). For the
%sake of speed many 
%implementations just overwrite older entries when a collision
%occurs. If such a transposition table is used, then re-searches of
%previously expanded nodes, that were overwritten 
%because of a collision, are highly likely.
%Only in the case that no relevant information is lost from the table
%does {\ABSSS} search exactly the same leaf nodes as {\SSS}. 
%In section~\ref{sec:coll} we will discuss whether collisions are a
%significant problem.

A potential drawback of most transposition-table implementations is
that they do 
not handle hash-key collisions well. Collisions occur since the hash
function maps more than one position to one table entry (since
there are many more possible positions than entries). For the
sake of speed many 
implementations just overwrite older entries when a collision
occurs. If such a transposition table is used, then re-searches of
previously expanded nodes are highly likely.
Only in the case that no relevant information is lost from the table
does {\ABSSS} search exactly the same leaf nodes as {\SSS}. 
In section~\ref{sec:coll} we discuss why collisions are not a
significant problem in practice.

The transposition table has a number
of important advantages:  \index{transposition table advantages}%
\begin{itemize}
\item
  It facilitates the identification of transpositions in the search space, 
  making it possible for tree-search algorithms to search a graph
  efficiently. 
%  This is an important advantage, since in many applications the high
%  number of transpositions makes the game tree several times larger than
%  a game graph \cite{Plaa94c}. 
  Other {\SSS} data structures do not readily support transpositions.
\item
    It takes a small constant time to add an entry to the table,
    and effectively zero time to delete an entry.
    There is no need for costly purges;
    old entries get overwritten with new ones.
    While at any time entries from old (inferior) solution trees may
    be resident, 
    they will be overwritten by newer entries when their space is
    needed.
    Since internal nodes are stored in the table, the {\AB} procedure
    has no problem of finding the critical leaf; it can traverse
    the principal variation. 
%   For this reason there is no danger of old
%   nodes interfering with the search.%, eliminating the second reason
 %   for purges.
\item
  The larger the table, the more efficient the search (because more
  information can be stored, the number of hask-key collisions diminishes).
  Unlike other storage proposals, the transposition-table size
  is easily adaptable to the available memory resources.
\item
  There are no constraints on the branching factor or depth of the search tree.
  Implementations that use an array  as an implicit data structure for
  the OPEN list are constrained to fixed-width, fixed-depth trees.
\item Most high-performance game-playing programs already use the
  {\AB} procedure with a transposition table.
  Consequently, no additional programming effort is required to implement it.
\end{itemize}
%Before turning to  experiments in the next section, 
We need to make
one last remark\label{sec:remark}\label{sec:boundsremark} on {\ABSSS}.
In this algorithm (and in all the tests that we will discuss further on) one
value was associated with each node, either \index{both bounds}%
an $f^+$, an $f^-$ or an $f$. {\ABSSS} manipulates one solution tree
at a time. The pseudo code for {\MT} stores only one bound per node. Other
algorithms may manipulate two solution trees at the 
same time (for example, {\MTDb} and {\MTDs}) and may need to store two
bounds for certain nodes.
In addition to this, there are some rare cases
where {\SSS} (and {\ABSSS}) unnecessarily expand some nodes. This can
be prevented by storing the value of both an upper and a lower bound
at the nodes, and using the value of all the children to update them.
This is the difference between Stockman's {\SSS} and Ibaraki's {\SSS}
(or Maxsearch)
from section~\ref{sec:gengame}. The difference is further analyzed in
\cite{Pijl91,Pijl95,Plaa94b}. 
%
%
% This version of {\AB} expands the same
%leaf nodes as {\SSS}. For the {\SSS} algorithm there is room
%for a minor optimization to preclude the expansion of dead nodes in
%some rare cases. This can be achieved by storing two bounds at each
%node. For other algorithms 
%that need to store both a max and a min solution tree at the same time
%({\MTDb} and {\MTDs}, to be discussed in section~\ref{sec:mt}) storing
%two bounds, and updating them to the minimax value of the bounds of
%their children, is even a necessity. For algorithms using a
%transposition table in general, two bounds can improve the
%performance. More on these points can be found
%in the appendix and in \cite{Pijl91,Pijl95,Plaa94b}.
%
%

\subsection{Experiment Design}
{\ABSSS} uses a standard
transposition table to store previous search results. If that table is
too small, previous results will be overwritten, requiring 
occasional re-searches. A small table will still provide the
correct minimax value, although the number of leaf expansions may be
high. To test the behavior of our algorithm, we
experimented with different transposition-table sizes for {\ABSSS}
and {\ABDUAL}. 

The questions we want to see answered are: ``Does {\SSS} fit in memory
in practical situations?''\ and\ ``How much memory is needed to
out-perform {\AB}?''.  We used iterative deepening versions of
{\ABSSS} and {\AB}, since these are used in practical applications too.
The experiments were conducted using game-playing programs of
tournament quality.
Our data has been gathered from three programs:
\Chinook\ (checkers) \cite{Scha92}, \index{Chinook}\index{Keyano}\index{Phoenix}%
\Keyano\ (Othello) \cite{Broc96} and
\Phoenix\ (chess) \cite{Schaeffer86}. With these programs we cover the
range from low to high branching factors.
Using three programs we ensure a higher degree of generality and
reliability of the results.
All three programs are well known in their respective domains.
%For our experiments, we used the original author's search algorithm
%which, presumably, has been highly tuned to the application.
The only change we made to the programs was to disable search extensions and
forward pruning, to ensure \index{forward pruning}\index{search!extensions}%
consistent minimax values for the different algorithms. For the
same reason we used the sequential versions of the programs.
For our experiments we used the original
program author's transposition-table data structures and code, without
modification.
At an interior node,
the move suggested by the transposition table is always searched first
(if known),
and the remaining moves are ordered before being searched.
{\Chinook} and {\Phoenix} use dynamic ordering based on the
history heuristic, \index{history heuristic}%
while our version of {\Keyano} uses static move ordering.

The {\AB} code given in figure~\ref{fig:mmab} differs from the one used
in practice in that the latter usually includes two details,
both of which are common practice in game-playing programs.
The first is a search-depth parameter.
This parameter is initialized to the depth of the search tree.
As {\AB} descends the search tree, the depth is decremented.
Leaf nodes are at depth zero.
The second is the saving of the best move at each node.
When a node is revisited, the best move from the previous search is
always considered first.

Conventional test sets in the literature  (such as \cite{Kope82})
proved to be inadequate to model real-life conditions.
%poor predictors of performance.
Positions in test sets are usually selected to test a particular
characteristic or property of the game (such as tactical combinations
in chess) and are not necessarily indicative of typical game conditions.
For our experiments, the programs were tested using a set
of 20 positions that mostly corresponded to move sequences from
tournament games (see appendix~\ref{app:pos}).\index{test positions}
By selecting move sequences rather than isolated positions,
we are attempting to create a test set that is representative
of real game search properties
(including positions with obvious moves, hard moves,
positional moves, tactical moves, different game phases, etc.). 
  Test runs were performed on a bigger test set and to a
  higher search depth to check that
  the 20 positions did not cause anomalies.
All three programs ran to a depth
so that all searched roughly for the same amount of time. The search
depths reached by the programs vary greatly 
because of the differing branching factors.
In checkers, the average branching factor is approximately 3
(there are typically 1.2 moves in a capture position
while roughly 8 in a non-capture position),
in Othello 10 and in chess 36.
Because of the low
branching factor Chinook was able to search to  depth 15 to 17,
iterating two ply at a time.
Keyano searched to 9--10 ply and Phoenix to 7--8, both one ply at a
time. % (a ply is one level in the tree, a half-move). 

\begin{figure}
\begin{center}
\includegraphics[width=10cm]{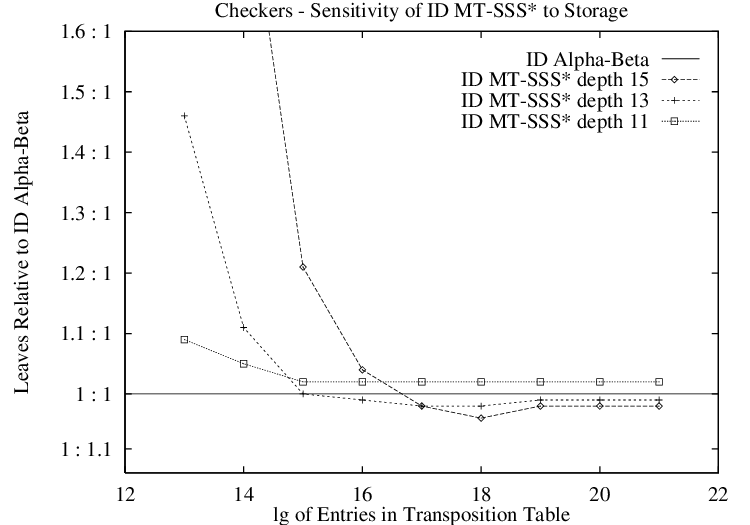}
\end{center}
\caption{Memory Sensitivity ID {\ABSSS} Checkers}\label{fig:ttsize1}
\end{figure}

\begin{figure}
\begin{center}
\includegraphics[width=10cm]{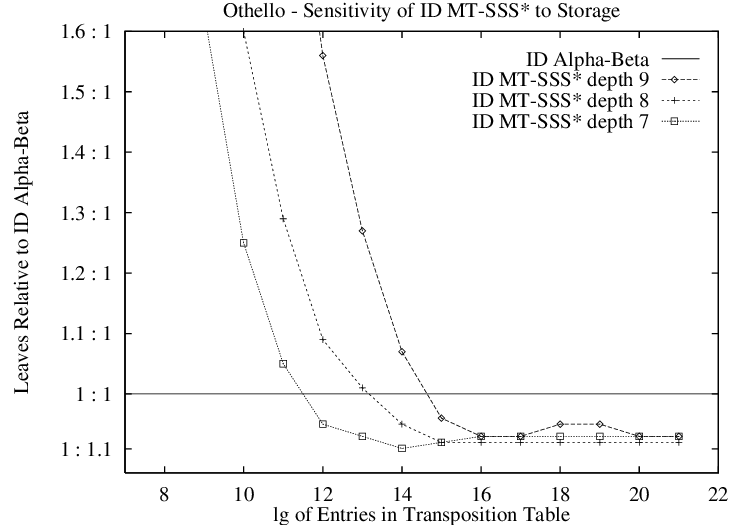}
\end{center}
\caption{Memory Sensitivity ID {\ABSSS} Othello}\label{fig:ttsize2}
\end{figure}

\begin{figure}
\begin{center}
\includegraphics[width=10cm]{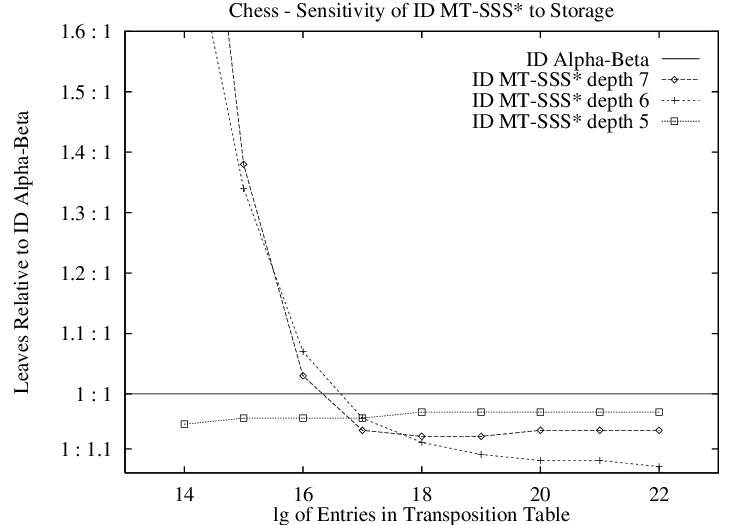}
\end{center}
\caption{Memory Sensitivity ID {\ABSSS} Chess}\label{fig:ttsize3}
\end{figure}

\begin{figure}
\begin{center}
\includegraphics[width=10cm]{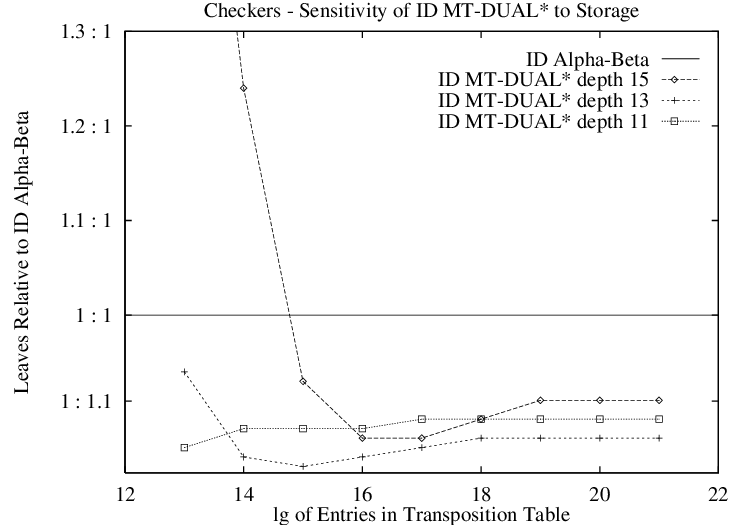}
\end{center}
\caption{Memory Sensitivity ID {\ABDUAL} Checkers}\label{fig:tt1}
\end{figure}

\begin{figure}
\begin{center}
\includegraphics[width=10cm]{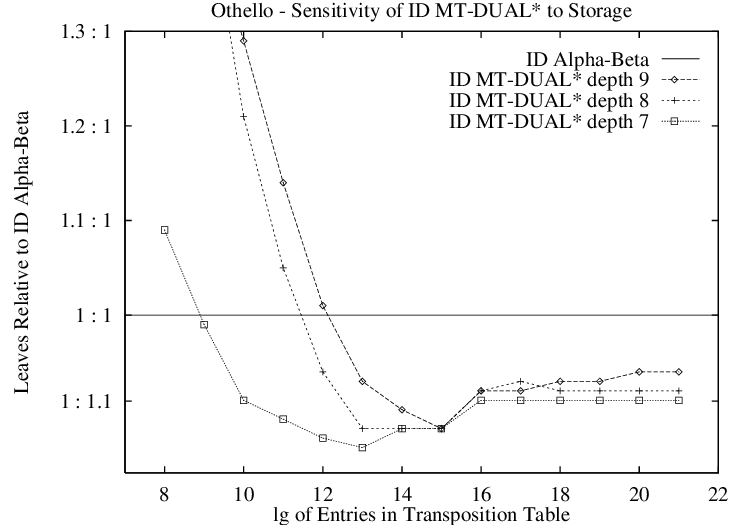}
\end{center}
\caption{Memory Sensitivity ID {\ABDUAL} Othello}\label{fig:tt2}
\end{figure}

\begin{figure}
\begin{center}
\includegraphics[width=10cm]{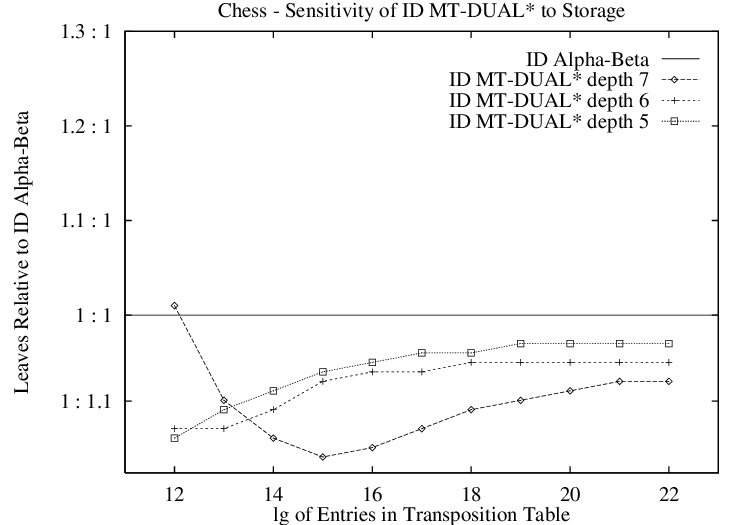}
\end{center}
\caption{Memory Sensitivity ID {\ABDUAL} Chess}\label{fig:tt3}
\end{figure}

\subsection{Results}
\label{sec:results}\index{memory sensitivity|see{transposition table size}} \index{transposition table size}%
Figures~\ref{fig:ttsize1}--\ref{fig:ttsize3} and
\ref{fig:tt1}--\ref{fig:tt3} show the number of 
leaf nodes expanded \index{iterative deepening}%
by ID {\ABSSS} and ID {\ABDUAL} relative to
ID {\AB} as a function of transposition-table size
(number of entries in powers of 2, $\lg$ denotes $\log$ base 2).
The graphs show that for small transposition tables, {\AB}
out-performs {\ABSSS}, and for very small sizes it out-performs {\ABDUAL} too.
However, once the  storage reaches a critical
level, {\ABSSS}'s performance levels off and
is generally better than {\AB}.
The graphs for {\ABDUAL} are similar to those of {\ABSSS}, except 
that the lines are shifted to the left.

%Smaller transposition tables affect the performance of {\ABSSS} and
%{\ABDUAL} more significantly than they affect {\AB}.
%{\AB} is a one-pass algorithm; in each iteration
%there are no re-searches.
%For each iteration, {\ABSSS} and {\ABDUAL} do many re-searches as
%they converge on the minimax value.
%This involves revisiting nodes and, if they are not in the table,
%re-expanding them.

Simple calculations and the empirical evidence
lead us to disagree with authors stating
that $O(w^{\lceil d/2 \rceil})$ is too much memory for practical purposes
\cite{Kain91,Mars87,Musz85,Rein89,Roiz83,Stoc79}.
For present-day 
search depths in applications like checkers, Othello and chess,
using present-day memory sizes, we see that {\ABSSS}'s
search trees fit in the available memory. The graphs in
figures~\ref{fig:ttsize1}--\ref{fig:ttsize3} show that {\ABSSS} needs
about $2^{17}$ table entries for the tested search depths. Assuming that
each entry is 16 bytes, a transposition table of 2 Megabyte is large
enough. 
For deeper searches a transposition table size of 10 Megabyte will be
more than adequate for {\ABSSS} under tournament
conditions.

The literature contains a number of
proposals for reducing the memory requirements of {\SSS}. This is done
either through 
having {\AB} search the top or bottom part of the tree in {\AB}/{\SSS}
hybrids
\cite{Leif85,Camp81,Camp83} or by phased or recursive versions of
{\SSS} such as 
Phased {\SSS} \cite{Mars86}, or
Staged {\SSS} (or Rsearch or GenGame)
\cite{Bhat94,Camp81,Camp83,Ibar91b,Pijls93}.\index{SSS*!Staged SSS*}\index{SSS*!Phased SSS*}\index{SSS*!recursive SSS*}\index{SSS*!hybrids}\index{Rsearch}\index{GenGame} 
Our tests show that neither hybrids nor staged/phased versions of {\SSS}
are necessary in practice, especially because
they achieve the memory reduction at the cost of a lower performance,
since
the best-first selection mechanism is not applied to the full tree.

%\subsection{Odd/Even: {\AB} Needs Memory Too}
\label{sec:coll}\label{sec:abmem}
There is an important difference between {\AB} as found in text books and
as it is used in game-playing programs. To achieve high performance,
{\AB} is enhanced with iterative deepening, transposition
tables, and various move-ordering enhancements. These enhancements
enlarge the memory requirements of the basic algorithms
significantly. Here we will analyze 
how much memory is needed to achieve high performance. We will do this
by comparing {\AB} to {\ABSSS} and {\ABDUAL}.

The graphs  provide a clear answer to the
main question: {\ABSSS} fits 
in memory, for practical search depths in games with both narrow and
wide branching factors. It out-performs {\AB} when given a reasonable
amount of storage.

%We have gained the possibility to formulate a strong answer on one of
%the most important issues concerning {\SSS}. In doing so, we have
%created a kind of compound experiment, in which the sum of a number of
%factors is tested, all of which are perhaps worth a
%careful investigation by themselves. 
% After having answered the main question, we will review some
% of the questions posed by 
% closer inspection of the graphs. We will try to answer the question
% using the ideas  of the preceding section of how {\ABSSS} 
% builds its trees. 

The shape of the graphs supports the idea that there is a table size
where {\ABSSS} does not have
to re-expand previously expanded nodes. This point lies
roughly at
the size of a max solution tree, which agrees with the statement
that {\ABSSS} needs memory to manipulate a single solution max
solution tree. As soon as there is enough memory to store essentially
the max solution tree, {\ABSSS} does not have to re-expand nodes in
each pass.
The graphs also support the notion that {\ABDUAL}
needs less memory since it manipulates a (smaller) min solution tree.
%(see also section~\ref{sec:dualbetter}, page~\pageref{sec:dualbetter}).

\subsubsection{\AB} \index{memory requirements!Alpha-Beta}%
The lines in the graphs show the leaf count of {\ABSSS} relative to
that of {\AB}. This poses the question as to what  degree  the shape
is influenced by the
sensitivity of {\AB} to the table size. If the table is too small,
then collisions will occur, causing deeper entries to be erased in
favor of nodes closer to the root. Thus, the move ordering and
transposition identification close to the leaves diminishes. 
The denominator of the lines in the graphs is not free from memory
effects. To answer the question, we have to look at the absolute
values. These show that {\AB} has
in principle 
the same curve as {\ABSSS}, only the curve is not as steep for small
table sizes. Interestingly, at roughly the same point as {\ABSSS}, does
{\AB}'s line stabilize, indicating that both need roughly the same amount 
of memory.
The numbers indicate that {\AB} needs about as much
memory to achieve high performance as {\ABSSS}.

%!!!!!!!!!!!!!!
%
%3.5: How much memory does AB need
%1 for mv ord: d-1
%2 for txpos: d
%where the mv ord has the biggest impact

To understand how much memory {\AB} needs for optimal performance,
we recall from section~\ref{sec:ttidx}
that {\AB} uses the 
transposition table for two purposes: 
\begin{enumerate} 
\item {\em Identification of transpositions}\\
 To store the nodes 
in a search to depth $d$, the
current search tree must 
be stored. For high-performance
 programs this is close to the minimal tree, whose
 size is
$O\big(w^{\lceil d/2\rceil}\big)$. 
\item  {\em Storing best-move information}\\
To store the best-move information in a search to depth $d$, for
use in the next iteration $d+1$,
the minimal tree {\em minus the leaf nodes\/} for that depth must fit
in memory, or size  $O\big(w^{\lceil (d-1)/2\rceil}\big)$. 
\end{enumerate}
Of these two numbers the transposition information is the biggest, 
 $O\big(w^{\lceil d/2\rceil}\big)$. To store
best-move information of the {\em previous\/} iteration only 
$O\big(w^{\lceil (d-2)/2\rceil}\big)$ is needed. Of these two factors, move 
ordering generally has the biggest impact on the search effort. We
conclude that 
{\AB} needs between $O\big(w^{\lceil d/2\rceil}\big)$ and 
$O\big(w^{\lceil (d-2)/2\rceil}\big)$ transposition-table entries.

\subsubsection{\ABSSS} \index{memory requirements!MT-SSS*}%
From section~\ref{sec:idtt} we recall that {\ABSSS} 
benefits from the transposition 
table in a third way: prevention of
re-expansion of nodes searched in previous passes.
The graphs show that this last aspect 
has the biggest impact on {\ABSSS}'s
memory sensitivity. 
For small table sizes collisions cause nodes near the leaves of the tree
to be overwritten constantly. We could ask ourselves whether collisions
remain more of a problem for {\ABSSS} than for {\AB} when more
memory is available. 

If the transposition table never loses any information except nodes
outside the max solution tree plus its direct descendants,
then {\ABSSS} builds exactly the same search tree as {\SSS}.
Conventional transposition tables, however,
are implemented as hash tables that resolve collisions by
over-writing entries. Usually, entries 
further away from the root are not allowed to overwrite entries closer to the
root, since these entries are thought to prevent the search of more
nodes. In the case of {\ABSSS} some of these nodes could be
useless---not belonging to the max solution tree---while some  nodes
that were searched to a shallow depth could be part 
of the principal variation,  and are thus needed for the next pass.
%Luckily, for trees with a wide branching factor, the vast majority of the
%nodes are on the lowest level, meaning that the likelihood of this
%phenomenon occurring is small, unless the table is much too small. The
%graphs show that in that case, a lot of re-expansions occur, caused by
%collisions overwriting old entries.

When information is lost, how does this affect {\ABSSS}'s
performance?
From our experiments with ``imperfect'' transposition 
tables we conclude that {\ABSSS}'s performance
does not appear to be negatively affected.
Inspection of the {\ABSSS} test results shows that after a certain
critical table size is reached, 
the lines stay relatively flat, just as in
figure~\ref{fig:ttsize1}--\ref{fig:ttsize3}. 
If collisions were having a significant impact,
then we would expect a downward slope,
since in a bigger table the number of collisions would drop.
(Maybe this happens close to the
point where the lines become horizontal, implying that choosing a
slightly bigger size for the table removes the need for
additional collision resolution mechanisms.)
We conclude that in practice the absence of elaborate collision resolution
mechanisms in transposition tables, such as chaining or rehashing, is
not an issue 
where {\ABSSS} is concerned.

\index{transposition table, uses of}%
How much memory does {\ABSSS} need for the third function: prevention
of re-searches of previous passes?
In section~\ref{sec:absss} we noted that {\ABSSS}
manipulates a max solution tree. The size of this tree is
$O\big(w^{\lceil d/2\rceil}\big)$. However, the benefit of storing
leaf nodes is small. The best-move information of their parents
causes just one call to the evaluation function, which will cause a
cutoff. Most
benefits from the transposition table are already achieved if it is of
size $O\big(w^{\lceil (d-1)/2\rceil}\big)$.  So, for high
performance in {\ABSSS} we need $O\big(w^{\lceil d/2\rceil}\big)$ for
the transpositions, $O\big(w^{\lceil (d-1)/2\rceil}\big)$ for the
multiple passes, and $O\big(w^{\lceil (d-2)/2\rceil}\big)$ for the
best moves of the previous iteration. 

Assuming that the impact of transpositions is less than that of move
ordering and multiple-pass re-searches, it seems that {\ABSSS} needs
a bit more memory than {\AB}, for high performance. However,
{\ABSSS} uses  null-window {\AB} calls, that generate
more cutoffs than the standard wide-window {\AB} algorithm. The null
windows cause {\ABSSS} to have, in effect, a
smaller $w$. Since both algorithms stabilize at roughly the same table
size, it appears that these effects compensate each other. 

% in this area can yieldHowever, this reasoning contains a number of
% (simplifying) assumptions 
% that lack  
% sufficient justification.

\subsubsection{\ABDUAL}
\index{memory requirements!MT-Dual*}%
The preceding reasoning is supported by the graphs for {\ABDUAL}. Here we
see an interesting phenomenon: compared to {\AB} first the node count drops sharply, and then
increases again slightly with growing table sizes. It appears there is
a point where the
best-first, ``memory hungry'' algorithm performs better when given
{\em less\/} memory. Again, the explanation is that the lines in the
graph show {\ABDUAL} {\em in relation to\/} {\AB}. Inspection
of the {\ABDUAL} test results reveals 
that the effect is caused by the fact that the {\ABDUAL} curve
stabilizes earlier than the {\AB} curve (which stabilizes at roughly
the same point  as {\ABSSS}). So, the increase at the end of the graph is not
caused by {\ABDUAL}, but by {\AB}.

The reason that {\ABDUAL} needs less
memory than {\ABSSS} is that it manipulates min solution trees, which
are of size  $O\big(w^{\lfloor d/2\rfloor}\big)$. For high
performance {\ABDUAL} needs $O\big(w^{\lfloor d/2\rfloor}\big)$ for
the transpositions, $O\big(w^{\lfloor (d-1)/2\rfloor}\big)$ for the
multiple passes, and $O\big(w^{\lceil (d-2)/2\rceil}\big)$---{\em not\/}
smaller---for the 
best-moves of the previous iteration (the size of the minimal tree
is the same for {\ABDUAL}). Since most of these figures are smaller
than for {\AB} and {\ABSSS}, this analysis provides an explanation why
{\ABDUAL} could perform 
better with less memory. Since we are using iterative deepening
versions of the algorithms, this advantage can also shine through in
{\em even\/} search
depths, where the floor or ceiling operators do not make a difference.

By examining the trees that are stored in the transposition table by
iterative deepening versions of
{\AB} and {\ABSSS}, and 
approximating the amount of memory that is needed, we were able to
find an
explanation why both algorithms need roughly the same amount of memory
to achieve high performance.
More research can provide further insight in this matter.
We conclude from the experiments that {\ABSSS} and {\ABDUAL} are practical
alternatives to {\AB}, as far as the 
transposition-table size is concerned. 
% Some of the finer points
% may benefit from further research.
%However, the experiments have also made clear that there are a number
%of factors that are not yet fully understood. We have not performed an
%exhaustive analysis into the relative 
%importance of these factors. This remains the subject of
%further research.

% In chapter~\ref{chap:anal} we will analyse some  technical
% questions more deeply.

%Also, {\ABSSS} and {\ABDUAL} do not use an explicit OPEN list,
%only an implicit search tree stored in a transposition table.
%The store and retrieve operations are just as
%fast for {\AB} as for {\SSS}.
%Thus, {\ABSSS}  and {\ABDUAL} execute as fast as {\AB}.
%
%Having seen that the {\ABSSS} reformulation effectively solves
%{\SSS}'s storage 
%problems, we will discuss two points in more detail.  The first
%concerns the problem of collisions, the second discusses the practical
%validity of the assumption that {\AB} has polynomial storage requirements.

\subsection{{\ABSSS} is a Practical Algorithm}\label{sec:sssprac}
Section~\ref{sec:space} cited two storage-related drawbacks of {\SSS}.
The first is the excessive memory
requirements.  We have shown that this is solved in practice.

The second drawback, the inefficiencies incurred in maintaining the
OPEN list, specifically the sort and purge operations, \index{OPEN list}%
were addressed in the RecSSS* algorithm \cite{Bhatta93,Rein94b}. Both
\index{RecSSS*}%
{\ABSSS} 
and RecSSS* store interior nodes and overwrite old entries to solve
this. The difference is that RecSSS* uses a restrictive data structure
to hold the OPEN list 
that has the disadvantage of requiring the search depth and width be known
{\em a priori}, and having no support for transpositions.
Programming effort (and ingenuity) are required to make RecSSS*
usable for high-performance game-playing programs.

In contrast,
since most game-playing programs already use the {\AB} procedure and
transposition tables, the effort to implement {\ABSSS} consists
only of adding a simple driver routine (figure~\ref{fig:sss2}). 
Implementing {\ABSSS} is as simple (or hard) as implementing {\AB}.
All the familiar {\AB} enhancements
(such as iterative deepening, transpositions and dynamic move ordering)
fit naturally into our new framework
with no practical restrictions
(variable branching factor, search extensions and forward pruning, for
example, cause no difficulties). 
%In summary,
%Stockman's vision of a fixed-depth {\SSS} algorithm,
%is just a special case of our formulation.

In {\ABSSS} and {\ABDUAL}, interior nodes are accessed
by fast hash-table lookups, to eliminate the slow operations. Execution time
measurements (not shown) confirm that in general the run time of {\ABSSS} and
{\ABDUAL} are proportional to the leaf count, as can be seen in
figure~\ref{fig:ttsize1}--\ref{fig:ttsize3} and
\ref{fig:tt1}--\ref{fig:tt3}, indicating that they 
are a few percent faster than {\AB}. However, in some programs where
interior node processing is slow, the high number of tree traversal by
{\ABSSS} and {\ABDUAL} can have a noticeable effect. For
real applications, in addition to leaf node count, the total node
count should also be checked (see section~\ref{sec:time}). 

We conclude that {\SSS} and {\DUAL}
have become practical, understandable, algorithms, when 
expressed in the new formulation.

% \section{Performance}
% Graphs
% experiments on performance
% Experiments search efficiency
% \input ai-sec6 % from MT sec4
\section{Performance}\label{sec:time}
To assess the performance of the proposed algorithms,
a series of experiments was performed. We present data for the
comparison of 
{\AB}, {\NS}, {\ABSSS}/{\MTDp}, {\ABDUAL}/{\MTDm}
and {\MTDf}.
Results for {\MTDb} and {\MTDs} are not shown; their results were
inferior to {\MTDf}. An in-depth treatment of the other algorithms is
deemed more interesting.

\subsection{Experiment Design}
We will assess the performance of the algorithms by counting leaf
\index{leaf node}\index{total node}\index{transposition node}%
nodes and total nodes (leaf nodes, interior nodes and nodes at which
a transposition occurred). For two algorithms we also provide data for
execution time. 
As before, experiments were conducted with
three
tournament-quality game-playing programs.
All three programs use a transposition table with
a maximum of $2^{21}$ entries.\index{transposition table size}
Tests like the ones in section~\ref{sec:space} showed that the solution trees
could comfortably fit in tables of this size for the depths used in 
our experiments, without any risk of
noise due to collisions. 
Since we implemented {\MT}
using null-window alpha-beta searches, we did not have to make any
changes at all to the code other than the
disabling of forward pruning and search extensions.
We only had to introduce the {\MTD} driver code.
%At an interior node,
%the move suggested by the transposition table is always searched first
%(if known),
%and the remaining moves are ordered before being searched.
%{\Chinook} and {\Phoenix} use dynamic ordering based on the
%history heuristic \cite{Scha89b},
%while {\Keyano} uses static move ordering.

%All three programs use
%transposition tables with only one bound at a time, either an $f^+, f$
%or $f^-$ (in contrast with our code in
%figure~\ref{fig:ttab}).

Many papers in the literature use {\AB} as the base-line for
comparing the performance of other algorithms
(for example, \cite{Camp83,Mars82a}).
The implication is that this is the standard data point which everyone
is trying to beat.
However, game-playing programs have evolved beyond simple {\AB} algorithms.
Most use {\AB} enhanced with null-window search ({\NS}), % \cite{Rein83}),
iterative deepening, transposition tables, move ordering and an
initial aspiration window.
Since this is the typical search algorithm used in high-performance programs 
(such as \Chinook, \Keyano, and \Phoenix),
it seems more reasonable to use the enhanced programs as our base-line
standard. 
The worse the base-line comparison algorithm chosen,
the better other algorithms appear to be.
By choosing NegaScout enhanced with aspiration searching 
(Aspiration NegaScout) as our performance metric,
we are emphasizing that it is possible to do better than the ``best''
methods currently practiced and that, contrary to published simulation
results, some algorithms---notably {\SSS}---turn out to be
inferior. To achieve high performance, the programs use the
following enhancements: all three programs use iterative deepening and
a transposition table, Chinook and Phoenix 
use the history heuristic and quiescence search, Chinook uses the ETC (see
chapter~\ref{chap:anal}), and Phoenix and Keyano use static move
ordering.

Because we implemented the {\MTD} algorithms %(like {\MTDp} and {\MTDm})
using MT %(null-window {\AB} calls with a transposition table) 
we were able to 
compare a number of algorithms that were previously seen as very
different. By using MT as a common proof-procedure, every algorithm
benefited from the same enhancements concerning iterative deepening,
transposition tables and move ordering code.
To our knowledge this is
the first comparison of fixed-depth depth-first and best-first
minimax search algorithms
where all the algorithms are given identical resources.
Through the use of large transposition tables, our base line, {\AspNS},
becomes for all practical purposes as effective as Informed NegaScout
\cite{Rein85}.    

\begin{figure}
% \begin{center}
\includegraphics[width=10cm]{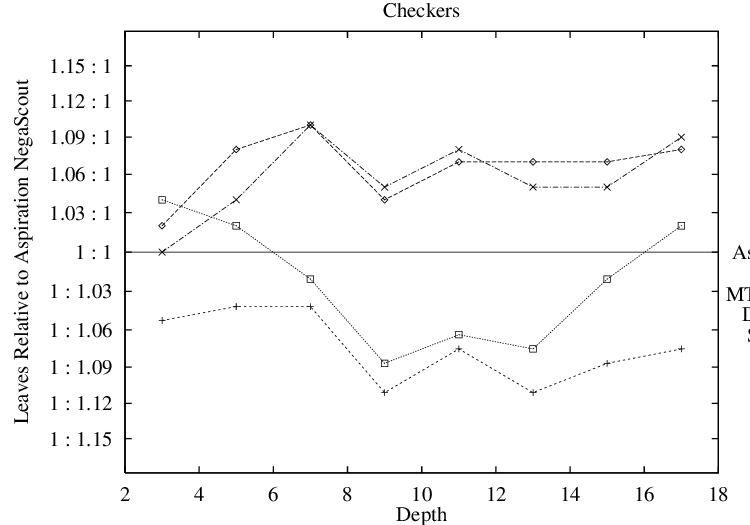}
% \end{center}
\caption{Leaf Node Count Checkers}\label{fig:pl1}
\end{figure}

\begin{figure}
% \begin{center}
\includegraphics[width=10cm]{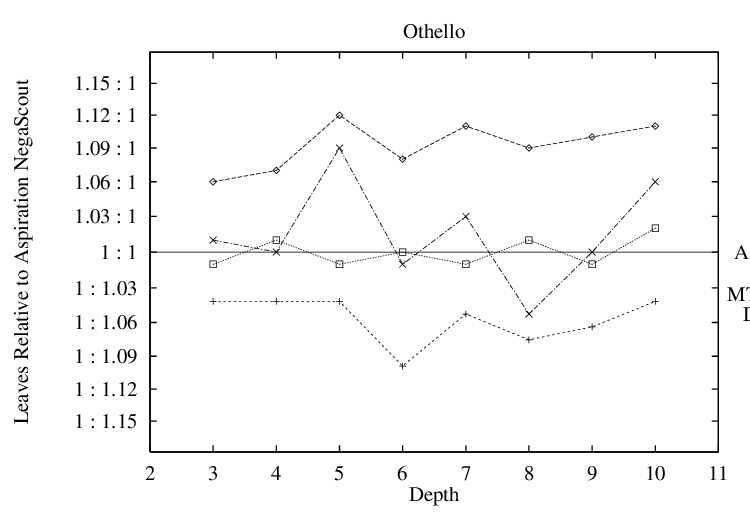}
% \end{center}
\caption{Leaf Node Count Othello}\label{fig:pl2}
\end{figure}

\begin{figure}
% \begin{center}
\includegraphics[width=10cm]{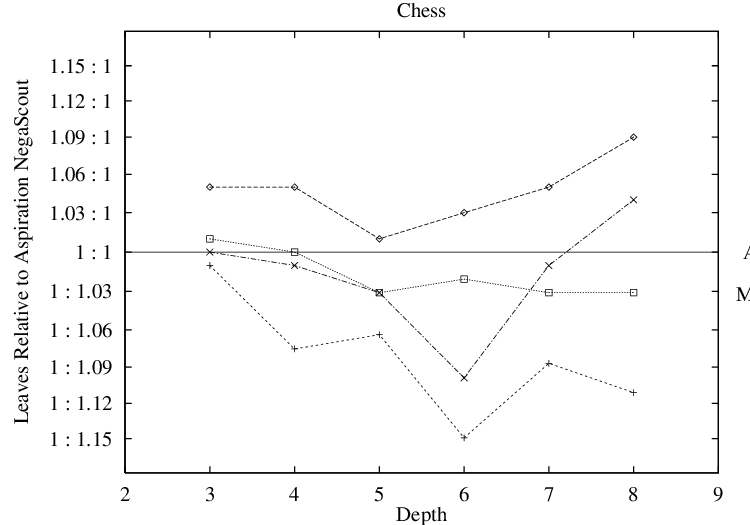}
% \end{center}
\caption{Leaf Node Count Chess}\label{fig:pl3}
\end{figure}

\begin{figure}
% \begin{center}
\includegraphics[width=10cm]{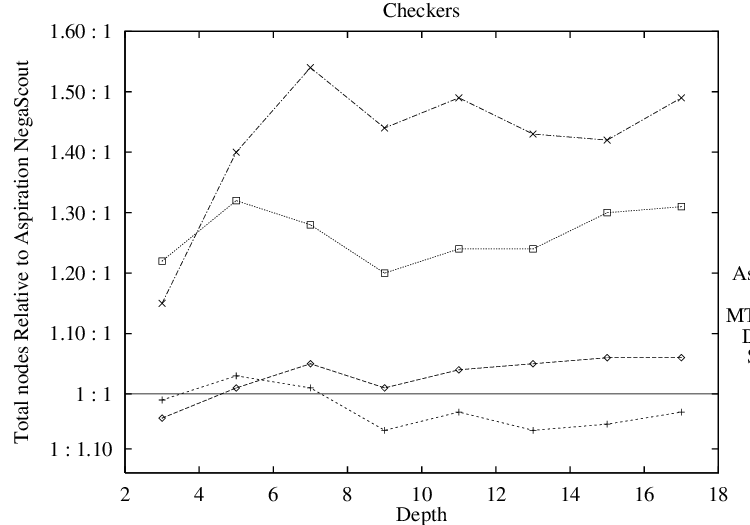}
% \end{center}
\caption{Total Node Count Checkers}\label{fig:pt1}
\end{figure}

\begin{figure}
% \begin{center}
\includegraphics[width=10cm]{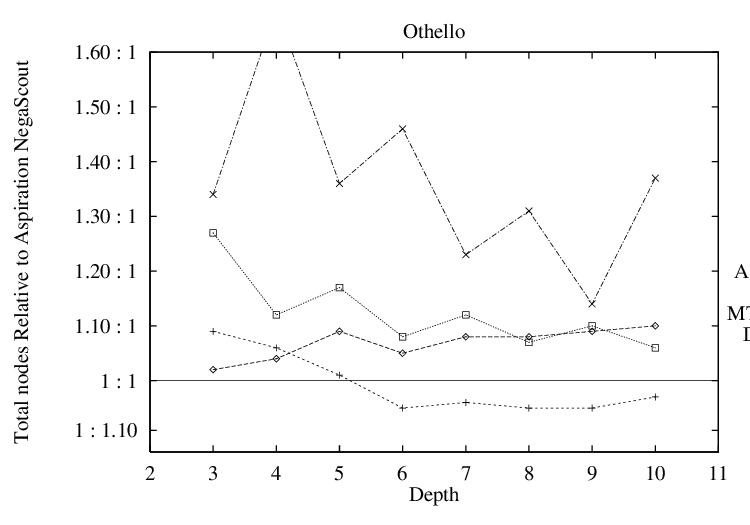}
% \end{center}
\caption{Total Node Count Othello}\label{fig:pt2}
\end{figure}

\begin{figure}
% \begin{center}
\includegraphics[width=10cm]{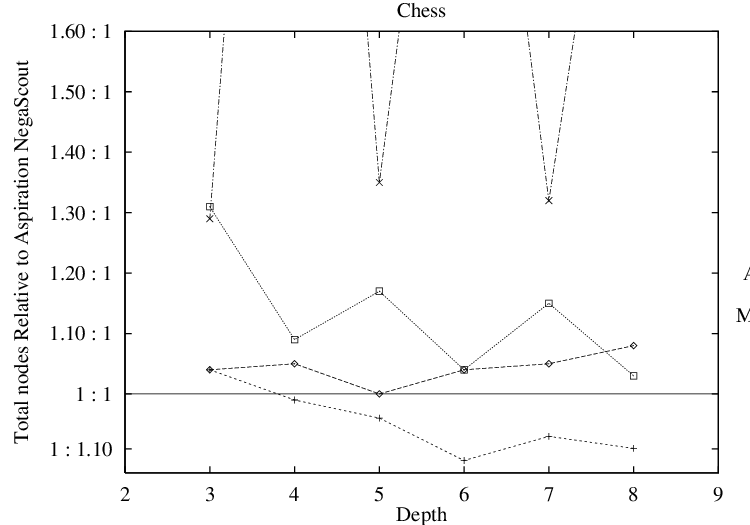}
% \end{center}
\caption{Total Node Count Chess}\label{fig:pt3}
\end{figure}
\subsection{Results} \index{performance graphs}%
Figures~\ref{fig:pl1}--\ref{fig:pl3} show the performance of
{\Chinook}, {\Keyano} and Phoe\-nix, respectively,
using the number of leaf evaluations
as the performance metric.
Figures~\ref{fig:pt1}--\ref{fig:pt3} show the performance of the programs
using the total number of nodes in the search tree
as the metric (note the different scale). 
The graphs show the geometric mean of the cumulative number
of nodes over all previous iterations for a certain depth
normalized to {\AspNS}
(which is realistic since iterative deepening is used).

The lines in the graphs are generally quite jagged. Many lines show an
odd/even oscillation \index{odd/even oscillation}%
depending on the
search depth, although the 
effect is not consistent. Also it appears that generally {\DUAL}
builds slightly smaller trees than {\SSS}, although here too there
are exceptions. These effects can to some extent be explained out of
the basic asymmetry of max and min solution trees: in a uniform max-rooted
minimax tree of odd depth, a min solution tree has less leaf nodes
than a max solution tree. The fact that the lines appear to fluctuate
wildly is partly due to the fine scale of the graphs and  the
irregularity of minimax trees that are generated in actual
applications. 

% old SSS part
\subsubsection{{\SSS} and {\DUAL}}
\index{SSS*!performance  of}%
\index{Dual*, performance  of}%
Contrary to many simulations, 
our results show that the difference in the number of leaves
expanded by {\SSS} and {\AB} is relatively small. 
Since game-playing programs use many search enhancements that reduce
the search effort---we used only iterative deepening, the history
heuristic, and transposition tables---the potential benefits of a
best-first search are greatly reduced. 
In practice, {\SSS} is a small improvement on {\AB}
(depending to some extent on the branching factor).
Claims that {\SSS} and {\DUAL} evaluate significantly fewer leaf
nodes than {\AB} are based on simplifying assumptions that have
little relation with what is used in practice.
In effect, the main advantage of {\SSS}
(point 5 in section~\ref{sec:viewsss} on page~\pageref{sec:viewsss})
has disappeared.
Reasons for this will be
discussed further in section~\ref{sec:discuss}.

\subsubsection{Odd/Even Effect in {\ABSSS} and {\ABDUAL}}
Looking at the graphs for total nodes \index{odd/even effect}%
(figures~\ref{fig:pt1}--\ref{fig:pt3}),  we see a clear odd/even
effect for {\ABSSS} and {\ABDUAL}. The reason is that the former
refines a max 
solution tree, whereas the latter refines a min solution tree. At even
depths the parents of the 
leaves are min nodes. With a wide branching factor, like in chess,
there are many leaves that will
initially cause cutoffs for a high bound, causing a return at their min parent
({\AB}'s cutoff condition at min nodes $g \leq \gamma$ is easily
satisfied when $\gamma$ is close to $+\infty$). Especially since the
move ordering near the leaves gets worse, it is likely
that {\ABSSS} will quickly find a slightly better bound to end each pass, 
causing it to make many traversals through the tree, perform many
hash-table lookups, and make many calls to the move generator. These
traversals 
show up in the total node count and interior node count (not
shown separately). For {\ABDUAL}, the reverse 
holds. At odd depths, many leaves cause a pass to end at the
max parents of the leaves when the bound is close to $-\infty$. (There
is room here for improvement, by remembering which moves have already
been searched. This will reduce the number of hash-table lookups, but
not the number of visits to interior and leaf nodes.)

%The graphs in figure~\ref{fig:leaf} show the number of leaves searched
%by {\ABSSS} and {\ABDUAL} relative to {\AB} (100\%) as a function of
%search depth. The transposition table size is $2^{20}$ entries, so
%that {\AB} is given all the memory it needs for a fair comparison
%against {\ABDUAL}, which performs relatively better than {\AB} with
%smaller transposition tables.
%Since iterative deepening is used,
%each data point is the cumulative figure over all
%the previously searched depths.
%

%Several simulations have pointed out the complementary behavior of
%{\SSS} and {\DUAL} for odd/even search depths.
%Both algorithms are said to significantly out-search {\AB} on leaf nodes,
%but which one is superior depends on the parity of the search depth.
%In our experiments, this effect is less pronounced.
%The graphs for leaf nodes indicate that {\ABDUAL} is slightly superior
%for small 
%branching factors, probably since min solution trees are smaller than
%max solution trees. This difference decreases as the branching factor
%increases. 
%For chess, {\ABSSS} and {\ABDUAL} perform comparably,
%contradicting the literature \cite{Mars87,Rein89,Rein94b}.
%
% ->>>>>> We're using ID!!!!

\subsubsection{Dominance Under Dynamic Move Reordering}
\index{domination}\index{surpassing}%
We see that for certain depths the iterative deepening version of 
{\SSS} expands more leaf nodes than iterative deepening {\AB}
in the case of checkers. This result appears to run counter to
Stockman's proof that {\AB} is dominated by {\SSS} \cite{Stoc79}.
%\subsubsection{Dominance under Dynamic Move Reordering}\label{sec:dom}
%Stockman proved that {\SSS}
%will never expand more leaf nodes than {\AB} \cite{Stoc79}.
How can this be? No one has questioned the assumptions under which
this proof was made. 
In general, game-playing programs do not perform single fixed-depth searches.
Typically, they use iterative deepening and other dynamic move
ordering schemes
to increase the likelihood that the best move is searched first.
%(the difference between the best and worst case ordered
%trees being exponential \cite{Knut75}).
The {\SSS} proof implicitly assumes that every time a node is visited,
its successor moves will $always$ be considered in the same order
(Coplan makes this assumption explicit in his proof of C*'s
dominance over {\AB} \cite{Copl82}). 

%\subsection{Non-dominance of Iterative Deepening {\SSS}}
\label{sec:app-b}
% This section presents an example to prove that {\SSS} with dynamic
% move reordering does not dominate {\AB}.
% Iterative deepening and move reordering are part of all state-of-the-art
% game-playing programs.

\Treestyle{%
  \addsep{1pt}%
  \minsep{3pt}%
  \vdist{26pt}%
  \nodesize{14pt}%
 }    % smaller than default 

 \begin{figure}[t]

{\small
  \begin{tabular}{ccc}\\
{\AB} Depth 2 & After Re-ordering & {\AB} Depth 3 \\
 \begin{Tree}
   \node{\external\type{square}\cntr{d}\bnth{4}}
   \node{\external\type{square}\cntr{e}\bnth{3}}
   \node{\lft{3}\cntr{b}}
   \node{\external\type{square}\cntr{f}\bnth{8}}
   \node{\external\type{square}\cntr{g}\bnth{9}}
   \node{\rght{8}\cntr{c}}
   \node{\type{square}\lft{8}\cntr{a}}
\end{Tree}
\hspace{\leftdist}\usebox{\TeXTree}\hspace{\rightdist}
& 
 \begin{Tree}
   \node{\external\type{square}\cntr{f}\bnth{8}}
   \node{\external\type{square}\cntr{g}\bnth{9}}
   \node{\lft{8}\cntr{c}}
   \node{\external\type{square}\cntr{e}\bnth{3}}
   \node{\external\type{square}\cntr{d}\bnth{4}}
   \node{\rght{3}\cntr{b}}
   \node{\type{square}\lft{8}\cntr{a}}
\end{Tree}
\hspace{\leftdist}\usebox{\TeXTree}\hspace{\rightdist}
  &
 \begin{Tree}
   \node{\external\cntr{h}\bnth{9}}
   \node{\external\cntr{i}\bnth{1}}
   \node{\type{square}\cntr{f}\lft{9}}
   \node{\external\cntr{j}\bnth{1}}
   \node{\external\cntr{k}\bnth{2}}
   \node{\type{square}\cntr{g}\lft{2}}
   \node{\lft{2}\cntr{c}}
   \node{\external\cntr{l}\bnth{3}}
   \node{\external\cntr{m}\bnth{2}}
   \node{\type{square}\cntr{e}\lft{3}}
   \node{\external\cntr{n}\bnth{4}}
   \node{\leftonly\type{square}\cntr{d}\rght{4, cutoff}}
   \node{\rght{3}\cntr{b}}
   \node{\type{square}\lft{3}\cntr{a}}
\end{Tree}
\hspace{\leftdist}\usebox{\TeXTree}\hspace{\rightdist}
\end{tabular}
 }
%\vspace{-4.5cm}
%\mbox{\hspace{-4cm}\epsfxsize=18cm  \epsffile{fig1.ps}}
%\vspace{-15.5cm}
 \caption{Iterative Deepening {\AB}}\label{fig:idab}
\end{figure}

 \begin{figure}[t]
{\small
  \begin{tabular}{ccc}\\
{\SSS} Depth 2 & After Re-ordering & {\SSS} Depth 3 \\
 \begin{Tree}
   \node{\external\type{square}\cntr{d}\bnth{4}}
   \node{\leftonly\lft{$f^+ = 4$}\cntr{b}}
   \node{\external\type{square}\cntr{f}\bnth{8}}
   \node{\external\type{square}\cntr{g}\bnth{9}}
   \node{\rght{8}\cntr{c}}
   \node{\type{square}\lft{8}\cntr{a}}
\end{Tree}
\hspace{\leftdist}\usebox{\TeXTree}\hspace{\rightdist}
& 
 \begin{Tree}
   \node{\external\type{square}\cntr{f}\bnth{8}}
   \node{\external\type{square}\cntr{g}\bnth{9}}
   \node{\rght{8}\cntr{c}}
   \node{\external\type{square}\cntr{d}\bnth{4}}
   \node{\leftonly\rght{$f^+ = 4$}\cntr{b}}
   \node{\type{square}\lft{8}\cntr{a}}
\end{Tree}
\hspace{\leftdist}\usebox{\TeXTree}\hspace{\rightdist}
  &
 \begin{Tree}
   \node{\external\cntr{h}\bnth{9}}
   \node{\external\cntr{i}\bnth{1}}
   \node{\type{square}\cntr{f}\lft{9}}
   \node{\external\cntr{j}\bnth{1}}
   \node{\external\cntr{k}\bnth{2}}
   \node{\type{square}\cntr{g}\lft{2}}
   \node{\lft{2}\cntr{c}}
   \node{\external\cntr{n}\bnth{4}}
   \node{\external\cntr{o}\bnth{5}}
   \node{\type{square}\cntr{d}\lft{5}}
   \node{\external\cntr{l}\bnth{3}}
   \node{\external\cntr{m}\bnth{2}}
   \node{\type{square}\cntr{e}\rght{3}}
   \node{\rght{3}\cntr{b}}
   \node{\type{square}\lft{3}\cntr{a}}
\end{Tree}
\hspace{\leftdist}\usebox{\TeXTree}\hspace{\rightdist}

\end{tabular}
 }
%\vspace{-4.5cm}
%\mbox{\hspace{-4cm}\epsfxsize=18cm  \epsffile{fig2.ps}}
%\vspace{-15.5cm}
 \caption{Iterative Deepening {\SSS}}\label{fig:idsss}
\end{figure}

\Treestyle{%
  \addsep{2pt}%
  \minsep{10pt}%
  \vdist{30pt}%
  \nodesize{14pt}%
 }    % smaller than default 

While building a tree to depth $d$, a node $n$ might consider the moves
in the order $1, 2, 3, \ldots, w$.
Assume move 3 is best.
When the tree is re-searched to depth $d+1$, the transposition table can 
retrieve the results of the previous search.
Since move 3 was successful at causing a cutoff previously,
albeit for a shallower search depth,
there is a high probability it will also work for the current depth.
Now move 3 will be considered first and, if it fails to cause a cutoff,
the remaining moves will be considered in the order $1, 2, 4, \ldots, w$
(depending on any other move ordering enhancements used).
The result is that prior history is used to $change$ the order in
which moves are considered.

Any form of move ordering violates the implied preconditions of
Stockman's proof.
In expanding more nodes than {\SSS} in a previous iteration,
{\AB} stores more information in the transposition table which may later
be useful.
In a subsequent iteration, {\SSS} may have to consider
a node for which it has no move-ordering information whereas {\AB} does.
Thus, {\AB}'s inefficiency in a certain iteration can actually
benefit it later in the search.
With iterative deepening,
it is possible for {\AB} to expand {\em fewer\/} leaf nodes than
{\SSS}.

When used with iterative deepening,
{\SSS} does not dominate {\AB}.
Figures~\ref{fig:idab} and \ref{fig:idsss} prove this point.
In the figures, the smaller depth 2 search tree causes
{\SSS} to miss information that would be useful for the
search of the larger depth 3 tree.
It searches a differently ordered depth 3 tree and,
in this case, misses the cutoff at node $o$ found by {\AB}.
If the branching factor at node $d$ is increased,
the improvement of {\AB} over {\SSS} can be made arbitrarily large.

That {\SSS}'s dominance proof does not hold for
dynamically ordered trees does not mean that
{\AB} is structurally better.
If {\SSS} expands more nodes for depth $d$, it will probably have
more information for the next depth, and it may well out-perform {\AB}
again at depth $d+1$. All it means is that under dynamic reordering
the theoretical 
superiority of {\SSS} over {\AB} does not apply.

The smaller the branching factor,
the more likely this phe\-nom\-e\-non is observed.
The larger the branching factor, the more opportunity there is
for best-first search to offset the benefits of increased information
in the transposition table.

%\input  ks-appb 
% In section~\ref{sec:app-b}
% an example is given that proves the non-dominance of iterative deepening
% {\SSS} over iterative deepening {\AB}.

We conclude that an advantage of {\SSS},
its domination over {\AB} (point 4 in 
section~\ref{sec:viewsss}), is wrong in practice.

% Section~\ref{sec:deeper} discusses more finer points that can be
% learned from these experiments.

\subsubsection{{\AspNS} and {\MTDf}}
\index{MTD$(f)$, performance of}%
\index{NegaScout, performance of}%
The results show that {\AspNS} is better than {\AB}.
This is consistent with \cite{Scha89b} which showed
{\AspNS} to be a small improvement over {\AB} when
transposition tables and iterative deepening were used.

Over all three games, the best results are from {\MTDf}. Not
surprisingly, the current algorithm of choice by the game programming
community, {\AspNS}, performs well too. 
The averaged {\MTDf}  leaf node counts are consistently better than
for {\AspNS},
averaging a 5 to 10\% improvement, depending on the game. 
More surprising is that {\MTDf} out-performs {\AspNS} on the 
total node measure as well.
%Since each iteration requires repeated calls to {\MT}
%(at least two and possibly many more), 
%one might expect
%{\MTDf} to perform badly by this measure
%because of the repeated traversals of the tree.
This suggests that {\MTDf} is calling {\MT}
close to the minimum number of times (which is 2---one for the upper
bound, one for the lower bound).
Measurements confirm that for all three programs, {\MTDf} calls {\MT} 
about 3 to 6 times per iteration on average.
In contrast, the {\ABSSS} and {\ABDUAL} results are poor compared
to {\AspNS} when all nodes in the search tree are considered.
Each of these algorithms usually performs 
hundreds of {\MT} searches. The wider the range of leaf values, the
smaller the steps with which they converge, and the more passes
they need.

\begin{figure}
\begin{center}
\includegraphics[width=10cm]{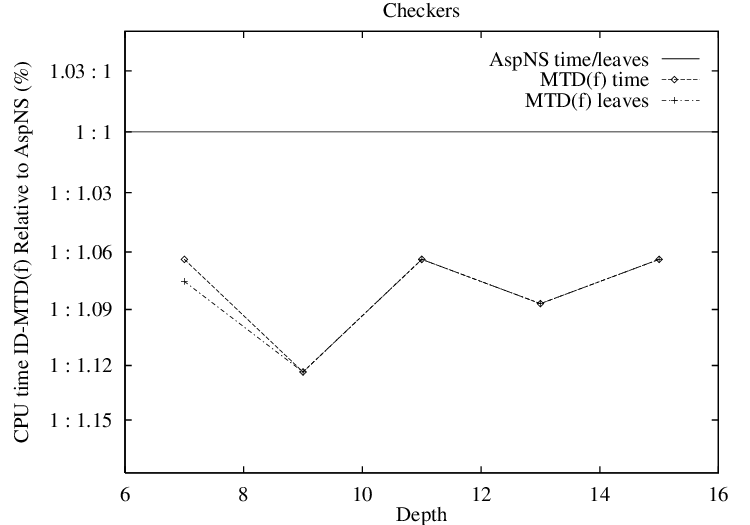}
\end{center}
\caption{Execution Time Checkers}\label{fig:cpu1}
\end{figure}

\begin{figure}
\begin{center}
\includegraphics[width=10cm]{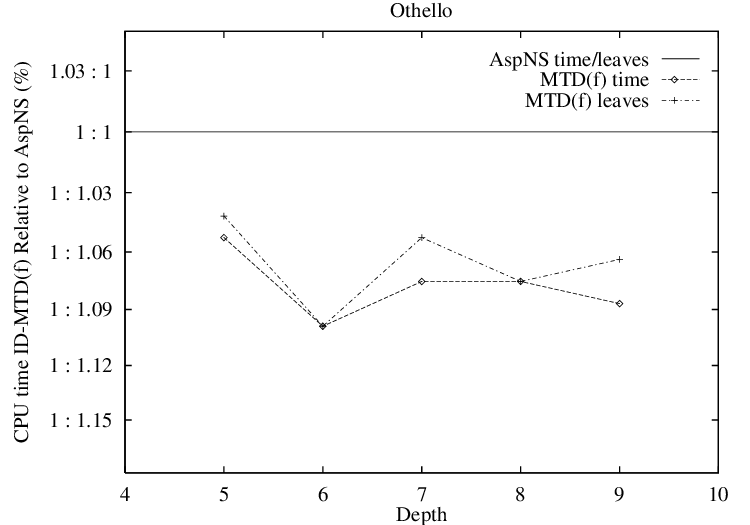}
\end{center}
\caption{Execution Time Othello}\label{fig:cpu2}
\end{figure}

\begin{figure}
\begin{center}
\includegraphics[width=10cm]{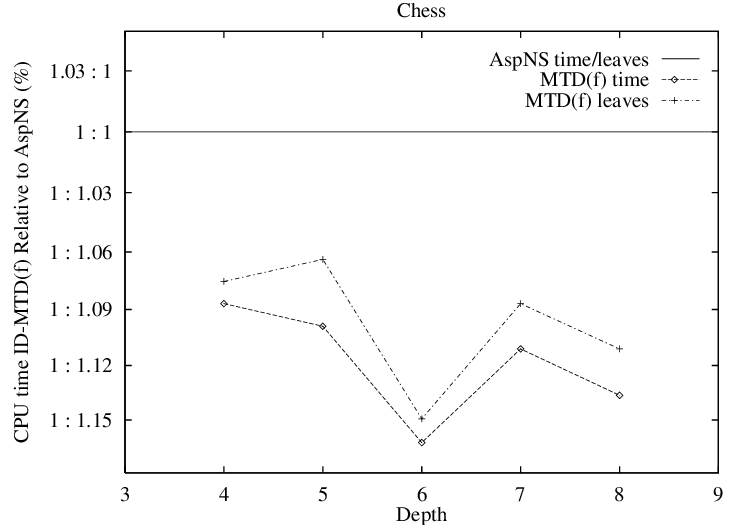}
\end{center}
\caption{Execution Time Chess}\label{fig:cpu3}
\end{figure}
\subsection{Execution Time}\index{execution time}%
The bottom line for practitioners is execution time.
This metric may vary considerably for different
programs. It is nevertheless included, to give evidence of the
potential of {\MTDf} (figures~\ref{fig:cpu1}--\ref{fig:cpu3}).  
%We did not have
%the resources to run all our experiments on identical and otherwise
%idle machines. 
%We only show execution time graphs for ID {\MTDf} and
%ID {\AspNS} (figure~\ref{fig:cpu}), the comparison that we think is
%the most interesting, since comparing results for the same machines we
%found that 
%{\MTDf} is consistently the fastest algorithm. 
We only show the deeper
searches, since the relatively fast shallower searches hamper accurate
timings. The runs shown are typical example runs on a Sun SPARC. We did
experience 
different timings when running on different machines. It may well be
that cache size plays an important role, and that tuning  the
program  has an impact as well. 
%After all, our programs we
%highly tuned for {\AspNS}.

The experiments showed that  for {\Chinook} and {\Keyano},
{\MTDf} was about 5\% faster in execution time than {\AspNS}; for
{\Phoenix} we found {\MTDf} 9 to 16\% faster. (Application dependent
tuning of {\MTDf} can improve this a few 
percentage points, see section~\ref{sec:swallow}.)
For other programs and other
machines these results will obviously differ, depending in part
on the quality of $f$ and on the test positions used. For programs of
lesser quality, the performance difference will be bigger, with
{\MTDf} out-performing {\AspNS} by a wider margin.
Also, since the tested algorithms
perform quite close together, the relative differences are quite
sensitive to variations in input parameters. In generalizing these
results, one should keep this sensitivity in mind. Using these numbers
as absolute predictors for other 
situations would not do justice to the complexities of real-life game
trees. The experimental data is better suited to provide insight into, or
guide and verify hypotheses about 
these complexities, as done, for example, in chapter~\ref{chap:anal}.

\section{Null-Windows and Performance} \index{narrow search window}
This section looks at some relations between the value of bounds and
the search 
effort that must be expended to compute them with a null-window {\AB}
search. 

\subsection{Start Value and Search Effort}\label{sec:start}
% The minimax wall occurs 
% with a single null-window call with a value somewhere in the middle of the
% scale, close to the minimax value. Next we will look at
% a {\em sequence\/} of null-window calls
% starting in 
% the middle of the scale, close to the minimax value.
% 
% {\MTDf} does. Regrettably, we were not able to find an analytical
% explanation, so we resort to empirical measurements.
% 
\label{sec:swallow}
% In this section we will investigate the relation between the size
% of the search tree and the start value of a sequence of {\MT} calls.
% 
The  biggest difference in the {\MTD} algorithms
is their first approximation of the minimax value:
{\SSS}/{\MTDp} is optimistic, {\DUAL}/ {\MTDm} is pessimistic and
{\MTDf} is realistic. 
It is clear that starting close to $f$, assuming integer-valued
leaves, should result in convergence in less steps, simply
because there are fewer discrete values in the range from the start
value to $f$. 
%On average, it turns out there is a relationship between the starting bound
%and the number of times that {\MT} makes a pass over the search tree.
%(In the light of Baudet's work on aspiration window searching, perhaps
%this is to be expected somehow \cite{Baud78b}.)
If each {\MT} call at the root expands roughly the same number of nodes, then 
doing less passes yields a better algorithm. However, this is not the
case.
Generally an
{\MT} call with a loose bound, like $+\infty$, is 
cheaper than an {\MT} call with a tight bound, like $f+4$. For a
loose bound the left-first solution tree suffices. For tighter bounds
it takes more work to get a cutoff, and hence the work to find 
the solution tree for the bound
is greater. Also, in well-ordered seach spaces, the construction of the first
solution tree is by far the most expensive. Refining it to yield a
slightly sharper bound costs only a few node expansions. 
% The
% construction of the matching solution tree of the other kind in the
% end is again an expensive computation, since there is no previous
% solution tree to build on. This process has been called ``hitting the
% minimax wall''  by Schaeffer \cite{Mars87,Schaeffer86}.
Furthermore, max solution trees contain
$w^{\lceil d/2 \rceil}$ leaf nodes, while min solution trees
%(lower bounds, fail high, provebest),\footnote{The terms disprovebest
%  and provebest were introduced by Hans Berliner \cite{Berl79}. We use
%  them here in a slightly different context, not to denote a strategy
%  to compute heuristic bounds, but 
%  max and min solution trees.} 
contain $w^{\lfloor d/2 \rfloor}$ leaf nodes (if trees are
of uniform width and depth).\label{sec:mmwall}\index{Schaeffer}

Schaeffer has looked at the size of the search tree of an isolated
null-window call, for different values \cite{Mars87,Schaeffer86}. 
% A related phenomenon is Schaeffer's minimax wall \cite{Schaeffer86},
% also called ``refutation wall'' in \cite{Mars87}. 
% performed
% experiments with  null-window 
% {\AB} searches, 
\index{minimax wall}%
The results pointed out that not all
null-windows are equal. Using Phoenix on the 24 Bratko-Kopec positions
\cite{Kope82}, 
a search with a null-window on the low side of $f$ 
was significantly cheaper than a search on the high side, for both odd
and even search depths. At $f$ the search effort jumped to a higher
level. This sharp increase in search effort has been called the
{\em minimax wall}.\index{Phoenix}
In a program, such as Phoenix, that searches well-ordered trees, a fail
high at the root ($f^-$, $T^-$) will occur before all children have 
been searched. An {\AB} call resulting in a fail low ($f^+$, $T^+$)
will have expanded  all children at the root. 

Thus, {\MT} calls generally do {\em not\/} expand the same
number of nodes. Since we could not find an analytical solution 
we
have conducted experiments to test the 
intuitively appealing idea that starting a search close to $f$ is
cheaper than starting far away.

\begin{figure}
\begin{center}
\includegraphics[width=10cm]{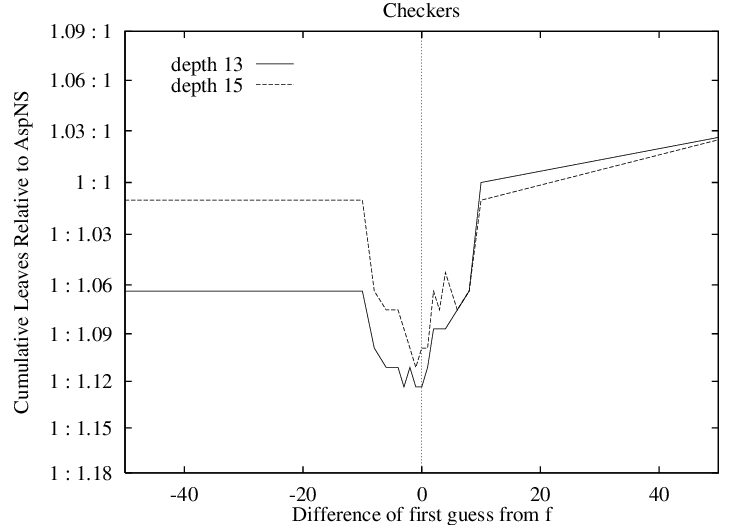}
\end{center}
\caption{Tree Size Relative to the First Guess $f$ in Checkers}\label{fig:swallow1}
\end{figure}
\begin{figure}
\begin{center}
\includegraphics[width=10cm]{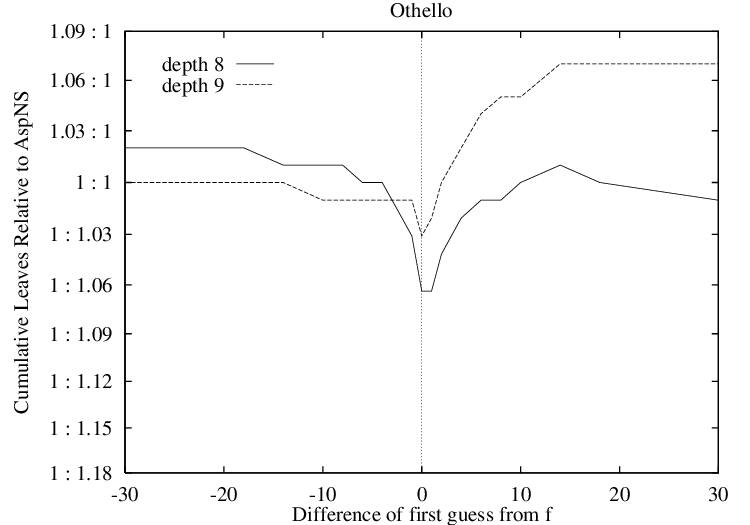}
\end{center}
\caption{Tree Size Relative to the First Guess $f$ in Othello}\label{fig:swallow2}
\end{figure}
\begin{figure}
\begin{center}
\includegraphics[width=10cm]{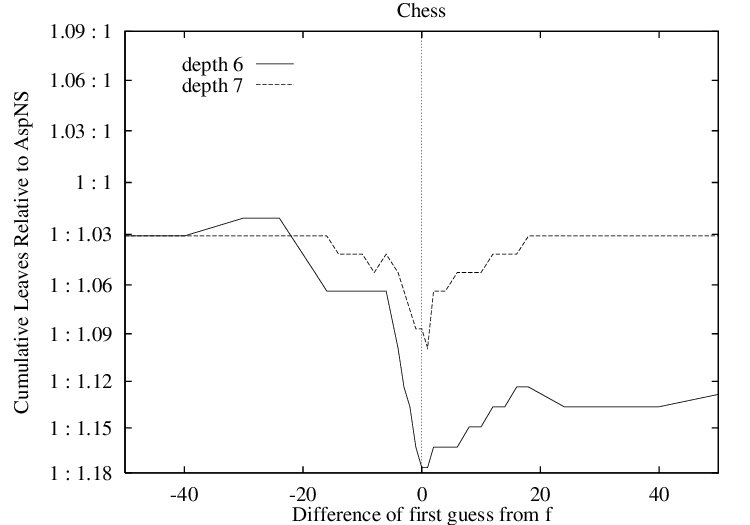}
\end{center}
\caption{Tree Size Relative to the First Guess $f$ in Chess}\label{fig:swallow3}
\end{figure}
Figures~\ref{fig:swallow1}--\ref{fig:swallow3} validate the choice of
a starting parameter 
close to the game value.
The figures show the efficiency of an iterative deepening search as a
function of the  
distance of the first guess from the correct minimax value for each
search depth.
The data points are given as a percentage of the size of the search tree 
built by {\AspNS}.  
\begin{figure}
%\begin{center}
\includegraphics[width=10cm]{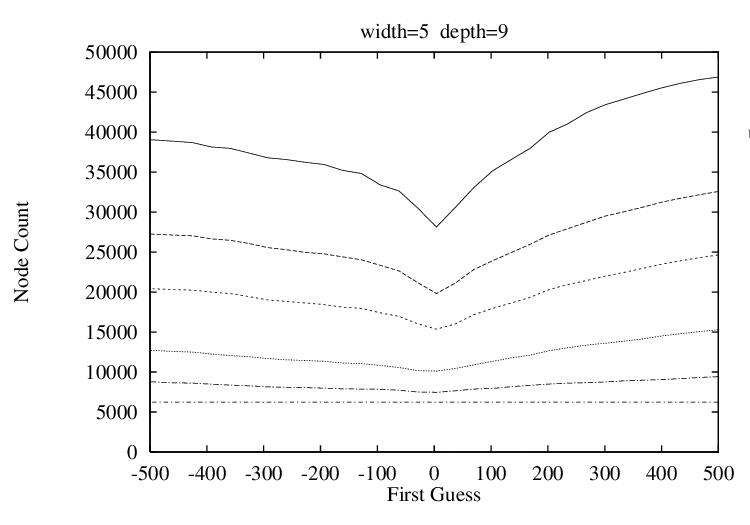}
%\end{center}
\caption{Effect of First Guess in Simulated Trees}\label{fig:swallowsim}
\label{fig:simswallow}
\end{figure}
\index{MTD$(f)$}\index{SSS*}\index{Dual*}%
\index{start value}%
To the left of the graph, {\MTDf} is closer to {\DUAL}/{\MTDm},
to the right it is closer to {\SSS}/{\MTDp}. It is instructive to
compare these figures with figure~\ref{fig:swallowsim}, which is based
on simulated trees---no iterative deepening or transposition tables
there. Now the  curves look reassuringly smooth. The graph is taken
from \cite{DeBruin94}. Each line in the graph is the average of 20
artificial trees of width 5 and depth 9. The top
line shows
unordered trees, where the first
move has a $\frac{1}{w} = 20\%$ probability of being best. The lower lines are
progressively better ordered. The bottom line  represents a
perfectly-ordered tree. Here all algorithms search the minimal tree,
independent 
of the start value of the search.
Figure~\ref{fig:swallowsim} shows that the closer a search  starts 
to the minimax value of a game tree, the less nodes are expanded, on
average. The gain in performance is less in ordered trees. 

The graphs in figures~\ref{fig:swallow1}--\ref{fig:swallow3} show that
the smaller the distortion, the smaller the search 
tree is. Our intuition that starting close to the minimax value is a
good idea is justified by these experiments.
A first guess close to $f$ makes {\MTDf} perform better than the 100\%
{\AspNS} baseline.
We also see that the guess must be quite close to $f$ for the effect to
become significant. 
Thus, if {\MTDf} is to be effective, the $f$ obtained from the previous
iteration must be a good indicator of the next iteration's value.
For programs with a
pronounced odd/even oscillation in their score, results are better if
the score from two iterations ago is used. Comparing the graphs in
figures~\ref{fig:pl1}--\ref{fig:pl3} and
\ref{fig:swallow1}--\ref{fig:swallow3}, we see that {\MTDf} is not  
achieving its lowest point, so there is room for improvement.
Indeed, we found that adjusting the first
guess by $\pm$ 1 to 4 points for each iteration can improve the
results for {\MTDf} in terms of 
leaf count by two to three percentage points. This can be regarded as
a form of
application-dependent fine tuning
%---albeit a very convenient and simple form---
of the {\MTDf} algorithm.
% Considering all this, it is not a surprise that both {\DUAL} and {\SSS}
% come out poorly. 
%Their initial bounds for the minimax value are generally poor
%($-\infty$ and $+\infty$ respectively),
%meaning that the many calls to {\MT} result in significantly more
%nodes.

When a {\em single\/} null-window {\AB} search for a bound is
performed, a search for a loose 
bound generally takes less effort than a search for a tight bound. The
figures~\ref{fig:swallow1}--\ref{fig:swallowsim}, however, show the
effort of {\em a number\/} of null-window searches, of all the
searches that are needed to prove the minimax value.

\begin{figure}
\begin{center}
\includegraphics[width=10cm]{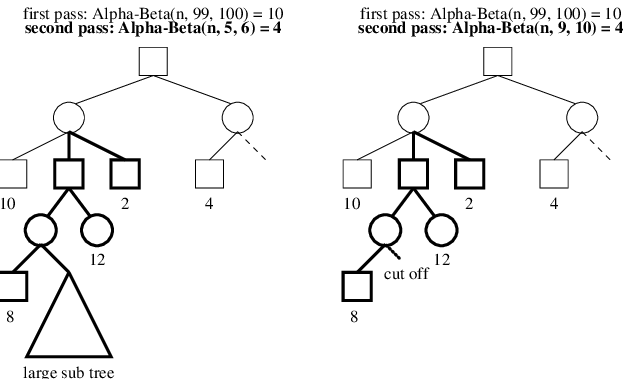}
\end{center}
  \caption{Two Counter Intuitive Sequences of {\MT} Calls}
  \label{fig:ce}
\end{figure}

In doing these experiments, the diversity of real-life game
trees became apparent.
It is not hard to construct a counter-example  \index{counter-example}%
where a bad null-window expands {\em less\/} nodes than a good first
guess. 
For example, figure~\ref{fig:ce} shows that {\AB}$(root, 99,
100)$ followed by the re-search {\AB}$(root, 9, 10)$ (in bold) skips
the large sub-tree, whereas the call
{\AB}$(root, 99, 100)$ followed by {\AB}$(root, 5, 6)$ (in bold) 
expands it, where the $\beta$ of 6 is closer to $2 \leq f \leq 4$ than
the $\beta$ of 10.
In the tests we also encountered 
some positions where  {\AspNS} performed better than {\MTDf}.

\subsection{Start Value and Best-First}

%\subsection{Algorithmic Factors}
\label{sec:bfdf}
One could ask the question how it is possible at all for a depth-first
algorithm like {\NS} to out-perform a best-first algorithm like
{\SSS}. The answer is given by the influence of the start value of an
{\MT} sequence. {\NS} derives its start value for later recursive
null-window calls
from the tree it is searching. If that tree has become relatively
well-ordered 
through the use of enhancements, such as iterative deepening and the
history heuristic, then the start 
value will become a reasonable guess. {\SSS} does not use this
idea; instead it uses best-first node selection based on the
information from a previous pass. In our tests, with programs using
many search enhancements, a good start value seems to be a bit more
effective. Best-first schemes tend to have an advantage on trees with a lower
quality of move ordering and a wider branching factor.
\begin{figure}
\begin{center}
\begin{tabular}{r|cc}
 & best-first & {\bf not} best-first \\ \hline 
good start value & {\MTDf} & {\NS} \\
{\bf not} good start value & {\SSS} & {\AB} 
\end{tabular}
\end{center}
\caption{Four Algorithms, Two Factors}\label{fig:bfdftable}
\end{figure}
The table in figure~\ref{fig:bfdftable} shows the two ideas:
\begin{enumerate}
\item {\em Best-First Selection}\\
The best-first node selection scheme that {\SSS} and other {\MT}
instances use, is essentially based on a traversal of 
solution trees from previous search passes to descend the principal
variation to the critical leaf, and then expand an open brother of
this leaf---the {\em best\/} node to expand in the {\SSS} sense
% find nodes that have to be searched
% in all cases---that cannot be killed by other nodes
(see figure~\ref{fig:best}).
\item {\em A Good Start Value}\\
Section~\ref{sec:swallow} showed that generally 
a start value closer to the final minimax value yields a more
efficient search for an {\MT} sequence.
\end{enumerate}
The table in figure~\ref{fig:bfdftable} suggests that {\MTDf} combines
the good parts of two existing algorithms: best-first expansion from
{\SSS} and a good start value from {\NS}.
\index{transposition table size, tiny}
{\NS} and {\MTDf} are the two algorithms that perform best in the
experiments. One is categorized as depth-first, the other as
best-first. Otherwise, they have much in common. By looking at their
behavior in very small memory situations, we will be able to see some
similarities and differences more clearly, since the best-first
behavior of {\MTDf} depends on the information of previous passes to
guide it through the tree.
% 
% {\MTDf} consists solely of null-window searches. In each pass, the
% previous search results are used to select the ``best'' node. The
% majority of  {\NS} searches are also done with a
% null-window. As we have seen, an important
% difference is the value of the input parameter to the null-window search.
% {\NS} derives this value from the tree itself,
% whereas {\MTDf} relies for the first guess on information from outside
% the tree. (In our
% experiments the minimax value from a previous 
% iteration was used for this purpose.) {\NS} is a recursive depth-first 
% algorithm, using null-window calls to construct minimal trees in a
% depth-first, left-most, bottom-up fashion. 
% {\MTDf} on the other hand,
% refines the solution tree built by a previous null-window call, using
% the critical path in the previous solution tree to select nodes best-first.
% Solution trees are smaller than minimal trees,
% providing the room for {\MTDf} to out-perform {\NS}.
% % In refining the solution tree, {\MTDf} 
% % traverses a principal variation of nodes with equal
% % bounds to perform a best-first expansion order based on upper or lower
% % bounds.

From section~\ref{sec:results} we recall that the many {\MT} calls of
{\ABSSS} and {\ABDUAL} make those algorithms perform badly when the
transposition table is too small to contain the previously expanded
solution tree. Since {\MTDf} performs significantly fewer calls,
re-expansions due to insufficient storage are not as big a
problem. 
% Compared to one-pass/wide-window
% {\AB}, the few-pass/null-window {\MTDf} performs even better than
% {\AB} when given less memory than needed for the solution tree.
% An explanation for this surprising behavior, a best-first algorithm
% using less memory than a depth-first algorithm, can be found in the
% literature on {\NS} \cite{Pear84,Rein89}.
% For {\NS}, the benefit of the cheaper null-window searches out-weighs
% a few re-searches, even if there is not enough memory to prevent the
% re-expansions \cite{Camp83,Mars87,Pear80}. This also holds for
% {\MTDf}'s behavior in small-memory situations.

\begin{figure}
  \begin{center}
    \includegraphics[width=10cm]{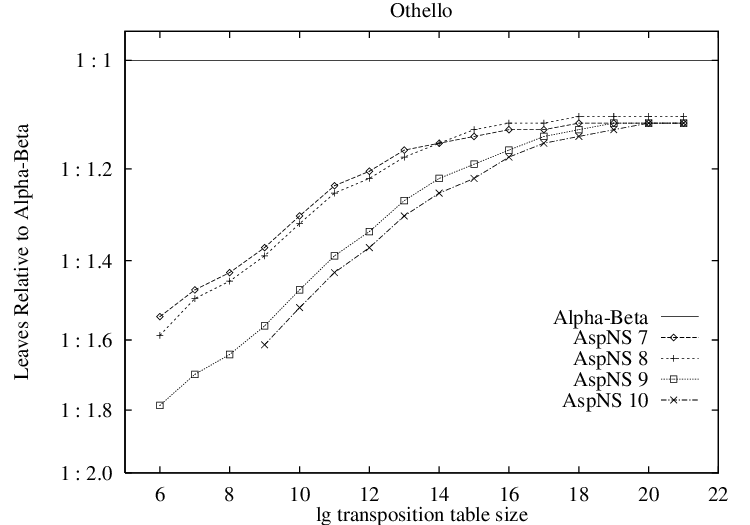}
  \end{center}
  \caption{{\AspNS} in Small Memory in Othello}\label{fig:tinykns}
\end{figure}
\begin{figure}
  \begin{center}
   \includegraphics[width=10cm]{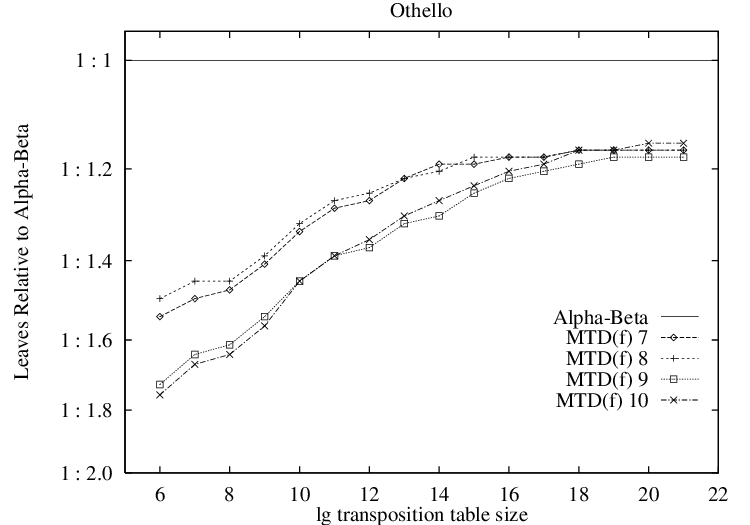}
  \end{center}
  \caption{{\MTDf} in Small Memory in Othello}\label{fig:tinykmtdf}
\end{figure}
\begin{figure}
  \begin{center}
  \includegraphics[width=10cm]{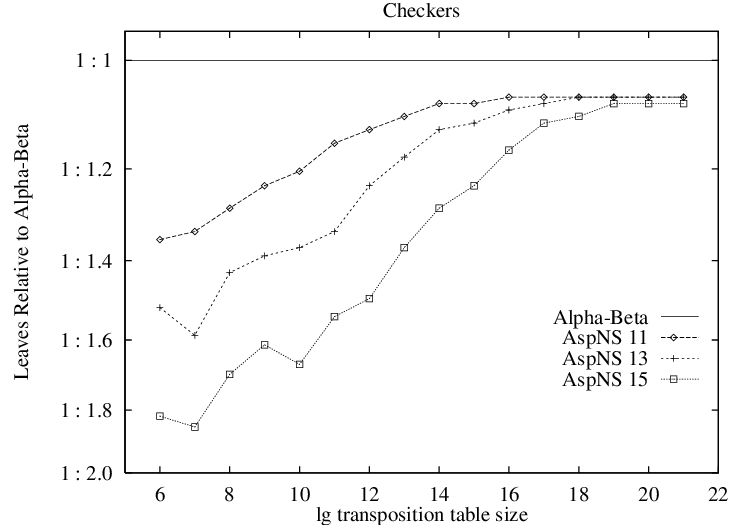}
  \end{center}
  \caption{{\AspNS} in Small Memory in Checkers}\label{fig:tinycns}
\end{figure}
\begin{figure}
  \begin{center}
  \includegraphics[width=10cm]{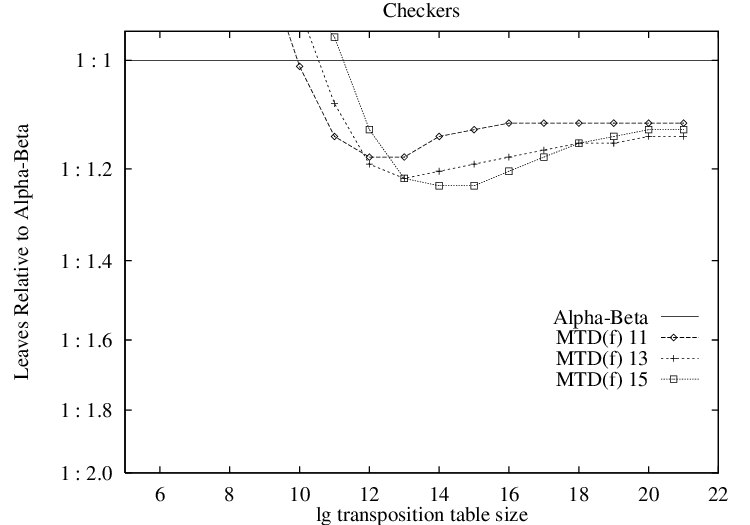}
  \end{center}
  \caption{{\MTDf} in Small Memory in Checkers}\label{fig:tinymtdf}
\end{figure}
\begin{figure}
  \begin{center}
  \includegraphics[width=10cm]{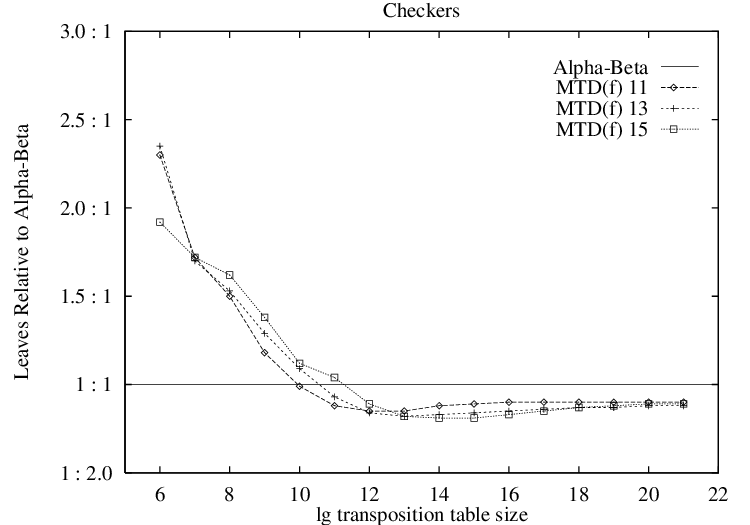}
  \end{center}
  \caption{{\MTDf} in Small Memory (Chinook) II}\label{fig:tinymtdfx}
\end{figure}

This point is illustrated in \label{sec:tiny}
figures~\ref{fig:tinykns}--\ref{fig:tinymtdfx}. In these graphs the
size of the transposition table has been reduced gradually to as low
as $2^6 = 64$ entries. The graphs show test results (leaf count) for
Othello ({\Keyano}) and checkers ({\Chinook}), with iterative
deepening versions of {\AB}, 
{\NS} and {\MTDf}. For {\Keyano} depth 7--10 is shown, for {\Chinook}
depths 11, 13 and 15. Note for {\Keyano} the odd/even effect: for small
memory the results for the depths  are paired. For clarity, the
Chinook result for {\MTDf} is shown again in
figure~\ref{fig:tinymtdfx} with a 
different scale on the $y$ axis. (The size of the transposition table
in {\Phoenix} 
cannot be reduced easily beyond $2^{12}$. In the range from
$2^{12}$--$2^{21}$ the results resemble {\Keyano}'s graphs.) \\

\noindent {\bf \AspNS}\\
We see that {\AspNS} out-performs {\AB} in small memory by a wide\index{NegaScout, performance of}
margin, like the literature on {\NS} 
predicted \cite{Mars87,Pear84,Rein89}. 
As the transposition table gets smaller, the margin grows wider.
Apparently, the fact that the move ordering information gradually
disappears, hurts 
{\AB} more than {\NS}.
% e move ordering
% deteriorates, causing slower convergence of $\alpha$ and
% $\beta$. 
{\NS} still benefits from the tight bounds of null-window calls, where
{\AB}'s search window converges more slowly.  
These graphs support the reasoning 
in section~\ref{sec:abmem}
that for high performance {\AB} has in fact exponential memory
requirements. \\

\noindent {\bf \MTDf}\\
For {\MTDf} the figures need more explanation. For {\Keyano}, {\MTDf}'s\index{MTD$(f)$, performance of}
sensitivity to memory is comparable to that of {\AspNS}. The graph for
Chinook is  different. As long as  the 
transposition table has more than $2^{12} = 4096$ entries, a smaller
transposition table hurts {\AB} more than {\MTDf}. As the table gets
extremely small, {\MTDf}'s relative performance deteriorates rapidly,
although it remains much better than for {\ABSSS}
(figure~\ref{fig:ttsize1}). The different memory
sensitivity of {\MTDf} in {\Chinook} and {\Keyano} is caused by a
different range of the evaluation function. {\Chinook}'s evaluation
function range is $\pm 9000$, {\Keyano}'s range is $\pm
64$.  The small range for {\Keyano} causes the value of the previous
iteration to be quite a good estimate. {\MTDf} rarely calls {\MT} more
\index{evaluation function range}
than 3 
times. The wider range in {\Chinook} causes {\MTDf} to call {\MT} more
often, in some positions  2 or 3 times, in some other 12 to  20
times. 
% (which is still roughly 20 times less than {\ABSSS}). 
The
absence of enough memory makes these {\MT} re-searches much 
more expensive, causing the number of leaf evaluations to rise to 2--2.5
times that of {\AB}, for a table with 64 entries.

This points to a difference between {\NS} and
{\MTDf}. {\NS} is fully recursive, in contrast to
{\MTDf}, which restarts the search a few times at the root.
%  in the way they find their ``good start value.''  {\MTDf} gets
% it externally, from a previous iteration. {\NS} finds it inside the
% tree, it is the value of a brother of the node that is to be
% searched. {\NS} uses it only to search the sub tree of this node, a
% search that is small 
% compared to {\MTDf}'s {\MT} call. 
% % {\MTDf} takes a bigger 
% % risk, 
\index{re-search}
By starting the  re-searches at the root, {\MTDf} takes the risk of
having to re-expand bigger trees than {\NS}, the impact
of which is  normally reduced by the presence of memory 
to the extent that {\NS} is out-performed. As a consequence, the
absence of memory removes the best-first nature of {\MTDf}'s node
selection scheme. With limited memory, there is not enough information
in the transposition table to guide {\MT} towards the
best node to select next. All the previous search information has to
be re-expanded in each pass.
The tests show that for depth 15 at least $2^{12}$
transposition-table entries should be present for Chinook, because of
its wide evaluation-function range (assuming 16
byte entries, 
this amounts to 64 kilobyte of memory). If there is almost no memory 
at all, then {\NS} performs better. 

It is possible to improve the low-memory behavior of {\MT} instances
by changing the replacement strategy of the transposition
table. Currently nodes close to the root are favored. For {\MT} a
preference for the last traversed solution tree would be
better. However, the test results indicate that there still is a wide
margin before tournament game-playing programs would benefit from
these changes. The problems of {\MTDf} only occur with extremely
small amounts of memory. Otherwise
{\MTDf} out-performs all tested algorithms in all three games.

\begin{figure}
\begin{center}
\includegraphics[width=10cm]{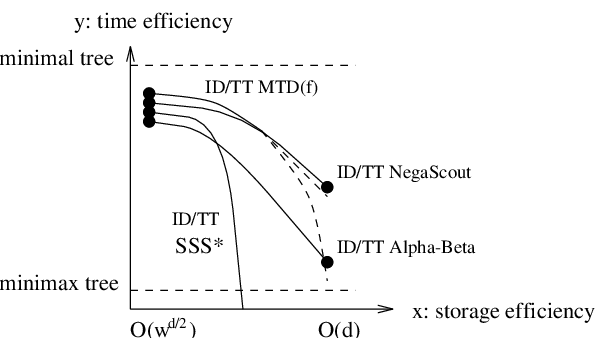}
\end{center}
\caption{Performance Picture of Practical
Algorithms}\label{fig:mmabnssssmtdfpic}
\end{figure}
\index{performance picture!practical algorithms}
Figure~\ref{fig:mmabnssssmtdfpic} shows a performance picture of
algorithms as they are used in a practical setting, with a transposition
table, iterative deepening, and other move ordering enhancements, in
contrast to figures~\ref{fig:abpic}, \ref{fig:abnspic} and
\ref{fig:mmabnsssspic}, that showed 
the un-enhanced, standard text-book versions of the algorithms. All
these enhancements reduce the number of nodes considered before
finding the cutoff to the extent that all algorithms
perform almost the same. 
% The memory usage of the algorithms is
% also determined by what they share, the enhancements: they need all
% much more storage for high performance. 
The enhancements also determine the memory needs for high performance.
The picture shows that more memory improves the effectiveness of the
enhancements. 
Using the same amount of storage
as {\AB} and {\ABSSS}, {\NS}
and {\MTDf} generally perform better than the other two. The null-windows  
cause more cutoffs to occur than with {\AB}$(n, -\infty, +\infty)$. For the
highest performance they  
need  $O(w^{d/2})$ as well, although their performance is less sensitive
to smaller transposition tables than {\AB} and {\ABSSS}.
Their good start value for the null-window {\AB} searches
makes them perform less re-searches than {\ABSSS}.
(The two
dotted lines indicate that {\MTDf}'s performance in low memory
situations varies. It depends on the number of re-searches, as with
{\ABSSS}. A narrow
evaluation-function range causes less re-searches and a better
performance, see
figures~\ref{fig:tinykmtdf} and \ref{fig:tinymtdf}, and their
explanation.)
Again, the experiments showed that memory requirements of $O(w^{d/2})$ 
are perfectly reasonable.

\section{{\SSS} and Simulations}\label{sec:simsucks}
% 
% 
% \section{Performance Results in Perspective}
\label{sec:discuss}\label{sec:sim}
\index{simulation} \index{artificial tree} \index{real tree}
The list in section~\ref{sec:sssview} 
summarized the general view on {\SSS} in five points. 
Three of these points were drawbacks that were declared 
``solved'' in section~\ref{sec:sssprac}.
The remaining two points were positive ones: {\SSS}
provably dominates {\AB}, and it expands significantly fewer leaf
nodes. 
With the disadvantages of the algorithm solved,
the question that remains is: what about the advantages in practice?

The first of the two advantages, theoretical domination, has disappeared. 
With dynamic move reordering,
Stockman's dominance proof for {\SSS} does not apply.
Experiments confirm that {\AB} can out-search {\SSS}.
% The likelihood of this scenario occurring is strongly tied to the
% branching factor, frequently occurring for checkers and rarely
% for chess.

The second advantage was that {\SSS} and {\DUAL} expand significantly
less leaf nodes. 
However, modern game-playing programs do a nearly optimal job of
move ordering, and employ other enhancements that are effective
at improving the efficiency of the search,
considerably reducing the advantage of
null-window-based
best-first strategies.
The experiments show that {\SSS} offers some search tree size advantages
over {\AB} for chess and Othello, but not for checkers. These small
advantages disappear when comparing to {\NS}.
Both {\ABSSS} and {\ABDUAL} compare unfavorably to {\AB} and {\NS} 
when all nodes in the search tree are considered.
%Each of these algorithms performs dozens and sometimes even
%hundreds of {\AB} searches,
%depending on how wide the range of leaf values is.
%
%Part of the problem
%is the high number of re-searches, which makes it less well suited for
%trees with many interior nodes and slow move generation. Luckily,
%there is a solution to {\SSS}'s problem.  The
%re-searches are used to decrease the upper bound on $f$ in tiny steps
%from $+\infty$ down to the minimax value. In section~\ref{sec:mtd} we
%have presented better ways to zoom in on $f$, like bigger steps,
%bisection, and heuristic guesses: {\MTDf}. These better ways have the
%additional advantage that starting closer to $f$ expands leaves as
%well. 

All algorithms, including {\MTDf}, perform 
within a few percentage points of each other's leaf counts.
Simulation results show that for fixed-depth searches,
without transposition tables and iterative deepening,
{\SSS}, {\DUAL} and {\NS} are major improvements over simple {\AB}
\cite{Kain91,Mars87,Musz85,Rein89}.
For example, one study shows {\SSS} and {\DUAL} building trees that are
about
half the size of those built by {\AB} \cite{Mars87}.
This is in sharp contrast to the results reported here.
The reason for this disparity with previously published work
is the difference between real and
artificial minimax trees.

%\subsection{Real and Artificial Game Trees}\label{sec:aim}
%{\bf Please make a choice of either this subsection, or the next one,
%  or a mix of the two, or something better of your own hand}
%
The literature on minimax search abounds with investigations into
the relative performance of algorithms. In many publications 
artificially-generated game trees are used to test these algorithms.
We argue that artificial trees are too simple
to form a realistic test environment.

Over the years researchers have uncovered a number of
interesting features of minimax trees as they are generated in actual
application domains like game-playing programs.
The following four features of real game trees can be exploited by
application-independent techniques to increase the performance
of search algorithms. 
\begin{itemize}
\item {\em Variable branching factor}\\
  The number of children of a node is \index{branching factor}
  often not a constant. Algorithms such as Proof Number and Conspiracy
  Number Search use this fact to guide the search in a
  ``least-work-first'' manner \cite{Allis94,McAl88,Scha90}.
\item {\em Value interdependence between parent and child nodes} \\
  A shallow search is often a good approximation of a deeper search.
  This notion is used in techniques like
  \index{correlation, parent-child value}
  iterative deepening, which---in conjunction with storing
  previous best moves---greatly increases the quality of move
  ordering. Value interdependence also
  facilitates forward 
  pruning and move ordering based on shallow searches \cite{Buro95}.
\item {\em Value independence of moves}\\
  In many domains there exists a \index{history heuristic}
  global partial move ordering: moves that are good in one position
  tend to be good in another as well. This fact is used by the history
  heuristic and the killer heuristic \cite{Scha89b}.
\item {\em Transpositions}\\
  The fact that the search space is most often a graph \index{transposition}
  has lead to the use of transposition
  tables. In some games, notably chess and checkers, they lead to a substantial
  reduction of the search effort \cite{Plaa94c}.
  Of no less importance is the better move 
  ordering, which drastically improves the effectiveness of {\AB}.
\end{itemize}
Furthermore, in many simulation experiments the nodes are  counted in
an inconsistent 
manner---for example, re-expansions are not counted in {\SSS}, but in 
\mbox{\NS} they are \cite{Mars87,Rein89}. The algorithms used in
simulations are 
often significantly different from those used in applications, making
reported node counts a bad predictor of execution time in practice.
The question 
of finding a set of representative test positions is a problem for
simulations as well. A large number of different artificial
trees is not 
necessarily a realistic test set.

The impact of the {\AB} enhancements is significant:
many state-of-the-art game-playing programs are reported to
approach their theoretical lower bound, the minimal tree
\cite{Ebel87,Feld90b,Plaa94c,Schaeffer86}. 
Regrettably, this high level of performance does not imply that we
have a clear 
understanding of the detailed structure of real-life game trees. 

%We are able to make a few high level observations regarding the general
%structure. For example, dynamic move reordering causes different
%algorithms to search 
%different trees. Searching more ``inefficiently'' on a shallow
%depth may be beneficial for the overall search.
%Furthermore, the use of iterative deepening causes the ordering of the
%tree to be higher in older parts, that is, towards the root, and in the
%branches that are more to the left of the tree.
%
Many points influence the search space in
different ways, although it is not exactly known what the effect is.
For example, transpositions, iterative deepening
and the history heuristic all cause the tree to be dynamically
re-ordered based on information that is gathered during the
search. 
The effectiveness of iterative deepening depends on many factors, such 
as on the strength of the
value interdependence, on the number of cutoffs in the previous iteration, and
on the
quality of the evaluation function.
The effectiveness of transposition tables
depends on game-specific parameters, the size of the transposition
table, the search depth, and possibly on move ordering and the phase
of the game. 
The effectiveness of the history heuristic also depends on game-specific
parameters, and on the quality of the evaluation function. 

The consequence is that  trees that are generated in practice are
highly complex and 
dynamic entities, whose structure is influenced by the techniques that
make use of (some of) the 
four listed features. 
Acquiring data on these factors and the way
they relate seems a 
formidable task.
It poses many problems for researchers attempting to model
the behavior of algorithms on realistic minimax trees reliably. 

All of the simulations that we know of include at most one
of the above four features 
\cite{Bhat90,Bhatta93,Camp83,DeBruin94,Hsu90,Kain91,Mars87,Musz85,Rein89,Rein94b,Stoc79}.
In the light of the highly-complex nature of real-life game trees,
simulations can only be regarded as approximations,
whose results may not be accurate for real-life applications.
We feel that simulations provide a feeble basis for conclusions on
the relative merit of search algorithms as used in practice.
The gap between the trees searched in practice and in simulations is large.
Simulating search on artificial trees that have little relationship
with real trees runs the danger of producing misleading or
incorrect conclusions.
It would take a considerable amount of work to build a program
that can properly simulate real game trees.
Since there are already a large number of quality
game-playing programs available,
we feel that the case for simulations of minimax
search algorithms is weak.\\

An often used approach to have simulations approximate the efficiency of 
real applications is to 
increase the quality of move ordering.
In the light of what has been said previously, 
just increasing the probability of first moves
causing a cutoff
to, say, 98\%, can only be viewed as
a naive solution, that is not sufficient to yield
realistic simulations. First of all, the move ordering is not uniform
throughout the tree (see figure~\ref{fig:mvord}). 
Second, and more important, the high level of move 
ordering is not a cause 
but an effect. It is caused by techniques (like the history heuristic)
that make use of  phenomena like a variable 
branching factor, value 
interdependence, value independence and transpositions. These causes and
effects are all interconnected, yielding a
picture of great complexity that does not look very inviting to
disentangle. 

As an example of what the differences between real and artificial
trees can lead to,
let us look at some statements in the literature concerning {\SSS}. In
section~\ref{sec:sssview} we mentioned five points 
describing the general 
view on {\SSS}: it  (1) is difficult to
understand, (2) has unreasonable memory requirements, (3) is slow, (4)
provably dominates {\AB} in expanded leaves, and (5)
expands significantly fewer leaf nodes than {\AB}.
The validity of these points has been examined by numerous researchers
in the past
\cite{Camp83,Kain91,Mars87,Musz85,Rein89,Roiz83,Stoc79}.
All come to roughly the same conclusion, that the answer to all five
points is ``true:'' {\SSS} searches less leaves than {\AB}, but it
is {\em not\/} a practical algorithm.
However, two publications contend that points
2 and 3 may be false, indicating that {\SSS} not
only builds 
smaller trees, but that the problem of the slow operations on the 
OPEN list
may be solved \cite{Bhat90,Rein94b}. This paints a favorable
picture for {\SSS}, since the negative points would be solved, while
the positive ones would still stand.  
Probably due to the complexity of the
{\SSS} algorithm the authors have restricted themselves 
to simulations. With our reformulation we were finally able to use real
programs to answer the five questions.
In practice {\em all\/}
five points are 
wrong, making it clear that, although {\SSS} is practical, in 
realistic programs it has {\em no\/} substantial advantage over
{\AB}, and is even worse than {\AB}-variants like {\AspNS}. 

This example may serve to illustrate our point that it is hard to
model real trees reliably.  
%We feel that simulations provide a feeble basis for conclusions on
%the relative merit of search algorithms as used in practice.
%The gap between  trees searched in practice and in simulations is large.
%Simulating search on artificial trees that have little relationship
%with real trees runs the danger of producing misleading or
%incorrect conclusions.
%It would take a considerable amount of work to build a program
%that can properly simulate real game trees.
%Since there are already a large number of quality
%game-playing programs available,
%we feel that the case for simulations of minimax
%search algorithms is weak.
%
In the past we have performed simulations too \cite{DeBruin94}.
We were quite shocked  when we
found out how easy it is to draw wrong conclusions based on what
appeared to be valid assumptions. We hope to have shown in this section
that 
the temptation of oversimplifying the structure of game trees can and
should be resisted. Whether this problem only occurs in minimax
search, or also in other domains of artificial intelligence,
is a question that we leave unanswered.

\subsubsection{Background}
Seeing that the null-window {\AB}  ideas stood up in practice, working
through the intricacies of solution trees and transposition tables, and 
finding  that solution trees do fit in memory,
has been joint work with Jonathan Schaeffer.  

It all started on a day in early October 1994 in Edmonton, just before
the first snow would fall, when we were having a quick lunch in a Korean
restaurant on campus.
After talking a bit about {\SSS}, solution trees, and transposition
tables, we tried to estimate how much memory a max solution tree would
take up in chess.  To our surprise, the number seemed entirely reasonable.
Since this contradicted all the papers that we knew of, we did some
experiments, in the hope that they would show us where our quick
calculation had gone wrong.  As has been shown in
section~\ref{sec:space}, they didn't; the estimates were correct.
The theoretical work on {\SSS} from chapter~\ref{chap:mt} turned out
to have a direct relevance to practice.\index{theory}\index{practice} 

Jonathan Schaeffer's\index{Schaeffer} insight and  experience were a
big asset in  
designing the experiments and explaining the results. Finding out that
{\SSS} turned out to be easily implementable in real programs, and
then finding that it did not bring significant gains over {\NS},
caused much excitement as well as long discussions.   Some of 
the ideas born in those discussions bore fruit, as the next chapter
will show.

 % stor/perf
\cleardoublepage
%\thispagestyle{empty}
%\chapter{Why? --- Analysis to Explain the Test Results}\label{chap:anal}
%\chapter{Analysis}\label{chap:anal}
\chapter{The Minimal Tree?}\label{chap:anal}\label{chap:bestall}
The experiments showed that all tested algorithms perform quite close
together. This raises the question whether 
the enhancements are causing the programs
to approach their theoretical limit on the performance, the minimal
tree. In this chapter we re-examine the concept of the
minimal tree in the light of the experiments.
Results from this chapter have been published in \cite{Plaa96b}.\\

% The experiments were performed to compare the storage requirements 
% and performance of the {\MT} instances to {\AB} and {\NS} in a practical
% setting. There are quite a number of finer points that surfaced in the
% results. In section~\ref{sec:deeper} we will attempt to analyze and
% explain these points.

% \section{Best Case of Minimax Algorithms in Practice}\label{sec:bestall}
% After having seen that the algorithms perform quite close together, 
% we will look at the limit on the performance of all
% algorithms, the minimal tree.

%\subsection{Introduction}
%
% K&M meet the real world
%
The search tree built by {\AB} is exponential in the depth of the tree.
With a constant branching factor $w$ and search depth $d$,
{\AB} builds a search tree ranging in size from $O(w^{\lceil
d/2\rceil})$ to $O(w^d$). 
Given the large gap between the best and worst case scenarios,
the research effort has concentrated on methods to ensure that
the search trees built in practice come as close to the best case as possible.
{\AB} enhancements such as minimal window searching,
move ordering and transposition tables have been
successful at achieving this.
Numerous authors have reported programs that build search trees
within 50\% of the optimal size
\cite{Ebel87,Feldmann93,Schaeffer86}.
This is quite a remarkable result,
given that a small error in the search can lead to large search
inefficiencies.\index{minimal tree}\index{Alpha-Beta!enhancements}

This section examines the notion of the minimal {\AB} search tree.
The notion {\em minimal tree\/} arises from Knuth and Moore's \index{Knuth
and Moore}
pioneering work on search trees with a constant branching factor,
constant search depth and no transpositions \cite{Knut75}.
In practice, real game trees have variable branching factor and
are usually searched to variable depth.
Since two search paths can transpose into each other, nodes
in the tree can have more than one parent, implying that
the search tree is more precisely referred to as a search {\em graph}.
For a real game, what is the minimal search graph?

We introduce the notion of the {\em left-most} minimal graph,
the minimal graph that a left-to-right {\AB} traversal of the tree
would generate.
The {\em real} minimal graph is too difficult to calculate,
but upper bounds on its size show it to be significantly smaller than
the left-most minimal graph.
The insights gained from these constructions lead to ideas for 
several {\AB} enhancements.
One of them, {\em enhanced transposition cutoffs},
results in significant search reductions that translate into
tangible program execution time savings.

\section{Factors Influencing Search Efficiency}\label{sec:ideas}
Several authors have attempted to approximate the minimal graph for
real applications 
(for example, \cite{Ebel87}).
In fact,
what they have been measuring is a minimal graph as generated by
a left-to-right, depth-first search algorithm.
Conventional {\AB} search considers nodes in a left-to-right manner,
and stops work on a sub-tree once a cutoff occurs.
However, there may be another move at that node capable of causing the same
cutoff, possibly achieving that result by building a smaller search tree.
A cutoff caused by move {\em A} may build a larger search
tree than a cutoff caused by move {\em B} because of three properties
of search trees:
\begin{enumerate}
\item  {\em Move ordering}\\ 
\index{move ordering, quality of (graph)}%
Move {\em B}'s search tree may be smaller because of better move ordering.
Finding moves that cause a cutoff early will significantly reduce the
tree size.
\item {\em Smaller branching factor}\\
\index{branching factor}%
Move {\em B} may lead to a search tree with a smaller average branching factor.
For example, in chess, a cutoff might be achieved with a forced series
of checking moves.
Since there are usually few moves out of check,
the average branching factor will be smaller.
\item {\em Transpositions}\\
Some moves may do a better job of maximizing the number
of transpositions encountered.
Searching move {\em B}, for example, may cause transpositions into
previously encountered sub-trees, thereby reusing available results.
\end{enumerate}
Note that while the last two points are properties of search trees built
in practice,
most search-tree models and simulations do not take them
into consideration
(for example,
\cite{Ibar86,Kain91,Mars87,Musz85,Pear84,Pijl91,Rein89,Rein94b,Roiz83}).

\begin{figure}
\begin{center}
\includegraphics[width=10cm]{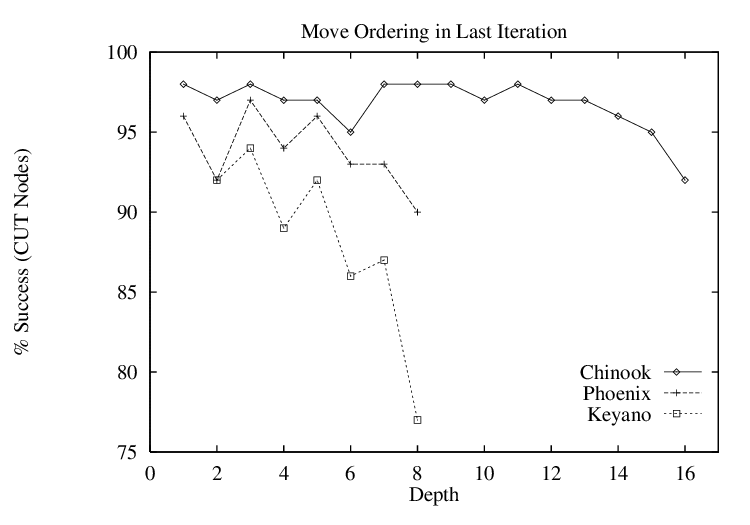}
\end{center}
\caption{Level of Move Ordering by Depth}\label{fig:mvord}
\end{figure}
\subsubsection{Move Ordering}
Considerable research effort has been devoted to
improving the move ordering,
so that cutoffs will be found as soon as possible
(for example, the history heuristic, killer heuristic,
iterative deepening and transposition tables \cite{Scha89b}).
Ideally, only one move should be considered at nodes where a cutoff is
expected. 

To see how effective this research has been,
we conducted measurements using the three programs, {\Chinook},
{\Keyano} and {\Phoenix}, covering 
the range of high (36 in chess) to low (3 in checkers)
branching factors (Othello has 10).
% The three programs use the {\NS} variant of {\AB} \cite{Rein83}
% enhanced with aspiration windows,
% iterative deepening and transposition tables ($2^{21}$ entries).
% {\Chinook} and {\Phoenix} also use the history heuristic.
% The programs were modified to search to a fixed depth,
% ensuring that changes to the search parameters would not
% alter the minimax value of the tree.
% All three programs have been finely tuned by their original authors and,
% presumably, achieve high performance. 
Data points were averaged
over 20 test positions.  
To be able to build reasonably sized trees, all tests used iterative
deepening. The  tests  in chapter~\ref{chap:exper} were concerned with
the performance of 
algorithms, and reported cumulative node counts over all
iterations. Here we are interested in comparing sizes of 
trees. Therefore we only report the tree size of the last
iteration. Including nodes of previous iterations could create a
disturbance.

The different branching factors of the three games affect
the depth of the search trees built within a reasonable amount of time.
%For checkers, our experiments were to 15--17 ply deep
%(one ply equals one move by one player),
%for Othello and chess, 9 ply.
For a $d$-ply search, the deepest nodes with move ordering information
are at depth $d-1$. Leaves have no move ordering information.

Figure~\ref{fig:mvord} shows how often, during the last iteration of
an {\AspNS} search, the first move considered
caused a cutoff at nodes where a cutoff occurred
(note the vertical scale).
For nodes that have been searched deeply, 
we see a success rate of over 90--95\%,
in line with results reported by others \cite{Feldmann93}.
Since the searches used iterative deepening,
all but the deepest nodes benefited from the presence of the
best move of the previous iteration in the transposition table.
Near the leaf nodes, 
the quality of move ordering decreases to roughly 90\% (75\% for {\Keyano}).
Here the programs do not benefit from the transposition table and
have to rely on their move-ordering heuristics
(dynamic history heuristic for Chinook; static knowledge for {\Keyano}).
Unfortunately, the majority of the nodes in the search tree
are at the deepest levels.
Thus, there is still some room for improvement.

Of the three programs, Chinook consistently has the best move ordering results.
The graph is misleading to some extent,
 since the high performance of Chinook is
partially attributable to the low branching factor.
The worst case is that a program has no knowledge about a position and
effectively guesses its choice of first move to consider.
With a lower branching factor (roughly 8 in non-capture positions),
Chinook has a much better chance of correctly guessing than does
Phoenix (branching factor of 36).

A phenomenon visible in the figure is an odd/even oscillation.
\index{odd/even oscillation}%
At even levels in the tree,
the move ordering appears to be less effective than at odd levels.
This is caused by the asymmetric nature of the search tree,
where nodes along a line alternate between those with cutoffs (one
child examined) and those where all children must be examined.
This is clearly illustrated by Knuth and Moore's formula for
the minimal search tree,
$w^{\lfloor d/2\rfloor} + w^{\lceil d/2 \rceil} - 1$ leaf nodes,
whose growth ratio depends on whether $d$ is even or odd. 
% In
% appendix~\ref{app:oddeven} this is analyzed in detail.

The evidence suggests that the research on move-ordering techniques
for {\AB} search has been very successful. 

\subsubsection{Variable Branching Factor}
Analyses of {\AB} often use the simplifying assumptions of a
fixed branching factor and depth to the search tree.
In practice, minimax trees have a less regular structure
with a variable branching factor and depth.
Algorithms like Conspiracy Number search \cite{McAl88,Scha90}
and Proof Number search \cite{Alli94b}
\index{conspiracy numbers}\index{proof numbers}\index{branching factor, irregular}% 
exploit this irregularity by using a
``least-work-first'' strategy. \index{least work first}%
For a number of application domains with a
highly irregular tree structure,
such as chess mating problems or the game of qubic, \index{qubic}%
these algorithms search more efficiently than {\AB}-like algorithms
\cite{Allis94}.

\begin{figure}
\begin{center}
\includegraphics[width=10cm]{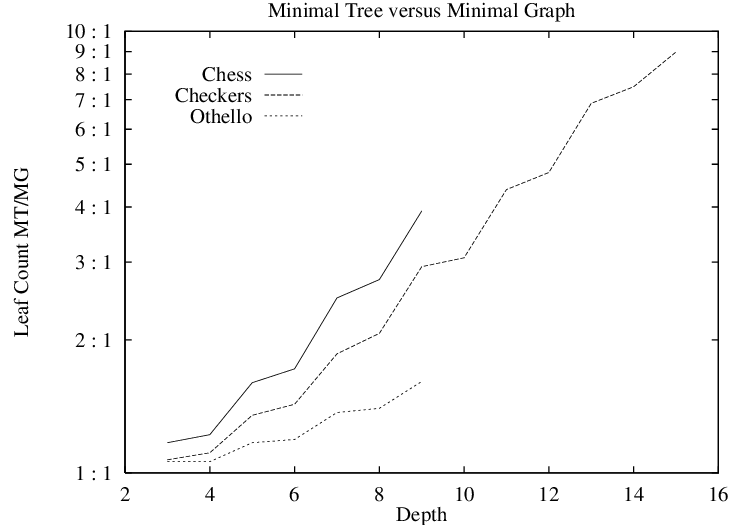}
\end{center}
\caption{Comparing the Minimal Tree and Minimal Graph}\label{fig:mtmg}
\end{figure}

\subsubsection{Transpositions}\label{sec:txpos}\index{transpositions, effect of (graph)}
In many application domains the search space is a graph:
nodes can have multiple parents.
To search this graph efficiently with a tree-search algorithm like {\AB},
nodes are stored in a transposition table.
% Before a node is searched,
% the table is queried to see if it has been examined previously.
% If the information is in the table and of sufficient accuracy,
% then no additional search need be performed.
For most games, transposition tables can be used to make
significant reductions in the search tree.

For our test programs,
we examined the size of the minimal tree with and without transpositions.
The result is shown in figure~\ref{fig:mtmg}
(the method used to compute this graph will be
explained in section~\ref{sec:numbers}).
Note the logarithmic vertical scale.
Identifying transpositions can reduce the size 
of the minimal tree for a chess program searching to depth
9 by a factor of 4.
In checkers searching to depth 15 yields a difference of a factor of 9.
Othello has less transpositions,
although there still is a clear advantage to identifying
transpositions in the search space.
In chess and checkers, a move affects few squares on the board,
meaning that a move sequence A, B, C often yields
the same position as the sequence C, B, A.
This is usually not true in Othello, where moves can affect many squares,
reducing the likelihood of transpositions occurring.

It is interesting to note that this figure also shows an odd/even effect,
for the same reasons as discussed previously.

To understand the nature of transpositions better we have gathered
some statistics for chess and checkers.
It turns out that roughly 99\% of the transpositions occur between
the nodes at the same depth in the tree.
Relatively few transposition nodes have parents of differing search depths.
(Although 1\% of 1,000,000 is not negligible, especially since the
number of transpositions does not indicate how big the sub-trees were
whose search was prevented.)
Another interesting observation is that the number of transpositions
is roughly linear to the number leaf nodes.
In checkers and chess,
identifying transpositions reduces the effective width of nodes
in the search tree by about 10 to 20\%,
depending primarily on characteristics of the test position.  
In endgame positions, characterized by having only a few pieces on the board,
the savings can be much more dramatic.

\subsubsection{Conclusion}
Having seen the impact of three factors on the efficiency of
minimax search algorithms,
we conclude that the often-used uniform game tree is not suitable
for predicting the performance of
minimax algorithms in real applications \cite{Plaa94a,Plaa95a}.
The minimal tree for fixed $w$ and $d$ must be an inaccurate upper bound
on the performance of minimax search algorithms.
In the next section we will discuss other ways to perform a best-case
analysis. 

\section{The Left-First Minimal Graph} 
\label{sec:numbers} 
\index{LFMG|see{minimal graph, left-first}}
\index{minimal graph!left-first}
\index{LFMT|see{minimal tree, left-first}}
\index{minimal tree, left-first}
Many simulations of minimax search algorithms have been performed
using a comparison with the size of the minimal tree as the performance metric
(for example, \cite{Kain91,Mars87}).
They conclude that some {\AB} variant is performing almost
perfectly, since the size of trees built is close to the size of the
minimal search tree.
Unfortunately, as pointed out previously,
simulated trees have little relation to those built in practice.

For most games, the search ``tree'' is a directed graph.
The presence of transpositions, nodes with more than one parent,
makes it difficult to calculate the size of
the minimal graph accurately.
However, by using the following procedure,
it is possible to compute the size of the graph traversed by a
left-to-right, depth-first search algorithm like {\AB} \cite{Ebel87}.
In the following, the transposition table is used to
store intermediate search results.
Trees are searched to a fixed depth.
\index{minimal graph!computing the left-first}

\begin{enumerate}
\item {\em Alpha-Beta:\/}
Compute the minimax value $f$ of the search using any {\AB}-based algorithm,
such as {\NS}.
At each node the best move
(the one causing a cutoff or, failing that,
the one leading to the highest minimax value)
is saved in the transposition table.

\item {\em Minimal Tree:\/}
Clear the transposition table so that only the positions and
their best moves remain
(other information, like search depth or value is removed).
Repeat the previous search
using the transposition table to provide only the best move
(first move to search) at each node
(no transpositions are allowed).
{\AB} will now traverse the minimal tree,
using the transposition table as an oracle to select
the correct move at cutoff nodes always.
Since our transposition table was implemented as a hash table,
a possibility of error comes from table collisions
(no rehashing is done).
In the event of a collision, searching with a window of
$\alpha = f - 1$ and $\beta = f + 1$ will reduce the impact of these errors.
Alternatively, a more elaborate collision-resolution scheme can be
used to eliminate this possibility.

\item {\em Minimal Graph:\/}
Clear the transposition table again (except for best moves).
Do another search, using the best-move information in the transposition table.
Allow transpositions, so that if a line of play transposes
into a previously seen position,
the search can re-use the previous result
(assuming it is accurate enough).
Again, a minimal search window ($\alpha = f - 1$, $\beta = f + 1$) is used.
The minimal tree is searched with transpositions,
resulting in a minimal graph.
\end{enumerate}
Of course, for this procedure to generate meaningful numbers,
the transposition table must be large enough to hold
at least the minimal graph.
Our table size was chosen to be consistent with
the results in section~\ref{sec:results}.

\begin{figure}
\begin{center}
\includegraphics[width=10cm]{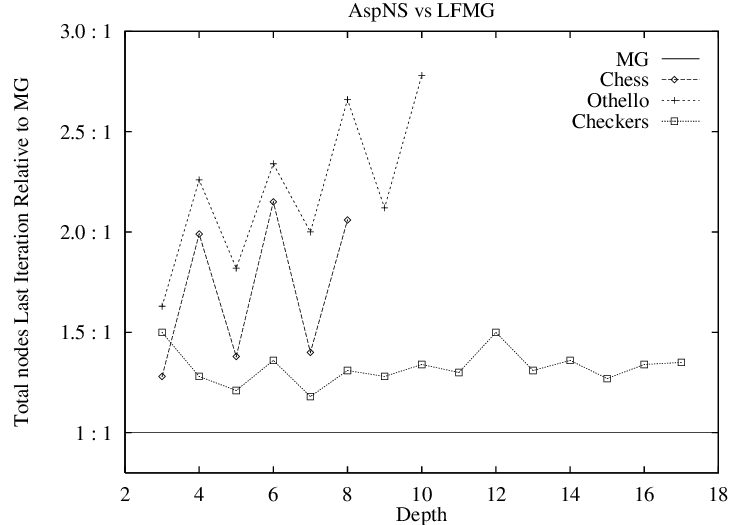}
\end{center}
\caption{Efficiency of Programs Relative to the Minimal Graph}\label{fig:nsmg}
\end{figure}

The minimal graph has been used by many authors as a yardstick to compare
the performance of their search algorithms in practice.
For example, in chess,
{\em Belle\/} is reported to be within a factor of 2.2
of the minimal {\AB} tree \cite{Ebel87}, \index{Hitech}
\index{Zugzwang} \index{Phoenix}
{\em Phoenix} within 1.4 \cite{Schaeffer86},
{\em Hitech} within 1.5 \cite{Ebel87} and
{\em Zugzwang} within 1.2 \cite{Feldmann93}.
Using the three-step procedure, we have measured the performance of 
{\Chinook}, {\Keyano} and {\Phoenix}.
The results of the comparison of {\NS} 
against this minimal graph are shown in figure~\ref{fig:nsmg}
(based on all nodes searched in the last iteration).
The figure confirms that the best programs are searching
close to the minimal graph (within a small factor).

An interesting feature is that all three games,
Othello and chess in particular,
have significantly worse performance for even depths.
The reason for this can be found in the structure of the
minimal tree.
In going from an odd to an even ply,
most of the new nodes are nodes where a cutoff is expected to occur.
For the minimal graph,
their children count as just one node access.
However, the search algorithm may have to consider a number of
alternatives before it finds one that causes the cutoff.
Therefore, at even plies, move ordering is critical to performance.
On the other hand, in going from an even to an odd ply,
most of these leaves are children of nodes where no cutoff is expected.
All of the leaves are part of the minimal graph.
Hence, at these nodes move ordering has no effect since
all children have to be searched anyway.

The preceding leads to an important point:
reporting the efficiency of a fixed-depth search algorithm based
on odd-ply data is misleading. \index{odd-ply data is misleading}
The odd-ply iterations give an inflated view of the search efficiency.
For odd-ply searches,
all three programs are searching with an efficiency similar 
to the results reported for other programs.
However,
the even-ply data is more representative of real program performance and,
on this measure, it appears that there is still room for improvement.
In light of this,
the {\em Hitech} results of 1.5 for 8-ply searches
seem even more impressive \cite{Ebel87}.

\subsubsection{The Left-First Minimal Graph and The Real Minimal
  Graph}
\index{TRMG|see{minimal graph, the real}}
\index{minimal graph!the real}
\index{real minimal graph}
The previous section discussed a minimal graph for the comparison of
algorithms that search trees in a left-to-right manner,
such as {\AB}, {\NS}, {\SSS}, {\DUAL} and {\MTDf}. Although the last
three are usually called ``best-first'', they expand new children at
each node in a left-to-right order. On a perfectly ordered tree, all
algorithms expand the same tree.

This minimal graph is not necessarily the smallest possible.
Consider the following scenario.
At an interior node $N$, there are two moves to consider, $A$ and $B$.
Searching $A$ causes a cutoff, meaning move $B$ is not considered.
Using iterative deepening and transposition tables,
every time $N$ is visited only move $A$ is searched,
as long as it continues to cause a cutoff.
However, move $B$, if it had been searched,
was also sufficient to cause a cutoff.
Furthermore, what if $B$ can produce a cutoff by building a smaller
search tree than for move $A$?
For example, in chess, $B$ might start a sequence of checking moves
that leads to the win of a queen.                                           
The smaller branching factor (because of the check positions) and the
magnitude of the search score will help reduce the tree size.        
In contrast, $A$ might lead to a series of non-checking moves that 
culminates in a small positional advantage.
The larger branching factor (no checking moves) and smaller score   
can lead to a larger search tree.                                
Most minimax search algorithms stop when they find a cutoff move,
even though there might be an alternative cutoff move that can
achieve the same result with less search effort.              

In real applications, where $w$ and $d$ are not uniform, 
the minimal graph defined in the previous section is not really minimal,
because at cutoff nodes no attempt has been made to achieve the cutoff
with the smallest search effort.
The ``minimal graph'' in the literature \cite{Ebel87,Feldmann93,Schaeffer86}
is really a
{\em left-first} minimal graph (LFMG), since only the left-most
move causing a cutoff is investigated.
The {\em real} minimal graph (RMG) must select the cutoff move
leading to the smallest search tree.

The preceding suggests a simple way of building the RMG,
by enhancing part~1 of the minimal graph construction algorithm:
\begin{enumerate}
\item Search all moves at a cutoff node,
counting the number of nodes in the sub-trees generated.
The move leading to a cutoff with the smallest number of nodes
in its search tree is designated ``best''.
\end{enumerate} 
In other words, explore the entire minimax tree, looking for the 
smallest minimal tree.

Obviously, this adds considerably to the cost of computing the
minimal graph. 
An optimization is to stop the search of a 
cutoff candidate as soon as its sub-tree size exceeds the size of the
current cheapest cutoff.

Unfortunately, finding the size of the RMG is not that simple. 
This solution would only work if there were no transpositions. 
In the presence of transpositions, the size of a search can
be largely influenced by the frequency of transpositions. 
Consider interior node $N$ again.
Child $A$ builds a tree of 100 nodes to generate the cutoff,
while child $B$ requires 200 nodes.
Clearly, $A$ should be part of the minimal graph.
Interior node $M$ has two children, $C$ and $D$, that cause cutoffs.
$C$ requires 100 nodes to find the cutoff, while $D$ needs 50.
Obviously, $D$ is the optimal choice.
However, these trees may not be independent.
The size of $D$'s tree may have been influenced by transpositions
into work done to search $B$.
If $B$ is not part of the minimal graph, then $D$ cannot benefit
from the transpositions.
In other words, minimizing the tree also implies maximizing the benefits
of transpositions. Since there is no known method to predict the
occurrence of transpositions, finding  the minimal graph involves 
enumerating all possible sub-graphs that prove the minimax value.

\index{minimal graph!computing the real}
Computing the real minimal graph is a computationally infeasible
problem for non-trivial search depths.
The number of possible minimal trees is exponential in the size of
the search tree. Transpositions increase the complexity of finding the 
RMG by making the size of sub-trees interdependent.
Choosing a smaller sub-tree
at one point may increase the size of the total solution. We have not found
a solution for finding the {\em optimal\/} RMG. The following
describes a method to approximate its size.

\section{Approximating the Real Minimal Graph}
% The extrapolation experiment hints at savings that would be, in the case
% for checkers, surprisingly large. Therefore w
%We have searched for methods
%to get a better approximation of the possible savings without
%disregarding transpositions. 
We have found two methods to approximate the RMG that find an
upper bound on its size that is smaller than the LFMG.
The first approach involves trying to maximize the number
of transpositions in the tree. 
%This method works both off-line and
%on-line, due to its low overhead.
The second approach is to exploit the variable branching factor of
some games, to select cutoff moves that lead to smaller search trees.
%This method works only off-line; the overhead is too big for use in
%on-line, real-time, algorithms.
We will call the graph generated using these ideas an approximate RMG (ARMG).
\index{minimal graph!approximating the real}
\index{ARMG|see{minimal graph, approximating the real}}

% 
% Miracle: returning at a TT cutoff tsize instead of 1 in Refine2 gives
% better results, in chinook.
% In other words: refine as a tree, count as a graph. This is a pretty
% pessimistic view of life.
% 

\begin{figure}
{\small
\begin{tabbing}
mmmmmmm\=mm\=mm\=mm\=mm\=mm\kill
\> {\bf function} alphabeta-ETC$(n, \alpha, \beta) \rightarrow g$;\\
\> \> {\bf if} retrieve$(n)$ = ok {\bf then}\\
\> \> \> {\bf if} $n.f^- \geq \beta$ {\bf then return} $n.f^-$;\\
\> \> \> {\bf if} $n.f^+ \leq \alpha$ {\bf then return} $n.f^+$;\\
\> \> {\bf if} $n$ = leaf {\bf then} $g := $ eval$(n)$;\\
\> \> {\bf else}\\
\>** \> \> $c := $ firstchild$(n)$;\\
\>** \> \> {\bf while}  $c \neq \bot$ {\bf do}\\
\>** \> \> \> {\bf if} retrieve$(c)$ = ok {\bf then}\\
\>** \> \> \> \> {\bf if} $n = $ max {\bf and} $c.f^- \geq \beta$ {\bf then return} $c.f^-$;\\
\>** \> \> \> \> {\bf if} $n = $ min {\bf and} $c.f^+ \leq \alpha$ {\bf then return} $c.f^+$;\\
\>** \> \> \> $c := $ nextbrother$(c)$;\\
\> \> \> {\bf if} $n$ = max  {\bf then}\\
\> \> \> \> $g := -\infty$;\\
\> \> \> \> $c := $ firstchild$(n)$;\\
\> \> \> \> {\bf while} $g < \beta$ {\bf and}  $c \neq \bot$ {\bf do}\\
\> \> \> \> \> $g := $ max$\big(g,$ alphabeta-ETC$(c, \alpha, \beta)\big)$;\\
\> \> \> \> \> $\alpha := \max(\alpha, g)$;\\
\> \> \> \> \> $c := $ nextbrother$(c)$;\\
\> \> \> {\bf else} /* $n$ is a min node */ \\ % {\bf then}\\
\> \> \> \> $g := +\infty$;\\
\> \> \> \> $c := $ firstchild$(n)$;\\
\> \> \> \> {\bf while} $g > \alpha$ {\bf and} $c \neq \bot$ {\bf do}\\
\> \> \> \> \> $g := $ min$\big(g,$ alphabeta-ETC$(c, \alpha, \beta)\big)$;\\
\> \> \> \> \> $\beta := \min(\beta, g)$;\\
\> \> \> \> \> $c := $ nextbrother$(c)$;\\
\> \> {\bf if} $g < \beta$ {\bf then} $n.f^+ := g$;\\
\> \> {\bf if} $g > \alpha$ {\bf then} $n.f^- := g$;\\
\> \> store $n.f^-, n.f^+$;\\ 
\> \> {\bf return} $g$;
\end{tabbing}
}
\caption{ETC pseudo code}\label{fig:etccode}
\end{figure}

\subsubsection{Maximizing Transpositions}\label{sec:practice}
\begin{figure}
\begin{center}
\includegraphics[width=4cm]{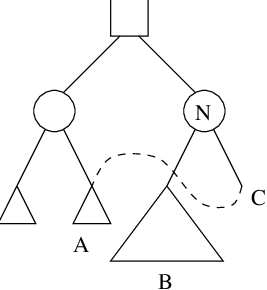} 
\end{center}
\caption{Enhanced Transposition Cutoff}\label{fig:etc}
\end{figure}

A simple and relatively cheap enhancement to improve search
efficiency is to try and 
make more effective use of the transposition table.
Consider interior node $N$ with children $B$ and $C$.
The transposition table suggests move $B$ and as long as it
produces a cutoff, move $C$ will never be explored.
However, node $C$ might transpose into a part of the tree, node $A$, that
has already been analyzed (figure~\ref{fig:etc}).
Before doing any search at an interior node,
a quick check of all the positions arising from this node
in the transposition table may result in finding a cutoff.
The technique to achieve  {\em Enhanced Transposition Cutoffs}, ETC,
performs transposition table lookups on successors of a node,
looking for transpositions into previously searched lines.
Figure~\ref{fig:etccode} shows the standard
{\AB}-with-transposition-table pseudo code, with the ETC code marked
by **.
In a left-to-right search, ETC encourages
\index{ETC}
\index{Enhanced Transposition Cutoff} 
sub-trees in the right part of the tree to transpose into the left.

\begin{figure}
\begin{center}
\includegraphics[width=10cm]{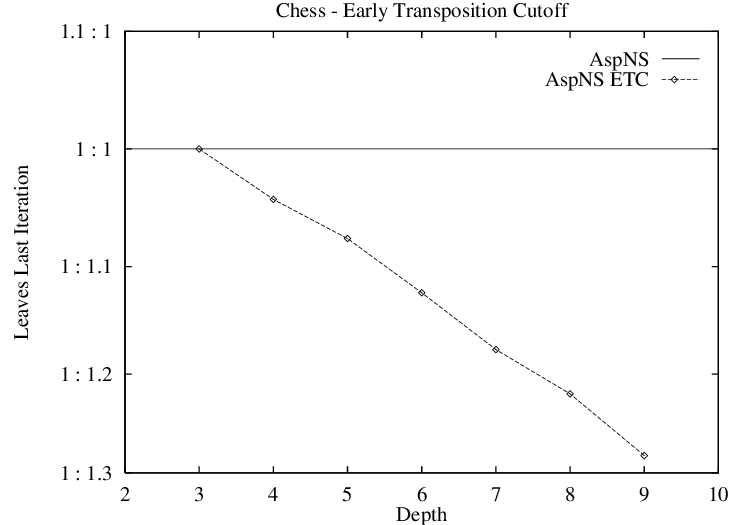}
\end{center}
\caption{Effectiveness of ETC in Chess}\label{fig:jsx}
\end{figure}

Figure~\ref{fig:jsx} shows the results of enhancing Phoenix with ETC.
For search depth 9, ETC lowered the number of expanded leaf nodes
by  a factor of 1.28 for {\NS} enhanced with aspiration searching.
Using {\MTDf},
the cumulative effect is  a factor of 1.33 fewer leaf nodes as compared to
Phoenix' original algorithm (not shown).

\begin{figure}
\begin{center}
\includegraphics[width=10cm]{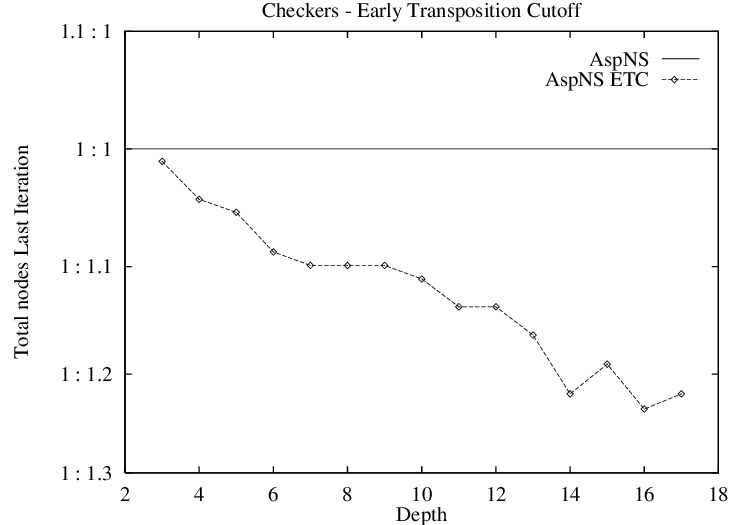}
\end{center}
\caption{Effectiveness of ETC in Checkers}\label{fig:jsxchin}
\end{figure}

Figure~\ref{fig:jsxchin} shows that the effect of ETC in {\Chinook} is of a
comparable magnitude.
The Othello results are not shown.
There are relatively few transpositions in the game and, hence, the
effect of ETC is small (roughly 4\% for depth 9).

The reduction in search tree size offered by ETC is, in part, offset
by the increased computation per node.
For chess and checkers, it appears that performing ETC at all interior nodes
is not optimal.
A compromise, performing ETC at all interior nodes that are more than 2 ply
away from the leaves, results in most of the ETC benefits with only a small
computational overhead, making it well suited for use in both on-line
and off-line algorithms.
Thus, ETC is a practical enhancement to most {\AB} search programs.

In addition, we have experimented with more elaborate lookahead
schemes involving shallow searches.
For example,
ETC can be enhanced to transpose also from left to right.
At an interior node,
all the children's positions are looked up in the transposition table.
If no cutoff occurs,
then check to see if one of the children leads to a position with
a cutoff score that has not been searched deep enough.
If so, then use the move leading to this score as the first
move to try in this position.
Unfortunately, several variations on this idea have failed to yield
a tangible improvement.

% Another way of reducing search tree size is to select moves at cutoff nodes
% that lead to smaller trees.
% For games with a relatively uniform branching factor
% (such as chess and Othello)
% it is difficult to exploit non-uniformity.
% In checkers, however, there is a non-uniform branching factor:
% 1.2 moves in a capture position and 7.8 in a non-capture position.
% The number of nodes in a search tree that begins with a sequence of
% two (forced) capture moves is smaller than for a tree that begins with
% two non-capture moves.
% This suggests that in checkers, smaller search trees might be obtained
% by having the program favour moves that lead to exchanges of pieces.
% So far, a variety of ideas to exploit this have led to mixed results.
% 
% We have tried many ideas to exploit transpositions and non-uniform
% branching factors.
% For example, shallow searches to improve the move ordering and adding a
% context of one or two moves to the history heuristic.
% All the ideas are interesting and show potential on individual positions.
% However, every one of our ideas (except ETC)
% fails to yield a consistent 
% improvement when averaged over a test set of 20 positions.
% 
% The ETC can be used to improve the performance of  game playing
% programs. In the next section we present a way to find a smaller
% minimal graph by using extra search effort. 

\subsubsection{Minimaxing to Exploit a Variable Branching Factor}
The ARMG can be further reduced by recognizing that all cutoffs are
not equal;
some moves may require less search effort.
Ideally, at all interior nodes the move leading to the cutoff that
builds the smallest search tree should be used.
Unfortunately, without an oracle, it is expensive to calculate
the right move.
In this section,
we present a method for finding some of the cheaper cutoffs,
allowing us to obtain a tighter upper bound on the ARMG.

Instead of performing a full minimax search to find the cheapest cutoff,
we perform a minimax search at the lowest plies
in the tree only.
The best moves at higher plies in the tree have already been optimized
by previous iterative deepening searches.
Whenever a cutoff occurs,
we record the size of the sub-tree that causes it.  
Then we continue searching at that node,
looking for cheaper cutoffs.
The cutoff move leading to the smallest sub-tree is added to the
transposition table.
A problem with this approach is that in discarding a sub-tree
because it was too big, we may also be throwing away
some useful transpositions.
Therefore,
an extra {\AB} pass must traverse the best moves again,
to count the real size of the approximated minimal graph. 
(For the best result every
transposition should be counted 
for the full size of the sub-tree that it represents. 
We assume as it were, that it will be removed and that the algorithm has
to search the sub-tree itself.)\index{minimal graph!minimaxing}

\begin{figure}
\begin{center}
\includegraphics[width=10cm]{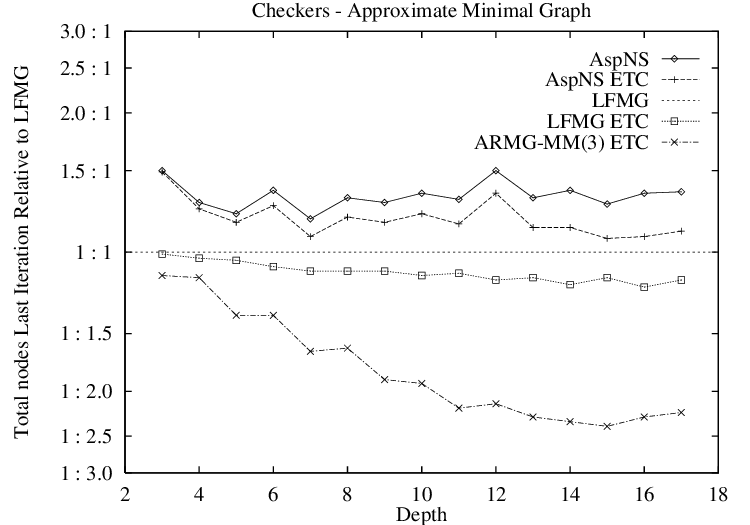}
\end{center}
\caption{LFMG Is Not Minimal in Checkers}\label{fig:armgchin}
\end{figure}

\begin{figure}
\begin{center}
\includegraphics[width=10cm]{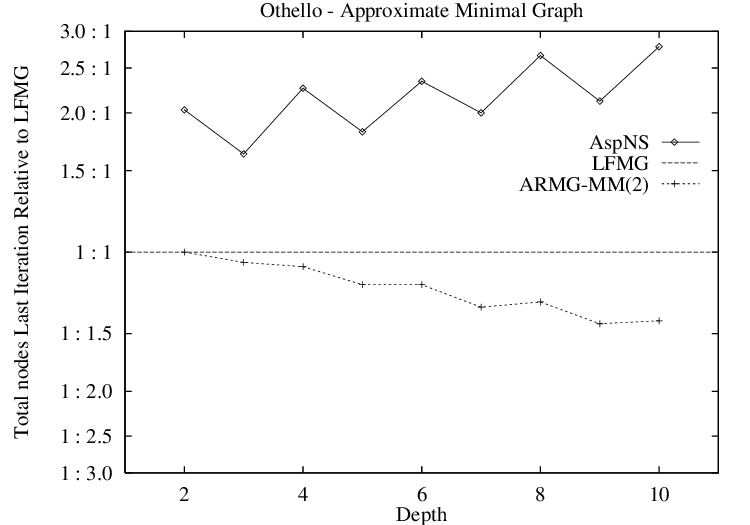}
\end{center}
\caption{LFMG Is Not Minimal in Othello}\label{fig:armgkey}
\end{figure}

\begin{figure}
\begin{center}
\includegraphics[width=10cm]{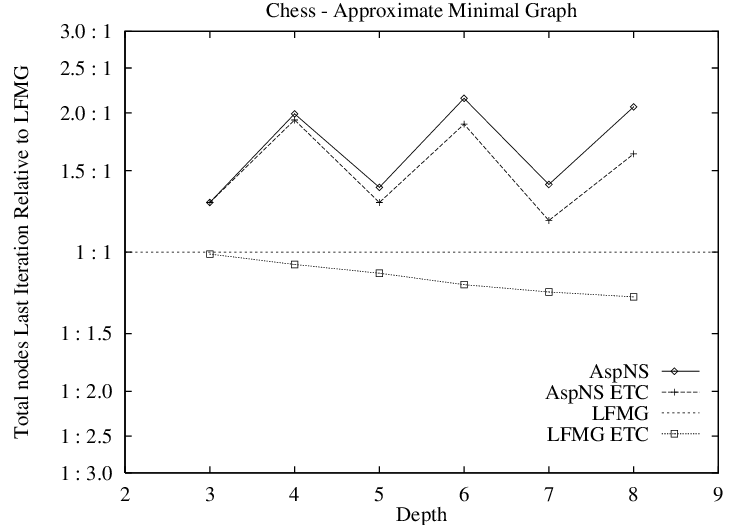}
\end{center}
\caption{LFMG Is Not Minimal in Chess}\label{fig:armgpho}
\end{figure}

The results for Chinook and Keyano are shown 
in figures~\ref{fig:armgchin} and \ref{fig:armgkey}.
ARMG-MM($d$) means that the last $d$ ply of the search tree
were minimaxed for the cheapest cutoff.
Chinook used MM(3), while Keyano used MM(2).
Othello has a larger branching factor than checkers,
resulting in MM(3) taking too long to compute.
The chess results are not reported since the branching factor
in the search tree is relatively uniform (except for replies to check),
meaning that this technique cannot improve the ARMG significantly
(as has been borne out by experiments). We do  show in
figure~\ref{fig:armgpho} the possible savings of ETC on the LFMG.
 

In checkers the forced capture rule creates trees with a diverse
branching factor.
The ARMG can take advantage of this.
The savings of ARMG-MM(3) are around a factor of 2. 
Minimaxing a bigger part of the graph (such as MM(4) or greater)
will undoubtedly create an even smaller ``minimal'' graph.

Othello's branching factor can vary,
but tends to be less volatile than for checkers,
accounting for the lower savings (a factor of 1.5--1.6).
In addition, since there are fewer transpositions possible in Othello,
there is less risk of throwing away valuable transpositions.

%Chess has a fairly regular branching factor, and the highest number of
%transpositions of the three games. These transpositions are probably
%the reason that the ARMG method does not work. Also, in chess the
%branching factor precludes deep searches, whereas the extrapolation
%results indicated that significant savings require deep searches.
%All this does not
%mean that the RMG for chess is about as big as the LFMG. The next
%section discusses the ETC method. ETC makes it possible to reduce the
%size of 
%the LFMG by 20\% (see figure~\ref{fig:armgpho}).

% We stress that this approximation is just a first cut. We do not know
% if this number is valid for bigger search trees, although our results
% indicated that the effect grows stronger with deeper searches. Also,
% well be that transpositions interact in ways that make RMG much
% smaller or larger than extrapolations would suggest. 
% However, it is an
% indication that there might still be quite some room for improvement
% of minimax search algorithms.
% We have spent a lot of effort to find better approximations of the
% RMG, as well as on using these ideas to enhance current search
% algorithms. This is the subject of ongoing research. 

Chess has a fairly uniform branching factor except for moving out of check.
Consequently, our test positions failed to show significant reductions
in the ARMG using our approach.
%The large branching factor precluded deep searches,
%whereas the extrapolation results indicated that significant savings
%required deep searches.
However, figure~\ref{fig:armgpho} shows that ETC still reduces the
LFMG substantially. More research is required to get a tighter bound for chess.

% We stress that this approximation is just a first cut. We do not know
% if this number is valid for bigger search trees, although our results
% indicated that the effect grows stronger with deeper searches. Also,
% well be that transpositions interact in ways that make RMG much
% smaller or larger than extrapolations would suggest. 
% However, it is an
% indication that there might still be quite some room for improvement
% of minimax search algorithms.

Thus minimaxing can find, in an off-line computation, a smaller
minimal graph. The overhead involved in minimaxing a few plies of the
tree makes this method unsuited for use in on-line, real-time, algorithms. 
We tried many ideas for exploiting transpositions and non-uni\-form
branching factors in real-time search.
All the ideas are interesting and show potential on individual positions.
However, every one of our ideas (except ETC) fails to yield a consistent 
improvement when averaged over a test set of 20 positions.

% This work should be interpreted as a first attempt at approximating
% the real minimal graph.
% Further refinements are the subject of ongoing research. 

\begin{figure}
\begin{center}
\includegraphics[width=10cm]{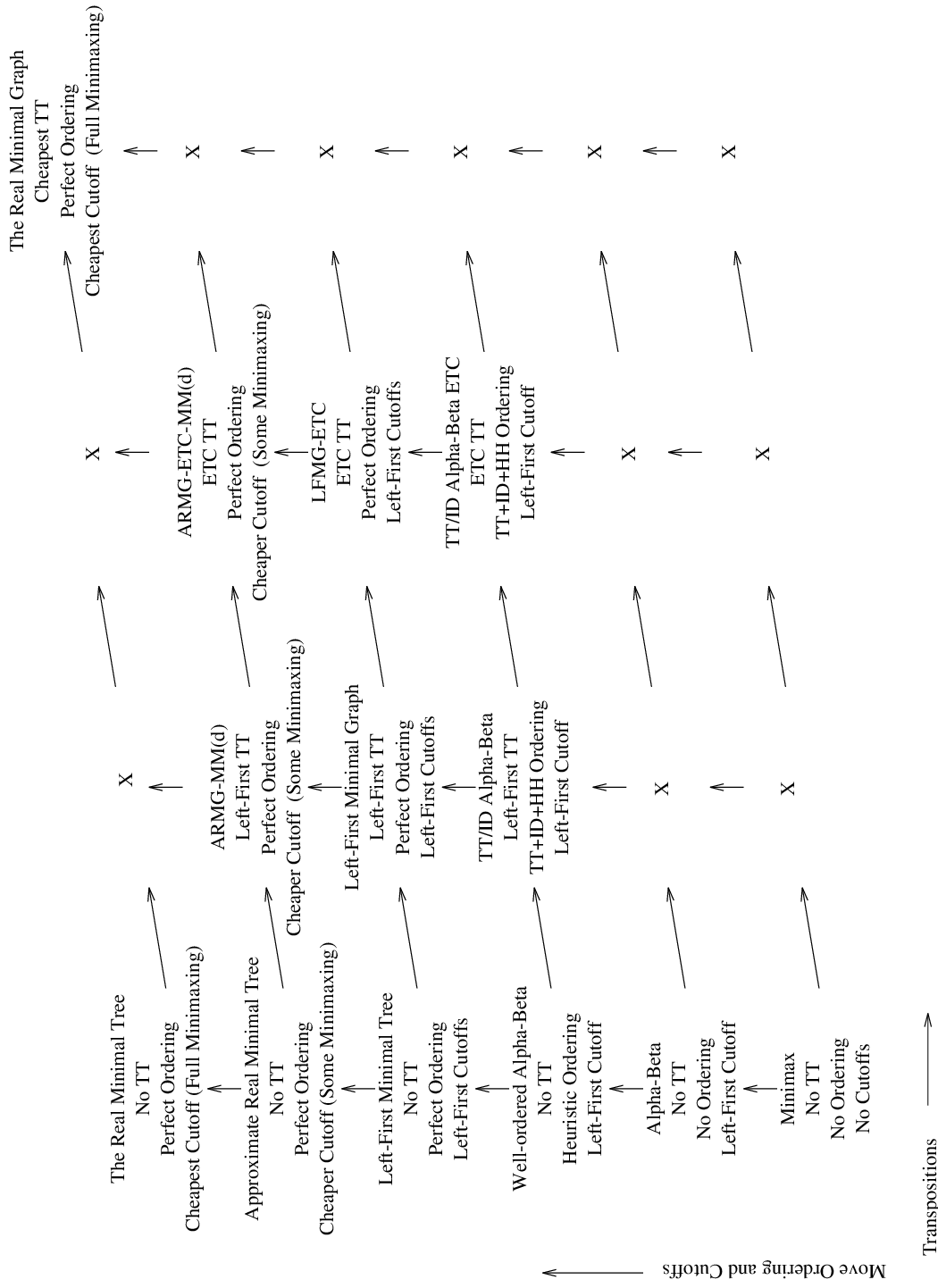} 
\end{center}
\caption{A Roadmap towards the Real Minimal Graph}\label{fig:roadmap}
\end{figure}

\subsubsection{Seeing the Forest through the Trees}\label{sec:road}
\index{minimal graph!roadmap}
Figure~\ref{fig:roadmap} gives a road map of the relations between
all the different kinds of approximations of the real minimal graph.
In the left bottom corner the worst case of minimax search can be found.
In the top right corner is the optimal case for real trees
(those with variable branching factor and transpositions).
From left to right in the diagram, the effectiveness of transpositions
is improved.
From bottom to top in the diagram, the quality of move ordering is improved.
The ``X''s represent data points in the continuum which are either
too difficult to calculate or are of no practical interest.
The top-right entry, the Real Minimal Graph, represents the theoretically
important, but unattainable, perfect search.
Abbreviations used include
TT (transposition table),
ID (iterative deepening),
HH (history heuristic, or some equally good ordering scheme),
ARMG (approximate real minimal graph),
LFMG (left-first minimal graph),
ETC (enhanced transposition cutoffs) and
MM(d) (minimax d-ply searches for finding cheapest cutoffs).

The figure illustrates how far game-tree searching has evolved since
the invention of minimax search.
The entry for {\AB} enhanced with TT, ID, HH and ETC data point is almost the new
state-of-the-art performance standard. To be complete, the figure
should contain extra entries, for null-window enhancements such as
{\NS} and {\MTDf}. The new state-of-the-art algorithm is the enhanced
version of {\MTDf}.
As this section shows, the gap between what can be achieved in practice
and the real minimal graph is larger than previously suspected.
Thus, there is still room for extending the road map to narrow the distance
between where we are and where we want to be.

\section{Summary and Conclusions}
\label{sec:concl}
The notion of the minimal search tree is a powerful tool to increase
our understanding of how minimax search algorithms work, and how they
can be improved.
One use of the minimal tree is as a yardstick for
the performance of minimax search algorithms and their enhancements.
However, trees as they are built by real applications,
such as game-playing programs, are neither uniform, nor trees.
Therefore, we have to be more precise in our definition of the
minimal search tree and graph.
This section defined two types of minimal graphs: 
\begin{enumerate}
\item The {\em Left-First Minimal Graph}
is constructed by a left-to-right, depth-first traversal of the search tree
(using, for example, {\AB}).
The use of a transposition table allows for the possibility of transpositions,
making the search tree into a search graph.
\item {\em The Real Minimal Graph} is the minimum effort required for a search.
However, this search requires an oracle so that at cutoff nodes the branch
leading to the cutoff requiring the least amount of search effort is selected.
Finding the size of the real minimal graph is difficult because the
utility of transpositions has to be maximized.
This involves enumerating all possible sub-graphs that prove the minimax value.
Lacking an ordering in the set of transpositions, 
this is a computationally intractable problem.
\end{enumerate}
We arrive at the following conclusions:
\begin{itemize}
\item For performance assessments of minimax search algorithms,
the minimal tree is an inadequate measure.
Some form of minimal graph that takes variable width and transpositions into
account should be used.
\item {\AB}-based search algorithms perform much closer to the
left-first minimal graph for odd depths than for even depths.
In performance comparisons, even ply results should be reported.
For odd search depths, the tree size is dominated by the last ply.
This ply consists largely of nodes where no cutoff occurs and, hence,
the move ordering is of no importance.
In effect, results for odd-ply search depths hide the
results at nodes where move ordering is important.
\item The real minimal graph is significantly smaller
than the left-first minimal graph,
the usual metric for search efficiency of minimax
algorithms. Therefore,
these algorithms are not as efficient as is generally
believed. 
\item Enhanced Transposition Cutoffs improve {\AB} searching
by trying to maximize the number of transpositions in the search.
The results indicate this to be a significant improvement.
\end{itemize}
Various publications indicate that game-playing programs are
almost perfect in their search efficiency.
Our results show that there remains room for improvement.

\subsubsection{Background}
Many of the analyses in the last two chapters were done in discussions with
\mbox{Jonathan} 
\mbox{Schaeffer}. The initial reason for the work on minimal graphs
was our disappointment with the relatively small gains that
{\MTDf} yielded---only a few percentage points. We assumed that the
three programs were all close to the minimal tree. As it turned out,
there was much to be learned. Our discussions and numerous
failed 
attempts at 
improving the algorithms helped creating much of the insight behind
the work in this chapter. Some ideas worked, notably the ETC, which
was suggested in 
an early stage by Jonathan Schaeffer\index{Schaeffer} in a discussion
on shallow searches, ``just to try as a first cut.'' As it turned out,
all the more elaborate ideas on 
shallow searches, as well as many improvements to the history
heuristic, did not yield a  consistent improvement. The simple first
cut, ETC, did.

 % optim
\cleardoublepage
\chapter{Concluding Remarks}\label{chap:concl}

This chapter concludes the research by briefly reviewing the main
issues. At the end, some directions for future research are
mentioned.

\section{Conclusions}
\subsubsection{Null-window {\AB} Search}
A widely-used enhancement to the {\AB} algorithm is the use of a
search window of reduced size. 
% The tighter bounds of this window
% cause {\AB} to cut off more nodes, but run the risk of not finding the
% minimax value if it happens to lie outside of the window. Then a
% re-search must establish its value.
Taking this idea to the limit is the reduction of the {\AB} search
window to a null window. A null-window {\AB} call never
finds the minimax value; it returns either
an upper or a lower bound on it. An algorithm consisting solely of
null-window {\AB} calls 
would have to perform many re-searches to home in on the minimax
value. This would reduce the efficiency of the null windows.

Chapter~\ref{chap:mt} discusses a solution to this problem. For all
the nodes that the null-window {\AB} call traverses, the return bounds
are stored in memory. Transposition tables (TT) provide an efficient way to do 
this. They allow for the
pruning power of null-window {\AB} calls to be retained over a sequence of
searches. In this way subsequent {\AB} calls  build on the work of previous
ones, using the information to find the node to expand
next. Interestingly, this results in a best-first expansion sequence.
The idea is formalized in the {\MT} framework, a framework for
null-window-only best-first minimax algorithms.  The memory needed to
store the tree that defines a bound is $O(w^{d/2})$, for a minimax
tree of uniform branching factor $w$ and uniform depth $d$.

The framework allows the formulation of a number of algorithms,
existing ones, such as {\SSS} \cite{Stoc79}, and new ones, such as
{\MTDf}. It 
focuses attention on the fundamental differences between algorithms. The
details of how to traverse trees are left to {\AB}. 

% An  advantage of using transposition tables for storage is that they are
% used in many applications such as game-playing programs. They are
% ``proven technology.'' Implementation of instances of the framework in
% most programs is straightforward.

\subsubsection{{\SSS}}
An instance of {\MT} called {\ABSSS} evaluates the same nodes (in
the same\index{SSS*}\index{MT-SSS*}\index{Alpha-Beta}\index{SSS*!reformulation} \index{SSS*!list of problems}
order) as {\SSS}. {\SSS} is an important algorithm because
of its claim to be ``better'' than 
{\AB}, since it provably never expands more nodes than {\AB}. Quite a
number of journal
publications try to decide whether {\SSS} is really ``better,'' 
using both analysis and simulations of the algorithm
\cite{Camp83,Ibar86,Kuma83,Mars87,Rein94b,Roiz83,Stoc79}. The general
view is mixed:
\begin{enumerate}
\item  {\SSS} is hard to understand.
\item  {\SSS} stores nodes in a sorted OPEN list. The operations on
this list make the algorithm slow.
\item  Being a best-first algorithm, {\SSS}'s exponential memory usage
makes it unsuited for practical use in game-playing programs.
\item  {\SSS} provably never expands more leaf nodes than {\AB}.
\item  {\SSS} expands on average significantly less nodes than {\AB}.
\end{enumerate}
The first three items are the disadvantages of {\SSS}, while the
last two are the advantages. Because of the three disadvantages,
{\SSS} is not used in practice, despite the promise of expanding fewer nodes.
In our framework, {\SSS} can be  re-expressed  as a single loop of
null-window {\AB} (+TT) calls.  
This makes {\SSS} easy to understand, solving
the first problem of the list.
Implementing the
algorithm is no longer a problem.
In the original publication {\SSS} was compared against the standard
text-book version of the {\AB} algorithm. Actual implementations of
{\AB} use many enhancements. For practical purposes, a 
comparison of {\SSS} against enhanced versions of {\AB} is much more
relevant.
Our experiments were performed with  tournament game-playing programs,
for chess, Othello and checkers. These games 
cover the range from high to low branching factor. Using multiple
programs, the
chance of unreliable answers, due to peculiarities of a single 
game or program, is reduced. All three programs are based on {\AB}
enhanced with iterative deepening, transposition tables, and move
ordering techniques. 

{\ABSSS} does not have a sorted OPEN \index{OPEN list}
list. The slow manipulation of the OPEN list is gone. 
Researchers trying to parallelize {\SSS} do no longer have
to try and find an efficient way to parallelize the OPEN list. Since
{\SSS} has been reformulated as a special case of {\AB},  the
\index{Alpha-Beta!parallel}
research on parallel {\AB} (see \cite{Broc95}\index{Brockington} for a detailed overview) is now
directly applicable; the 
framework may lead to some new ideas for parallel best-first {\AB}
versions. 
In short, the second problem of the list is now solved.
 
The experiments show that the memory needs of {\SSS} in current
\index{MT framework}
game-playing programs are reasonable. Instances of the {\MT} framework 
need $O(w^{d/2})$ transposition table entries to store the intermediate
search results. Contrary to what the literature says, for game-playing
programs under tournament conditions, this is a practical size. 
An additional argument is that an analysis of {\AB}'s memory needs
shows that to achieve high performance {\AB} needs memory of
about this size too. Thus, the third problem is solved too.

Having shown the first three points wrong, the reformulation of {\SSS}
is now a 
practical  {\AB} variant. However, we have also examined the
remaining two positive points. As it turns out they are wrong as well. 
Game-playing programs that use {\AB} enhance it with forms of dynamic
move reordering, such as iterative deepening (ID). This violates the
assumptions of {\SSS}'s dominance proof, making it possible for
ID-{\AB} to search {\em less\/} leaf nodes than ID-{\SSS}, which has
actually happened in our experiments.

The last point states that {\SSS} expands significantly less nodes
than {\AB}. For simulations this may be true, but not for real
applications. In game-playing programs {\SSS} expands on average
a few percent less leaf nodes than {\AB}. Furthermore, these
programs generally use an enhanced version of {\AB} called
{\NS}. Using this algorithm as the base line, even these few percents
disappear (with {\NS} getting a clear advantage when interior nodes are
counted as well).

Thus {\em all\/} five points are wrong: {\SSS} is a practical
algorithm, but there is no point in using it, since practical {\AB}
versions out-perform it.
% 
% {\NS} is  a null-window enhancement to {\AB}. The
% reformulation of {\SSS}, using null-window {\AB} search, and the fact
% that its performance is comparable to that of \mbox{\NS}, make it
% reasonable to view {\SSS} as another  null-window {\AB}
% enhancement.

\subsubsection{MTD$\mathbf{(f)}$} 
{\ABSSS} is an instance of the {\MT} framework that starts the
sequence of null-window {\AB} searches at $+\infty$, descending down
to the minimax value. Another instance is {\ABDUAL}, which starts at
$-\infty$, going up to the minimax value. Intuition says that a start
value closer to the 
minimax value should perform better. It gets a head start on the
algorithms starting at $\pm\infty$. 
Experiments show that this intuition is true. Creating
bounds for an uninteresting value is wasteful. On average it is more
efficient to start establishing bounds that are closer to the target.
% (figures~\ref{fig:swallow1}--\ref{fig:swallowsim}). 

\index{MTD$(f)$}
{\MTDf} is a new algorithm embodying this idea. It performs
better than {\NS}, the current 
algorithm of choice for game-playing programs. In our tests
{\MTDf}'s improvement
over {\NS} is slightly bigger than {\NS}'s improvement over
{\AB}. The efficiency comes at no extra algorithmic complexity:
just add a single control loop to a standard {\AB}-based program. 

% \begin{figure}
% \begin{center}
% \begin{tabular}{r|cc}
%  & best-first & {\bf not} best-first \\ \hline 
% good start value & {\MTDf} & {\NS} \\
% {\bf not} good start value & {\SSS} & {\AB} 
% \end{tabular}
% \end{center}
% \caption{Four Algorithms, Two Factors}\label{fig:bfdftable2}
% \end{figure}
% In the table in figure~\ref{fig:bfdftable2} we see the two techniques for
% constructing efficient 
% algorithms, as they are used by {\MTDf} and the other algorithms.

One of the most interesting outcomes of our experiments is that the
performance of all algorithms differs only by a few percentage
points. The search enhancements used in high-performance game-playing
programs  improve the search efficiency to such a high degree, that 
the question of which algorithm to use, be it {\AB}, {\NS}, {\SSS} or
{\MTDf}, is no longer of prime importance. (For programs of
lesser quality, the performance difference will be bigger, with
{\MTDf} out-performing {\NS} by a wider margin. There the best-first
nature of {\MTDf} will make more of a difference. Also, in some cases
{\SSS} does not perform very well.) 
Judging by the performance, the
enhancements have become more important than the base algorithm.
Furthermore, seen from an algorithmic viewpoint, in the new
framework a number of
best-first algorithms become enhanced versions of {\AB}.

The
{\AB} paradigm is versatile enough to be used for creating a
best-first node selection sequence.
Given that the new formulation for best-first full-width  minimax
search is more general, clearer, and allows a better performance, we
believe that \index{SSS*!footnote}
the old {\SSS} should become a footnote in the history of minimax search.
% 
% Considering, that  {\SSS} is equivalent to a special case of {\AB},
% that is out-performed by other {\AB} variants, both best-first and
% depth-first, and that there is a new, clearer, framework for best-first
% search of fixed-depth minimax trees, we think that  the old
% {\SSS} is now a closed book.
% There is no need anymore to spend effort on
% understanding the {\SSS} formulation; future research  but can (for
% example) be 
% directed towards finding ways to make better use of  irregularities
% of the search tree.

\index{search!best-first}
\index{search!depth-first}
Traditionally depth-first and best-first search are considered to be two
fundamentally different search strategies. Depth-first search uses little
memory, but 
has a simplistic, rigid, left-to-right node expansion
strategy. Best-first search, on the other hand, uses (too) much memory, but
has a smart, flexible, node expansion strategy.  Best-first
algorithms typically recalculate after each node expansion which node
appears the most promising to expand next. They jump from one part of
the tree to another, a pattern that is markedly different from the
fixed left-to-right sweep through the tree of a depth-first search.

Our work shows that in reality the picture is less clear cut. Simple
enhancements to depth-first search create a node selection sequence
that is identical to that of best-first algorithms. It is
not hard to make depth-first search transcend the rigid left-to-right
strategy. The view of best-first search as a memory-hungry monster %strategy
with an inefficient OPEN list needs refinement too. The new {\MT}
framework  solves the OPEN list problem and shows that memory
requirements are not a problem for real-time search.

Thus, positioning depth-first and best-first as two irreconcilable,
fundamentally different strategies, is an oversimplification. In
reality, they are not that different at all. A few simple enhancements
show how close they really are---at least in minimax search.

\subsubsection{The Minimal Tree}
Knuth and Moore  reported a limit on the performance of any
algorithm that finds the minimax value by searching a tree of uniform
$w$ and $d$ \cite{Knut75}. Any such   algorithm has to search at least
the tree that proves the minimax value.
Many authors of game-playing programs compare the size of their search
tree to the size of the minimal tree, to see how much improvement of
their search algorithm is still possible. 
\index{minimal tree}

However, game playing programs do not search uniform trees. And since
transpositions occur, the search {\em tree\/} is really a {\em graph}.
One solution, adopted by several authors,
is to redefine the minimal ``tree'' as the graph that is
searched by {\AB} when all cutoffs are caused by the first child at a
node. We call this the Left-First Minimal Graph or LFMG. 
However, the LFMG is not the smallest graph that proves the minimax
value. Because of transpositions and variances in $w$ and $d$
alternative children may cause cutoffs that are cheaper to
compute. The entity where each cutoff is backed-up by the smallest sub
tree is called the Real Minimal Graph, or RMG. 

\index{minimal graph!the real}
\index{RMG|see{minimal graph, the real}}
Finding the RMG is a computationally infeasible task for non-trivial
search depths, due to transpositions. Using approximation
techniques that exploit
irregular branching factors or transpositions, we have found
that the RMG is at least a factor of 1.25 (for chess) to 2 (for
checkers) smaller than the LFMG.
% (figures~\ref{fig:armgchin}--\ref{fig:armgpho}). 

Current game-playing programs are further from the minimal graph
that proves the minimax value than is generally assumed. There is more
room for improvement. One such improvement is the Enhanced Transposition
Cutoff, or ETC. For games with many transpositions this technique
reduces the search effort significantly.
% (figures~\ref{fig:jsx}--\ref{fig:jsxchin}). 

\section{Future Work}\index{future work}
The research described in the previous chapters has uncovered a
number of interesting avenues for further research. We
list the following: 
\begin{itemize}
\item {\em Node Expansion Criteria for {\MTDf}}\\
The literature describes static node expansion criteria for  {\AB}
and {\SSS} \cite{Pijl91,Rein89}.  Finding these criteria for
{\MTDf} can give additional insight in the relation between the start
value of a search and the size of the search tree, supplementing the
experimental evidence in  section~\ref{sec:swallow}. 

\item {\em Value/Size Experiments}\\
In addition to more analysis, more experiments are needed to gain
a deeper insight into the effect of the size and value of the {\AB}
window on the size of the search tree, extending the work on the
minimax wall \cite{Mars87} and section~\ref{sec:swallow}. 

\item {\em {\NS} and {\MTDf}}\\
The same---more analysis, more experiments---is needed to gain a better
understanding of the relation between {\NS} and {\MTDf}.
Section~\ref{sec:bfdf} only scratches the surface of this relation. 

\item {\em Variable Depth {\MTDf}}\\
All experiments in this work were performed for fixed-depth searches,
to make sure that different algorithms searched the same tree.
We believe that since both algorithms use search windows that are
comparable in size and value, the reaction of {\MTDf} will not be very
different from that of {\NS} when search extensions and forward pruning are
turned on again. However, experiments are needed to show whether this is indeed
the case.

\item {\em Effect of Individual Enhancements}\\
Section~\ref{sec:simsucks} describes the hazards of simulating minimax
algorithm performance for real games.  Although we believe that
real trees are too hard to model realistically, it might be worthwhile to
try to gain insight into the effect of individual aspects of real
trees, such as iterative deepening, the history heuristic, search
extensions, transpositions, narrow search windows, the distribution of
leaf values, and the correlation between parent and child positions. 

\item {\em Exploiting Irregularity of the Branching Factor}\\
Chapter~\ref{chap:anal} showed that the ARMG is 
significantly smaller than the LFMG.  The chapter suggests that there
is room (at least in some games) to exploit the irregularity of the
branching factor, by devising a suitable {\AB} enhancement. (See
\cite{Plaa96b} for some early results.)

\item {\em Replacement Schemes}\\
The work on the relation between {\AB} and {\SSS} (chapters~\ref{chap:mt}
and \ref{chap:exper}) makes heavy use of the transposition table to
store information from previous search passes.  Conventional
transposition table implementation choices \cite{Schaeffer86}  appear
to be well-suited for this task.  However, analysis of and experiments
with storing additional information and applying different replacement
schemes (see \cite{Breu94,Knut73c}) can be fruitful.

\item {\em Search Inconsistencies}\\
On page~\pageref{searchinconsistencies} the problem of search
inconsistencies is mentioned.  Search inconsistencies occur when, using
variable deepening or transposition tables, search results of
different search depths (different accuracies) are compared.  Every
now and then 
a game is lost due to this problem and programmers are painfully
reminded of it.  However, although it is a well-known problem, it is,
to the best of our knowledge, still unsolved.

\item {\em Parallelism}\\
Much research effort has been devoted to finding ways to parallelize
minimax algorithms efficiently. 
\index{parallelism}\index{Alpha-Beta!parallel}%
Parallelization of the {\MTD} family of algorithms seems viable along
two lines. First of all, the conventional tree-splitting/work-stealing
\index{parallelism!work stealing}%
parallelization techniques used for {\AB} (see, for example,
\cite{Blum94,Broc95,Feldmann93,Kusz94,Mars82a}) are obvious
candidates to try, since the {\AB} procedure forms the heart of the {\MTD}
algorithms. Second, the calls to {\AB} in the main loop of
{\MTD} can be run in parallel, each with different values
for the null-window.  This creates a Baudet-like scheme
\cite{Baud78b}, with the exceptions that now the parallel aspiration
\index{Baudet}\index{parallelism!Baudet}%
windows have become null windows, and
that in {\MTD} the parallel {\AB}-instances share information, through
their transposition table.  However, there is evidence that such
a parallelization is not well-suited for strongly ordered search
spaces (see, for example, \cite{Baud78b,Mars82a}).  
%
%{\AB}-instances are independent, each searching parts of the tree that
% the others also have to search---for example, each processor has to
% construct at least 
% the minimal tree. Analysis showed that this duplication of efforts gave
% rise to a bound on the speed-up of about 5 or 6 \cite{Baud78b,Mars82a}. 
% With information sharing through a transposition table, this kind of
% duplication would be aleviated to a certain extent.
% not arise since processors would benefit from each
% others work, and the speed-up bound of 5 or 6 would not apply (which
% does not mean .
% A relation with the range of leaf values is more likely.

Passing information from one {\AB}-pass to another via the
transposition table is a cornerstone in the design of the {\MTD}
algorithms.  Therefore, we believe that an efficient implementation of
a (logically) shared transposition is a necessary condition for a
succesful parallelization of {\MTD}. 
\end{itemize}

\subsubsection{Understanding Trees}
This work was driven by a desire to understand better what is going on
in  search trees as they are being searched by full-width
minimax algorithms.
For algorithms like {\AB}, {\NS}, {\SSS} and
{\MTDf}, we have found two notions central to our understanding:
bounds and solution trees. Realizing that all these
algorithms---and the minimal tree---could be understood in these terms
created a new perspective on  best-first and depth-first full-width
minimax search. They have been a guide throughout this research.

\index{bound|see {solution tree}}
\index{solution tree}
Solution trees and the structure of 
the minimal tree can help explain many interesting phenomena 
witnessed in the experiments.
Our experiments also show the limitations of our model and the
danger of putting too much trust in models of reality. 
For example, although the minimal tree may be a good model to
understand certain 
phenomena in search trees, it is not an accurate limit on the
performance of game-playing programs.
Analyses and simulations of {\SSS} and {\AB} turned out
to mispredict their relative performance in real applications severly.
% (section~\ref{sec:simsucks}). 
The reason for the difference between our results and that of 
simulations is that the trees generated in actual applications are complex.

\index{simulation}
It is hard to create reliable models for simulations. 
The field of minimax search is fortunate to have  a
large number of game-playing
programs available. These should be used in preference to
artificially-constructed simulations. 
Future research should try to identify factors 
that are of importance in real game trees, and use them as a guide in
the construction of better search algorithms, instead of
artificial models with a weak link to reality.
For example, in pursuing the real minimal graph existing notions like bounds
and solution trees
% ---although
% they are well suited to reason about uniform trees---
proved
inadequate to explain many results. 
Better concepts to help in reasoning
about irregularity and transpositions are dearly needed. Finding them
would be very useful for research on the minimal graph and on improving
full-width algorithms further.

 % concl

\appendix

\ \newpage

%\cleardoublepage

\thispagestyle{empty}

\ \newpage

%\cleardoublepage
\chapter{Examples}
% \chapter{{\AB} Example}
\label{app:ex}
\section{{\AB} Example}
\label{app:abex}
% \markboth{Appendix A}{{\AB} Example}
% \markboth{Appendix A}{{\AB} Example}
% \markboth{Appendix A}{Appendix A}
%\subsection{{\AB} Example}
\label{sec:abex}
\index{Alpha-Beta!example}
To give an idea of how {\AB} works, this appendix illustrates how it traverses
the tree of figure~\ref{fig:mmtree} in detail and
concept. For convenience, figure~\ref{fig:abcodeex} shows the code of
the {\AB} function again.
The two items of most interest are how and why cutoffs are performed,
and seeing how {\AB} constructs the minimal tree that proves $f_{root}$.
In this example {\AB} will find some cutoffs, but it will traverse more
than the minimal tree, since the children of some nodes are not ordered
best first. (For example, in max node $a$ the left-most child $b$ is
not the highest and in min node $k$ the left-most child is not the
lowest.) 
The values for $\alpha$, $\beta$ and $g$ as {\AB} traverses them are
shown next to the nodes in figure~\ref{fig:abtree}. Children that are
cut off are shown as a small black dot.

\begin{figure}
{\small
\begin{tabbing}
mmmmmmmm\=mm\=mm\=mm\=mm\kill
\> {\bf function} alphabeta$(n, \alpha, \beta) \rightarrow g$;\\
\> \> {\bf if} $n$ = leaf {\bf then return} eval$(n)$;\\
\> \> {\bf else if} $n$ = max  {\bf then}\\
\> \> \> $g := -\infty$;\\
\> \> \> $c := $ firstchild$(n)$;\\
\> \> \> {\bf while} $g < \beta$ {\bf and}  $c \neq \bot$ {\bf do}\\
\> \> \> \> $g := $ max$\big(g,$ alphabeta$(c, \alpha, \beta)\big)$;\\
\> \> \> \> $\alpha := \max(\alpha, g)$;\\
\> \> \> \> $c := $ nextbrother$(c)$;\\
\> \> {\bf else} /* $n$ is a min node */ \\ % {\bf then}\\
\> \> \> $g := +\infty$;\\
\> \> \> $c := $ firstchild$(n)$;\\
\> \> \> {\bf while} $g > \alpha$ {\bf and} $c \neq \bot$ {\bf do}\\
\> \> \> \> $g := $ min$\big(g,$ alphabeta$(c, \alpha, \beta)\big)$;\\
\> \> \> \> $\beta := \min(\beta, g)$;\\
\> \> \> \> $c := $ nextbrother$(c)$;\\
\> \> {\bf return} $g$;
\end{tabbing}
}
\caption{The {\AB} Function}\label{fig:abcodeex}
\end{figure}

\Treestyle{%
%  \addsep{1pt}%
  \minsep{2pt}%
  \vdist{30pt}%
  \nodesize{11pt}%
 }    % smaller than default 

\begin{figure}
{\footnotesize
% {\scriptsize
 \begin{Tree}
   \node{\external\type{square}\cntr{$e$}\bnth{$41$}\lft{$^{+\infty}_{-\infty}$}}
   \node{\external\type{square}\cntr{$f$}\bnth{$5$}\lft{$^{41}_{-\infty}$}}
   \node{\cntr{$d$}\lft{$^{+\infty}_{-\infty}$}\rght{$5$}}
   \node{\external\type{square}\cntr{$h$}\bnth{$12$}\lft{$^{+\infty}_{5}$}}
   \node{\external\type{square}\cntr{$i$}\bnth{$90$}\lft{$^{12}_{5}$}}
   \node{\cntr{$g$}\lft{$^{+\infty}_{5}$}\rght{$12$}}
   \node{\type{square}\cntr{$c$}\lft{$^{+\infty}_{-\infty}$}\rght{$12$}}
   \node{\external\type{square}\cntr{$l$}\bnth{$101$}\lft{$^{12}_{-\infty}$}}
   \node{\external\type{square}\cntr{$m$}\bnth{$80$}\lft{$^{12}_{-\infty}$}}
   \node{\cntr{$k$}\lft{$^{12}_{-\infty}$}\rght{$80$}}
   \node{\external\type{dot}}
   \node{\type{square}\cntr{$j$}\lft{$^{12}_{-\infty}$}\rght{$\geq 80$}}
   \node{\cntr{$b$}\lft{$^{+\infty}_{-\infty}$}\rght{$12$}}

   \node{\external\type{square}\cntr{$q$}\bnth{$10$}\lft{$^{+\infty}_{12}$}}
   \node{\external\type{dot}}
   \node{\cntr{$p$}\lft{$^{+\infty}_{12}$}\rght{$\leq 10$}}
   \node{\external\type{square}\cntr{$s$}\bnth{$36$}\lft{$^{+\infty}_{12}$}}
   \node{\external\type{square}\cntr{$t$}\bnth{$35$}\lft{$^{36}_{12}$}}
   \node{\cntr{$r$}\lft{$^{+\infty}_{12}$}\rght{$35$}}
   \node{\type{square}\cntr{$o$}\lft{$^{+\infty}_{12}$}\rght{$35$}}
   \node{\external\type{square}\cntr{$w$}\bnth{$50$}\lft{$^{35}_{12}$}}
   \node{\external\type{square}\cntr{$x$}\bnth{$36$}\lft{$^{35}_{12}$}}
   \node{\cntr{$v$}\lft{$^{35}_{12}$}\rght{$36$}}
   \node{\type{dot}\external}
   \node{\type{square}\cntr{$u$}\lft{$^{35}_{12}$}\rght{$\geq 36$}}
   \node{\cntr{$n$}\lft{$^{+\infty}_{12}$}\rght{$35$}}
   \node{\type{square}\cntr{$a$}\lft{$^{\beta\ =\ +\infty}_{\alpha\ =\ -\infty}$}\rght{$g = 35$}}
\end{Tree}
\hspace{6.25cm}\usebox{\TeXTree}
}
\caption{Example Tree for {\AB}}\label{fig:abtree}
\end{figure}

\Treestyle{%
  \addsep{2pt}%
  \minsep{10pt}%
  \vdist{30pt}%
  \nodesize{14pt}%
 }    % smaller than default 

\begin{figure}
 {\footnotesize
%{\scriptsize
\begin{tabular}{r|cccccclcl}
\# & $n$ & $\alpha_n$ & $\beta_n$ & $g_n$ & cutoff? & $f^-_n$ & $T^-_n$         & $f^+_n$   & $T^+_n$ \\ \hline
1  & $a$ &$-\infty$ &$+\infty$&$-\infty$& &$-\infty$ &               &$+\infty$&       \\
2  & $b$ &$-\infty$ &$+\infty$&$+\infty$& &$-\infty$ &               &$+\infty$&       \\
3  & $c$ &$-\infty$ &$+\infty$&$-\infty$& &$-\infty$ &               &$+\infty$&       \\
4  & $d$ &$-\infty$ &$+\infty$&$+\infty$& &$-\infty$ &               &$+\infty$&       \\
5  & $e$ &$-\infty$ &$+\infty$& 41  & & 41           & $e$           &  41     &   $e$ \\
6  & $d$ &$-\infty$ &  41& 41&$41>-\infty$&$-\infty$ &          & 41      &   $d,e$ \\
7  & $f$ &$-\infty$ &   41    & 5   &     &5         &  $f$          & $f$     &   $f$ \\
8  & $d$ &$-\infty$ &   5     & 5   & $\bot$&5     & $d,e,f$       & 5       &   $d,f$ \\
9  & $c$ &   5      &$+\infty$& 5   &$5<+\infty$&5   & $c,d,e,f$     &$+\infty$& \\
10 & $g$ &   5      &$+\infty$&$+\infty$&&$-\infty$  &               &$+\infty$& \\
11 & $h$ &   5      &$+\infty$& 12  &&  12           & $h$           &  12     &   $h$ \\
12 & $g$ &   5      & 12      & 12  &$12>5$&$-\infty$&            & 12      &   $g,h$ \\
13 & $i$ &   5      & 12      & 90  & & 90           & $i$           &  90     &   $i$ \\
14 & $g$ &   5      & 12      & 12  & $\bot$& 12     & $g,h,i$    &  12     &   $g,h$ \\
15 & $c$ &  12      &$+\infty$& 12  & $\bot$& 12     &$c,g,h,i$  &  12     &$c,d,f,g,h$\\
16 & $b$ &$-\infty$ & 12   &12&$12>-\infty$&$-\infty$&       & 12      &$b,c,d,f,g,h$ \\
17 & $j$ &$-\infty$ & 12      &$-\infty$&&$-\infty$  &               &$+\infty$&  \\
18 & $k$ &$-\infty$ & 12      &$+\infty$&&$-\infty$  &               &$+\infty$&  \\
19 & $l$ &$-\infty$ & 12      &101  &  &101          & $l$           & 101     &   $l$ \\
20 & $k$ &$-\infty$ & 12&101 &$101>-\infty$&$-\infty$&               & 101     & $k,l$ \\
21 & $m$ &$-\infty$ & 12      & 80  & & 80           & $m$           &  80     &   $m$ \\
22 & $k$ &$-\infty$ & 12      & 80  &$\bot$& 80      & $k,l,m$       &  80     &$k,m$ \\
23 & $j$ &80& 12      & 80  &$80\not<12$&  80& $j,k,l,m$     &$+\infty$&    \\
24 & $b$ &$-\infty$ &12& 12  &$\bot$&12&$b,c,g,h,         $&  12     &   $b,c,d,f,g,h$ \\
   &     &          &  &     &      &  &$        i,j,k,l,m$&         &
                 \\
25  & $a$ &12 &$+\infty$& 12&$12<+\infty$&12&$a,b,c,g,           $&$+\infty$&    \\
    &     &   &         &   &            &  &$        h,i,j,k,l,m$&
        &    \\
26 & $n$ &12&$+\infty$&$+\infty$& &$-\infty$& &$+\infty$& \\
27 & $o$ &12&$+\infty$&$-\infty$& &$-\infty$& &$+\infty$& \\
28 & $p$ &12&$+\infty$&$+\infty$& &$-\infty$& &$+\infty$& \\
29 & $q$ &12&$+\infty$&10& &10&$q$&10&$q$\\
30 & $p$ &12&10&10&$10\not>12$&$-\infty$&&10&$p,q$\\
31 & $o$ &12&$+\infty$&10&$10<+\infty$&$-\infty$&&$+\infty$&\\
32 & $r$ &12&$+\infty$&$+\infty$&&$-\infty$&&$+\infty$&\\
33 & $s$ &12&$+\infty$&36&&36&$s$&36&$s$\\
34 & $r$ &12&36&36&$36>12$&$-\infty$&&36&$r,s$\\
35 & $t$ &12&36&35&&35&$t$&35&$t$\\
36 & $r$ &12&35&35&$\bot$&35&$r,s,t$&35&$r,t$\\
37 & $o$ &35&$+\infty$&35&$\bot$&35&$o,r,s,t$&35&$o,p,q,r,t$\\
38 & $n$ &12&35&35&$35>12$&$-\infty$&&35&$n,o,p,q,r,t$\\
39 & $u$ &12&35&$-\infty$&&$-\infty$&&$+\infty$&\\
40 & $v$ &12&35&$+\infty$&&$-\infty$&&$+\infty$&\\
41 & $w$ &12&35&50&&50&$w$&50&$w$\\
42 & $v$ &12&35&50&$50>12$&$-\infty$&&50&$v,w$\\
43 & $x$ &12&35&36&&36&$x$&36&$x$\\
44 & $v$ &12&35&36&$\bot$&36&$v,w,x$&36&$v,x$\\
45 & $u$ &36&35&36&$36\not<35$&36&$u,v,w,x$&$+\infty$&\\
46 & $n$ &12&35&35&$\bot$&35&$n,o,r,s,         $&35&$n,o,p,q,r,t$\\
   &     &  &  &  &      &  &$        t,u,v,w,x$&  &             \\
47 & $a$ &35&$+\infty$&35&$\bot$&35&$a,n,o,r,  $&35&$a,b,c,d,f,g,$\\
   &     &  &         &  &      &  &$        s,t,u,v,w,x$&  &$h,n,o,p,q,r,t$
\end{tabular}
\caption{{\AB} Example}\label{fig:abex}
}
\end{figure}

The table in figure~\ref{fig:abex} gives a step-by-step account of
{\AB}'s progress. Cutoffs are indicated in the table, and explained in
the text below. Since {\AB} works towards finding a max and a min
solution tree of equal value, we have shown these in the table as well.
Recall from the postcondition of {\AB} that if a node returns a $g$
value that lies within the search window, then both a $T^+$ and a $T^-$
have been traversed. This case applies to most of the nodes. Only nodes
$j, p$ and $u$ return a value outside their window. Their value is
determined by a single solution tree.
Upper bounds for a node are indicated as $f^+$, lower bounds
as $f^-$. Max solution trees are shown as $T^+$, min solution trees as $T^-$.
Figure~\ref{fig:critabtree} gives the final minimal tree that proves
the minimax value.

\begin{figure}
{\small
 \begin{Tree}
   \node{\external\type{square}\cntr{$t$}\bnth{$35$}}
   \node{\external\type{square}\cntr{$s$}\bnth{$36$}}
   \node{\cntr{$r$}}
   \node{\external\cntr{$q$}\type{square}\bnth{$10$}}
   \node{\cntr{$p$}\leftonly}
   \node{\type{square}\cntr{$o$}}
   \node{\external\type{square}\cntr{$x$}\bnth{$36$}}
   \node{\external\type{square}\cntr{$w$}\bnth{$50$}}
   \node{\cntr{$v$}}
   \node{\type{square}\cntr{$u$}\leftonly}
   \node{\cntr{$n$}}

   \node{\external\type{square}\cntr{$h$}\bnth{$12$}}
   \node{\cntr{$g$}\leftonly}
   \node{\external\type{square}\cntr{$f$}\bnth{$5$}}
   \node{\cntr{$d$}\leftonly}
   \node{\type{square}\cntr{$c$}}
   \node{\cntr{$b$}\leftonly}
   \node{\type{square}\cntr{$a$}}
\end{Tree}
\hspace{6.4cm}\usebox{\TeXTree}
}
\caption{Minimal {\AB} Tree}\label{fig:critabtree}
\end{figure}

The root is called with $\alpha = -\infty$ and $\beta = +\infty$. Its
first child $b$ is expanded with the same parameters. The same holds for
node $c, d$ and $e$. Node $e$ is a leaf, which calls the evaluation
function. It returns its minimax value
of 41 to its parent. Here, at node $d$, the values of $g$ and $\beta$
are updated to 
41. 
% Recall that the variable $g$ in a min node represents an upper
% bound on its return value. This upper bound is
% used in the search of the other children by passing them the new $\beta$
% value. 
Node $d$ performs a cutoff check; since $g > \alpha$
(because $41 > -\infty$) the search continues to the next child, $f$.
Since $\beta$ was updated in its parent, this node is searched with a
window of $\langle -\infty, 41\rangle$. Node $f$ returns its minimax
value.
% Parent $d$ updates $g$ and $\beta$ to the minimum of $41$ and $5$: it
% lowers the upper bound $g$ from 41 to 5. 
% Since there are no more children
% and $\alpha < g < \beta$ (the success part of the postcondition)
% this is also the 
% minimax value of node $d$. 
% The table in figure~\ref{fig:abex} shows
% the max and min solution tree that determine this value in line~8. 

Parent $d$
returns 5, the minimum of 41 and 5. Parent $c$ updates $g$ and 
$\alpha$ to 5. 
% Node $c$ is a max node, so the value of $g$ is a lower
% bound on the final return value of node $c$. 
Node $c$ continues to search node $g$, 
since $5 < +\infty$. The lower bound is
used to search 
node $g$ and its children with the smaller search window of
$\langle 5, +\infty\rangle$. 
% Its child $h$ returns 12, which does not cause a
% cutoff in node $g$, since $12 > 5$. (A cutoff would have been wrong,
% because $g$ is a min node, whose value can still go below 12.)
% Node $i$ is searched, which returns 90. 
Node $g$ returns the minimum of 12 and
90 to $c$, which returns the maximum of 5 and 12 to $b$.
In node $b$ the search is continued to expand $j$. Node $b$ is a min
node. The $g$-value 12 is an upper bound, substantiated by a max
solution tree containing node $c, d, f, g$ and $h$. See line 16 in the
table. The search window
for node $j$ is reduced to $\langle -\infty, 12\rangle$, indicating
that parent $b$ already has an upper bound of 12, so that if in any
of the children of $b$
a lower bound $\geq 12$ appears, the search can be stopped.
% , since
% that node can never influence the outcome of $b$.
Node $j$ expands
the sub-tree rooted in its child $k$, which returns 80. This causes
a cutoff of 
its brother in node $j$, since $80 \not< 12$. The search of node $j$,
whose value, being a max node,  is a lower bound of 80 that can only
increase, is 
no longer useful. The value of parent $b$ is already as low as
12, and since it is a min node, it will never increase. Because $b$
has no other children to lower his value further, it returns 
his value too. The return value of $b$ is defined by min solution tree
$b, c,g, h, i, j, k, l, m$ and max solution tree $b, c, d, f, g, h$.
% Line 24 in the table lists the minimal tree that
% defines $b$'s return value.

At the root the value $g$ is updated to the new lower bound of 12.
% , substantiated
% by the min solution tree consisting of nodes $b, c, g, h$ and
% $i$.
%  (line 25). 
Consequently, $\alpha$ is updated to 12. Since $12 <
+\infty$, no cutoff occurs. Searching the
sub-tree below $n$ can still increase the $g$ value of the root. Node
$n$ is expanded. Node $o, p$ and $q$ are searched with
window $\langle 12, +\infty\rangle$. Node $q$ returns 10. At its parent
$p$ this causes a cutoff, since $10 \not> 12$. In contrast to the previous
cutoff, the reason for it cannot be found by looking at the value of
the direct
parent, since $o$ does not have a $g$ value yet other than $-\infty$. This
kind of cutoff is\index{cutoff, shallow}\index{cutoff, deep}
known as a {\em deep\/} cutoff (in contrast to the previous
cutoff, which is said to be shallow). It is called deep because the
reason for the cutoff lies further away than the direct parent of the
node. The sub-tree
below $b$ has returned 12, so the value of $a$, which is a max node, 
will be at least 12. Node $p$, a min node, can at most return a value of
10. The value of node $q$ ensures that none of its brothers can ever
influence the value of the root. Node $p$ returns 10. 

Next nodes $r, s, t, u, v, w$ and $x$ are searched.
% 
% This value does
% not cause a cutoff at node $o$, nor does it cause a change in the window
% with which node $r$ is searched. The sub tree below $r$ is searched and
% returns 35. Node $o$ returns the maximum of 10 and 35 to $n$, which
% searches node $u$ with window $\langle 12, 35\rangle$. Node $u$ searches
The sub-tree below $v$ returns 36. This causes a cutoff (one of
the shallow kind) in node $u$, since $36 \not< 35$. 
% The cutoff prevents
% useless further expansions below $u$, because its value will not drop
% below 36, and can thus never influence the value of $n$ anymore. 
Node
$u$ returns 36, node $n$ returns the minimum of 35 and 36, and the root
returns the maximum of 12 and 35. The minimax value of the tree has
been found, it is 35.

The example  illustrates that {\AB} can miss
some cutoffs. 
Although pruning occurs in the example, {\AB} builds a tree that is
larger than the minimal tree, as can be seen
by comparing figures~\ref{fig:abtree} and \ref{fig:critabtree}. {\AB}
expands 11 leaves, less than the minimax tree of 16, but more than the
minimal tree of 7.

% \cleardoublepage
% \chapter{{\SSS} Example}
\section{{\SSS} Example}
% \markright{{\SSS} Example}
\label{app:sssex}
% \markboth{Appendix B}{Appendix B}
% \markboth{Appendix B}{{\SSS} Example}
% \markboth{Appendix A}{{\SSS} Example}
In the next example we will give an
idea of how {\SSS} works.

\index{SSS*!Stockman's view}
\index{solution tree, min}
\index{SSS*!example}
Stockman originally used min solution trees to 
explain his algorithm. Here {\SSS} is explained using upper bounds and max
solution trees, since we think that the algorithm is easier to
understand that way and it makes the connection to {\ABSSS} easier
to see. 
The two key concepts in the explanation are
an {\em upper bound\/} 
on the minimax value, and a {\em max solution tree}. {\SSS} starts off
with an  
upper bound of $+\infty$, and works by successively lowering this
upper bound until it 
is equal to the minimax value. 
The max solution trees are constructed
to compute the value of each upper bound.

{\SSS} works by manipulating a list of 
nodes, the OPEN list. \index{OPEN list}
The nodes have a status associated with them, 
either {\em live\/} (L) or {\em solved\/} (S), and a merit, denoted $\hat{h}$.
The OPEN list is sorted in descending order, 
so that the entry with highest merit (the ``best'' node) is at the front
and will be selected for expansion.

\Treestyle{%
%  \addsep{1pt}%
  \minsep{3pt}%
  \vdist{30pt}%
  \nodesize{13pt}%
 }    % smaller than default 

\begin{figure}
{\small
 \begin{Tree}
   \node{\external\type{square}\cntr{$e$}\bnth{$41$}}
   \node{\external\type{square}\cntr{$n$}\bnth{$5$}}
   \node{\cntr{$d$}}
   \node{\external\type{square}\cntr{$g$}\bnth{$12$}}
   \node{\external\type{square}\bnth{$90$}}
   \node{\cntr{$f$}}
   \node{\type{square}\cntr{$c$}}
   \node{\external\type{square}\bnth{$101$}}
   \node{\external\type{square}\bnth{$80$}}
   \node{}
   \node{\external\type{square}\bnth{$20$}}
   \node{\external\type{square}\bnth{$30$}}
   \node{}
   \node{\type{square}}
   \node{\cntr{$b$}}

   \node{\external\type{square}\cntr{$k$}\bnth{$10$}}
   \node{\external\type{square}\bnth{$80$}}
   \node{\cntr{$j$}}
   \node{\external\type{square}\cntr{$m$}\bnth{$36$}}
   \node{\external\type{square}\cntr{$o$}\bnth{$35$}}
   \node{\cntr{$l$}}
   \node{\type{square}\cntr{$i$}}
   \node{\external\type{square}\cntr{$s$}\bnth{$50$}}
   \node{\external\type{square}\cntr{$t$}\bnth{$36$}}
   \node{\cntr{$q$}}
   \node{\external\type{square}\bnth{$25$}}
   \node{\external\type{square}\bnth{$3$}}
   \node{\cntr{$r$}}
   \node{\type{square}\cntr{$p$}}
   \node{\cntr{$h$}}
   \node{\type{square}\cntr{$a$}}
\end{Tree}
\hspace{5.5cm}\usebox{\TeXTree}
}
\caption{Example Tree for {\SSS}}\label{fig:ssstree}
\end{figure}

\Treestyle{%
  \addsep{2pt}%
  \minsep{10pt}%
  \vdist{30pt}%
  \nodesize{14pt}%
 }    % smaller than default 

%\subsection{{\SSS} Example}
\label{sec:sssexample}
We will examine how {\SSS}  searches the tree in
figure~\ref{fig:ssstree} for its minimax value (the same tree as in
the {\AB} example). 
% The following contains a
% detailed description of how {\SSS} works.
% One of the reasons to create
% the {\ABSSS} reformulation was 
% the sense of confusion that the complexity of {\SSS} brings about. By
% using standard concepts from the {\AB} literature we try to alleviate
% this problem. 
% Although instructive, going
% through the example step-by-step is not necessary (or advisable) on a
% first pass through this thesis.
% to follow most of
% the rest of this thesis. 
Since the tree is almost the same as the one used by Pearl in his
explanation \index{Pearl}
of {\SSS} \cite{Pear84}, it may be instructive to compare his min with our
max solution tree explanation. 
In the trees in figures~\ref{fig:pass1}--\ref{fig:pass4} the
nodes are numbered $a$ to $t$ in the order in which {\SSS} visits them
first.  
A number of stages, or passes, can be distinguished in the traversal of
this tree. At the end of each pass the OPEN list consists of {\em solved\/}
nodes only. In appendix~\ref{app:equiv} we will see that after
$\Gamma$ operations 1 and 3 {\SSS} has constructed a min solution tree
for the solved node. Appendix~\ref{app:equiv} also stresses the fact
that at any time the nodes in the OPEN list are the leaves of a
(partial) max solution tree rooted at the root, defining an upper
bound on it. These two solution trees are shown in the tables.
We will start by examining how {\SSS} traverses the first pass.\\

\noindent {\em First pass:\/} (see figures~\ref{fig:pass1} and
\ref{fig:ssstab1})\\ 
In the first pass the left-most max solution tree is expanded to
create the first non-trivial upper bound on the minimax value of the
root. 
The code in figure~\ref{fig:ssscodeex} places an entry in the OPEN list
containing the root 
node $a$, with a trivial upper
bound: $+\infty$. This entry, 
containing an internal max node which has as first child a min node,
matches $\Gamma$ case 6, causing $a$ to be replaced by its children (in
left-first order). The OPEN list now contains nodes $b$ and $h$, with a
value of 
$+\infty$. The left-most of these children, $b$, is at the front of the
list. It matches $\Gamma$ case 5, causing it to be replaced by its
left-most child, $c$. The list now contains nodes $c$ and $h$. Next, case 6
replaces $c$ 
by $d$ and $f$, and case 5 replaces $d$ by $e$, giving as list nodes
$e, f$ and $h$. Now a 
new $\Gamma$-case comes into play, since node $e$ has no children.
Node $e$
does not match case 6, but
case 4, causing its state to change from {\em live\/} to {\em solved}, and
its $\hat{h}$ value to go from $+\infty$ to $41$. Since the list is
kept sorted in descending order, the next entry on
the list appears at the front, $f$, the left-most entry with highest
$\hat{h}$ value. It matches case 5, $g$ is inserted, which matches
case 4, causing it to enter the list with value 12. The list is now
$\big(\langle h, L, \infty\rangle, \langle e, S, 41\rangle, \langle g, S,
12\rangle\big)$. 
Next the right subtree below $h$ is expanded in the same way, through
a sequence of $\Gamma$ cases 5 (node $i$ is visited), 6 ($j$ and $l$ 
enter the list), 5 ($k$ replaces $j$), 4 ($k$ gets value 10), 5 ($l$
is replaced by $m$), and 4 ($l$ gets value 36). The OPEN list is now
$\big(\langle e, S, 41\rangle, \langle m, S, 36\rangle,  \langle g, S,
12\rangle, \langle k, S, 10\rangle \big)$.
\begin{figure}
{\small
 \begin{Tree}
   \node{\external\type{square}\cntr{$e$}\bnth{$41$}}
   \node{\leftonly\cntr{$d$}\lft{$f^+ = 41$}}
   \node{\external\type{square}\cntr{$g$}\bnth{$12$}}
   \node{\leftonly\cntr{$f$}\rght{$f^+ = 12$}}
   \node{\type{square}\cntr{$c$}\lft{$f^+ = 41$}}
   \node{\leftonly\cntr{$b$}\lft{$f^+ = 41$}}

   \node{\external\type{square}\cntr{$k$}\bnth{$10$}}
   \node{\leftonly\cntr{$j$}\lft{$f^+ = 10$}}
   \node{\external\type{square}\cntr{$m$}\bnth{$36$}}
   \node{\leftonly\cntr{$l$}\rght{$f^+ = 36$}}
   \node{\type{square}\cntr{$i$}\rght{$f^+ = 36$}} 
   \node{\leftonly\cntr{$h$}\rght{$f^+ = 36$}}

   \node{\type{square}\cntr{$a$}\lft{$f^+ = 41$}}
\end{Tree}
\hspace{5.5cm}\usebox{\TeXTree}
}
\caption{{\SSS} Pass 1}\label{fig:pass1}
\end{figure}
\begin{figure}
%{\scriptsize
{\footnotesize
\begin{tabular}{r|cclclcl}
\# & $n$ & $\Gamma$  & OPEN list after $\Gamma$ & $f^-_{top}$ & $T^-_{top}$ &  $f^+_{root}$   & $T^+_{root}$ \\ \hline
0  &     &   & $\big(\langle a, L, +\infty\rangle\big)$  &  & &$+\infty$ & $a$     \\
1  & $a$ & 6 & $\big(\langle b, L, +\infty\rangle,
                     \langle h, L, +\infty\rangle\big)$   &$-\infty$ &
 &$+\infty$ & $a,b,h$  \\
2  & $b$ & 5 & $\big(\langle c, L, +\infty\rangle,
                     \langle h, L, +\infty\rangle\big)$  &$-\infty$ &
 & $+\infty$ & $a,b,c,h$  \\
3  & $c$ & 6 & $\big(\langle d, L, +\infty\rangle,
                     \langle f, L, +\infty\rangle,
                     \langle h, L, +\infty\rangle\big)$  &$-\infty$ &
 & $+\infty$ & $a,b,c,d,f,h$   \\
4  & $d$ & 5 & $\big(\langle e, L, +\infty\rangle,
                     \langle f, L, +\infty\rangle,
                     \langle h, L, +\infty\rangle\big)$  &$-\infty$ &
 & $+\infty$ & $a,b,c,d,e,f,h$    \\
5  & $e$ & 4 & $\big(\langle f, L, +\infty\rangle,
                     \langle h, L, +\infty\rangle,
                     \langle e, S, 41\rangle\big)$ & $-\infty$ & 
& $+\infty$ & $a,b,c,d,e,f,h$     \\
6  & $f$ & 5 & $\big(\langle g, L, +\infty\rangle,
                     \langle h, L, +\infty\rangle,
                     \langle e, S, 41\rangle\big)$ & $-\infty$ & 
 & $+\infty$ & $a,b,c,d,e,f,g,h$     \\
7  & $g$ & 4 & $\big(\langle h, L, +\infty\rangle,
                     \langle e, S, 41\rangle,
                     \langle g, S, 12\rangle\big)$ & $-\infty$ & 
 & $+\infty$ & $a,b,c,d,e,f,g,h$     \\
8  & $h$ & 5 & $\big(\langle i, L, +\infty\rangle,
                     \langle e, S, 41\rangle,
                     \langle g, S, 12\rangle\big)$ & $-\infty$ & 
 & $+\infty$ & $a,b,c,d,         $     \\
   &     &   &                                     
                                                   &           & 
 &            &$        e,f,g,h,i$     \\
9  & $i$ & 6 & $\big(\langle j, L, +\infty\rangle,
                     \langle l, L, +\infty\rangle,  
                                                 $ & $-\infty$         & 
 & $+\infty$          & $a,b,c,d,e,         $     \\
   &     &   & $                                  
                     \langle e, S, 41\rangle,
                     \langle g, S, 12\rangle\big)$ &           & 
 &           & $          f,g,h,i,j,l$     \\
10  & $j$ & 5 & $\big(\langle k, L, +\infty\rangle,
                     \langle l, L, +\infty\rangle,
                                                 $ & $-\infty$ & 
 & $+\infty$ & $a,b,c,d,e,           $     \\
    &     &   & $                                  
                     \langle e, S, 41\rangle,
                     \langle g, S, 12\rangle\big)$ &           & 
 &           & $          f,g,h,i,j,k,l$     \\
11 & $k$ & 4 & $\big(\langle l, L, +\infty\rangle,
                     \langle e, S, 41\rangle,
                                                 $ & $-\infty$ & 
 & $+\infty$ & $a,b,c,d,e,           $     \\
   &     &   & $                                    
                     \langle g, S, 12\rangle,
                     \langle k, S, 10\rangle\big)$ &           & 
 &           & $          f,g,h,i,j,k,l$     \\
12 & $l$ & 5 & $\big(\langle m, L, +\infty\rangle,
                     \langle e, S, 41\rangle,
                                                 $ & $-\infty$ & 
 & $+\infty$ & $a,b,c,d,e,             $     \\
   &     &   & $                                    
                     \langle g, S, 12\rangle,
                     \langle k, S, 10\rangle\big)$ &           & 
 &           & $          f,g,h,i,j,k,l,m$     \\
13 & $m$ & 4 & $\big(\langle e, S, 41\rangle,
                     \langle m, S, 36\rangle,
                                                 $ &  41& $e$ 
 & 41 & $a,b,c,d,e,             $     \\
   &     &   & $                             
                     \langle g, S, 12\rangle,
                     \langle k, S, 10\rangle\big)$ &    &     
 &    & $          f,g,h,i,j,k,l,m$     
\end{tabular}
\caption{{\SSS} Table Pass 1}\label{fig:ssstab1}
}
\end{figure}

We have seen so far that at max nodes all children were expanded
(case 6),
while at min nodes only the first child was added to the OPEN list
(case 5).
Case 4 evaluated the leaf nodes of the tree. Maintaining  the list
 in sorted order guaranteed
that the entry with the highest upper bound was at front. Note that
the sub-tree expanded thus far is a max solution tree
(compare figure~\ref{fig:pass1} to figure~\ref{fig:maxsoltree}).
The minimax value of this tree is 41, 
which is  also the $\hat{h}$
value of the first entry of the OPEN list. \\

\begin{figure}
{\small
 \begin{Tree}
   \node{\external\type{square}\cntr{$e$}\bnth{$41$}}
   \node{\external\type{square}\cntr{$n$}\bnth{$5$}}
   \node{\cntr{$d$}\lft{$f^+ = 5$}}
   \treesymbol{\lvls{1}\cntr{$f$}\bnth{skipped}\rght{$f^+ = 12$}}
   \node{\type{square}\cntr{$c$}\lft{$f^+ = 12$}}
   \node{\leftonly\cntr{$b$}\lft{$f^+ = 12$}}

   \treesymbol{\lvls{3}\cntr{$h$}\lft{$f^+ = 36$}\bnth{skipped}}

   \node{\type{square}\cntr{$a$}\lft{$f^+ = 36$}}
\end{Tree}
\hspace{5.5cm}\usebox{\TeXTree}
}
\caption{{\SSS} Pass 2}\label{fig:pass2}
\end{figure}
\begin{figure}
%{\scriptsize
 {\footnotesize
\begin{tabular}{r|cclclcl}
\# & $n$ & $\Gamma$  & OPEN list after $\Gamma$ & $f^-_{top}$ & $T^-_{top}$ & $f^+_{root}$   & $T^+_{root}$ \\ \hline
14 & $e$ & 2 & $\big(\langle n, L, 41\rangle,
                     \langle m, S, 36\rangle,
                                                 $ & $-\infty$ &    
 & 41 & $a,b,c,d,e,f,g,h,i,j,k,l,m$     \\
   &     &   & $
                     \langle g, S, 12\rangle,
                     \langle k, S, 10\rangle\big)$ &           &    
 &    &                                 \\
15 & $n$ & 4 & $\big(\langle m, S, 36\rangle,
                     \langle g, S, 12\rangle,
                                                $ & 36 & $m$ 
 & 36 & $a,b,c,d,e,n,f,g,h,i,j,k,l,m$     \\
   &     &   & $
                     \langle k, S, 10\rangle,
                     \langle n, S, 5\rangle\big)$ &    &     
 &    &                                 
\end{tabular}
\caption{{\SSS} Table Pass 2}\label{fig:ssstab2}
}
\end{figure}

\noindent {\em Second pass:\/} (see figures~\ref{fig:pass2} and \ref{fig:ssstab2})\\
% The first pass ended with  $\big(\langle e, S,
% 41\rangle, \langle m, S, 36\rangle, \langle g, S, 12\rangle, \langle k, S, 
% 10\rangle \big)$.  
In the second pass, {\SSS} will try to lower the upper bound of 41 to
come closer 
to $f$. The next upper bound will be computed by expanding a
brother of the critical leaf $e$. The critical leaf has a min parent, 
node $d$,
so expanding this brother can lower $d$'s value, which will, in turn,
 lower the minimax 
value at the root of the max solution tree. Since this value is the maximum 
of its leaves, there is no point in expanding brothers
of non-critical leaves,
since then node $e$ will keep the value of the root at $41$.
Thus, node $n$ is in a sense the best node to expand.
The entry for node $e$ matches $\Gamma$ case 2,
which replaces $e$ by 
the brother $n$, giving it state {\em live}, and the value 41, the sharpest
(lowest) upper bound of the previous pass. The $n$ entry matches $\Gamma$ case 
4. Case 4 evaluates the leaf, and assigns to $\hat{h}$ either this value
(5), or the 
sharpest upper bound so far, if that happens to be lower. Node $n$
gets value 5. In general, $\Gamma$ case 4 performs the minimizing operation of
the minimax 
function, ensuring that the $\hat{h}$ of the first (highest) entry of
the OPEN 
list will always be the sharpest upper bound on the minimax value of
the root, based on the previously expanded nodes. The OPEN list has become
$\big(\langle m, S, 36\rangle, \langle g, S, 12\rangle, \langle k, S,
10\rangle, \langle n, S, 5\rangle \big)$.
Thus, the upper bound on the root has been lowered to 36. Its value is
determined 
by a new (sharper) max solution tree, whose leaves are contained in
the OPEN list.
\\

\begin{figure}
{\small
 \begin{Tree}
   \treesymbol{\lvls{3}\cntr{$b$}\lft{$f^+ = 12$}\bnth{skipped}}

   \treesymbol{\lvls{1}\cntr{$j$}\bnth{skipped}\lft{$f^+ = 10$}}
   \node{\external\type{square}\cntr{$m$}\bnth{$36$}}
   \node{\external\type{square}\cntr{$o$}\bnth{$35$}}
   \node{\cntr{$l$}\rght{$f^+ = 35$}}
   \node{\type{square}\cntr{$i$}\rght{$f^+ = 35$}} 
   \node{\leftonly\cntr{$h$}\rght{$f^+ = 35$}}

   \node{\type{square}\cntr{$a$}\lft{$f^+ = 35$}}
\end{Tree}
\hspace{5.5cm}\usebox{\TeXTree}
}
\caption{{\SSS} Pass 3}\label{fig:pass3}
\end{figure}
\begin{figure}
%{\scriptsize
{\footnotesize
\begin{tabular}{r|cclclcl}
\# & $n$ & $\Gamma$  & OPEN list after $\Gamma$ & $f^-_{top}$ & $T^-_{top}$ & $f^+_{root}$   & $T^+_{root}$ \\ \hline
16 & $m$ & 2 & $\big(\langle o, L, 36\rangle,
                     \langle g, S, 12\rangle,
                    $ & $-\infty$ &   
 & 36 & $a,b,c,d,e,n,f,g,$\\
   &    &    &  $\langle k, S, 10\rangle,
                     \langle n, S, 5\rangle\big)   $                              &           &
  &   & $h,i,j,k,l,m,o$     \\
17 & $o$ & 4 & $\big(\langle o, S, 35\rangle,
                     \langle g, S, 12\rangle,
                    $ & 35 & $o$ 
 & 35 & $a,b,c,d,e,n,f,g,$\\
   &    &    &   $\langle k, S, 10\rangle,
                     \langle n, S, 5\rangle\big)             $                   &           &
  &   &  $h,i,j,k,l,m,o$     \\
\end{tabular}
\caption{{\SSS} Table Pass 3}\label{fig:ssstab3}
}
\end{figure}

\noindent {\em Third Pass:} (see figures~\ref{fig:pass3} and \ref{fig:ssstab3})\\
In the third pass, the goal of the search is to get the upper bound
below 36. 
Just as in the second pass, the first entry of the OPEN list, $m$,
matches $\Gamma$ case 2, and its
brother is inserted. It matches $\Gamma$ case 4, so it gets evaluated.
The new brother is node $o$, with value 35.
Again, a sharper upper bound has been found. The new OPEN list is
$\big(\langle o, S, 35\rangle, \langle g, S,
12\rangle, \langle k, S, 10\rangle, \langle n, S, 5\rangle \big)$. 
\\

\Treestyle{%
%  \addsep{1pt}%
  \minsep{3pt}%
  \vdist{30pt}%
  \nodesize{13pt}%
 }    % smaller than default 

\begin{figure}
{\small
 \begin{Tree}
   \treesymbol{\lvls{3}\cntr{$b$}\lft{$f^+ = 12$}\bnth{skipped}}

   \treesymbol{\lvls{1}\cntr{$j$}\bnth{skipped}\lft{$f^+ = 10$}}
   \node{\external\type{square}\cntr{$m$}\bnth{$36$}}
   \node{\external\type{square}\cntr{$o$}\bnth{$35$}}
   \node{\cntr{$l$}\rght{$f^- = 35$}}
   \node{\type{square}\cntr{$i$}\lft{$f^- = 35$}} 
   \node{\external\type{square}\cntr{$s$}\bnth{$50$}}
   \node{\external\type{square}\cntr{$t$}\bnth{$36$}}
   \node{\cntr{$q$}\rght{$f^- = 36$}}
   \node{\external\cntr{$r$}}
   \node{\type{square}\cntr{$p$}\rght{$f^- = 36$}}
   \node{\cntr{$h$}\rght{$f^- = 35$}}

   \node{\type{square}\cntr{$a$}\rght{$f^- = 35$}}
\end{Tree}
\hspace{5.2cm}\usebox{\TeXTree}
}
\caption{{\SSS} Pass 4}\label{fig:pass4}
\end{figure}

\Treestyle{%
  \addsep{2pt}%
  \minsep{10pt}%
  \vdist{30pt}%
  \nodesize{14pt}%
 }    % smaller than default 

\begin{figure}
%{\scriptsize
 {\footnotesize
\begin{tabular}{r|cclclcl}
\# & $n$ & $\Gamma$  & OPEN list after $\Gamma$ & $f^-_{top}$ & $T^-_{top}$ & $f^+_{root}$   & $T^+_{root}$ \\ \hline
18 & $o$ & 3 & $\big(\langle l, S, 35\rangle,
                     \langle g, S, 12\rangle,
                     $ & 35 & $l,m,o$ 
 & 35 & $a,b,c,d,e,n,f,g,$\\
   &     &    &  $\langle k, S, 10\rangle,
                     \langle n, S, 5\rangle\big)   $                              &    & & & $h,i,j,k,l,m,o$     \\
19 & $l$ & 1 & $\big(\langle i, S, 35\rangle,
                     \langle g, S, 12\rangle,
                     $ & 35 & $i,l,m,o$ 
 & 35 & $a,b,c,d,e,n,f,g,$\\
   &     &    &  $ \langle n, S, 5\rangle\big)$                                &    & & & $h,i,j,k,l,m,o$     \\
20 & $i$ & 2 & $\big(\langle p, L, 35\rangle,
                     \langle g, S, 12\rangle,
                     $ & $-\infty$ &  
 & 35 & $a,b,c,d,e,n,f,g,$\\
   &     &    &    $  \langle n, S, 5\rangle\big)$                             &    & & & $h,i,j,k,l,m,o,p$     \\
21 & $p$ & 6 & $\big(\langle q, L, 35\rangle,
                     \langle r, L, 35\rangle,
                     $ & $-\infty$ &  
 & 35 & $a,b,c,d,e,n,f,g,$\\
   &     &    &     $  \langle g, S, 12\rangle,
                     \langle n, S, 5\rangle\big)$                            &    & & & $h,i,j,k,l,m,o,p,q,r$     \\
22 & $q$ & 5 & $\big(\langle s, L, 35\rangle,
                     \langle r, L, 35\rangle,
                     $ & $-\infty$ &  
 & 35 & $a,b,c,d,e,n,f,g,$\\
   &     &    &     $  \langle g, S, 12\rangle,
                     \langle n, S, 5\rangle\big)$                            &    & & & $h,i,j,k,l,m,o,p,q,r,s$     \\
23 & $s$ & 4 & $\big(\langle s, S, 35\rangle,
                     \langle r, L, 35\rangle,
                     $ & 50 & $s$ 
 & 35 & $a,b,c,d,e,n,f,g,$\\
   &     &    &      $\langle g, S, 12\rangle,
                     \langle n, S, 5\rangle\big)$                             &    & & & $h,i,j,k,l,m,o,p,q,r,s$     \\
24 & $s$ & 2 & $\big(\langle t, L, 35\rangle,
                     \langle r, L, 35\rangle,
                    $ & $-\infty$ &
 & 35 & $a,b,c,d,e,n,f,g,$\\
   &     &    &      $\langle g, S, 12\rangle,
                     \langle n, S, 5\rangle\big)$                              &    & & & $h,i,j,k,l,m,o,p,q,r,s,t$     \\
25 & $t$ & 4 & $\big(\langle t, S, 35\rangle,
                     \langle r, L, 35\rangle,
                   $ & 36 & $t$ 
 & 35 & $a,b,c,d,e,n,f,g,$\\
   &     &    &        $  \langle g, S, 12\rangle,
                     \langle n, S, 5\rangle\big)$                           &    & & & $h,i,j,k,l,m,o,p,q,r,s,t$     \\
26 & $t$ & 3 & $\big(\langle q, S, 35\rangle,
                     \langle r, L, 35\rangle,
                     $ & 36 & $q,s,t$ 
 & 35 & $a,b,c,d,e,n,f,g,$\\
   &     &    &      $\langle g, S, 12\rangle,
                     \langle n, S, 5\rangle\big)$                             &    & & & $h,i,j,k,l,m,o,p,q,r,s,t$     \\
27 & $q$ & 1 & $\big(\langle p, S, 35\rangle,
                     \langle g, S, 12\rangle,
                     $ & 35 & $p,q,s,t$ 
 & 35 & $a,b,c,d,e,n,f,g,$\\
   &     &    &     $\langle n, S, 5\rangle\big)$                              &    & & & $h,i,j,k,l,m,o,p,q,r,s,t$     \\
28 & $p$ & 3 & $\big(\langle h, S, 35\rangle,
                     \langle g, S, 12\rangle,
                     $ & 35 & $h,i,l,m,$ &
35 &$a,b,c,d,e,n,f,g,$\\
   &     &    &     $\langle n, S, 5\rangle\big)$                              &    & $o,p,q,s,t$ 
 &  & $h,i,j,k,l,m,o,p,q,r,s,t$     \\
29 & $h$ & 1 & $\big(\langle a, S, 35\rangle\big)$ & 35 & $a,h,i,l,m,$
& 35 &$a,b,c,d,e,n,f,g,$\\
   &      &   &                                   &    & $o,p,q,s,t$
 &  & $h,i,j,k,l,m,o,p,q,r,s,t$     \\
\end{tabular}
\caption{{\SSS} Table Pass 4}\label{fig:ssstab4}
}
\end{figure}

\noindent {\em Fourth Pass:} (see figures~\ref{fig:pass4} and \ref{fig:ssstab4})\\
The previous search lowered the bound from 36 to 35.
In the fourth pass the first entry has
no immediate brother. It matches a $\Gamma$ case that is used to backtrack,
case 3, which replaces node $o$ by its parent $l$. Case 3 is always
followed by case 1, which replaces $l$  by its parent $i$ and, in
addition, deletes all child-entries from the list---only node $k$ in
this case. Each time
case 1 applies, all children of a min node and its max parent have
been expanded and the 
search of the subtree has been completed. 
%See also the explanation of case 1 on page~\pageref{sec:caseone}.
% In this example the minimax
% value is known, but in general an upper bound has been found. 
To avoid
having old nodes interfere with the 
remainder of the search, they must be removed from the OPEN list.
%These nodes {\em can\/} be deleted, since all children of the min nodes on the
%principal variation have been expanded. The value of $i$ has been
%determined. 35 is not only an upper bound, but also a lower bound (the
%value of a lower bound is determined by a {\em min\/} solution tree,
%where all descendants of min nodes are included). Both a max and a min
%solution tree have been constructed, a minimal tree in Knuth \&
%Moore's terminology. In other words, node $i$'s minimax value is known
%to be 35.
%The nodes {\em must\/} be deleted since their relatively high value
%could otherwise interfere with the attempts to lower the upper bound
%in other subtrees.
The list now contains: $\big(\langle i, S, 35\rangle, \langle g, S,
12\rangle, \langle n, S, 5\rangle \big)$. Next, case 2 matches entry
$i$, and expansion of the brother of $i$ commences. Node $p$ is
inserted into the list with state {\em live}. It matches case 6, which
inserts $q$ and $r$ into the list. Node $q$ matches case 5, which
inserts its left-most child $s$, still with $\hat{h}$ value 35. This leaf is
then evaluated by case 4. The evaluation of 50 is not less than the
sharpest upper bound of 35, so $\hat{h}$ is not changed. The OPEN
list is now: $\big(\langle s, S, 35\rangle, \langle r, L, 35\rangle,
\langle g, S, 12\rangle, \langle n, S, 5\rangle \big)$. Node $s$ is a
max node with a brother. It matches case 2, which replaces $s$ by its
brother $t$. Node $t$ is evaluated to value 36 by case 4, which again
does not lower the sharpest upper bound of 35. The OPEN
list is now: $\big(\langle t, S, 35\rangle, \langle r, L, 35\rangle,
\langle g, S, 12\rangle, \langle n, S, 5\rangle \big)$.
Node $t$ matches case 3, which is followed by case 1, inserting $p$
and purging the OPEN list of entry $r$. The list is now: $\big(\langle p, S,
35\rangle, \langle g, S, 12\rangle, \langle n, S, 5\rangle \big)$.
Since max node $p$ has no brothers, case 3 applies, which is followed
by case 1. Case 1 inserts the root $a$ into the list, and purges the
list of all the children of $a$. The list now becomes the single {\em 
solved\/}
entry $\langle a, S,35\rangle$, which satisfies  the termination
condition of {\SSS}. The minimax value is~35.

% \cleardoublepage
% \chapter{{\ABSSS} Example}
\section{{\ABSSS} Example}
\label{app:absssex}
% \markboth{Appendix C}{Appendix C}
% \markboth{Appendix C}{{\ABSSS} Example}
% \markboth{Appendix A}{{\ABSSS} Example}

% SSS* is practical -- our reformulation -- bf/df same mem
% \input ai-sec2
\label{sec:newsss}
%{\bf The idea is to use the example to introduce how {\ABSSS} works.
%  The example can also be used to show what a max solution tree is,
%  and how {\AB} constructs one.}
%
%In 1979 Stockman introduced a best first minimax algorithm,
%called {\SSS} \cite{Stoc79}. 
%{\SSS} is  difficult to understand.
% 
% Here an example will be used to show how {\ABSSS} traverses a
% tree. 

Now we will use the tree from the previous examples to
show how the reformulation of {\SSS} works.

\index{MT-SSS* example}%
{\SSS} finds $f_{root}$ by determining a sequence of upper bounds on it. The idea
behind {\ABSSS} is that these upper bounds can also be found using a
null-window {\AB} call. The null-window call creates  the
solution tree. This solution tree is stored in memory, so that it can
be refined in later passes. It turns out that in this way {\AB} will
expand the 
same solution trees as {\SSS}. We will show how this  works in detail
using an example. % in section~\ref{sec:absssex}.
Comparing it to the {\SSS} example illustrates that both
formulations expand the same trees.
(In appendix~\ref{app:equiv} the equivalence of the two
formulations is discussed in some detail.)

{\AB} is used to construct solution trees.
The
postcondition of the {\AB} procedure in section~\ref{sec:pcab} suggests that 
using outcome 2, we can have it return an upper
bound if we make it fail low. To create a fail low, {\AB} must be called
with a search window greater than any
possible leaf node value. 
{\AB}, when called
with such a window, will find the same upper
bound, and expand the same max solution tree, as {\SSS}. This can be
seen intuitively because both {\AB} and {\SSS}
expand the children of a node in a left-to-right order. 

%\subsection{{\ABSSS} Example}
\label{sec:absssexample}\label{sec:absssex}
The example tree in
figure~\ref{fig:abtree} 
is searched to determine its minimax value. 
\index{SSS*!passes}%
A number of stages, or passes, can be distinguished in the traversal of
this tree. At the end of each pass a full max solution tree exists,
which determines a better upper bound on the minimax value.
Also, solved nodes
represent min solution trees.  These two solution trees are shown in the tables.
Note that in the figures the
nodes are still numbered $a$ to $t$ in the order in which {\SSS} visits
them first, not {\ABSSS}.\\ 

\noindent {\em First pass:\/} (see figures~\ref{fig:abssspass1} and \ref{fig:abssstab1})\\
In the first pass the left-most max solution tree is expanded to
create the first non-trivial upper bound on the minimax value of the
root. 
\begin{figure}
{\small
 \begin{Tree}
   \node{\external\type{square}\cntr{$e$}\bnth{$41$}}
   \node{\leftonly\cntr{$d$}\lft{$f^+ = 41$}}
   \node{\external\type{square}\cntr{$g$}\bnth{$12$}}
   \node{\leftonly\cntr{$f$}\rght{$f^+ = 12$}}
   \node{\type{square}\cntr{$c$}\lft{$f^+ = 41$}}
   \node{\leftonly\cntr{$b$}\lft{$f^+ = 41$}}

   \node{\external\type{square}\cntr{$k$}\bnth{$10$}}
   \node{\leftonly\cntr{$j$}\lft{$f^+ = 10$}}
   \node{\external\type{square}\cntr{$m$}\bnth{$36$}}
   \node{\leftonly\cntr{$l$}\rght{$f^+ = 36$}}
   \node{\type{square}\cntr{$i$}\rght{$f^+ = 36$}} 
   \node{\leftonly\cntr{$h$}\rght{$f^+ = 36$}}

   \node{\type{square}\cntr{$a$}\lft{$f^+ = 41$}}
\end{Tree}
\hspace{5.5cm}\usebox{\TeXTree}
}
\caption{{\ABSSS} Pass 1}\label{fig:abssspass1}
\end{figure}
\begin{figure}
%{\scriptsize
 {\footnotesize
\begin{tabular}{r|ccccclcl}
\# & $n$ & $\gamma_n$  & $g_n$ & cutoff? & $f^-_n$ & $T^-_n$         & $f^+_n$   & $T^+_n$ \\ \hline
1  & $a$ &$+\infty$&$-\infty$&           &$-\infty$ &               &$+\infty$&       \\
2  & $b$ &$+\infty$&$+\infty$&           &$-\infty$ &               &$+\infty$&       \\
3  & $c$ &$+\infty$&$-\infty$&           &$-\infty$ &               &$+\infty$&       \\
4  & $d$ &$+\infty$&$+\infty$&           &$-\infty$ &               &$+\infty$&       \\
5  & $e$ &$+\infty$&     41  &           &$-\infty$       & $e$           &  41     &   $e$ \\
6  &$d$&$+\infty$&41&$41\not\geq +\infty$&$-\infty$ &               & 41      &   $d,e$ \\
7  &$c$&$+\infty$&41&$41 < +\infty$   &$-\infty$ &               &$+\infty$       &    \\
8  &$f$&$+\infty$&$+\infty$&   &$-\infty$ &               &$+\infty$       &    \\
9  &$g$&$+\infty$&12&   &$-\infty$ & $g$              &12    &  $g$  \\
10 &$f$&$+\infty$&12&$12\not\geq +\infty$&$-\infty$ &               &12    &  $f,g$  \\
11 &$c$&$+\infty$&41&$\bot$  &$-\infty$ &               &41       & $c,d,e,f,g$   \\
12 &$b$&$+\infty$&41&$41\not\geq +\infty$  &$-\infty$ &               &41       & $b,c,d,e,f,g$   \\
13 &$a$&$+\infty$&41&$41 < +\infty$  &$-\infty$ &               & $+\infty$     &    \\
14 & $h$ &$+\infty$&$+\infty$&           &$-\infty$ &               &$+\infty$&       \\
15 & $i$ &$+\infty$&$-\infty$&           &$-\infty$ &               &$+\infty$&       \\
16 & $j$ &$+\infty$&$+\infty$&           &$-\infty$ &               &$+\infty$&       \\
17 & $k$ &$+\infty$&     10  &           & $-\infty$       & $k$           &  10     &   $k$ \\
18 &$j$&$+\infty$&10&$10\not\geq +\infty$&$-\infty$ &               & 10      &   $j,k$ \\
19 &$i$&$+\infty$&10&$10 < +\infty$   &$-\infty$ &               &$+\infty$       &    \\
20 &$l$&$+\infty$&$+\infty$&   &$-\infty$ &               &$+\infty$       &    \\
21 &$m$&$+\infty$&36&   &$-\infty$ & $m$              &36    &  $m$  \\
22 &$l$&$+\infty$&36&$36\not\geq +\infty$&$-\infty$ &               &36    &  $l,m$  \\
23 &$i$&$+\infty$&36&$\bot$  &$-\infty$ &               &36       & $i,j,k,l,m$   \\
24 &$h$&$+\infty$&36&$36\not\geq +\infty$  &$-\infty$ &               &36       & $h,i,j,k,l,m$   \\
25 &$a$&$+\infty$&41& $\bot$  &$-\infty$ &               & 41
& $a,b,c,d,e,f,g,h,i,j,k,l,m$ 
\end{tabular}
\caption{{\ABSSS} Table Pass 1}\label{fig:abssstab1}
}
\end{figure}{\SSS} builds the max solution 
tree shown in figure~\ref{fig:abssspass1}, using cases 4, 5, and 6 of the $\Gamma$
operator. 
Instead of using $\Gamma$ cases 4, 5 and 6 and a sorted OPEN list,
{\ABSSS} uses {\MT} to compute the bound, by traversing solution trees.
At unexpanded nodes, $f^-$ is $-\infty$, and $f^+$ is
$+\infty$. 
A call {\MT}$(G, \infty)$ will cause an alpha cutoff at all
min nodes, since  all internal calls return values $g < \gamma =
\infty$. No beta cutoffs at max nodes will occur, since all $g <
\gamma$.  
We see that the call {\MT}$(a, \infty)$ on the tree in
figure~\ref{fig:abtree} will traverse the tree in
figure~\ref{fig:abssspass1}, conforming to {\AB}'s postcondition.
Due to the ``store'' operation in
figure~\ref{fig:mmab}, this tree is saved in memory so that its
backed-up values can be used in a later pass. 
The max solution tree stored at the end of this pass consists of the 
nodes $a,b,c,d,e,f,g,h,i,j,k,l$ and $m$, yielding an upper bound of 41.
% For Stockman's formulation,
% the leaves of this tree are
% stored in the OPEN list, which is 
% $\big(\langle e, S, 41\rangle, \langle m, S,
% 36\rangle,  \langle g, S, 12\rangle, \langle k, S, 10\rangle \big)$.
% Note that the entry at the head of the list is also 41, {\SSS}'s upper 
% bound on the minimax value.
\\

\begin{figure}
{\small
 \begin{Tree}
   \node{\external\type{square}\cntr{$e$}\bnth{$41$}}
   \node{\external\type{square}\cntr{$n$}\bnth{$5$}}
   \node{\cntr{$d$}\lft{$f^+ = 5$}}
   \treesymbol{\lvls{1}\cntr{$f$}\bnth{skipped}\rght{$f^+ = 12$}}
   \node{\type{square}\cntr{$c$}\lft{$f^+ = 12$}}
   \node{\leftonly\cntr{$b$}\lft{$f^+ = 12$}}

   \treesymbol{\lvls{3}\cntr{$h$}\lft{$f^+ = 36$}\bnth{skipped}}

   \node{\type{square}\cntr{$a$}\lft{$f^+ = 36$}}
\end{Tree}
\hspace{5.5cm}\usebox{\TeXTree}
}
\caption{{\ABSSS} Pass 2}\label{fig:abssspass2}
\end{figure}
\begin{figure}
 {\footnotesize
%{\scriptsize
\begin{tabular}{r|ccccclcl}
\# & $n$ & $\gamma_n$  & $g_n$ & cutoff? & $f^-_n$ & $T^-_n$         & $f^+_n$   & $T^+_n$ \\ \hline
26  & $a$ &41&$-\infty$&           &$-\infty$ &               &41&  $a,b,c,d,e,f,g,h,i,j,k,l,m$      \\
27  & $b$ &41&$+\infty$&           &$-\infty$ &               &41&  $b,c,d,e,f,g$      \\
28  & $c$ &41&$-\infty$&           &$-\infty$ &               &41&  $c,d,e,f,g$     \\
29  & $d$ &41&$+\infty$&           &$-\infty$ &               &41&  $d,e$     \\
30  & $e$ &41&     41  & $41 \neq +\infty$   & 41     & $e$           &  $+\infty$    &   $e$ \\
31  & $n$ &41&     5  &           &$-\infty$        & $n$           &  5     &   $n$ \\
32  &$d$&41&5&$\bot$&$-\infty$ &  $d,e,n$             & 5      &   $d,n$ \\
33  &$c$&41&5&$5 < 41 $   &$-\infty$ &               &$+\infty$       &    \\
34  &$f$&41&12&$12 \not\geq 41$   &$-\infty$ &               & 12       &  $f,g$  \\
35 &$c$&41&12&$\bot$  &$-\infty$ &    $c,d,e,n$           &12       & $c,d,n,f,g$   \\
36 &$b$&41&12&$12\not\geq 41$  &$-\infty$ &               &12       & $b,c,d,n,f,g$   \\
37 &$a$&41&12&$12 < 41$  &$-\infty$ &               & 12    &  $a,b,c,d,n,f,g,h,i,j,k,l,m$   \\
38 & $h$ &41&36& $36 \not\geq 41$  &$-\infty$ &               &36& $h,i,j,k,l,m$      \\
39 &$a$&41&36&$\bot$  &$-\infty$ &               & 36    &  $a,b,c,d,n,f,g,h,i,j,k,l,m$   
\end{tabular}
\caption{{\ABSSS} Table Pass 2}\label{fig:abssstab2}
}
\end{figure}
\noindent {\em Second pass:\/} (see figures~\ref{fig:abssspass2} and
\ref{fig:abssstab2})\\ 
This pass lowers the upper bound on $f$ from 41 to 36.
How can we use {\AB} to do this? Since
the max solution tree defining the upper bound of 41 has been stored
by the previous {\MT} call, {\AB} can re-traverse the nodes on the
principal variation $(a, b, c, d, e)$ to find the critical leaf $e$,
and see whether expanding its brother will yield a search tree with a
lower minimax value. Finding this critical leaf, and selecting its
brother for expansion is the essence of the ``best-first'' behavior of
{\SSS} (and {\ABSSS}). \label{sec:best} The critical leaf $e$ has a
min parent,  node $d$,
so expanding the brother can lower its value, which will, in turn,
 lower the minimax 
value at the root of the max solution tree. Since this value is the maximum 
of its leaves, there is no point in expanding brothers
of non-critical leaves,
because then node $e$ will keep the value of the root at $41$.
Thus, based on the information that the max solution tree provides,
there is only one node ($n$) whose expansion makes sense. Other nodes are
worse,  since they cannot change the bound at the
root.

To give {\AB} the task of returning a value lower than 
$f^+ = 41$, we give it a search window which will cause it to fail low. The
old window of $\langle \infty-1, \infty\rangle$ will not do, since the
code in figure~\ref{fig:mmab} will cause it
to return from both nodes $b$ and $h$, with a value of 41, 
lower than $+\infty$, but not the lower upper bound. A better choice
is the search window 
$\langle f^+-1, f^+\rangle$, or $\langle 40, 
41\rangle$, which
prompts {\MT} to 
descend the principal variation and return as soon as a lower $f^+$ on 
node  $a$
is found. It descends to nodes $b, c, d, e$ and continues to
search node $n$. It will back up value 5 to node $d$ and cause a cutoff.
The value of $d$ is no longer determined by $e$
but by $n$. Node $e$ is no
longer part of 
the max solution tree that determines the sharpest upper
bound. It has been proven that $e$ can be erased from memory as long
as we remember 
that $n$ is the new best child (not shown in the {\MT} code). 
The value 5 is backed up to $c$. No beta cutoff
occurs at $c$, so $f$'s bound is retrieved. Since $f^+ \not\geq \gamma$ at
node $f$, {\MT} does not enter it, but uses $c.f^+ = 12$ for $g{\prime}$. 
12 is backed up to $b$, where it
causes an alpha cutoff. Next, 12 is backed up to $a$. Since $g <
\gamma$, node $h$ is probed, but since $c.f^+ \not\geq \gamma$ ($36
\not\geq 41$) it is not entered.
The call {\MT}$(a, 41)$ fails low with value 36, the sharper
upper bound. The max solution tree defining this bound consists of
nodes $a, b, c, d, n, f, g, h, i, j, k, l$ and $m$ (that is, node $e$ has been
replaced with $n$). 

By storing previously expanded nodes in memory, and calling {\MT} with
the right search window, we can make it traverse the principal
variation, and expand brothers of the critical leaf, to get a better
upper bound on the minimax value of the root, in exactly the same way
as {\SSS} does. 
\\

\begin{figure}
{\small
 \begin{Tree}
   \treesymbol{\lvls{3}\cntr{$b$}\lft{$f^+ = 12$}\bnth{skipped}}

   \treesymbol{\lvls{1}\cntr{$j$}\bnth{skipped}\lft{$f^+ = 10$}}
   \node{\external\type{square}\cntr{$m$}\bnth{$36$}}
   \node{\external\type{square}\cntr{$o$}\bnth{$35$}}
   \node{\cntr{$l$}\rght{$f^+ = 35$}}
   \node{\type{square}\cntr{$i$}\rght{$f^+ = 35$}} 
   \node{\leftonly\cntr{$h$}\rght{$f^+ = 35$}}

   \node{\type{square}\cntr{$a$}\lft{$f^+ = 35$}}
\end{Tree}
\hspace{5.5cm}\usebox{\TeXTree}
}
\caption{{\ABSSS} Pass 3}\label{fig:abssspass3}
\end{figure}
\begin{figure}
%{\scriptsize
 {\footnotesize
\begin{tabular}{r|ccccclcl}
\# & $n$ & $\gamma_n$  & $g_n$ & cutoff? & $f^-_n$ & $T^-_n$         & $f^+_n$   & $T^+_n$ \\ \hline
40  & $a$ &36&$-\infty$&           &$-\infty$ &               &36&  $a,b,c,d,n,f,g,h,i,j,k,l,m$      \\
41  & $b$ &36&12&$12 \not\geq 36$&$-\infty$ &               &12&  $b,c,d,n,f,g$      \\
42  & $h$ &36&$+\infty$&           &$-\infty$ &  &36&$h,i,j,k,l,m$     \\
43 & $i$ &36&$-\infty$&           &$-\infty$ &               &36&$i,j,k,l,m$      \\
44 & $j$ &36&10&$10 \not\geq 36$  &$-\infty$ &               &10&$j,k$\\
45 &$l$&36&$+\infty$&   &$-\infty$ &               &36 &  $l,m$  \\
46 &$m$&36&36& $36 \neq +\infty$  & 36 & $m$              & $+\infty$    &  $m$  \\
47 &$o$&36&35&      &$-\infty$ &   $o$            &35    &  $o$  \\
48 &$l$&36&35&$\bot$&$-\infty$ &  $l,m,o$             &35    &  $l,o$  \\
49 &$i$&36&35&$\bot$  &$-\infty$ & $i,l,m,o$              &35       &$i,j,k,l,o$   \\
50 &$h$&36&35&$35\not\geq 36$  &$-\infty$ &               &35       & $h,i,j,k,l,o$   \\
51 &$a$&36&35&$\bot$  &$-\infty$ &               & 35& $a,b,c,d,n,f,g,h,i,j,k,l,o$ 
\end{tabular}
\caption{{\ABSSS} Table Pass 3}\label{fig:abssstab3}
}
\end{figure}
\noindent {\em Third Pass:} (see figures~\ref{fig:abssspass3} and \ref{fig:abssstab3})\\
In the previous pass, the upper bound was lowered from 41 to 36.
A call {\MT}$(a, 36)$ is performed.
From the previous search,
we know that $b$ has an $f^+ \not\geq 36$ so it is not entered; $h.f^+
\geq 36$, so it is.
The algorithm follows the principal variation to the node giving the 36
($h$ to $i$ to $l$ to $m$). The brother of $m$ is expanded.
The bound on the minimax value at the root has now been improved from 36 to 35.
The max solution tree defining this bound consists of
nodes $a, b, c, d, n, f, g, h, i, j, k, l$ and  $o$.\\

\Treestyle{%
%  \addsep{1pt}%
  \minsep{3pt}%
  \vdist{30pt}%
  \nodesize{13pt}%
 }    % smaller than default 

\begin{figure}
{\small
 \begin{Tree}
   \treesymbol{\lvls{3}\cntr{$b$}\lft{$f^+ = 12$}\bnth{skipped}}

   \treesymbol{\lvls{1}\cntr{$j$}\bnth{skipped}\lft{$f^+ = 10$}}
   \node{\external\type{square}\cntr{$m$}\bnth{$36$}}
   \node{\external\type{square}\cntr{$o$}\bnth{$35$}}
   \node{\cntr{$l$}\rght{$f^- = 35$}}
   \node{\type{square}\cntr{$i$}\lft{$f^- = 35$}} 
   \node{\external\type{square}\cntr{$s$}\bnth{$50$}}
   \node{\external\type{square}\cntr{$t$}\bnth{$36$}}
   \node{\cntr{$q$}\rght{$f^- = 36$}}
   \node{\external\cntr{$r$}}
   \node{\type{square}\cntr{$p$}\rght{$f^- = 36$}}
   \node{\cntr{$h$}\rght{$f^- = 35$}}

   \node{\type{square}\cntr{$a$}\rght{$f^- = 35$}}
\end{Tree}
\hspace{5.2cm}\usebox{\TeXTree}
}
\caption{{\ABSSS} Pass 4}\label{fig:abssspass4}
\end{figure}

\Treestyle{%
  \addsep{2pt}%
  \minsep{10pt}%
  \vdist{30pt}%
  \nodesize{14pt}%
 }    % smaller than default 

\begin{figure}
%{\scriptsize
 {\footnotesize
\begin{tabular}{r|ccccclcl}
\# & $n$ & $\gamma_n$  & $g_n$ & cutoff? & $f^-_n$ & $T^-_n$         & $f^+_n$   & $T^+_n$ \\ \hline
52  & $a$ &35&$-\infty$&           &$-\infty$ &               &35&
$a,b,c,d,n,f,             $      \\
    &     &  &         &           &          &               &  &  $
             g,h,i,j,k,l,o$      \\
53  & $b$ &35&12&$12 \not\geq 35$&$-\infty$ &               &12&  $b,c,d,e,f,g$      \\
54 & $h$ &35&$+\infty$&           &$-\infty$ &  &35&$h,i,j,k,l,o$     \\
55 & $i$ &35&$-\infty$&  &$-\infty$ &    $i,l,m,o$           &35&$i,j,k,l,o$      \\
56 & $j$ &35&34&$34 \not\geq 35$  &$-\infty$ &               &34&$j,k$\\
57 &$l$&35&$+\infty$&   &$-\infty$ &               &36 &  $l,m$  \\
58 &$m$&35&36& $36 \not < 35$  & 36 & $m$              & $+\infty$    &  $m$  \\
59 &$o$&35&35& $35 \neq +\infty$     & 35&   $o$            & $+\infty$    &  $o$  \\
60 &$l$&35&35&$\bot$& 35 &  $l,m,o$             & $+\infty$ &  $l,o$  \\
61 &$i$&35&35&$\bot$  & 35 & $i,l,m,o$              & $+\infty$       &$i,j,k,l,o$   \\
62 &$p$&35& $-\infty$&   &$-\infty$ &               & $+\infty$       &   \\
63 &$q$&35& $+\infty$&   &$-\infty$ &               & $+\infty$       &   \\
64 &$s$&35& 50&   & 50 &   $s$            & $+\infty$       &  $s$ \\
65 &$t$&35& 36&   & 36 &     $t$          & $+\infty$       & $t$  \\
66 &$q$&35& 36& $\bot$  & 36 &  $q,s,t$        & $+\infty$       &  $q,t$ \\
67 &$p$&35& 36& $36 \not < 35$  & 36 &   $p,q,s,t$            & $+\infty$       &   \\
68 &$h$&35&35&$\bot$  & 35 & $h,i,l,m,o,p,q,s,t$  & $+\infty$  & $h,i,j,k,l,o$   \\
69 &$a$&35&35&$\bot$  & 35 & $a,h,i,l,m,         $ & $+\infty$&
$a,b,c,d,n,f,             $ \\
   &   &  &  &        &    & $          o,p,q,s,t$ &          & $         g,h,i,j,k,l,o$ 
\end{tabular}
\caption{{\ABSSS} Table Pass 4}\label{fig:abssstab4}
}
\end{figure}

\noindent {\em Fourth Pass:} (see figures~\ref{fig:abssspass4} and \ref{fig:abssstab4})\\
This is the last pass of {\ABSSS}. We will find that the
upper bound cannot be 
lowered. 
A call with window $\langle f^+-1, f^+\rangle$, or
{\MT}$(a, 35)$, is performed. In this 
pass we will not find a fail low as usual, but a fail high with return
value 35. The return value is now a lower bound, backed-up by a min
solution tree (all children of a min node included, only one for each
max node).

How does {\AB} traverse this min solution tree?
The search follows the critical path $a, h, i, l$ and $o$.
At node $l$,
child $m$ has $c.f^- \not< \gamma$, so it is not
entered. Now $o$ is entered but not evaluated since it is no longer open.
Node $o$ returns the stored bound $o.f^+ = 35$.
The value of the children is retrieved from storage.
Note that the previous pass stored an $f^+$ value for $l$ and $o$,
while this pass  stores an $f^-$.
% The value of $l$ does not change, $j$'s bound of 10 precludes it from
% being searched, so $i$'s value remains unchanged.
Node $i$ cannot lower $h$'s value ($g \geq \gamma, 35 \geq 35$,
 no cutoff occurs),
so the search explores $p$.
Node $p$ expands $q$ which, in turn,  searches $s$ and $t$.
Since $p$ is a maximizing node, the value of $q$ (36) causes a cutoff:
$g \not< \gamma$, node $r$ is not searched. 
Both of $h$'s children are $\geq 35$. Node $h$
returns 35, and so does $a$.
Node $a$ was searched attempting to show whether its value
was $<$ or $\geq$  $35$.
Node $h$ provides the answer: greater than or equal.
This call to {\MT} fails high, meaning we have a
lower bound of 35 on the search.
The previous call to {\MT} established an upper bound of 35.
Thus the minimax value of the tree is proven to be 35.

We see that nothing special is needed to have {\AB} traverse the min
solution tree $a, h, i, l, m, o, p, q, s$ and $t$. The ordinary cutoff
decisions cause its traversal, when $\alpha = f^+(a)-1$ and $\beta =
f^+(a)$.\\

\noindent
In the previous four passes we called {\MT} ({\AB}) with a special search
window to have it emulate {\SSS}. This sequence of calls, creating a 
sequence of fail lows until the final fail high, can be captured in a
single loop, given by the 
pseudo code of figure~\ref{fig:sss2}.

\cleardoublepage
\chapter{Equivalence of {\SSS} and {\ABSSS}}
\label{app:equiv}
% \markboth{Appendix D}{Appendix D}
% \markboth{Appendix B}{Equivalence of {\SSS} and {\ABSSS}}

This appendix discusses the ideas behind the equivalence of Stockman's
{\SSS} and the new 
reformulation {\ABSSS}.  We do not present a
formal proof in this appendix. That can be found in \cite{Pijl95}. An
outline of the full proof can be found in appendix A of
\cite{Plaa95b,Plaa95e}. The aim of the current 
appendix is to convey the idea behind the equivalence. It can be
regarded as an ``informal proof,'' or as an illustration supporting the formal
proof, which requires a certain amount of familiarity with the old
formulation of {\SSS}.
\index{SSS*!equivalence}\index{MT-SSS* equivalence}\index{MT with list-ops}\index{SSS*!list operations}

\begin{figure}
{\small%\footnotesize
  \begin{center}
    \leavevmode
    \begin{tabbing}
m\=mm\=mm\=mm\=mm\=mmmmm\=mmmmmmmmmmmmmmmmmmmmmmmm\=  \kill
/* For equivalence with SSS* this function must be called by MT-SSS*\>\>\>\>\>\>\>*/\\
/* (see figure~\ref{fig:sss2app}) \>\>\>\>\>\>\>*/\\ 
/* MT: storage enhanced null-window {\AB}$(n, \gamma-1, \gamma)$.\>\>\>\>\>\>\>*/\\
/* $n$ is the node to be searched, $\gamma-1$ is $\alpha$ ,
$\gamma$ is $\beta$ in the call.\>\>\>\>\>\>\>*/\\
%/* 'Store' saves search bound information in memory;\>\>\>\>\>\>\>*/\\
%/* 'Retrieve' accesses this 
%information.\>\>\>\>\>\>\>*/\\
{\bf function} MT$(n, \gamma) \rightarrow g$; \\

\> {\bf if} $n$ = leaf {\bf then } \\
\>  \> retrieve $n.f^-,n.f^+$; /* non-existing bounds are $\pm\infty$ */\\
\>  \> {\bf if} $n.f^- = -\infty$ {\bf and} $n.f^+ = +\infty$ {\bf then} \\
\>   \> \> {\bf \{* } List-op$(4, n)$; {\bf *\} }\\ 
\>   \> \> $g :=$ eval$(n)$;\\
\>   \> {\bf else if} $n.f^+ = +\infty$ {\bf then} $g := n.f^-$; {\bf else}
                                                 $g := n.f^+$;\\
 
\> {\bf else if} $n$ = max {\bf then}\\
\>   \> {\bf \{* } retrieve $n.f^-,n.f^+$; {\bf if} $n.f^+ = +\infty$
            {\bf and} $n.f^- = -\infty$ {\bf then}  List-op$(6,n)$;
            {\bf *\} }\\   
\>   \> $g := -\infty$;\\
\>   \> $c := $ firstchild$(n)$;\\
\>   \> /* $g \geq \gamma$ causes a beta cutoff ($\beta = \gamma$) */ \\
\>   \> {\bf while} $ g < \gamma$ {\bf and} $c \neq \bot$ {\bf do}\\
 
\>   \> \> retrieve $c.f^+$;\\
\>   \> \> {\bf if} $c.f^+ \geq \gamma $ {\bf then} \\
\>   \> \> \> $g{\prime}:=$ MT$(c, \gamma)$;\\ 
\>   \> \> \> {\bf \{* if} $g{\prime} \geq \gamma$ {\bf then}
               List-op$(1,c)$; {\bf *\} } \\ 
\>   \> \> {\bf else}  $g{\prime}:= c.f^+$;\\ 
\>   \> \> $g :=$ max$(g,g{\prime})$;\\
\>   \> \> $c := $ nextbrother$(c)$;\\
 
\> {\bf else} /* $n$ is a  min node */ \\ %{\bf then}\\
\>   \> {\bf \{* } retrieve $n.f^-,n.f^+$; {\bf if} $n.f^+ = +\infty$
            {\bf and} $n.f^- = -\infty$ {\bf then}  List-op$(5,n)$;
            {\bf *\} }\\   
\>   \> $g := +\infty$;\\
\>   \> $c := $ firstchild$(n)$;\\
\>   \> /* $g < \gamma $ causes an alpha cutoff ($\alpha = \gamma
            -1$) */ \\
\>   \> {\bf while} $ g \geq \gamma$ {\bf and} $c \neq \bot$ {\bf do}\\
\>   \> \> retrieve $ c.f^-$;\\
\>   \> \> {\bf if} $c.f^- < \gamma$ {\bf then} \\
\>   \> \> \> $g{\prime}:=$ MT$(c, \gamma)$; \\
\>   \> \> \> {\bf \{* if} $g{\prime} \geq \gamma$ {\bf then}\\
\>   \> \> \> \> {\bf if} $c <$ lastchild$(n)$ {\bf then} List-op$(2,c)$; 
               {\bf else} List-op$(3,c)$; {\bf *\} }  \\ 
\>   \> \> {\bf else} $g{\prime} := c.f^-$;\\
\>   \> \> $g :=\min(g,g{\prime})$; \\
\>   \> \> $c := $ nextbrother$(c)$;\\
\> /* Store one bound per node. Delete possible old bound (see
                page~\pageref{sec:boundsremark}). */ \\ 
\> {\bf if} $g \geq \gamma$ \={\bf then} $n.f^- := g$; store $n.f^-$;\\
\>                          \>{\bf else} $n.f^+ := g$; store $n.f^+$;\\

\> {\bf return} $g$;
\end{tabbing}
  \end{center}
}
  \caption{{\MT} with {\SSS}'s List-Ops}
  \label{fig:mmabapp}\label{fig:mtcode}
\end{figure}

\begin{figure}
{\small
\begin{tabbing}
mmmmmmmmmmm\=mm\=mm\=mmmmmmmmmmmmmmmmmmmmm\=mm\=mm\=  \kill
\>{\bf function} MT-SSS*$(n) \rightarrow f$; \\
\>   \>$g := +\infty$;\\
\>   \>{\bf repeat}                      \\
\>   \>   \>$\gamma := g$;        \\ 
\>   \>   \>$g := $ MT($n,  \gamma$);\\
\>   \>{\bf until} $g = \gamma$;          \\
\>   \>{\bf return} $g$;                 
\end{tabbing}}
  \caption{{\SSS} as a Sequence of {\MT} Searches} \label{fig:sss2app}
\end{figure}

\section{{\MT} and List-ops}
Figure~\ref{fig:mmabapp} shows an extended version of {\MT}, to be
called by {\ABSSS}, shown in figure~\ref{fig:sss2app}. 
The list operations (list-ops) between {\bf \{*}
and {\bf *\} } are 
inserted to show  the equivalence of {\ABSSS} and Stockman's
{\SSS}. The value of $\hat{h}$ is the current value of $\gamma$. (In
implementations of {\ABSSS} the list-ops should not be 
included.) The claim is that, when called by {\ABSSS}, the list operations  in
{\MT} cause the same $\Gamma$ operations to be applied as in 
Stockman's original 
formulation. These list-ops are the core of the equivalence
proof \cite{Pijl95}. Their place in the {\MT} code shows in a clear
and concise way 
where and how the {\SSS} operators fit in the way {\AB} traverses a tree.
The example in appendix~\ref{app:absssex} can be used to check this.
Without the list-ops, the 
call {\MT}$(n, \gamma)$ is an ordinary null-window
 {\AB}$(n, \gamma-1, \gamma)$ search, except that
{\MT} uses one bound, making the code a bit simpler.
% It traverses the same nodes as a null-window {\AB}
% call that uses storage.

%\subsection{Equivalence of {\ABSSS} and {\SSS}}
\label{sec:proofy}
{\ABSSS} and {\SSS} are equivalent in the sense that they evaluate the
same leaf nodes in the same order, 
when called on the same minimax tree. 
In the previous chapters and examples {\SSS} and {\ABSSS} have been
treated primarily as algorithms that manipulate max solution trees.
The original {\SSS} formulation
stresses the min-solution-tree view. 
The full story is, of course, that both max and min solution trees are
manipulated by {\SSS}, just as {\AB}'s postcondition in
section~\ref{sec:pcab} shows that {\AB} also constructs both types of
trees. 

The critical tree that proves the minimax value of a game tree is a
union of the max solution tree with the lowest possible
upper bound, and the min solution tree with the highest possible lower bound. 
{\SSS} can be regarded as an interleaving of two processes, one
working downward on the max solution tree part of the final minimal
tree, and one working upward on the min solution tree part.  First a
max solution tree is 
expanded downwards from the root.  Next, {\SSS}  constructs
a min solution tree growing upwards off of the critical leaf. The aim
is to create a min solution tree that reaches all the way up to the
root, and has the same value as 
the max solution tree, because that signals that the critical tree that
proves the minimax value has been constructed. 

However, often this process is ended prematurely, when {\SSS} finds
that the value of the min solution tree drops below that of another
node/min solution tree, which causes the  whole process to start over
again, giving {\SSS} its interleaved nature.

\section{{\MT} and the Six Operators}
Turning to figures~\ref{fig:sss2app}, \ref{fig:mtcode}, and
\ref{fig:sss}, the construction of solution trees by the six operators
will be discussed in a little more detail, focusing on the 
circumstances in which {\SSS} and {\MT} invoke the six $\Gamma$
operators. 

Many of the operations that {\SSS} performs have a minor, local, effect:
a max solution tree is being expanded, or {\SSS} backtracks to the node
where the next one will be grown. However, at some points operations
with a more major, global,  nature are performed. When
operator 4 has 
evaluated the last leaf of a max solution tree, and all the entries in
the OPEN list are {\em solved}, the highest $\hat{h}$
value drops. The selection of the next entry, through {\SSS}'s
sorting of the OPEN list, is a major decision point, where {\SSS}
selects a node which is in a certain sense globally ``best''
(figure~\ref{fig:best}).  

In {\ABSSS} the minor, local, operations are performed by the {\MT} code,
by having a null-window {\AB} search construct a solution tree. The
major decision to find the next node which is globally ``best''
occurs in {\ABSSS} when {\MT} is called at the root to traverse the
critical path to the critical leaf. In {\ABSSS}  the 
global decisions are more visible than in {\SSS}, since they coincide
with a new pass of the main loop.

A number of features of {\SSS} play an important role in this
discussion. At any time, the highest $\hat{h}$ value in the OPEN list
is an upper bound on the minimax value of the root of the game tree.
The entries are the leaves of a partial max solution tree.
When the OPEN list consists only of {\em solved\/}
entries, a pass of the main loop of {\ABSSS} has ended (assuming that
only one critical path exists). The entries in
the list are the leaves of a total max solution tree, that is rooted at the
root of the game tree.  

Concerning {\MT}, the following pre- and postcondition hold (following
{\AB}'s postcondition in section~\ref{sec:pcab}):
at a fail high, {\MT} has traversed a min solution tree, whose value
$f^-(n)$ is stored with node $n$. 
At a fail low, {\MT} has traversed a max solution tree, whose value
$f^+(n)$ is stored with node $n$.
Since {\MT} is called by {\ABSSS} with $\gamma = f^+(root)$, the
node on which {\MT} is called is either open, or part of a max
solution tree.

The {\SSS} notions {\em live\/} and {\em solved\/} correspond to the
{\AB} notions {\em open\/} (children not yet expanded) and {\em
  closed\/} (all children expanded).
The term {\em sub\/} max solution tree is used for a max solution tree whose
root is an interior node, not necessarily the root of the game
tree, in other words, a smaller, deeper, max solution tree. A sub max
solution tree may even be as small as a single leaf. The
same applies to the term sub min solution trees.

% Show that MT is equiv to SSS* by showing that the six list-ops in MT
% are called in the same circumstances as in SSS*, and put the same
% nodes in the list as SSS*

It will be shown that the list-ops in {\MT} cause the same operations
to be performed on the OPEN list as the original {\SSS} operators.  
One of the key points is that {\SSS} always selects
the entry with the highest $\hat{h}$ value. (Other preconditions of the
operators, such as being applied to an open node or to a max node, are
easily verified from the {\MT} code in figure~\ref{fig:mtcode}.)
Therefore, the focus of attention in the following discussion will be
on how {\MT} performs all the list-ops on the (left-most) node
with the highest $\hat{h}$ value.

%Note that the aim of this appendix is to convey a high level idea of
%how this works. The full, detailed, proof is to be found in
%\cite{Pijl95}. 
%
% Induction on the sequence of operators.
%
% In {\SSS} operator 6 is either performed on the root, or after operator
% 2 or 5.  If it is performed on the root, $f^+(root) = +\infty =
% \hat{h}$. Operator 2 and 5 do not change the value of $\hat{h}$, so
% assuming that it was the highest, it remains. Operator 5 selects the
% first child of a node, which Operator 2 selects the
% next brother, after a fail high of the inner {\MT} call. Since a fail
% high does not cause a cutoff at the min 
% node, {\MT} also selects this node. 
% 
% hte
% remains is performed when the inner {\MT} call failed
% high, 
% so the 6after If list-op 6 is performed, 

The key circumstances in which the six operators are invoked are
discussed from the viewpoint of {\SSS} and {\MT}. 

{\em {\SSS}:\/}
At the start of the algorithm, the OPEN list is empty. 
The entry with the highest $\hat{h}$ value is the root.  {\SSS}
now expands the left-most max solution tree using the operators 4, 5, and
6.\\
{\em {\MT}:\/}
{\MT} selects this entry, below which the left-most max solution
tree is expanded (by {\AB}'s postcondition),
applying list-op's 4, 5, and 6. Examination of the {\SSS} operators on the one
hand, and the {\MT} code on the other, shows easily that the list-op's
in {\MT} follow the same expansion sequence as the operators in
{\SSS}. 

{\em {\SSS}:\/}
At the start of another pass of {\ABSSS}, the OPEN list contains only
{\em solved\/} 
entries, that form the leaves of a total max solution tree. The
highest entry is an upper bound on the minimax value of the root, and
contains the critical leaf of the max solution tree. \\ 
{\em {\MT}:\/}
The conventional {\AB} cutoff decisions cause
{\MT} to traverse the left-most critical path of the max solution
tree, along which  $\gamma = f^+(n) = f^+(root)$ holds, until the
left-most critical leaf is reached.  
(In closed max nodes the left-most child with $c.f^+ = \gamma$ is
entered; the highest child among the children in the max solution
tree. In closed min nodes the left-most child with $c.f^- = -\infty 
< \gamma$ is entered; the only child in the max solution tree.) 
Thus, both {\SSS} and {\MT} select the critical leaf to be operated
upon.

Next, {\SSS} selects the node where a sub max solution tree will be
expanded, in order to try to lower the upper bound on the minimax
value of the root.\\
{\em {\SSS}:\/}
Depending on the parity of the depth of the tree, either operator 1, or
2/3 are performed on the critical leaf. (Turning to the {\MT} code,
the critical leaf's value is 
equal to $\gamma$, so the inner {\MT} call fails high with $g{\prime}
\geq \gamma$.) Assuming an even depth of the tree, operator 2 or 3 are
performed. These operators expand the ``best'' node: the node (or
rather, the sub max solution tree which happens to be only a single node) that can
influence the value at the root (see also figure~\ref{fig:best}). If
the critical leaf has an unexpanded brother, then operator 2 selects
it. \\
{\em {\MT}:\/}
In {\MT} the unexpanded brother is selected because the return value $g{\prime}
\geq \gamma$ does not cause a cutoff. (The {\ABSSS} code ensures that
$\gamma = f^+(root)$ is equal to the highest $\hat{h}$ value). \\
{\em {\SSS}:\/}
If all open brothers of the critical leaf have been expanded with
decreasing their $\hat{h}$ value, operator
3 applies, which performs half of the back-up operation to start
expanding the next smallest sub max solution tree. Operator 1 applies
next, which does the other half of the back-up motion, putting a max
node at the front of the list, to be replaced by operator 2 by
its next brother, starting the expansion of the sub max solution tree by
operator 4, 5 and 6 just as when the sub max solution tree were only a
single node, only it is bigger. (For odd depth trees, the part of the
back-up process starting with operator 1 also applies.)

{\em {\MT}:\/}
In {\MT} the back-up  motion of operator 3 occurs because there
are no more children, causing {\MT} to automatically back-up to its
parent. The back-up 
motion of operator 1 occurs because the return value of the inner {\MT}
call is still a fail high ($g{\prime} \geq \gamma$), which causes a
cutoff in the max parent because $g \not< \gamma$, causing {\MT}
to back-up. In the min node where {\MT} ends up, the return value is
still a fail high, which causes no cutoff, so the next child is
expanded, just like operator 2 does. (Of course, if no brother exists
for operator 2, operator 3 applies. And if operator 1 puts the root on the
list, which has no brothers, {\MT}/{\ABSSS} and {\SSS} both terminate.)
Thus {\MT} and {\SSS} also select the same nodes when the place to
expand the sub max solution tree is to be determined.

Expansion of the sub max solution tree succeeds if the value of
$\hat{h}$ is lowered, or in {\MT} terms, if the inner {\MT} calls fail
low. If it fails high, a min solution tree will be constructed. 
(In the {\MT} case this follows from the {\AB} postcondition.)

{\em {\SSS}:\/}
The question of failing high or low occurs when operator 4 is
performed. A ``fail high'' causes the $\hat{h}$ value of
the first entry to stay the same, so the backing-up
operators for solved entries 1, 2 and 3 are applied, in the same
fashion as described previously. A ``fail low'' causes another entry
to have the highest $\hat{h}$ value. This entry has been inserted by
operator 6, the only operator that inserts multiple entries into the
list. All of the entries inserted by operator 6 have the same $\hat{h}$
value. The 
only operator that can lower an $\hat{h}$ value is number 4. When this
happens, the next brother (or rather, (great) uncle) appears at the front. (No
entries from outside the sub max solution tree to be constructed can
enter, because their $\hat{h}$ value is lower.)  In
this way, as long as the expansions ``fail low'', a max solution tree
is expanded in the same way as the left-most max solution tree in the
first pass.\\
{\em {\MT}:\/}
In {\MT}, construction of a sub max solution tree after a fail low is
easy to see from {\AB}'s postcondition, or from the code.  (The same
applies for a sub min solution tree after a fail high).

This concludes the description of a single pass of {\ABSSS}. Next, the
globally ``best'' node is selected---in {\SSS} since the OPEN list is
sorted, in {\ABSSS} because the stored $f^+(n)$ values 
together with the $\gamma$ value cause {\MT} to traverse the
critical path. 

The previous explanation indicated that {\SSS} and {\MT} perform the
same operations on 
the OPEN list for the following cases: 
\begin{itemize}
\item At the first call to {\MT} from  {\ABSSS}, where the left-most max
  solution tree is constructed.
\item When finding the critical leaf in subsequent calls to {\MT} from the
  main loop of {\ABSSS}, 
  where the OPEN list contains the leaves of a max solution tree.
\item When further expanding parts of the tree in order to get a
  better upper bound on the root (fail low), or finding that the current upper
  bound is the minimax value (fail high).
\end{itemize}
Comparing {\SSS} to {\MT}/{\ABSSS}, the ingenuity with which the six
$\Gamma$ 
cases emulate {\AB}'s behavior in {\ABSSS} is  extraordinary. Why 
{\SSS} was created in its old form is easier to see if one realizes
that its roots lie in AO*, an algorithm for the search of
problem-reduction spaces \cite{Nils71}.

\cleardoublepage
\chapter{Test Data}
\label{app:testpos}
\label{app:pos}
% \markboth{Appendix E}{Appendix E}
% \markboth{Appendix C}{Test Positions}

%\vspace{-0.5cm}
This appendix gives the test positions with which most of the reported
test results were computed. (The extra positions that were used to
verify that the base test set did not cause anomalies are not shown.)
Furthermore we give  the
absolute node counts (the best in bold) for each position for
 the deepest search
depth. 

\section{Test Results}

\subsubsection{Checkers}

%\begin{figure}[h]
\begin{center}
Cumulative Leaf Nodes Checkers Depth 17 %}\label{fig:cleaf17}
{\footnotesize
%{\scriptsize
\begin{tabular}{r|rrrrr}
position & {\AB} & {\NS} & {\SSS} & {\DUAL} & {\MTDf} \\ \hline
1 & 126183 & 111189    &    106907   &  {\bf  98465}   &          114964        \\
2 & 177188       & 155517    &  192163     &136168        &   {\bf 128615}  \\
3 & 271082       & 285198    & 301536      & 302170       &   {\bf 232238} \\
4 & 945249       & 888190    & 779443      &  {\bf 771386}  & 794280        \\
5 & 1978773        &1884569     &2213865       & 1869080       & {\bf 1661477}  \\
6 & 1368018       & {\bf 1228123} &  1504601      & 1350407       &  1283644  \\
7 & 544271       & {\bf 481183} & 550118      &    538703    &    510826  \\
8 & 905571       & 917252    & 1034249      &  794801      &  {\bf  746069}  \\
9 & 1518870       &1344296     & 1510397      &1490632        &  {\bf 981401}  \\
10 & 1763143       &{\bf 1869436} &  2260341      &  2176786      &     1998528 \\
11 & 221754       & 193222    & {\bf 190221}  & 194031       &      199732 \\
12 &  207108       & {\bf 189654} & 206880       &    197521    &       195524 \\
13 & 203402       & 182482    & 192235      & 217626 &    {\bf  178357}   \\
14 & 253300       & 261241    & 248463      & {\bf 212910}  & 239799    \\
15 & 265731       & 217568    & 268320      & 228979       & {\bf 205280} \\
16 & 248657       & 250896    & 248168      & 241268       & {\bf 238638} \\
17 & 357382       & {\bf 320275} & 398641      & 382976 & 383503 \\
18 &  718949       & 672121    & 767435      &   726969     &     {\bf 623981}        \\
29 &  1319426       & 1040648    & 1244843      &{\bf 815568} &  946250    \\
20 &  1159464      &  1066905   &  1106151     &  1632830 & {\bf 886752}    \\
\end{tabular}
}
\end{center}
% \caption{Cumulative Leaf Nodes Depth 17}\label{fig:cleaf17}
% \end{figure}

% \begin{figure}[h]
\begin{center}
Cumulative Total Nodes Checkers Depth 17 %}\label{fig:cleaf17}
{\footnotesize
%{\scriptsize
\begin{tabular}{r|rrrrr}
position & {\AB} & {\NS} & {\SSS} & {\DUAL} & {\MTDf} \\ \hline
1 & 413369 & {\bf 367200}   &  570840     &  398806  & 459440       \\
2 & 462424       &417435     & 1193339      & {\bf 405075} & 412199   \\
3 &  620027       &667162     & 1052940 & 728422 & {\bf 539190} \\
4 & 2352550  & 2253635 & 2926647 & 2426232 & {\bf 2016395} \\
5 & 4800897       &4716967     &7103273       & 5598219 & {\bf 4124889}   \\
6 & 3227618       & {\bf 2937729}    & 5584087       & 4074537 & 3124792  \\
7 & 1321635       & {\bf 1191009}    & 1982897       &  2074539 & 1305613 \\
8 & 2173069       &  2217950    & 3041809 & 3097442 & {\bf 1903700} \\
9 & 3658536       & 3346953    & 4208761 &  5059490 & {\bf 2587216} \\
10 & 3916120       & {\bf 4130046}     & 6597076      & 7265863       & 4554377  \\
11 & 567607       & {\bf 503305}   & 633484      & 531527       & 523338  \\
12 & 535462       & {\bf 507324}    & 736516      &  585877      &  535851           \\
13 & 511587       & {\bf 470443}   & 641281      &  602930      & 482941             \\
14 &  681776       &697866     & 833239      & 709308       & {\bf 634044}         \\
15 &  707039      & 611411    & 970302      &    864961& {\bf 540321} \\
16 & 656466       & 658201     & 977425      &  901367 &  {\bf 639200} \\
17 & 906151   & {\bf 844337}   &  1279247 &  1227917 & 1018729 \\
18 & 1820759  & {\bf 1738552} & 2528449 & 2711356   & 1755457  \\
29 & 3290314  & 2680018     & 4024730  & 3266939 & {\bf 2491371}  \\
20 & 3124797  & 2870023    & 3197680  & 6197403 & {\bf 2690201}   \\
\end{tabular}
}
\end{center}
% \caption{Cumulative Total Nodes Depth 17}\label{fig:ctotal17}
% \end{figure}

\newpage
\subsubsection{Othello}
\begin{center}
Cumulative Leaf Nodes Othello Depth 10
% \begin{figure}[h]
{\footnotesize
%{\scriptsize
\begin{tabular}{r|rrrrr}
position & {\AB} & {\NS} & {\SSS} & {\DUAL} & {\MTDf} \\ \hline
1 & 1077545 & 1002150 & {\bf 756286} & 1044572 &973214  \\
2 &342372  & 309981 & 400052 & 309283 & {\bf 291823}\\
3 & 469632 & 422933 & 482805  & 435760 & {\bf 410908}\\
4 &810910  & 748961 & {\bf 594689} &774421  &754279  \\
5 & 554011 & 499434 &528925  &507417  & {\bf 493501}\\
6 & 533061 & {\bf 481959}&487469  & 506341 & 496562 \\
7 & 511596 & {\bf 455207}&499995  &472456  & 459245 \\
8 & 550836 & 505633 & 577386 & 509321 & {\bf 502990}\\
9 & 443277 &  {\bf 365431}& 606806 & 419831 & 394294 \\
10 &541548  &519187  &545360  &512457  & {\bf 504142}\\
11 & 675550 & 660471 &{\bf 463393} &681217  & 466876 \\
12 &734587  &609609  & 719514 &{\bf 570924} & 635436 \\
13 & 578096 & {\bf 489623} & 517986 & 536282 & 495934 \\
14 &517719  & 479973 &462101  & 447383 & {\bf 407302 }\\
15 &684021  & 595341 & 738158 & 583758 & {\bf 558692} \\
16 & 2103805 & 1874684  & 2336257  &2146225  & {\bf 1785653} \\
17 & 444027 & {\bf 381167} &497447  &387313  & 381482 \\
18 & 693612 & 665237  & {\bf 610446} & 618819 &660550  \\
19 & 643278 & 562910  &596583  & 577897 & {\bf 518339} \\
20 & 816195 & {\bf 700857} &768269  & 707928 &704796  \\
\end{tabular}
}
\end{center}
% \caption{Cumulative Leaf Nodes Depth 10}\label{fig:cleaf10}
% \end{figure}

\begin{center}
Cumulative Total Nodes Othello Depth 10
% \begin{figure}[h]
{\footnotesize
%{\scriptsize
\begin{tabular}{r|rrrrr}
position & {\AB} & {\NS} & {\SSS} & {\DUAL} & {\MTDf} \\ \hline
1 &1720270  &1617060  &{\bf 1531266}  & 1768573 & 1588262 \\
2 &580163  &541031  & 884386 & 564478 & {\bf 524143} \\
3 &779481  & 715389 &996805  & 781187 & {\bf 699880 }\\
4 & 1380704 &{\bf 1300157}  & 1377430 & 1361942 & 1310286 \\
5 &955789  & {\bf 881531} & 1283857 & 960307 & 882869 \\
6 &975970  &{\bf 900137} & 1099004 &961814  & 927423 \\
7 & 766394 &{\bf 687075} & 1037853 &775328  & 688037 \\
8 &936839  & {\bf 863573} &1265701  &  944988& 871360 \\
9 & 783713 & {\bf 668933} & 1244449 &807264  & 723717 \\
10 &1006877  & 973282 & 1329432 & 971021 & {\bf 940991} \\
11 & 958764 & 941033 &764296  & 1004557 & {\bf 679615} \\
12 &1241744  &1033213  & 1503985 & {\bf 1022343} & 1094600 \\
13 & 963782 & {\bf 845619} & 1292204  & 943211 & 873842 \\
14 & 856701 &807066  & 896673 & 780930 & {\bf 661439} \\
15 & 1105961 &986293  & 1513070 & 1019654 & {\bf 915995} \\
16 &3207698  &2902247  &4565011  &3552642  &{\bf 2753854}  \\
17 & 769704 & {\bf 662209} & 1277545 &693966 & 666419 \\
18 & 1104089 & {\bf 1059736}  &1359768  &986829  & 1092532 \\
19 & 1015573 & 896343 &1281957  & 982266 & {\bf 834850} \\
20 & 1375433 & {\bf 1164485} & 1670091  &1236846  & 1181879 \\
\end{tabular}
}
% \caption{Cumulative Total Nodes Depth 10}\label{fig:ctotal10}
% \end{figure}
\end{center}

\subsubsection{Chess}
\begin{center}
Cumulative Leaf Nodes Chess Depth 8
% \begin{figure}[h]
{\footnotesize
%{\scriptsize
\begin{tabular}{r|rrrrr}
position & {\AB} & {\NS} & {\SSS} & {\DUAL} & {\MTDf} \\ \hline
1 &2139803  & 1985777 & {\bf 1785412} & 1978323 & 1812787 \\
2 &2644826  & 2476699 & 2789442 & {\bf 2084553} & 2204018 \\
3 & 2579806 & 2427222 & 2605278 & 2543878 & {\bf 1772911} \\
4 & 2981122 & 2534212 & 3068148 & 2622208 & {\bf 2186732} \\
5 & 2704505 & 2384329 & 2581678 &2417116  & {\bf 2306988} \\
6 &3144654  & 2608290 & 2959795 & 2619766 & {\bf 2593478} \\
7 & 1519760 & 1477958 & {\bf 1467838} & 1504933 & 1473132 \\
8 & 5341059 & 5317246 & 5008482 & {\bf 3414294} & 3862095 \\
9 & 4886358 & 4053813 & 5356186 & 3906343 & {\bf 3722252} \\
10 &7567304  & 6327725 & 5662936 &5934283  & {\bf 5413685} \\
11 &2604254  & {\bf 2424632} & 2930218 & 2568160 &  2433724 \\
12 & 3751738 & 3383563 & 2420545 &3321493  & {\bf 2205522} \\
13 & 2751157 & 2660323 & 2396603 & 2604666 & {\bf 2064121} \\
14 &2970794  & 2517235 & 3161045 & {\bf 2362827} & 2697990 \\
15 & 1301380 & 1210901 & 1213741 & 1308549 & {\bf 1174032} \\
16 & 1256342 & 1224924 &  1402361& 1207203 & {\bf 1178989} \\
17 & 1120616 & 1111323 &1161968  & {\bf 1065096} & 1118400 \\
18 &  1287345& 1238906 & 1277886 & 1278149 & {\bf 1173362} \\
19 & 2560676  & 2363614 & 2540931 & 2393803 &  {\bf 2132042} \\
20 & 2095458 & 1977936 &  2220139& {\bf 1951231} & 1970067 \\
\end{tabular}
}
% \caption{Cumulative Leaf Nodes Depth 8}\label{fig:cleaf8}
% \end{figure}
\end{center}

\begin{center}
Cumulative Total Nodes Chess Depth 8
% \begin{figure}[h]
{\footnotesize
%{\scriptsize
\begin{tabular}{r|rrrrr}
position & {\AB} & {\NS} & {\SSS} & {\DUAL} & {\MTDf} \\ \hline
1 &3303219  &3103360  & 5902747 &3485589  & {\bf 2920250} \\
2 &3867251  & 3648085 & 6695953 & 3318056  & {\bf 3267966} \\
3 &3970801  & 3723289 & 7834576 & 3958079 &  {\bf 2882594}\\
4 &4670576  & 4031181 & 8728375 & 4283256 & {\bf 3546285} \\
5 &4393422  & 3863494 & 6477649 & 4041426 & {\bf 3742236} \\
6 & 4937220 & {\bf 4178390} & 10703081 & 4298577 & 4188882 \\
7 &2731907  & 2682243 & 3948657 & 2993329 & {\bf 2665386} \\
8 & 7604601 & 7487371 & 12764983 & {\bf 5498919}  &  5728846 \\ 
9 & 7160185 & 6056626 & 14830764 & 6217694 & {\bf 5588940} \\
10 &10339575  &8620155  &21572998  & 8714449 & {\bf 7470636} \\
11 & 4068559 & {\bf 3819434} & 11293758 &4140659  & 3907065 \\
12 &5625271  & 5062261 & 8174419 &5292912  & {\bf 3464844} \\
13 & 4319258 & 4178469 & 9141174 & 4243100 & {\bf 3309548} \\
14 & 4518683 & 3837605 & 9609051 & {\bf 3782715} & 4175020 \\
15 & 2107983 & 1988196 & 2616457 & 2475369 & {\bf 1951313} \\
16 & 2240408 & 2187733 & 4827452 & 2315079 & {\bf 2127931} \\
17 &1980620  & {\bf 1973707} & 4868220 & 2046282 & 2110399 \\
18 & 2164189 & 2110233 & 7863155 & 2226773 & {\bf 2020108} \\
19 &4011716  & 3708911 & 12487591 & 3905568 & {\bf 3389902} \\
20 & 3447220 & {\bf 3272595} & 12443353 & 3407857 & 3276117 \\
\end{tabular}
}
% \caption{Cumulative Total Nodes Depth 8}\label{fig:ctotal8}
% \end{figure}
\end{center}

\newpage
\section{Test Positions}
\index{test positions}

\subsubsection{Checkers}
% 1
% b...
% bb.b
% ....
% w..w
% w...
% .w..
% w.b.
% ....
% B
% 
% 2
% b...
% bb.b
% ....
% w..w
% w..b
% ...b
% ww..
% ..w.
% B
% 
% 3
% b.b.
% bbbb
% ....
% www.
% wb.b
% ...b
% www.
% ..ww
% B
% 
% 4
% b.b.
% bbbb
% ....
% www.
% wbbb
% ....
% w.w.
% .www
% B
% 
% 5
% bbb.
% bb.b
% ....
% www.
% .bbb
% w...
% w.w.
% .www
% B
% 
% 6
% bbb.
% bb.b
% ....
% w.wb
% wbb.
% w...
% w.w.
% .www
% B
% 
% 7
% bbb.
% bb.b
% b...
% wwwb
% .bb.
% w...
% .ww.
% wwww
% B
% 
% 8
% bbb.
% bb.b
% b..b
% www.
% .bb.
% ....
% www.
% wwww
% B
% 
% 9
% bbb.
% bb.b
% b..b
% .wwb
% wb..
% ....
% www.
% wwww
% B
% 
% 10
% bbbb
% bb..
% b..b
% .wwb
% .b..
% w...
% www.
% wwww
% B
% 
% 11
% ....
% .Wb.
% ..bb
% bb.b
% .b.b
% b.ww
% wwww
% .w..
% W
% 
% 12
% W...
% ..b.
% ..bb
% bbbb
% ...b
% b.ww
% wwww
% .w..
% W
% 
% 13
% W...
% ..b.
% .bbb
% b.bb
% ...b
% b.ww
% www.
% .w.w
% W
% 
% 14
% ....
% W.bb
% .bb.
% b.bb
% ...b
% b.ww
% www.
% .w.w
% W
% 
% 15
% bb..
% w..b
% .bb.
% w.bb
% ...b
% b.ww
% www.
% .w.w
% W
% 
% 16
% bb..
% .b.b
% w.b.
% w.bb
% ...b
% b.ww
% www.
% .w.w
% W
% 
% 17
% bbb.
% .b..
% ..b.
% wwbb
% ...b
% b.ww
% www.
% .w.w
% W
% 
% 18
% bbb.
% .b..
% ..b.
% wwbb
% b..b
% ..ww
% .ww.
% ww.w
% W
% 
% 19
% bbb.
% .b..
% ..bb
% wwb.
% b..b
% ..w.
% .www
% ww.w
% W
% 
% 20
% bbb.
% .b.b
% ..b.
% w.b.
% bw.b
% ..w.
% .www
% ww.w
% W

%\vspace{-4cm}
%\begin{figure}
\mbox{
%\vspace{-4cm}
%\hspace{-3cm}
% \epsfbox[138 -227 714 598]{chk1-20.ps}
\includegraphics[width=12cm]{xchk1-12.epsi}
 }
%\end{figure}
\newpage
\ \ \\

%\vspace{-4cm}
\mbox{
%\vspace{-6cm}
%\hspace{-3.5cm} 
%  \epsfbox[138 -227 714 598]{chk1-20.ps}
  \includegraphics[width=12cm]{xchk13-20.epsi}
\vspace{5cm}
}
\vspace{5cm}

\subsubsection{Othello}
\mbox{\vspace{-4cm}%\hspace{-2.1cm}
%  \epsfbox[166 -287 743 532]{oth1-20.ps}
  \includegraphics[width=12cm]{xoth1-12.epsi}
 }
\newpage
\ \ \\

%\vspace{-4.5cm}
\mbox{\vspace{-6cm}
%\hspace{-2.5cm}  
%  \epsfbox[166 -287 743 532]{oth1-20.ps}
  \includegraphics[width=12cm]{xoth13-20.epsi}
 }
\newpage

\subsubsection{Chess}
For all 20 positions: White to move.\\
%\vspace{-2.6cm}
\mbox{
%\vspace{-6cm}
%\hspace{-3.8cm}

  \includegraphics[width=13cm]{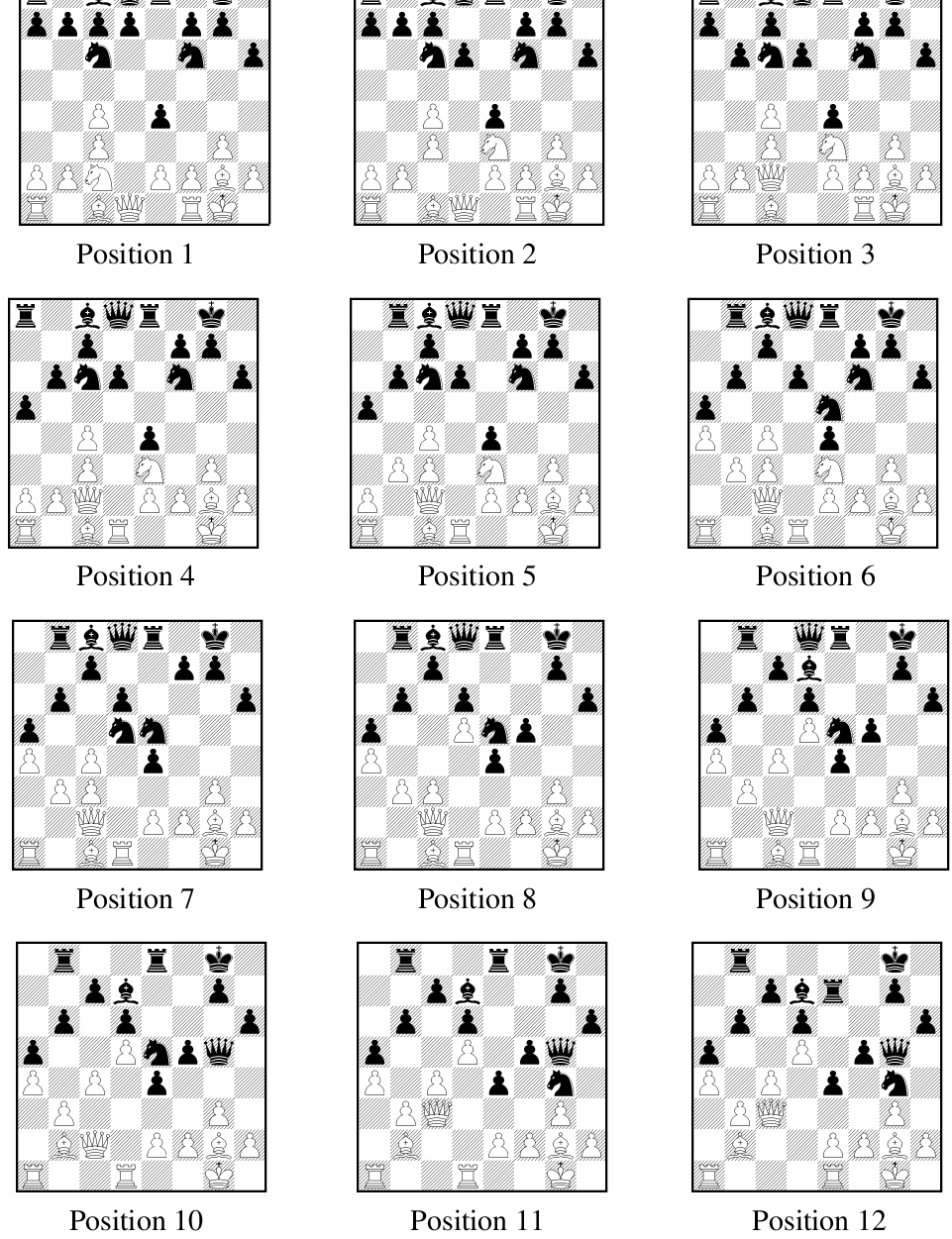}
%\vspace{-5cm}
 }

%\vspace{-3cm}
\mbox{
%\vspace{-4cm}
%\hspace{-4cm}
  \includegraphics[width=13cm]{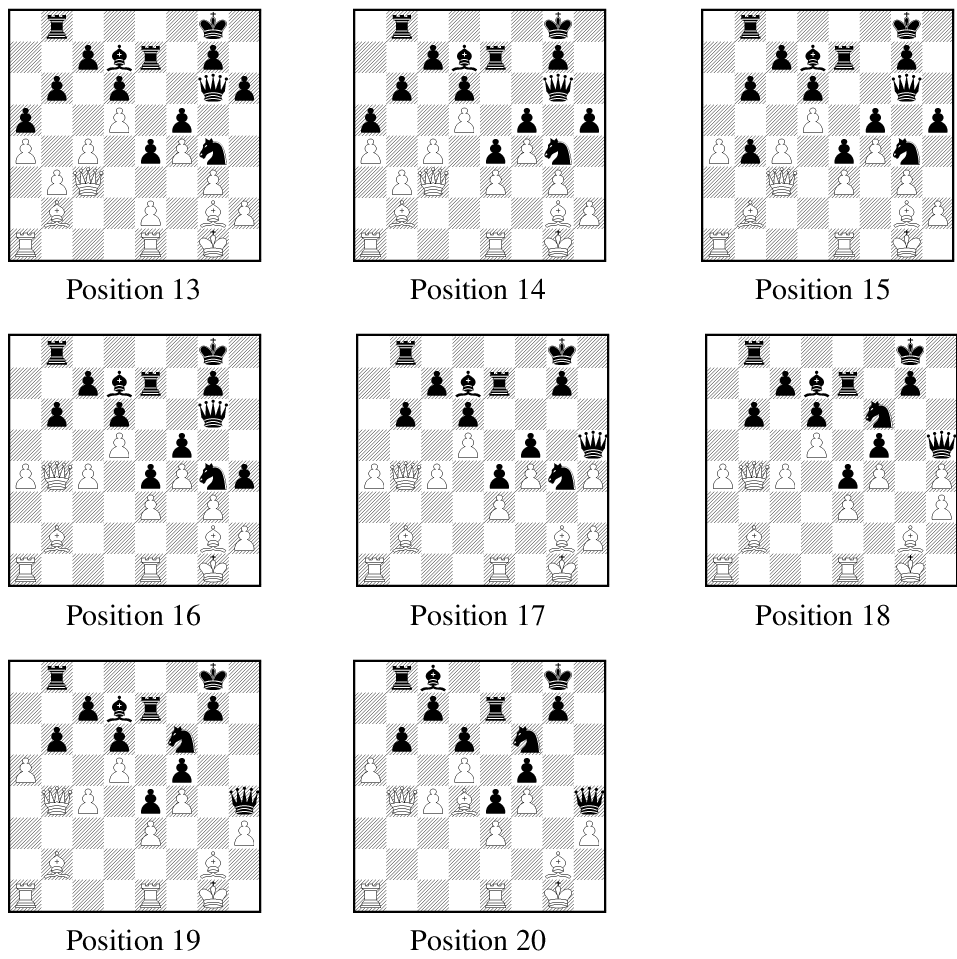}
 }
% \begin{figure}
% YYYYYYYYY
% \vspace{-3cm}
% \mbox{%\hspace{-4cm}
% %\epsfxsize=16cm 
% \epsfbox[180 150 600 700]{pos1-4.ps}}
% %\epsfbox{pos1-4.ps}}
% \vspace{-10cm}
% YYYYYYYY
% \end{figure}
% XXXXXXXXXXx
% \begin{figure}[h]
% \vspace{-3cm}
% \mbox{\hspace{-4cm}\epsfxsize=17cm \epsffile{pos1-4.ps}}
% \vspace{-17cm}
% \end{figure}
% \begin{figure}[h]
% \vspace{-3cm}
% \mbox{\hspace{-4cm}\epsfxsize=17cm \epsffile{pos5-8.ps}}
% \vspace{-17cm}
% \end{figure}
% \begin{figure}[h]
% \vspace {-3cm}
% \mbox{\hspace{-4cm}\epsfxsize=17cm \epsffile{pos9-12.ps}}
% \vspace{-17cm}
% \end{figure}
% \begin{figure}[h]
% \vspace{-3cm}
% \mbox{\hspace{-4cm}\epsfxsize=17cm \epsffile{pos13-16.ps}}
% \vspace{-17cm}
% \end{figure}
% \begin{figure}[h]
% \vspace{-3cm}
% \mbox{\hspace{-4cm}\epsfxsize=17cm \epsffile{pos17-20.ps}}
% \vspace{-17cm}
% \end{figure}

\cleardoublepage
%\markright{Bibliography} 
\markboth{}{}
\addcontentsline{toc}{chapter}{References}
\bibliography{minimax,aske}

\cleardoublepage
\markboth{}{}
\addcontentsline{toc}{chapter}{Index}
{\small
\printindex
}

\cleardoublepage
\chapter*{Samenvatting}
\addcontentsline{toc}{chapter}{Abstract in Dutch}

\vspace{-0.5cm}
%\section*{\centering{Onderzoek naar zoeken \& na-zoeken}}
\section*{\centering{Zoeken \& Her-zoeken Onderzocht}}
\markboth{Samenvatting}{Samenvatting}
Dit proefschrift gaat over zoekalgoritmen.
Zoekalgoritmen worden vaak aan de hand van hun expansie-strategie
gekarakteriseerd.  Een mogelijke strategie is de depth-first strategie, een
eenvoudig backtrack mechanisme waarbij de volgorde
waarin knopen gegenereerd worden bepaalt hoe de zoekruimte doorlopen
wordt.  Een alternatief is de best-first strategie, ontworpen om het
gebruik van domein-afhankelijke heuristische informatie mogelijk te
maken.  Door veelbelovende wegen eerst te bewandelen, zijn best-first
algoritmen in het algemeen effici\"enter dan depth-first algoritmen. 

Voor minimax spelprogramma's (zoals voor schaken en dammen) is de
effici\"entie van het zoek algoritme van cruciaal belang.  Echter,
alle goede programma's zijn gebaseerd op een depth-first 
algoritme, niettegenstaande het succes van best-first strategie\"en
bij andere toepassingen.

In dit onderzoek worden een depth-first algoritme, {\AB},
en een best-first algoritme, {\SSS}, nader beschouwd. De heersende
opinie zegt dat {\SSS} in potentie effici\"enter zoekt, maar dat de
ingewikkelde formulering en het exponenti\"ele geheugengebruik het tot een
onpraktisch algoritme maken. Dit onderzoek laat zien dat er een
verrassend eenvoudige relatie tussen de twee algoritmen bestaat:
{\SSS} is te beschouwen als een speciaal geval van {\AB}.  Empirisch
vervolgonderzoek toont aan dat de heersende opinie over {\SSS} fout
is: het is geen ingewikkeld algoritme, het gebruikt niet te veel
geheugen, maar het is ook niet effici\"enter dan depth-first
algoritmen. 

In de loop der jaren heeft onderzoek naar {\AB} vele verbeteringen
opgeleverd, zoals transpositietabellen en minimale zoekwindows met her-zoeken.
Deze verbeteringen maken het mogelijk dat een depth-first procedure
gebruikt kan worden om de zoekruimte op een best-first manier te
doorlopen. 
Op basis van deze inzichten wordt een nieuw algoritme gepresenteerd,
{\MTDf}, dat beter presteert dan zowel {\SSS} als {\NS}, de in de
praktijk meest gebruikte {\AB} variant.

Naast best-first strategie\"en behandelt dit proefschrift ook andere 
mo\-ge\-lijk\-he\-den om minimax algoritmen te verbeteren. In het
algemeen wordt 
aan\-ge\-no\-men dat geen enkel algoritme dat de minimax waarde wil vaststellen
effici\"enter kan zijn dan de best-case van {\AB} -- dit is de zogeheten
minimale boom. In de praktijk is deze opvatting niet juist.
De echte minimale graaf die de minimax waarde bepaalt blijkt aanmerkelijk
kleiner te zijn dan {\AB}'s best-case. Dien\-ten\-ge\-vol\-ge is er meer
ruimte voor verbetering van minimax zoekalgoritmen dan in het
algemeen wordt aangenomen.

\ \newpage

\thispagestyle{empty}

\ \newpage

\thispagestyle{empty}

\pagestyle{empty}

\ \newpage

\thispagestyle{empty}

\ \newpage

{\small
% first version Andre Lucas
% Updated by Aske Plaat 10 March 1996
%
\pagestyle{empty}\newpage

\noindent{}The Tinbergen Institute is the Netherlands Research Institute
and Graduate School for Economics and Business, which was founded in
1987 by the Faculties of Economics and Econometrics of the Erasmus
University in Rotterdam, the University of Amsterdam and the Free
University in Amsterdam. The Institute is named after the late
Professor Jan Tinbergen, Dutch Nobel Prize laureate in economics in
1969. The Tinbergen Institute is located in Amsterdam and
Rotterdam.

\noindent{}Copies of the books which are published in the Tinbergen
Institute Research Series can be ordered through Thesis Publishers, P.O.
Box 14791, 1001 LG Amsterdam, The Netherlands, phone: +3120 6255429;
fax: +3120 6203395.\\[1.5ex]
The following books already appeared in this series:\\[1.5ex]

%\begin{enumerate}
\newcounter{backboek}
\begin{list}%
{\arabic{backboek}.\hfill\/}{\usecounter{backboek}
   \settowidth{\labelwidth}{108.\/}
   \setlength{\rightmargin}{0pt}
   \setlength{\itemsep}{0pt}
   \setlength{\leftmargin}{\labelwidth}
   \addtolength{\leftmargin}{\labelsep}}
\item O.H. SWANK, {\it Policy makers, voters and optimal control,
   estimation of the preferences behind monetary and fiscal policy in
   the United States.}
\item J. VAN DER BORG, {\it Tourism and urban development. The impact of
   tourism on urban development: towards a theory of urban tourism, and
   its application to the case of Venice, Italy.}
\item A. JOLINK, {\it Libert\'e, Egalit\'e, Raret\'e. The evolutionary
   economics of L\'eon Walras.}
\item R.B. BUITENDIJK, {\it Towards an effective use of relational database
   management systems.}
\item R.M. VERBURG, {\it The two faces of interest. The problem of order and
   the origins of political economy and sociology as distinctive fields
   of inquiry in the Scottish Enlightenment.}
\item H.P. VAN DALEN, {\it Economic policy in a demographically divided
   world.}
\item P.J. VERBEEK, {\it Two case studies on manpower planning in an
   Airline.}
\item M.W. HOFKES, {\it Modelling and computation of general equilibrium.}
\item T.C.R. VAN SOMEREN, {\it Innovatie, emulatie en tijd. De rol van de
   organisatorische vernieuwingen in het economische proces.}
\item M. VAN VLIET, {\it Optimization of manufacturing system design.}
\item R.M.C. VAN WAES, {\it Architectures for Information Management. A
   pragmatic approach on architectural concepts and their application in
   dynamic environments.}
\item K. NIMAKO, {\it Economic change and political conflict in Ghana
   1600-1990.}
\item J.M.C. VOLLERING, {\it Care services for the elderly in the
   Netherlands. The PACKAGE model.}
\item S. ZHANG, {\it Stochastic queue location problems.}
\item C. GORTER, {\it The dynamics of unemployment and vacancies on regional
   labour markets.}
\item P. KOFMAN, {\it Managing primary commodity trade (on the use of
   futures markets).}
\item P.Th. VAN DE LAAR, {\it Financieringsgedrag in de Rotterdamse
   maritieme sector, 1945-1960.}
\item P.H.B.F. FRANSES, {\it Model selection and seasonality in time
   series.}
\item  P.W. VAN WIJCK, {\it Inkomensverdelingsbeleid in Nederland. Over
   individuele voorkeuren en distributieve effecten.}
\item A.E. VAN HEERWAARDEN, {\it Ordering of risks. Theory and actuarial
   applications.}
\item J.C.J.M. VAN DEN BERGH, {\it Dynamic models for sustainable
   development.}
\item H. XIN, {\it Statistics of bivariate extreme values.}
\item C.P. VAN BEERS, {\it Exports of developing countries. Differences
   between South-South and South-North trade and their implications for
   economic development.}
\item L. BROERSMA, {\it The relation between unemployment and interest rate.
   Empirical evidence and theoretical justification.}
\item E. SMEITINK, {\it Stochastic models for repairable systems.}
\item M. DE LANGE, {\it Essays on the theory of financial intermediation.}
\item S.J. KOOPMAN, {\it Diagnostic checking and intra-daily effects in time
series models.}
\item R.J. BOUCHERIE, {\it Product-form in queueing networks.}
\item F.A.G. WINDMEIJER, {\it Goodness of fit in linear and
   qualitative-choice models.}
\item M. LINDEBOOM, {\it Empirical duration models for the labour market.}
\item S.T.H. STORM, {\it Macro-economic considerations in the choice of an
   agricultural policy.}
\item H.E. ROMEIJN, {\it Global optimization by random walk sampling
   methods.}
\item R.W. VAN ZIJP, {\it Austrian and new classical business cycle
   theories.}
\item J.A. VIJLBRIEF, {\it Unemployment insurance and the Dutch labour
   market.}
\item G.E. HEBBINK, {\it Human capital, labour demand and wages. Aspects of
   labour market heterogeneity.}
\item J.J.M. POTTERS, {\it Lobbying and pressure: theory and experiments.}
\item P. BOSWIJK, {\it Cointegration, identification and exogeneity.
   Inference in structural error correction models.}
\item M. BOUMANS, {\it A case of limited physics transfer. Jan Tinbergen's
resources for re-shaping economics.}
\item J.B.J.M. DE KORT, {\it Edge-disjointness in combinatorial
   optimization: problems and algorithms.}
\item J.F.J. DE MUNNIK, {\it The valuation of interest rates derivative
   securities.}
\item J.C.A. POTJES, {\it Empirical studies in Japanese retailing.}
\item J.-K. MARTIJN, {\it Exchange-rate variability and trade: Essays on the
   impact of exchange-rate variability on international trade flows.}
\item J.B.L.M. VERBEEK, {\it Studies on economic growth theory. The role of
   imperfections and externalities.}
\item R.H. VAN HET KAAR, {\it Medezeggenschap bij fusie en ontvlechting.}
\item F. KALSHOVEN, {\it Over Marxistische economie in Nederland,
   1883-1939.}
\item W. SWAAN, {\it Behaviour and institutions under economic reform. Price
   regulation and market behaviour in Hungary.}
\item J.J. CAPEL, {\it Exchange rates and strategic decisions of firms.}
\item M.F.M. CANOY, {\it Bertrand meets the fox and the owl. Essays in the
   theory of price competition.}
\item H.M. KAT, {\it The efficiency of dynamic trading strategies in
   imperfect markets.}
\item E.A.M. BULDER, {\it The social economics of old age: strategies to
   maintain income in later life in the Netherlands 1880-1940.}
\item J. BARENDREGT, {\it The Dutch Money Purge. The monetary consequences
   of German occupation and their redress after liberation, 1940-1952.}
\item Nanda PIERSMA, {\it Combinatorial optimization and empirical processes.}
\item M.J.C. SIJBRANDS, {\it Strategische en logistieke besluitvorming. Een
   empirisch onderzoek naar informatiegebruik en instrumentele
   ondersteuning.}
\item H.J. VAN DER SLUIS, {\it Heuristics for complex inventory systems.
   Deterministic and stochastic problems.}
\item E.F.M. WUBBEN, {\it Markets, uncertainty and decision-making. A
   history of the introduction of uncertainty into economics.}
\item V.J. BATELAAN, {\it Organizational culture and strategy. A study of
   cultural influences on the formulation of strategies, goals, and
   objectives in two companies.}
\item R.M. DE JONG, {\it Asymptotic theory of expanding parameter space
   methods and data dependence in econometrics.}
\item D.P.M. DE WIT, {\it Portfolio management of common stock and real
   estate assets. An empirical investigation into the stochastic
   properties of common stock and equity real-estate.}
\item A. LAGENDIJK, {\it The internationalisation of the Spanish automobile
   industry and its regional impact. The emergence of a
   growth-periphery.}
\item B.M. KLING, {\it Life insurance, a non-life approach.}
\item J.H. GROOTENDORST, {\it De markthuur op kantorenmarkten in Nederland.}
\item M. DINGENA, {\it The creation of meaning in advertising. Interaction
   of figurative advertising and individual differences in processing
   styles.}
\item R.T. LIE, {\it Economische dynamiek en toplocaties.
   Locatiekarakteristieken en prijsontwikkeling van kantoren in een
   aantal grote Europese steden.}
\item R.L.M. PEETERS, {\it System identification based on Riemannian
   geometry: theory and algorithms.}
\item O. MEMEDOVIC, {\it On the theory and measurement of comparative
   advantage. An empirical analysis of Yugoslav trade in manufactures
   with the OECD countries, 1970-1986.}
\item S. FISCHER, {\it The paradox of information technology management.}
\item P.A. STORK, {\it Modelling international financial markets: an
   empirical study.}
\item R.A. BELDERBOS, {\it Strategic trade policy and multinational
   enterprises: essays on trade and investment by Japanese electronics
   firms.}
\item I.T. VAN DEN DOEL, {\it Dynamics in cross-section and panel data
   models.}
\item M.W. DELL, {\it Maximum price regulations and resulting parallel and
   black markets.}
\item Bo CHEN, {\it Worst case performance of scheduling heuristics.}
\item P.W. CHRISTIAANSE, {\it Strategic advantage and the exploitability of
   information technology. An empirical study of the effects of IT on
   supplier-distributor relationships in the US airline industry.}
\item L. LEI, {\it User participation and the success of information system
   development. An integrated model of user-specialist relationships.}
\item J.H. BAGGEN, {\it Duurzame mobiliteit. Duurzame ontwikkeling en de
   voor\-zie\-ningen\-struc\-tuur van het personenvervoer in de Randstad.}
\item R.A. BOSCHMA, {\it Looking through a window of locational opportunity.
   A long-term spatial analysis of techno-industrial upheavals in
   Great-Britain and Belgium.}
\item C.A.G. SNEEP, {\it Innovation management in the Agro-food industry.}
\item F.R. KLEIBERGEN, {\it Identifiability and nonstationarity in classical
   and Bayesian econometrics.}
\item R.F. VAN DE WIJNGAERT, {\it Trade Unions and collective bargaining in
   the Netherlands.}
\item M. BOOGAARD, {\it Defusing the software crisis: Information systems
   flexibility through data independence.}
\item E.M. VERMEULEN, {\it Corporate risk management. A multi-factor
   approach.}
\item R.A. ZUIDWIJK, {\it Complementary triangular forms for pairs of
   matrices and operators.}
\item M.H. GOEDHART, {\it Financial planning in divisionalised firms: models
   and methods.}
\item E. EGGINK, {\it A symmetric approach to the labor market: an
   application of the sem method.}
\item A.F. CORRELJ\'E, {\it The Spanish Oil Industry: Structural change and
   modernization.}
\item A.H.M. LELIVELD, {\it Social security in developing countries;
   operation and dynamics of social security mechanisms in rural
   Swaziland.}
\item Y.M. PRINCE, {\it Price cost margins in Dutch manufacturing: with an
   emphasis on cyclical and firm size effects.}
\item W.E. KUIPER, {\it Farmers, prices and rational expectations.}
\item B. ROORDA, {\it Global total least squares. A method for the
   construction of open approximate models from vector time series.}
\item A.I. MARTINS BOTTO DE BARROS, {\it Discrete and fractional programming
   techniques for location models.}
\item J.A. DOS SANTOS GROMICHO, {\it Quasiconvex optimization and location
   models.}
\item R.B. KOOL, {\it Aspects of enlarging the market effects in the Dutch
   health care.}
\item B. LEEFTINK, {\it The desirability of currency unification theory and
   some evidence.}
\item Andr\'e VAN VLIET, {\it Heuristic and optimal methods for the
   three-dimensional bin packing problem.}
\item C.C.J.M. HEYNEN, {\it Models for option evaluation in alternative
   price-movements.}
\item Yvonne VAN EVERDINGEN, {\it Adoption and diffusion of the European
currency unit. An empirical study among European companies.}
\item Robin DE VILDER, {\it Endogenous business cycles.}
\item Joachim J. STIBORA and Albert DE VAAL, {\it Services and services
   trade: A theoretical inquiry.}
\item Ronald VAN DER BIE, {\it "Een doorlopende groote roes". De economische
   ontwikkeling van Nederland, 1913-1921.}
\item W.J. JANSEN, {\it International capital mobility and asset demand. Six
   empirical studies.}
\item Natasha E. STROEKER, {\it Second-hand markets for consumer durables.}
\item Annette VAN DEN BERG, {\it Trade union growth and decline in The
   Netherlands.}
\item Patrick VAN DER LAAG, {\it An analysis of refinement operators in
   inductive logic programming.}
\item Euro BEINAT, {\it Multiattribute value functions for environmental
   management.}
\item H.A. VAN KLINK, {\it Towards the borderless mainport Rotterdam. An
   analysis of functional, special and administrative dynamics in port
   systems.}
\item W.H.J. HASSINK, {\it Worker flows and the employment adjustment of
   firms. An empirical investigation.}
\item S. LIU, {\it Contributions to matrix calculus and applications in
   econometrics.}
\item Mirjam VAN PRAAG, {\it Determinants of successful entrepreneurship.}
\item Erik VERHOEF, {\it Economic efficiency and social feasibility in the
   regulation of road transport externalities.}
\item Ilaria BRAMEZZA, {\it The competitiveness of a European city and the
   role of urban management in improving the city's performance. The
   cases of the Central Veneto and Rotterdam regions.}
\item Guido BIESSEN, {\it East European foreign trade and system changes.}
\item Andr\'e LUCAS, {\it Outlier robust unit root analysis.}
\item Tineke FOKKEMA, {\it Residential moving behaviour of the
    elderly: an explanatory analysis for the Netherlands.}
\item Turan EROL, {\it Exchange rate policy in Turkey. External
    competitiveness and internal stability studied through a
    macromodel.}
\item Thijs DE RUYTER VAN STEVENINCK, {\it The impact of capital
    imports on the Argentine economy, 1970--1989} 
\item Dennis DANNENBURG, {\it Actuarial credibility models:
    evaluations and extensions.}
\item Carsten FOLKERTSMA, {\it On equivalence scales.}
\item Aske PLAAT, {\it Research Re:\,search \& Re-search.}
\end{list}
%\end{enumerate}

}

\end{document}